# DECISIVE Benchmarking Data Report
## sUAS Performance Results from Phase I

Adam Norton, Reza Ahmadzadeh, Kshitij Jerath, Paul Robinette,
Jay Weitzen, Thanuka Wickramarathne, Holly Yanco, Minseop Choi,
Ryan Donald, Brendan Donoghue, Christian Dumas, Peter Gavriel,
Alden Giedraitis, Brendan Hertel, Jack Houle, Nathan Letteri,
Edwin Meriaux, Zahra Rezaei Khavas, Rakshith Singh, Gregg Willcox, Naye Yoni

University of Massachusetts Lowell
U.S. Army Combat Capabilities Development Command Soldier Center (DEVCOM-SC)

**Abstract**: This report reviews all results derived from performance benchmarking conducted during Phase I of the Development and Execution of Comprehensive and Integrated Subterranean Intelligent Vehicle Evaluations (DECISIVE) project by the University of Massachusetts Lowell, using the test methods specified in the DECISIVE Test Methods Handbook v1.1 for evaluating small unmanned aerial systems (sUAS) performance in subterranean and constrained indoor environments, spanning communications, field readiness, interface, obstacle avoidance, navigation, mapping, autonomy, trust, and situation awareness. Using those 20 test methods, over 230 tests were conducted across 8 sUAS platforms: Cleo Robotics Dronut X1P (P = prototype), FLIR Black Hornet PRS, Flyability Elios 2 GOV, Lumenier Nighthawk V3, Parrot ANAFI USA GOV, Skydio X2D, Teal Golden Eagle, and Vantage Robotics Vesper. Best in class criteria is specified for each applicable test method and the sUAS that match this criteria are named for each test method, including a high-level executive summary of their performance.

# Table of Contents

# Executive Summary

Using the 20 test methods specified in the DECISIVE Test Methods Handbook v1.1 [Norton et al., 2022], over 230 tests were conducted across 8 sUAS platforms: Cleo Robotics Dronut X1P (P = prototype), FLIR Black Hornet PRS, Flyability Elios 2 GOV, Lumenier Nighthawk V3, Parrot ANAFI USA GOV, Skydio X2D, Teal Golden Eagle, and Vantage Robotics Vesper. The table below provides a high-level review of each system designated as best in class per test method, followed by a summarization of best in class per each test method category. Some tests are not shown in the below tables if best in class criteria is not appropriate (e.g., Logistics and OCU Characterization).

It should be noted that the results contained in this report should be interpreted as benchmarks for each system at this particular moment in time and that their performance may differ in future evaluations due to system updates.

✓ = Best in class per test method
* = Not enough tests conducted or sUAS evaluated to determine best in class
- = Not evaluated due to sUAS inability, logistic issues, or availability
X = Not evaluated due to safety concerns

| Test Method | Test | Cleo Robotics Dronut X1P | FLIR Black Hornet PRS | Flyability Elios 2 GOV | Lumenier Nighthawk V3 | Parrot ANAFI USA GOV | Skydio X2D | Teal Golden Eagle | Vantage Robotics Vesper |
|---|---|---|---|---|---|---|---|---|---|
| NLOS Communications | Horizontal, NERVE | | ✓ | ✓ | | ✓ | | | ✓ |
| | Vertical, NERVE | | ✓ | | | | | | ✓ |
| | Horizontal, MUTC | - | | ✓ | ✓ | ✓ | ✓ | X | |
| | Vertical, MUTC | | ✓ | ✓ | ✓ | ✓ | X | X | |
| NLOS Video Latency | Horizontal, NERVE | | ✓ | ✓ | ✓ | ✓ | ✓ | ✓ | - |
| | Vertical, NERVE | | ✓ | | | | | ✓ | - |
| Runtime Endurance | Indoor Movement | | | | X | ✓ | ✓ | X | - |
| | Hover and Stare | | | | | ✓ | ✓ | | |
| | Perch and Stare | | - | - | X | ✓ | | | X |
| Takeoff and Land/Perch | Takeoff | | - | ✓ | ✓ | ✓ | | ✓ | |
| | Land/Perch | | - | ✓ | | ✓ | | | |
| | Dark Operations | | | ✓ | - | ✓ | | X | - |
| Room Clearing | In-Situ, Static Cam | | | | | ✓ | ✓ | X | |
| | In-Situ, Zoom Cam | - | | - | - | ✓ | ✓ | X | |
| | Post-Hoc, Static Cam | | | ✓ | | ✓ | ✓ | X | |
| | Post-Hoc, Zoom Cam | - | | - | - | ✓ | ✓ | X | |
| Indoor Noise Level | | | ✓ | | | | | | |
| Obstacle Avoidance and Resilience | Collision Resilience | | - | ✓ | - | - | - | - | |
| Position and Traversal Accuracy | Path Deviation | | - | ✓ | ✓ | ✓ | | | ✓ |
| | Waypoint Navigation | | - | ✓ | ✓ | ✓ | | ✓ | ✓ |
| Navigation Through Apertures | Doorway | ✓ | ✓ | ✓ | ✓ | ✓ | ✓ | X | ✓ |
| | Window | | ✓ | ✓ | | | | X | ✓ |
| Navigation Through Confined Spaces | Hallway | ✓ | ✓ | ✓ | ✓ | ✓ | ✓ | X | |
| | Tunnel | - | * | * | - | - | - | X | - |
| | Stairwell | | | ✓ | - | ✓ | ✓ | | |
| | Shaft | ✓ | ✓ | ✓ | | | - | X | - |
| Indoor Mapping Resolution | | - | - | * | - | - | - | - | - |
| Indoor Mapping Accuracy | Horizontal, NERVE | - | - | * | - | - | - | - | - |
| | Vertical, NERVE | - | - | * | - | - | - | - | - |
| | 3D, MUTC Fire Trainer | - | - | * | - | - | - | - | - |
| | 3D, MUTC Hotel Trainer | - | - | * | - | - | - | - | - |
| Non-Contextual Autonomy Ranking | | | - | | | ✓ | ✓ | ✓ | ✓ |
| Contextual Autonomy Ranking | | | - | ✓ | ✓ | | | | ✓ |
| Characterizing Factors of Trust | Protective Hardware | ✓ | | ✓ | | | | | ✓ |
| | Low Light Performance | | | ✓ | ✓ | | | | |
| Interface-Afforded Attention Allocation | | | - | ✓ | | | ✓ | | - |
| SA Survey Comparison | | - | - | * | - | * | - | - | - |

|  |  |  |  |
|---|---|---|---|
| **Communications** | **NLOS Communications** | 3 or more best in class tests | FLIR Black Hornet PRS; Flyability Elios 2 GOV; Parrot ANAFI USA GOV |
| | **NLOS Video Latency** | 2 best in class tests | FLIR Black Hornet PRS; Teal Golden Eagle |
| **Field Readiness** | **Runtime Endurance** | 2 or more best in class tests | Parrot ANAFI USA GOV; Skydio X2D |
| | **Takeoff and Land/Perch** | 3 best in class tests | Flyability Elios 2 GOV; Parrot ANAFI USA GOV |
| | **Room Clearing** | 4 best in class tests | Parrot ANAFI USA GOV; Skydio X2D |
| **Obstacle Avoidance** | **Collision Resilience** | Best in class | Flyability Elios 2 GOV |
| **Navigation** | **Position and Traversal Accuracy** | 2 best in class tests | Flyability Elios 2 GOV; Lumenier Nighthawk V3; Parrot ANAFI USA GOV; Vantage Robotics Vesper |
| | **Navigation Through Apertures** | 2 best in class tests | FLIR Black Hornet PRS; Flyability Elios 2 GOV; Vantage Robotics Vesper |
| | **Navigation Through Confined Spaces** | 2 or more best in class tests | Cleo Robotics Dronut X1P; FLIR Black Hornet PRS; Flyability Elios 2 GOV; Parrot ANAFI USA GOV; Skydio X2D |
| **Autonomy** | **Non-Contextual Autonomy Ranking** | Best in class | Parrot ANAFI USA GOV; Skydio X2D; Teal Golden Eagle; Vantage Robotics Vesper |
| | **Contextual Autonomy Ranking** | Best in class | Flyability Elios 2 GOV; Lumenier Nighthawk V3; Vantage Robotics Vesper |
| **Trust** | **Protective Hardware** | Best in class | Cleo Robotics Dronut X1P; Flyability Elios 2 GOV; Vantage Robotics Vesper |
| | **Low Light Performance** | Best in class | Flyability Elios 2 GOV; Lumenier Nighthawk V3 |
| **Situation Awareness** | **Interface-Afforded Attention Allocation** | Best in class | Flyability Elios 2 GOV; Skydio X2D |

# sUAS Platforms Evaluated

The sUAS platforms evaluated for this project were selected due to matching some of the desired performance capabilities for operating in subterranean and constrained indoor environments. These desired capabilities are derived from various Army reference documents, including the U.S. Army Subterranean and Dense Urban Environment MATDEV CoP Future Materiel Experiment (MATEx) Planning: Dense Urban Materiel Concepts and Capabilities RFI, as well as guidance from DEVCOM-SC. These capabilities include (in order of decreasing importance): GPS-denied operation, collision avoidance, ability to perch and stare, ability to operate in lowlight conditions, and small enough to comfortably fit through a typical door threshold.

A set of 8 platforms were evaluated, 4 of which are from the Blue sUAS list and the remaining 4 are NDAA compliant systems. While not all systems match all selection criteria, the 8 that were selected initially claimed to meet the minimum defined criteria for GPS-denied operation and being physically small enough to fit through a typical door threshold. The systems are listed below; full configuration details are provided in the **Logistics Characterization** test method results.

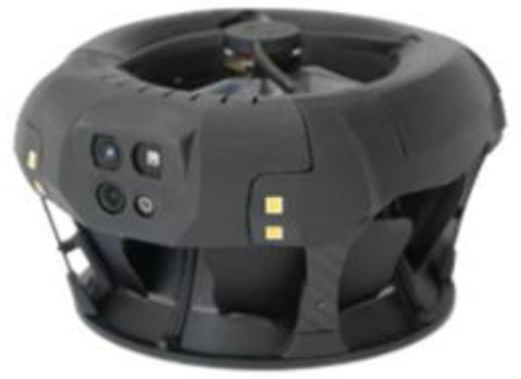

**Cleo Robotics Dronut X1P**
(P = prototype)

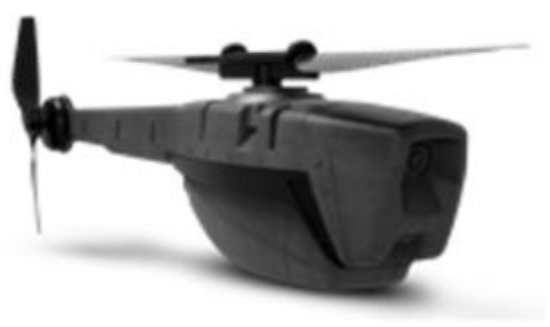

**FLIR Black Hornet PRS**

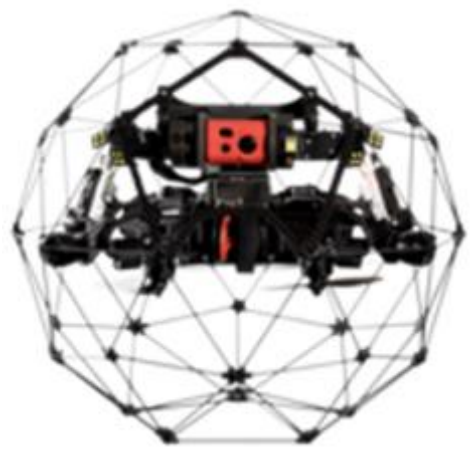

**Flyability Elios 2 GOV**

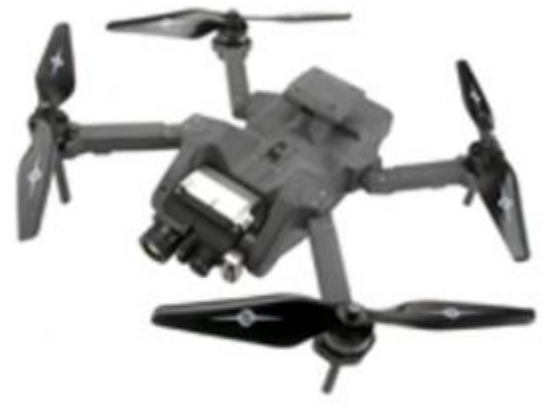

**Lumenier Nighthawk V3**

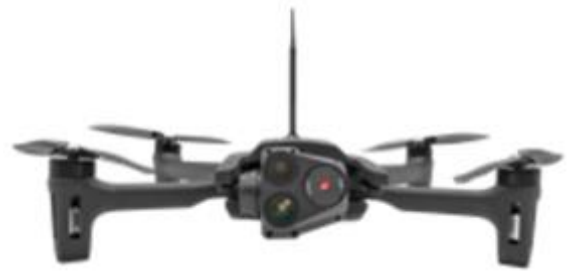

**Parrot ANAFI USA GOV**

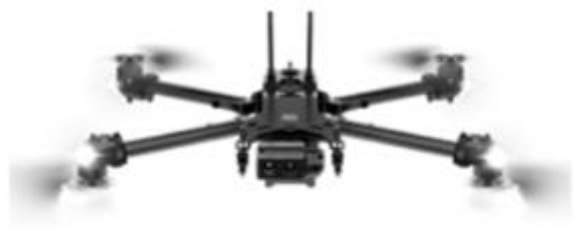

**Skydio X2D**

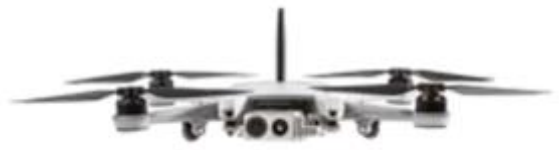

**Teal Golden Eagle**
(with custom prop guards)

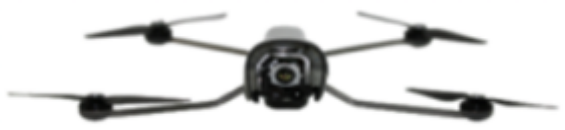

**Vantage Robotics Vesper**

# Communications

## Non-Line-of-Sight (NLOS) Communications

### Summary of Test Method

This test method consists of connecting the sUAS and the OCU at an initial position and then moving the sUAS to other positions that are in NLOS with the initial point due to a wall or floor obstruction. The NLOS communication range for each position is recorded, measured as a straight path through one or more floors and walls between the sUAS and the OCU. The position at which communication fails is indicated by a lack of ability to transmit video, control signal, or command the sUAS to perform tasks. This measure provides an approximate scenario at which the sUAS would be expected to lose communications signal in a real-world deployment.

For each OCU position in the test, the sUAS starts on the ground while the operator attempts to make initial connection to confirm video and control signals (i.e., static connection test). Once confirmed, the operator attempts the following tasks: takeoff, hovering in place, yawing, pitching forward and back, rolling left and right, ascending and descending, camera movement, and landing.

This test method can be run concurrently with the NLOS Video Latency test method.

## Benchmarking Results

Tests were conducted at the UMass Lowell NERVE Center and Muscatatuck Urban Training Center (MUTC) in the Fire Trainer and Hotel Trainer test sites, both comprised of Conex container structures. Results from both test locations are shown below.

## UMass Lowell NERVE Center

### Environment characterization

**Horizontal, through walls**

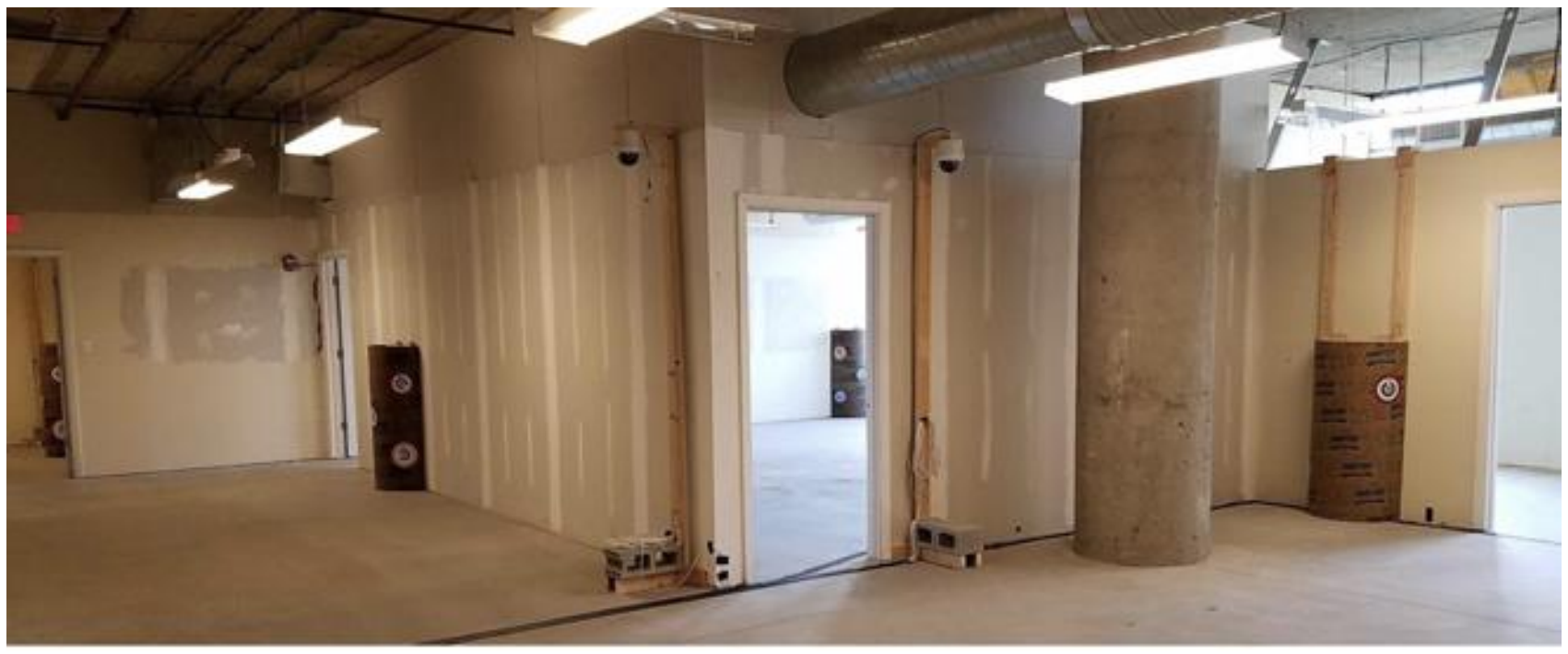

**Vertical, through floors**

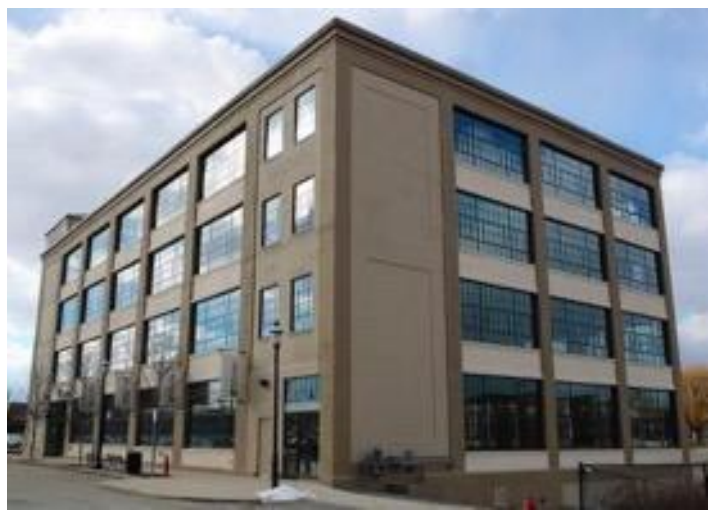

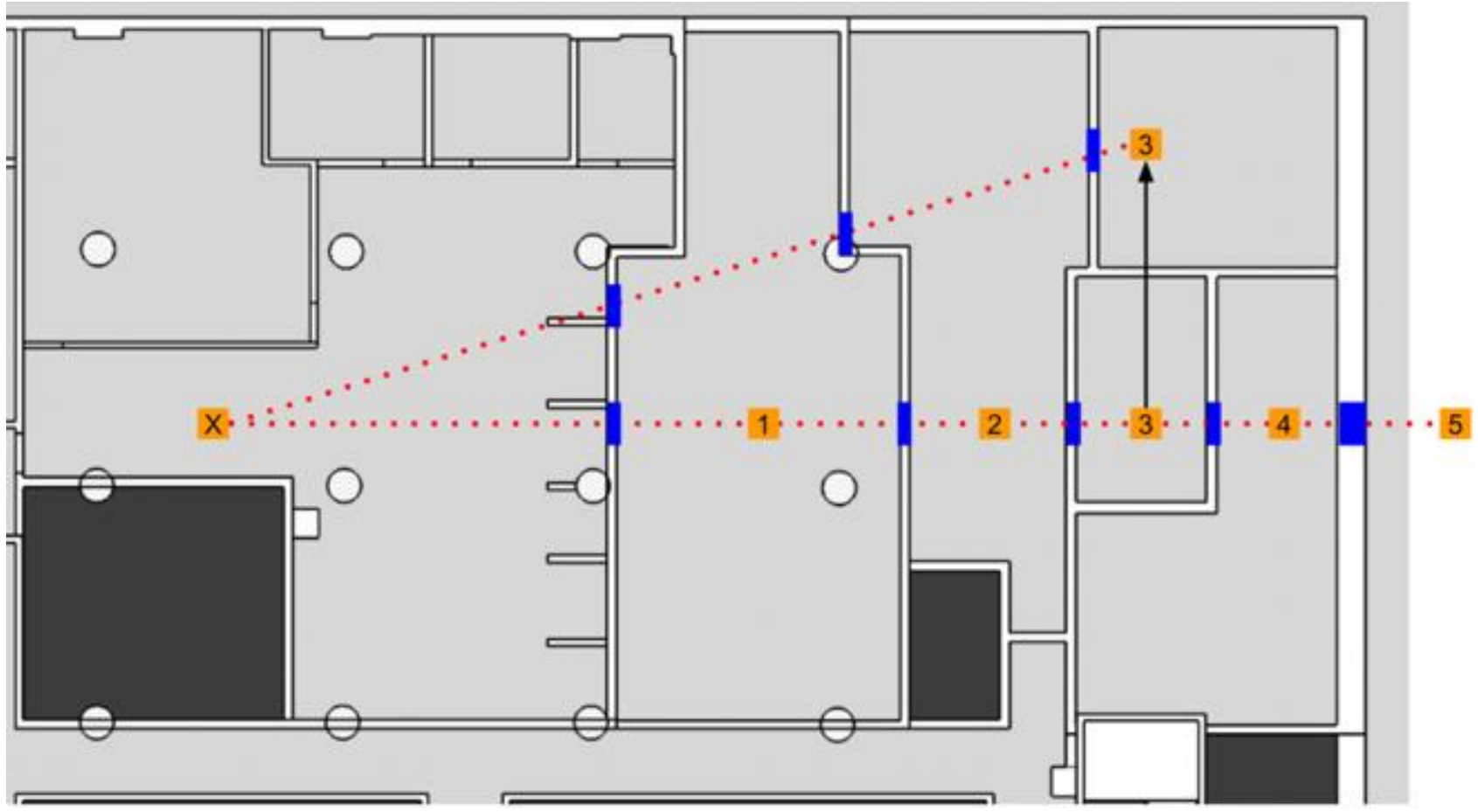

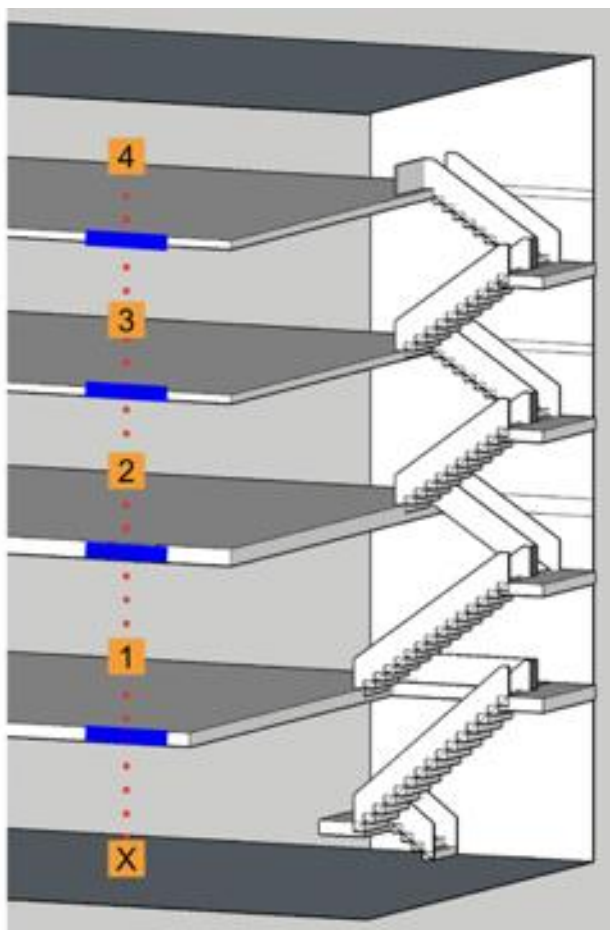

| Metrics | Horizontal, through walls | | | | | Vertical, through floors | | | |
|---|---|---|---|---|---|---|---|---|---|
| | X-1 | X-2 | X-3 | X-4 | X-5 | X-1 | X-2 | X-3 | X-4 |
| Position distance (m) | 14 | 20 | 25 | 27 | 31 | 5 | 9 | 13 | 17 |
| Position obstructions | 1 drywall | 2 drywall | 3 drywall | 4 drywall | 4 drywall<br>1 concrete | 1 concrete | 2 concrete | 3 concrete | 4 concrete |

## Performance data

Best in class = maximum NLOS performance through 4 walls or 2 floors (i.e., 1 less than the maximum number of walls or floors successfully transmitted through across all systems)

| | | Horizontal, through walls | | | | | | Vertical, through floors | | | | |
|---|---|---|---|---|---|---|---|---|---|---|---|---|
| **sUAS and communications frequency** | **Metrics** | **X** | **1** | **2** | **3** | **4** | **5** | **X** | **1** | **2** | **3** | **4** |
| Cleo Robotics Dronut X1P<br>2.4 GHz | Connect | ✓ | ✓ | ✓ | X | X | X | ✓ | X | X | X | X |
| | Fly | ✓ | ✓ | ✓ | X | X | X | ✓ | X | X | X | X |
| | Maximum NLOS performance | 20 m, 2 walls | | | | | | 0 | | | | |
| FLIR Black Hornet PRS<br>355-385 MHz | Connect | ✓ | ✓ | ✓ | ✓ | ✓ | ✓ | ✓ | ✓ | ✓ | ✓ | ✓ |
| | Fly | ✓ | ✓ | ✓ | ✓ | ✓ | ✓ | ✓ | ✓ | ✓ | ✓ | ✓ |
| | Maximum NLOS performance | 31 m, 5 walls | | | | | | 17 m, 4 floors | | | | |
| Flyability Elios 2 GOV<br>2.4 GHz | Connect | ✓ | ✓ | ✓ | ✓ | ✓ | X | ✓ | ✓ | X | X | X |
| | Fly | ✓ | ✓ | ✓ | ✓ | ✓ | X | ✓ | ✓ | X | X | X |
| | Maximum NLOS performance | 27 m, 4 walls | | | | | | 5 m, 1 floor | | | | |
| Lumenier Nighthawk V3<br>2.4 GHz | Connect | ✓ | ✓ | ✓ | ✓ | X | X | ✓ | ✓ | X | X | X |
| | Fly | ✓ | ✓ | ✓ | ✓ | X | X | ✓ | ✓ | X | X | X |
| | Maximum NLOS performance | 25 m, 3 walls | | | | | | 5 m, 1 floor | | | | |
| Parrot ANAFI USA GOV<br>2.4 GHz | Connect | ✓ | ✓ | ✓ | ✓ | ✓ | ✓ | ✓ | ✓ | X | X | X |
| | Fly | ✓ | ✓ | ✓ | ✓ | ✓ | ✓ | ✓ | ✓ | X | X | X |
| | Maximum NLOS performance | 31 m, 5 walls | | | | | | 5 m, 1 floor | | | | |
| Skydio X2D<br>1.8 GHz | Connect | ✓ | ✓ | ✓ | X | X | X | ✓ | ✓ | X | X | X |
| | Fly | ✓ | ✓ | ✓ | X | X | X | ✓ | ✓ | X | X | X |
| | Maximum NLOS performance | 20 m, 2 walls | | | | | | 5 m, 1 floor | | | | |
| Teal Golden Eagle*<br>2.4 GHz | Connect | ✓ | ✓ | ✓ | ✓ | ✓ | X | ✓ | ✓ | ✓ | X | X |
| | Fly | X | X | X | X | X | X | X | X | X | X | X |
| | Maximum NLOS performance | 27 m, 4 walls<br>(Connect only) | | | | | | 9 m, 2 floors<br>(Connect only) | | | | |
| Vantage Robotics Vesper<br>1.8 GHz | Connect | ✓ | ✓ | ✓ | ✓ | ✓ | ✓ | ✓ | ✓ | ✓ | X | X |
| | Fly | ✓ | ✓ | ✓ | ✓ | ✓ | X | ✓ | ✓ | ✓ | X | X |
| | Maximum NLOS performance | 27 m, 4 walls | | | | | | 9 m, 2 floors | | | | |
| | **Best in class:** | **Horizontal, through walls** | | | | | | **Vertical, through floors** | | | | |
| | | FLIR Black Hornet PRS<br>Flyability Elios 2 GOV<br>Parrot ANAFI GOV USA<br>Vantage Robotics Vesper | | | | | | FLIR Black Hornet PRS<br>Vantage Robotics Vesper | | | | |
| | | **NLOS Communications (NERVE), overall** | | | | | | | | | | |
| | | FLIR Black Hornet PRS<br>Vantage Robotics Vesper | | | | | | | | | | |

*Note: Due to instability of the Teal Golden Eagle when flying indoors, we did not attempt to fly it to confirm communications performance at each location.

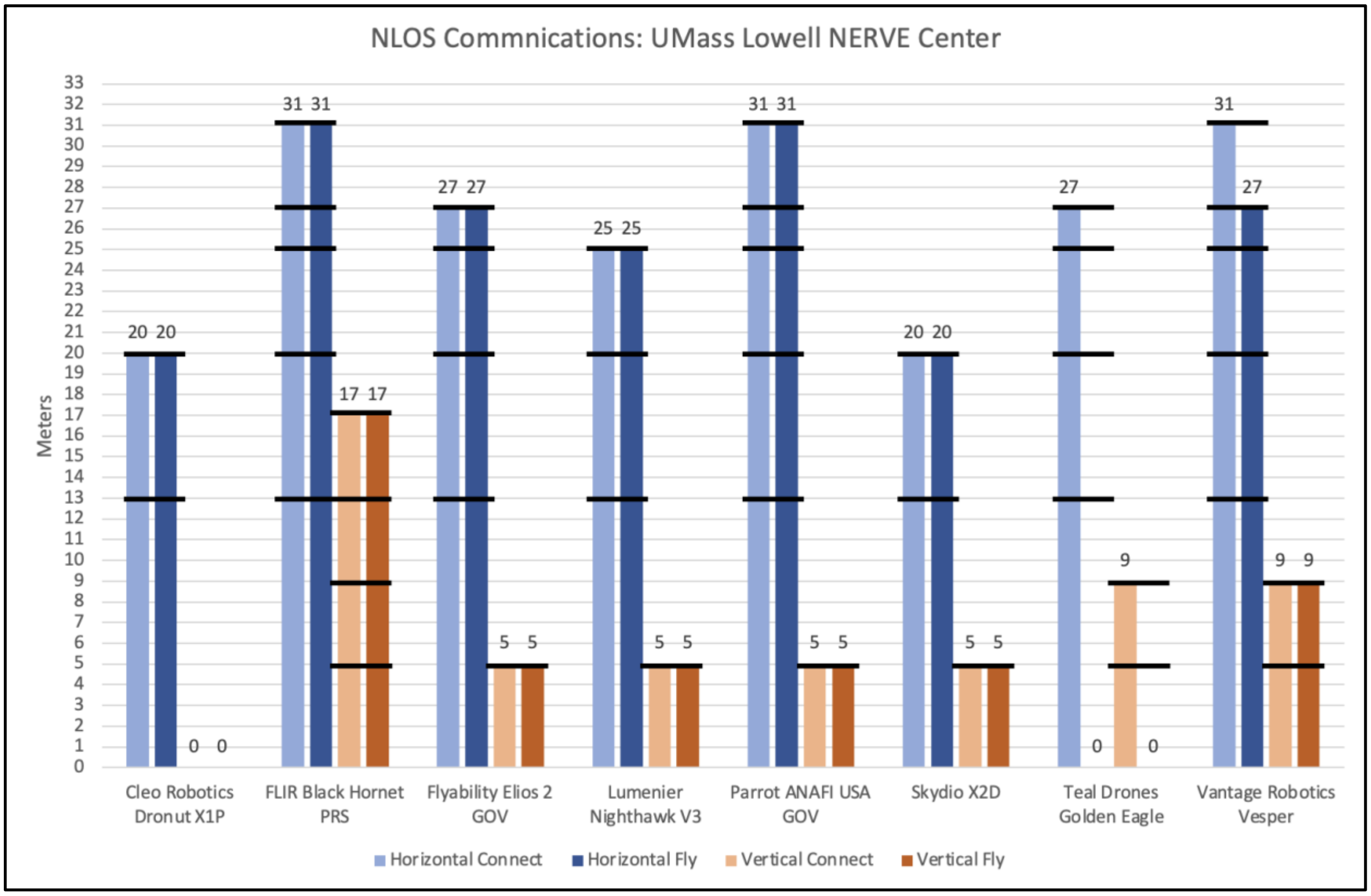
NLOS Commnications: UMass Lowell NERVE Center
Meters
Cleo Robotics Dronut X1P
FLIR Black Hornet PRS
Flyability Elios 2 GOV
Lumenier Nighthawk V3
Parrot ANAFI USA GOV
Skydio X2D
Teal Drones Golden Eagle
Vantage Robotics Vesper
Horizontal Connect
Horizontal Fly
Vertical Connect
Vertical Fly

# MUTC Fire Trainer and Hotel Trainer

## Environment characterization

**Horizontal, through walls** **Vertical, through floors**

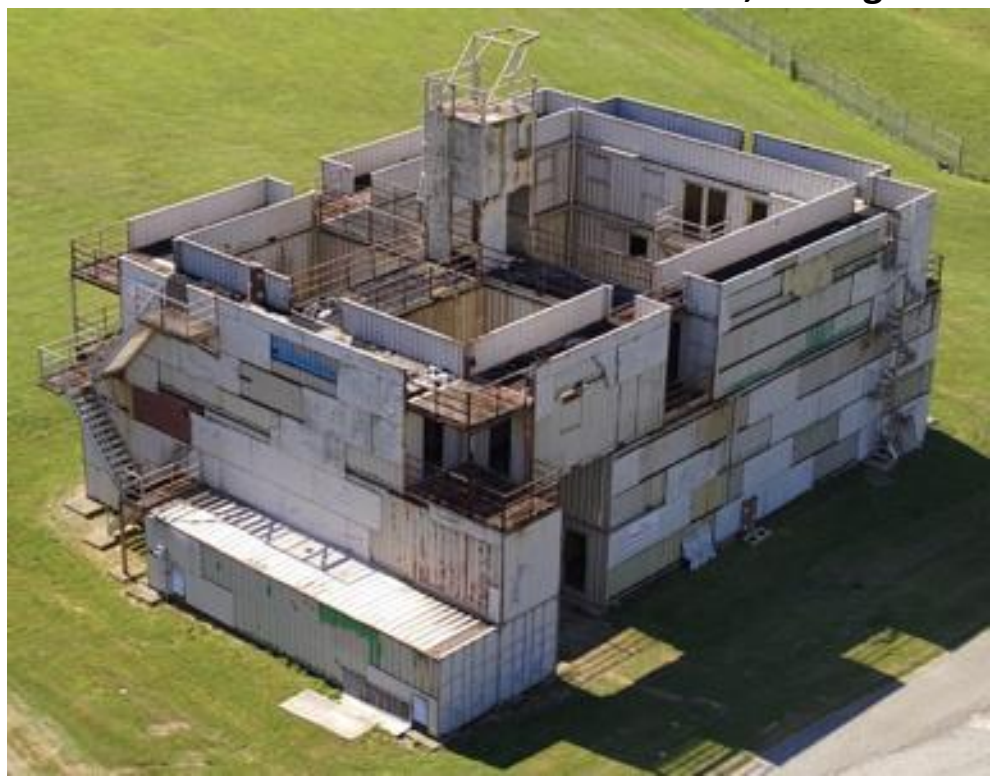
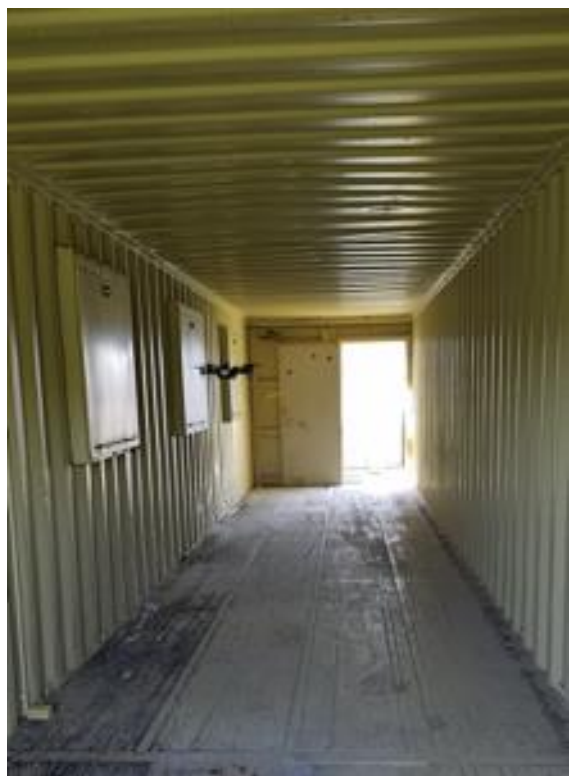
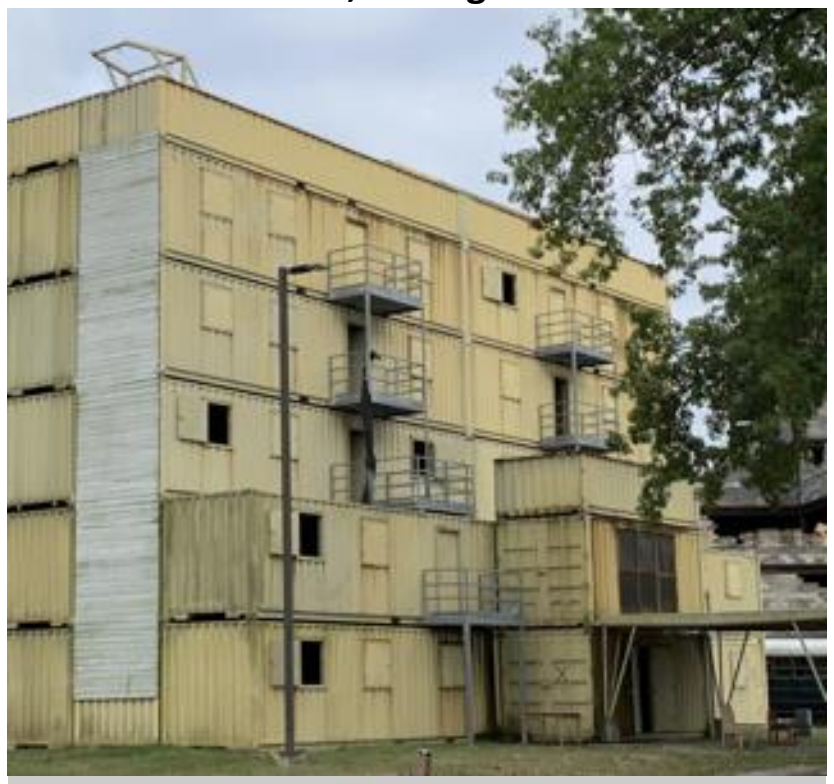

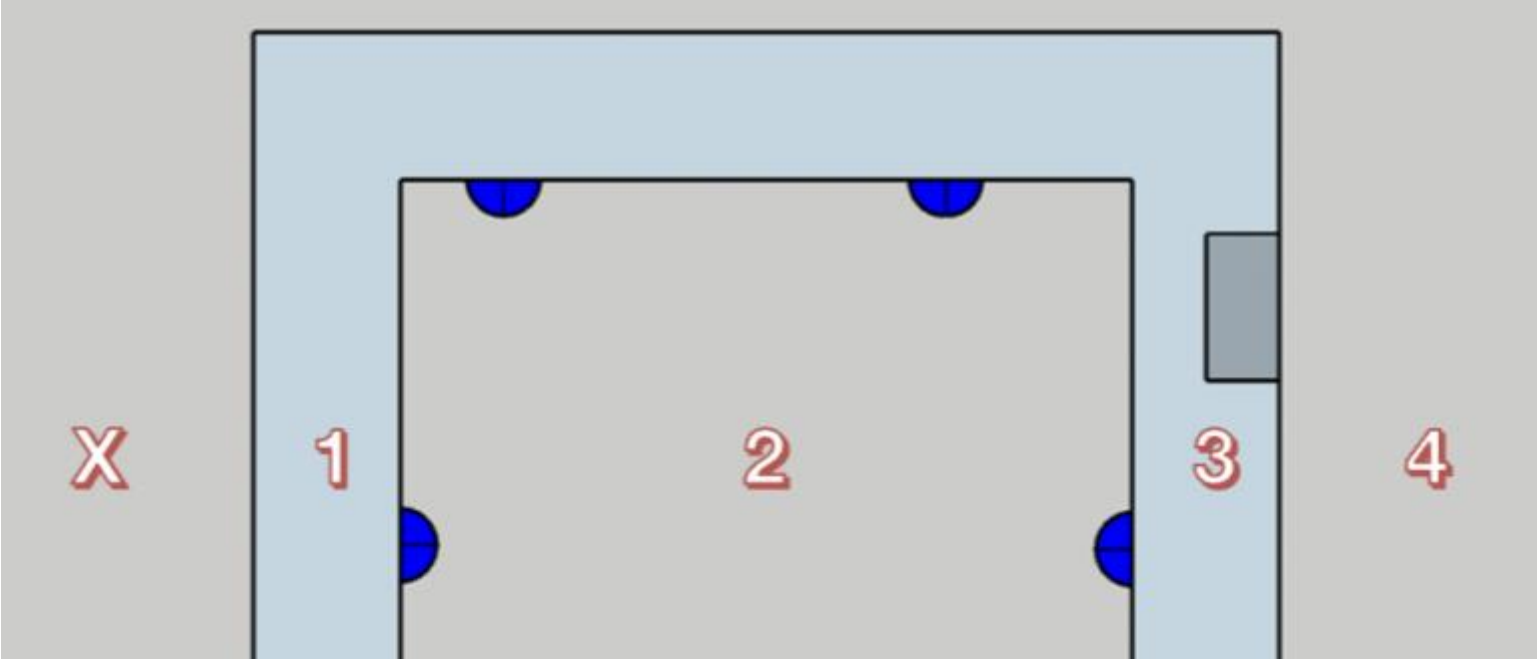

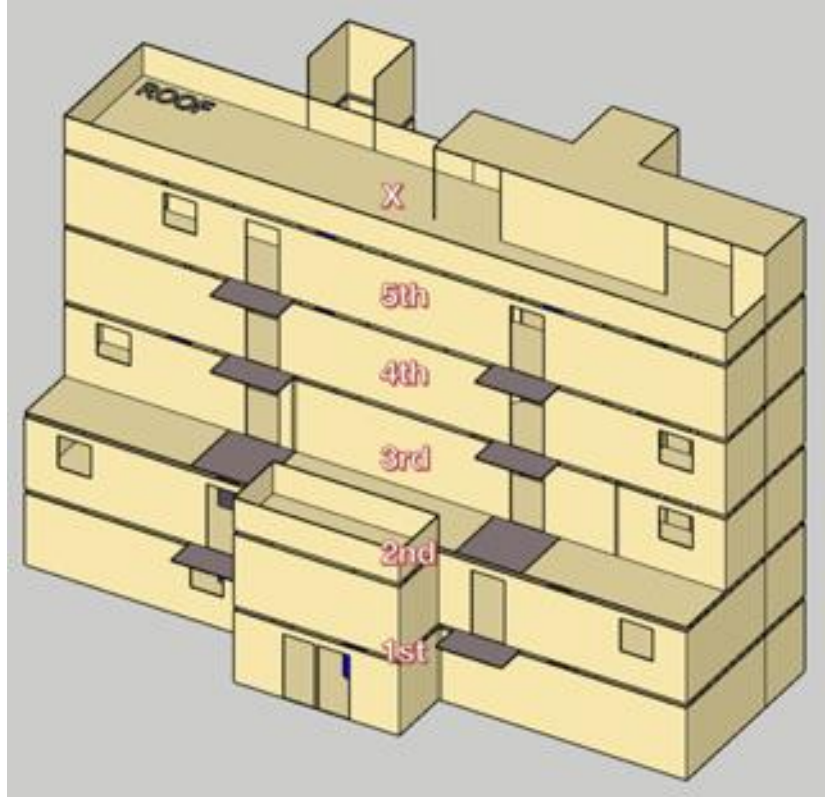

| Metrics | Horizontal, through walls | | | | Vertical, through floors | | | | |
|---|---|---|---|---|---|---|---|---|---|
| | X-1 | X-2 | X-3 | X-4 | X-1 | X-2 | X-3 | X-4 | X-5 |
| Position distance (m) | 6 | 13 | 20 | 25 | 3 | 6 | 8 | 11 | 14 |
| Position obstructions | 1 steel | 2 steel | 3 steel | 4 steel | 2 steel* | 4 steel* | 6 steel* | 8 steel* | 10 steel* |

*The building is made of stacked Conex containers, so there are 2 steel panels (floor and ceiling) between containers

## Performance data

Best in class = maximum NLOS performance through 3 walls or 4 floors (i.e., 1 less than the maximum number of walls or floors successfully transmitted through across all systems)

| | | Horizontal, through walls | | | | | Vertical, through floors | | | | | |
|---|---|---|---|---|---|---|---|---|---|---|---|---|
| **sUAS and communications frequency** | **Metrics** | **X** | **1** | **2** | **3** | **4** | **X** | **1** | **2** | **3** | **4** | **5** |
| Cleo Robotics Dronut X1P<br>2.4 GHz | Connect | - | - | - | - | - | ✓ | ✓ | ✓ | ✓ | X | X |
| | Fly | - | - | - | - | - | ✓ | ✓ | ✓ | ✓ | X | X |
| | Maximum NLOS performance | N/A | | | | | 8 m, 3 floors | | | | | |
| FLIR Black Hornet PRS<br>355-385 MHz | Connect | ✓ | ✓ | ✓ | ✓ | ✓ | ✓ | ✓ | ✓ | ✓ | ✓ | ✓ |
| | Fly | ✓ | ✓ | - | X | - | ✓ | ✓ | ✓ | ✓ | ✓ | ✓ |
| | Maximum NLOS performance | 6 m, 1 wall | | | | | 14 m, 5 floors | | | | | |
| Flyability Elios 2 GOV<br>2.4 GHz | Connect | ✓ | ✓ | ✓ | ✓ | ✓ | ✓ | ✓ | ✓ | ✓ | ✓ | ✓ |
| | Fly | ✓ | ✓ | ✓ | ✓ | ✓ | ✓ | ✓ | ✓ | ✓ | ✓ | ✓ |
| | Maximum NLOS performance | 25 m, 4 walls | | | | | 14 m, 5 floors | | | | | |
| Lumenier Nighthawk V3<br>2.4 GHz | Connect | ✓ | ✓ | ✓ | ✓ | ✓ | ✓ | ✓ | ✓ | ✓ | ✓ | ✓ |
| | Fly | ✓ | ✓ | ✓ | ✓ | ✓ | ✓ | ✓ | ✓ | ✓ | ✓ | ✓ |
| | Maximum NLOS performance | 25 m, 4 walls | | | | | 14 m, 5 floors | | | | | |
| Parrot ANAFI USA GOV<br>2.4 GHz | Connect | ✓ | ✓ | ✓ | ✓ | X | ✓ | ✓ | ✓ | ✓ | ✓ | ✓ |
| | Fly | ✓ | ✓ | ✓ | ✓ | X | ✓ | ✓ | ✓ | ✓ | ✓ | ✓ |
| | Maximum NLOS performance | 20 m, 3 walls | | | | | 14 m, 5 floors | | | | | |
| Skydio X2D*<br>1.8 GHz | Connect | ✓ | ✓ | ✓ | ✓ | ✓ | X | X | X | X | X | X |
| | Fly | ✓ | ✓ | ✓ | ✓ | ✓ | X | X | X | X | X | X |
| | Maximum NLOS performance | 25 m, 4 walls | | | | | N/A | | | | | |
| Teal Golden Eagle†<br>2.4 GHz | Connect | X | X | X | X | X | X | X | X | X | X | X |
| | Fly | X | X | X | X | X | X | X | X | X | X | X |
| | Maximum NLOS performance | N/A | | | | | N/A | | | | | |
| Vantage Robotics Vesper<br>1.8 GHz | Connect | ✓ | X | X | X | X | ✓ | X | X | X | X | X |
| | Fly | ✓ | X | X | X | X | ✓ | X | X | X | X | X |
| | Maximum NLOS performance | 0 | | | | | 0 | | | | | |
| | **Best in class:** | **Horizontal, through walls** | | | | | **Vertical, through floors** | | | | | |
| | | Flyability Elios 2 GOV<br>Lumenier Nighthawk V3<br>Parrot ANAFI USA GOV<br>Skydio X2D | | | | | FLIR Black Hornet PRS<br>Flyability Elios 2 GOV<br>Lumenier Nighthawk V3<br>Parrot ANAFI USA GOV | | | | | |
| | | **NLOS Communications (MUTC), overall** | | | | | | | | | | |
| | | Flyability Elios 2 GOV<br>Lumenier Nighthawk V3<br>Parrot ANAFI USA GOV | | | | | | | | | | |

*Note: The Skydio X2D was not able to takeoff inside of the hallways of the Conex building (horizontal positions 1 and 3) due to the ceiling height (2.4 m [8 ft]) and lateral distance to the walls (1.2 m [4 ft]), causing it to crash when takeoff was attempted. However, the system could still operate in this environment if flown it from the outside, so successful performance can still be claimed. Due to this issue, it was not attempted to be flown in the vertical test, all of which took place inside of Conex containers of the same measurements.

†Note: Due to instability of the Teal Golden Eagle when flying indoors, we did not attempt to confirm its communications performance at each location.

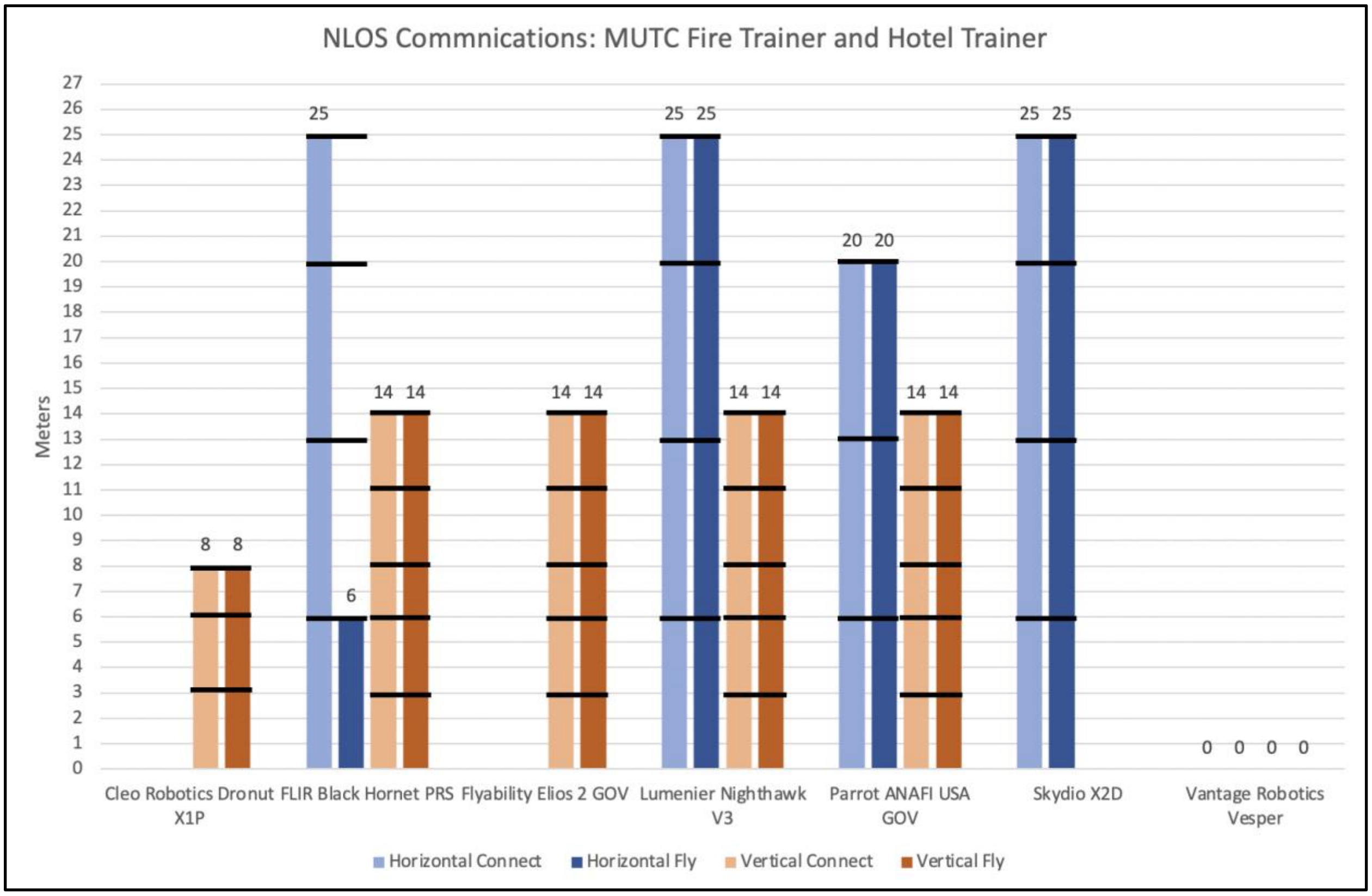

# Non-Line-of-Sight (NLOS) Video Latency

## Summary of Test Method

This test method is an expansion of an existing test method currently under development by NIST for standardization through the ASTM E54.09 Committee on Homeland Security; Subcommittee on Response Robots. In that test method, a flashing light is placed within view of the sUAS and an external camera is used to record the flashing light and the OCU display of the flashing light as seen by the sUAS camera in the same view. The sUAS and light are positioned further and further apart from the OCU while still maintaining that the light and OCU screen are visible in the external camera view to evaluate the impact of range on video latency. The external camera records while the light flashes several times. The video is then exported and the amount of delay between when the light actually flashes compared to when it is seen flashing on the OCU screen is calculated by counting video frames and converting to milliseconds (based on the frames per second of the recorded video).

This test method adapts the existing method for NLOS operations by instead using two synchronized stopwatches (with millisecond displays) rather than camera flashing lights, which will move between the different rooms and floors that separate the sUAS and the OCU. An external video camera captures one of the stopwatches and the OCU display in a single frame. Once the watches are synchronized, the sUAS and other stopwatch are moved into position, pointing the sUAS camera at the stopwatch such that both stopwatches can be seen in the external camera view back at the starting point. After all positions are completed, the video from that camera is then exported and evaluated the same as previously described (i.e., counting time difference between the two).

This test method can be run concurrently with the NLOS Communications test method.

## Environment characterization

**Horizontal, through walls** | **Vertical, through floors**

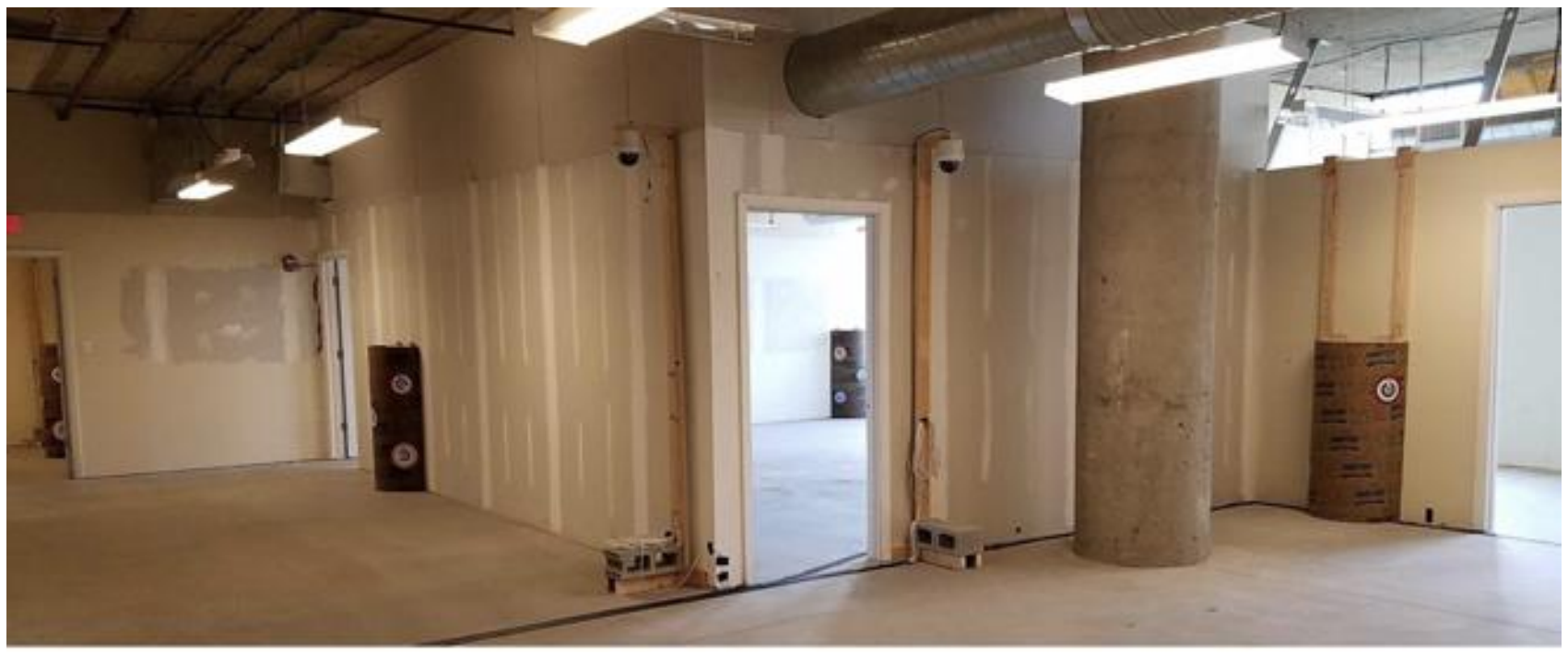

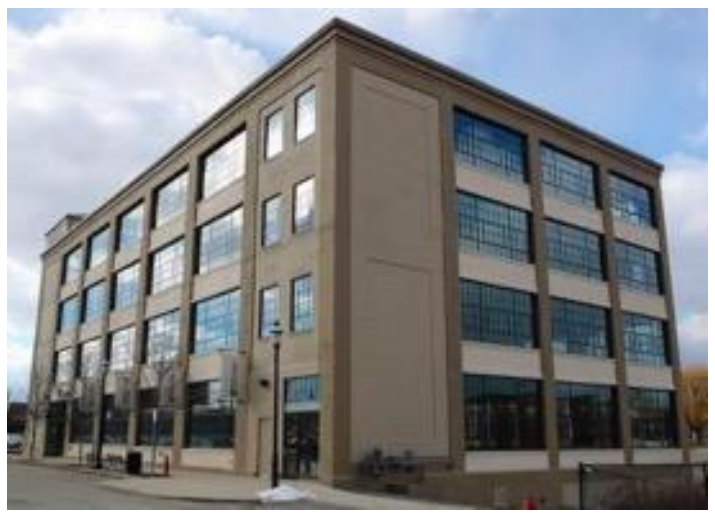

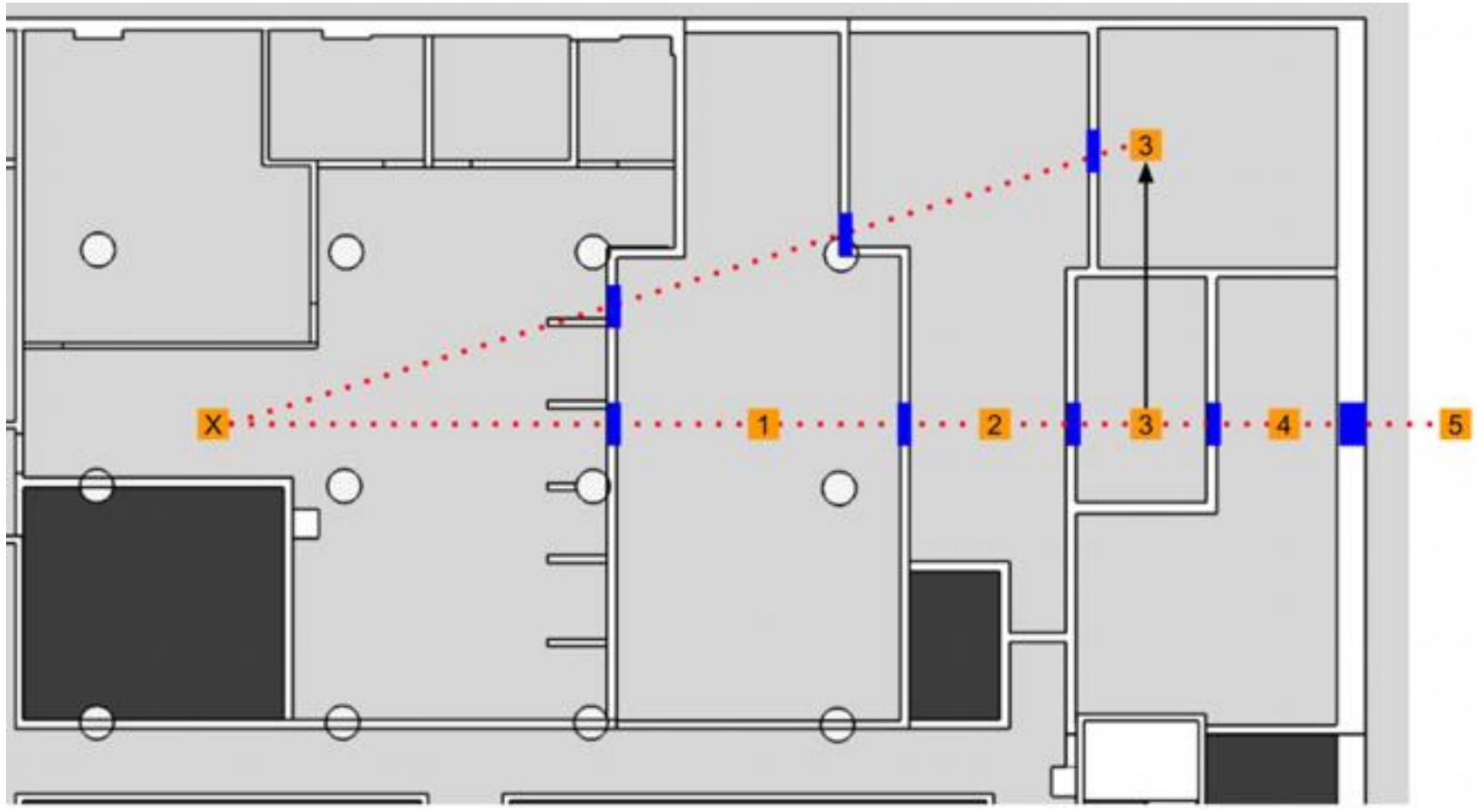

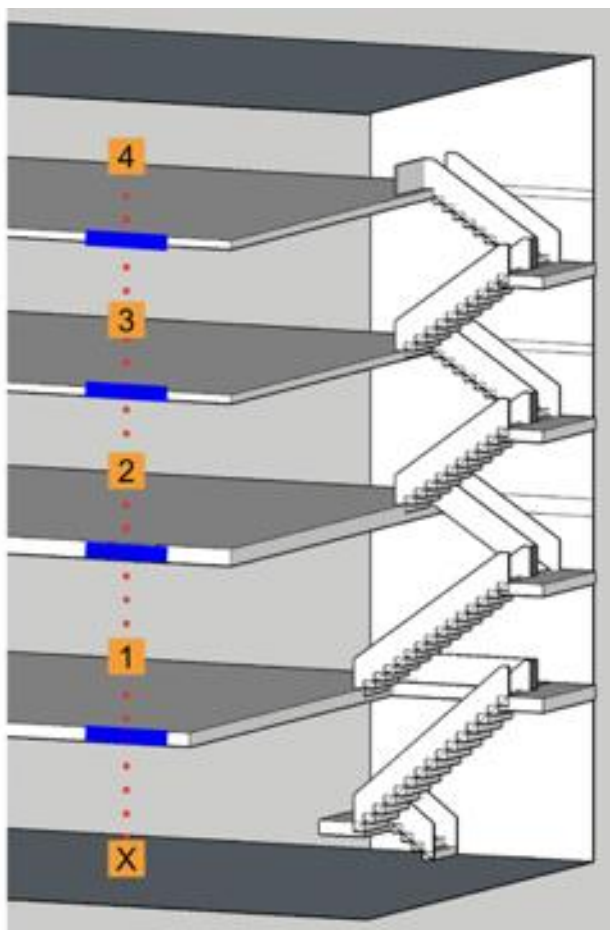

| Metrics | Horizontal, through walls | | | | | Vertical, through floors | | | |
|---|---|---|---|---|---|---|---|---|---|
| | X-1 | X-2 | X-3 | X-4 | X-5 | X-1 | X-2 | X-3 | X-4 |
| Position distance (m) | 14 | 20 | 25 | 27 | 31 | 5 | 9 | 13 | 17 |
| Position obstructions | 1 drywall | 2 drywall | 3 drywall | 4 drywall | 4 drywall<br>1 concrete | 1 concrete | 2 concrete | 3 concrete | 4 concrete |

## Performance data

Best in class = maximum NLOS latency through 4 walls or 2 floors (i.e., 1 less than the maximum number of walls or floors successfully transmitted through across all systems) is within 2 standard deviations of the average latency across that all of that system's measurements in the same direction (i.e., horizontal or vertical)

| sUAS, communications frequency, and OCU signal indication | Metrics (ms) | Horizontal, through walls | | | | | | Vertical, through floors | | | | |
|---|---|---|---|---|---|---|---|---|---|---|---|---|
| | | X | 1 | 2 | 3 | 4 | 5 | X | 1 | 2 | 3 | 4 |
| Cleo Robotics Dronut X1P<br>2.4 GHz | Latency | 200 | 200 | 200 | X | X | X | 200 | X | X | X | X |
| | Maximum NLOS latency | 200 ms<br>20 m, 2 walls | | | | | | 200 ms<br>0 | | | | |
| FLIR Black Hornet PRS<br>355 - 385 MHz | Latency | 200 | 200 | 200 | 300 | 300 | 500 | 200 | 300 | 300 | 2500 | X |
| | Maximum NLOS latency | 500 ms<br>31 m, 5 walls | | | | | | 2500 ms<br>13 m, 3 floors | | | | |
| Flyability Elios 2 GOV<br>2.4 GHz | Latency | 400 | 400 | 400 | 400 | 400 | X | 400 | 400 | X | X | X |
| | Maximum NLOS latency | 400 ms<br>27 m, 4 walls | | | | | | 400 ms<br>5 m, 1 floor | | | | |
| Lumenier Nighthawk V3<br>2.4 GHz | Latency | 200 | 200 | 200 | 200 | 200 | 200 | 200 | 200 | X | X | X |
| | Maximum NLOS latency | 200 ms<br>31 m, 5 walls | | | | | | 200 ms<br>5 m, 1 floor | | | | |
| Parrot ANAFI USA GOV<br>2.4 GHz | Latency | 300 | 300 | 300 | 500 | 300 | X | 300 | 300 | X | X | X |
| | Maximum NLOS latency | 300 ms<br>27 m, 4 walls | | | | | | 300 ms<br>5 m, 1 floor | | | | |
| Skydio X2D<br>1.8 GHz | Latency | 180 | 180 | 180 | 180 | 180 | 7980 | 180 | 180 | X | X | X |
| | Maximum NLOS latency | 7980 ms<br>31 m, 5 walls | | | | | | 180 ms<br>5 m, 1 floor | | | | |
| Teal Golden Eagle<br>2.4 GHz | Latency | 300 | 400 | 400 | 400 | 400 | X | 300 | 400 | 2200 | X | X |
| | Maximum NLOS latency | 400 ms<br>27 m, 4 walls | | | | | | 2200 ms<br>9 m, 2 floors | | | | |
| Vantage Robotics Vesper*<br>1.8 GHz | Latency | - | - | - | - | - | - | - | - | - | - | - |
| | Maximum NLOS latency | - | | | | | | - | | | | |
| | **Best in class:** | **Horizontal, through walls** | | | | | | **Vertical, through floors** | | | | |
| | | FLIR Black Hornet PRS<br>Flyability Elios 2 GOV<br>Lumenier Nighthawk V3<br>Parrot ANAFI USA GOV<br>Skydio X2D<br>Teal Golden Eagle | | | | | | FLIR Black Hornet PRS<br>Teal Golden Eagle | | | | |

*Note: The Vantage Robotics Vesper was not available for testing due to being with the vendor for repairs.

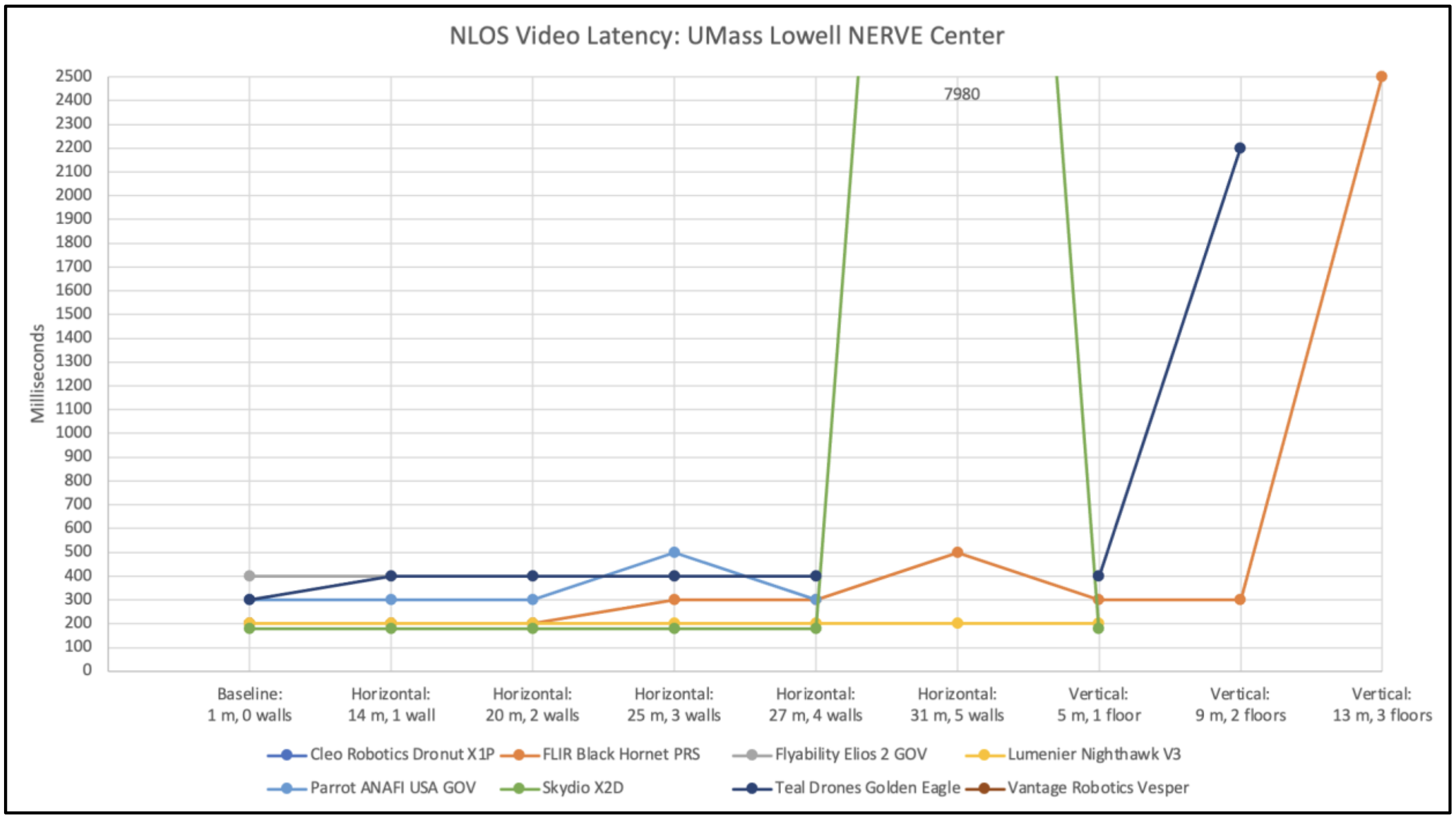
NLOS Video Latency: UMass Lowell NERVE Center
Milliseconds
7980
Baseline: 1 m, 0 walls
Horizontal: 14 m, 1 wall
Horizontal: 20 m, 2 walls
Horizontal: 25 m, 3 walls
Horizontal: 27 m, 4 walls
Horizontal: 31 m, 5 walls
Vertical: 5 m, 1 floor
Vertical: 9 m, 2 floors
Vertical: 13 m, 3 floors
Cleo Robotics Dronut X1P
FLIR Black Hornet PRS
Flyability Elios 2 GOV
Lumenier Nighthawk V3
Parrot ANAFI USA GOV
Skydio X2D
Teal Drones Golden Eagle
Vantage Robotics Vesper

## Interference Reaction

### Summary of Test Method

The test consists of generating an interfering radio-frequency signal whose frequency will fall within the sUAS communication camera or control channels (i.e., jamming its communication channel). The possible outcomes of sUAS behavior once jammed include exhibiting lost or degraded communication functionality (e.g., landing, return to home), automatic channel hopping to deconflict with the interfering signal, or inability to reconnect after interference has ceased (i.e., sUAS needs to be restarted before connection is regained). There are multiple types of interference tests that are performed (note that each test serves as a prerequisite for running the subsequent tests; e.g., run the Hovering test before running the Command Input test):

Frequency Characterization: Before the sUAS signal can be interfered with, a receiver antenna connected to a spectrum analyzer can be used to determine exactly which frequencies are being used by the sUAS to operate.

Grounded Interference: While the sUAS is grounded, transmit the interfering signal and attempt to take off.

Hovering Interference: While the sUAS is hovering, transmit the interfering signal and attempt to continue hovering in place, yaw, pitch forward and back, roll left and right, ascend and descend, move the camera, and then land.

Command Input Interference: Command the sUAS to continuously yaw in place while hovering, then proceed to transmit the interfering signal. Note whether the sUAS either continues to turn or stops if and when control is lost.

Note: running these tests may severely degrade existing WiFi networks in the area where testing is conducted.

## Environment characterization

| NERVE Basement | | | |
|---|---|---|---|
| **Lighting** | Well Lit | **sUAS to Transmitter Distance** | 2 m (6.6 ft) |
| **Walls** | Concrete, Drywall | **sUAS to Receive Distance** | 2 m (6.6 ft) |
| **Floor** | Concrete | **sUAS to OCU Distance** | 8 m (26.2 ft) |
| **Type** | Indoor | **Transmitter to Receiver Distance** | 0.7 m (2.3 ft) |
| **Image** | 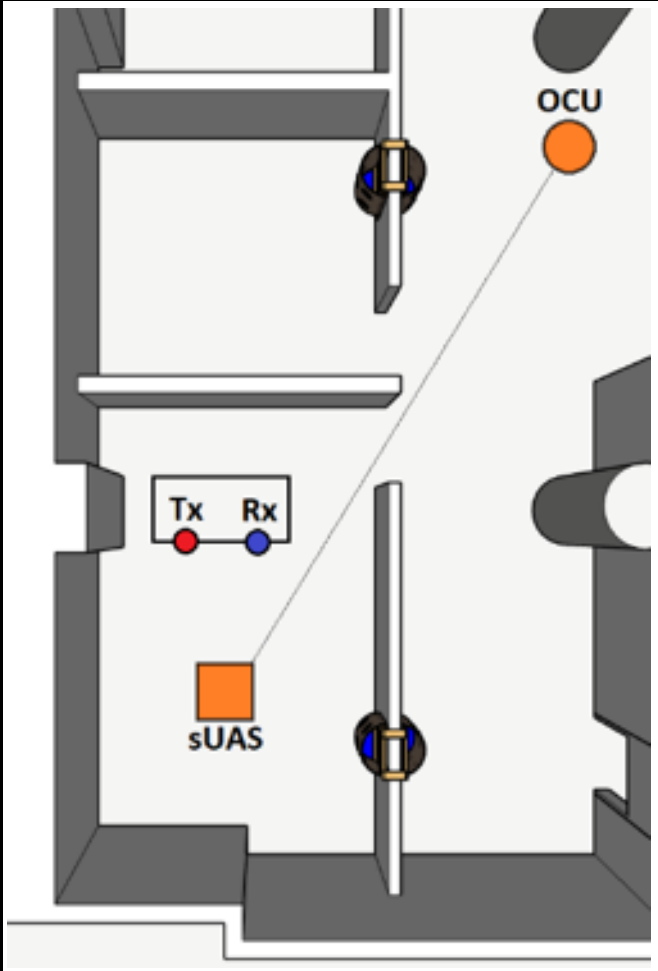  | 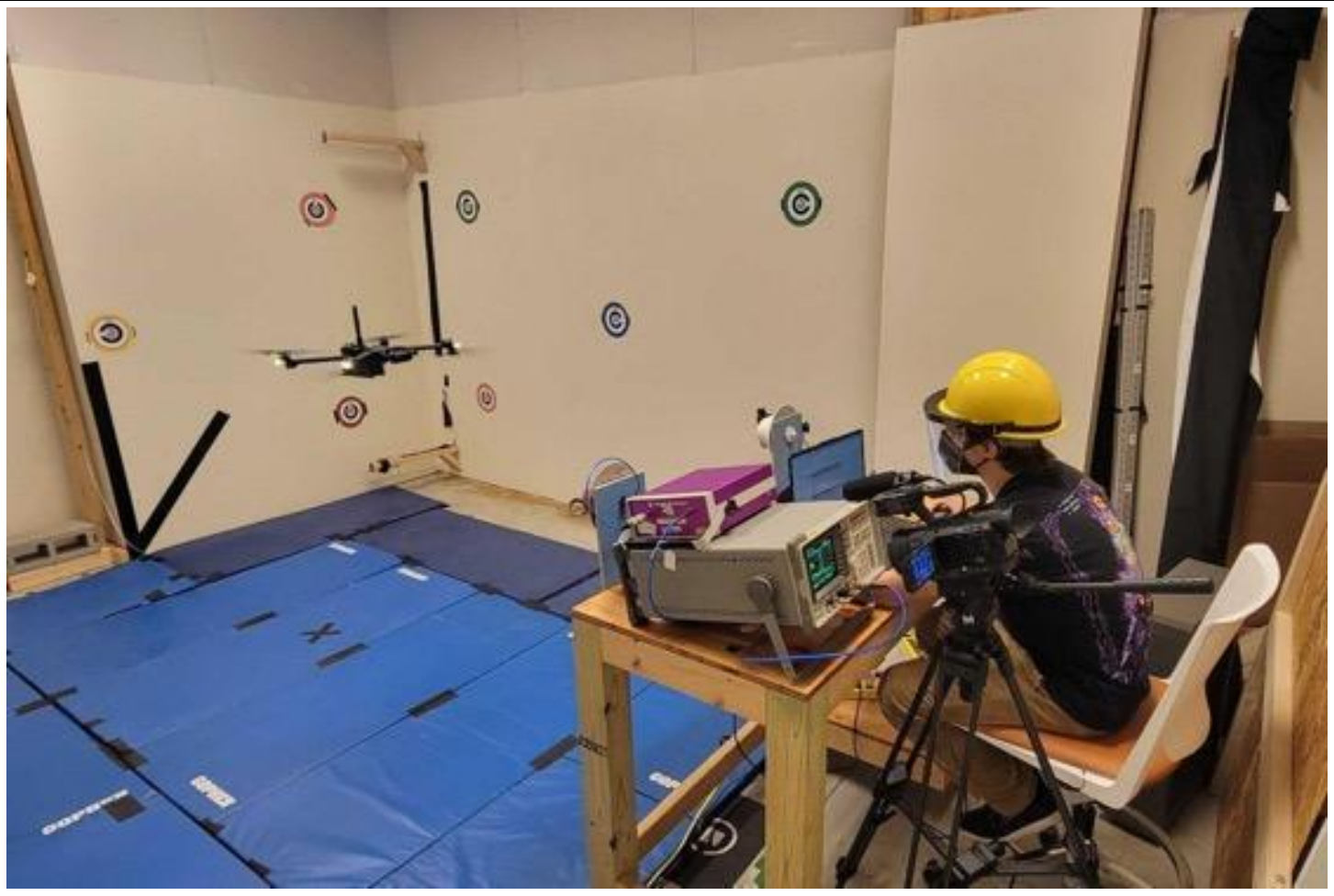 | |

## Interference characterization

| Equipment Used | Purpose | Range |
|---|---|---|
| [2x] CommScope Andrew CELLMAX-O-CPUSEI | Signal receiver and transmitter | 698-960 MHz and 1710-2700 MHz |
| HP 8596E Spectrum Analyzer | Spectrum analysis, data collection | 9 kHz - 12.8 GHz |
| Stealth Microwave SM0825-36HR | Amplify receiver antenna, send to analyzer | 800-2500 MHz |
| ADALM PLUTO SDR | RF interference signal generation | 325-3800 MHz |
| Mini Circuits ZHL-42W+ Amplifier | Amplify the transmitted signal from SDR | 10-4200 MHz |
| HP 6214A Power Supply | Power the ZHL amplifier | N/A |

| Low Power Signal | High Power Signal |
|---|---|
| -25.61 dBm | 13.00 dBm (85x stronger) |
| 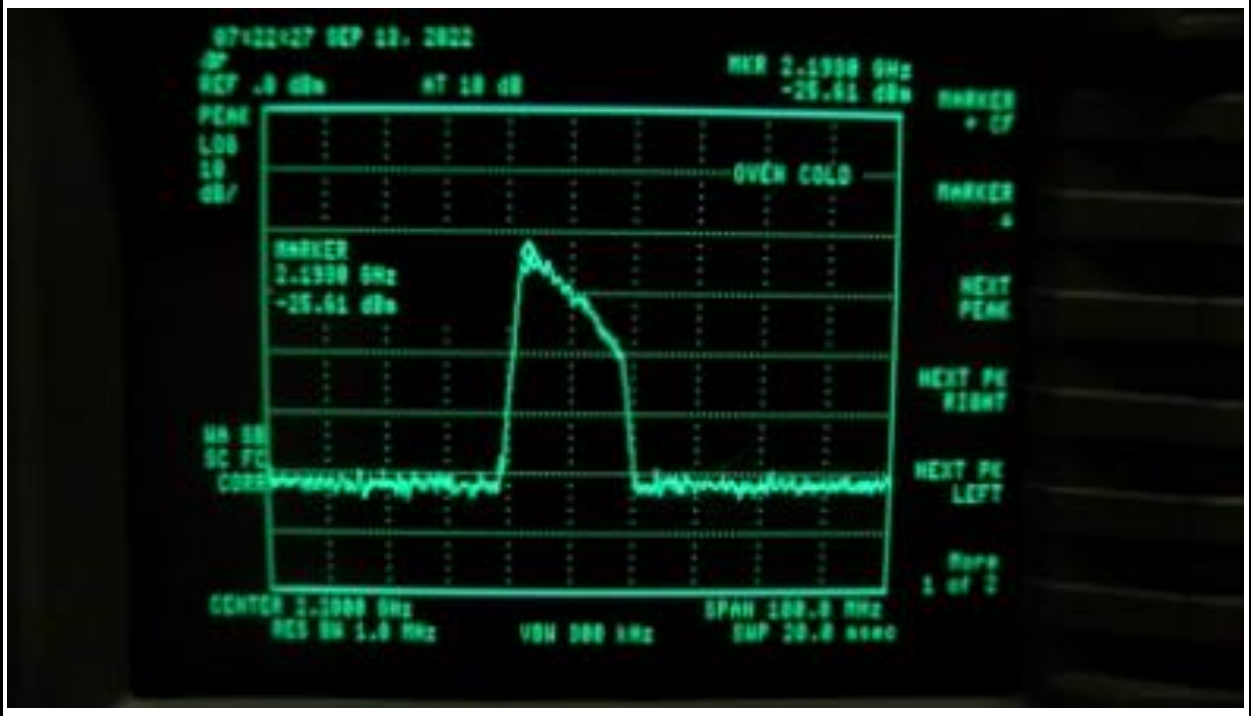 | 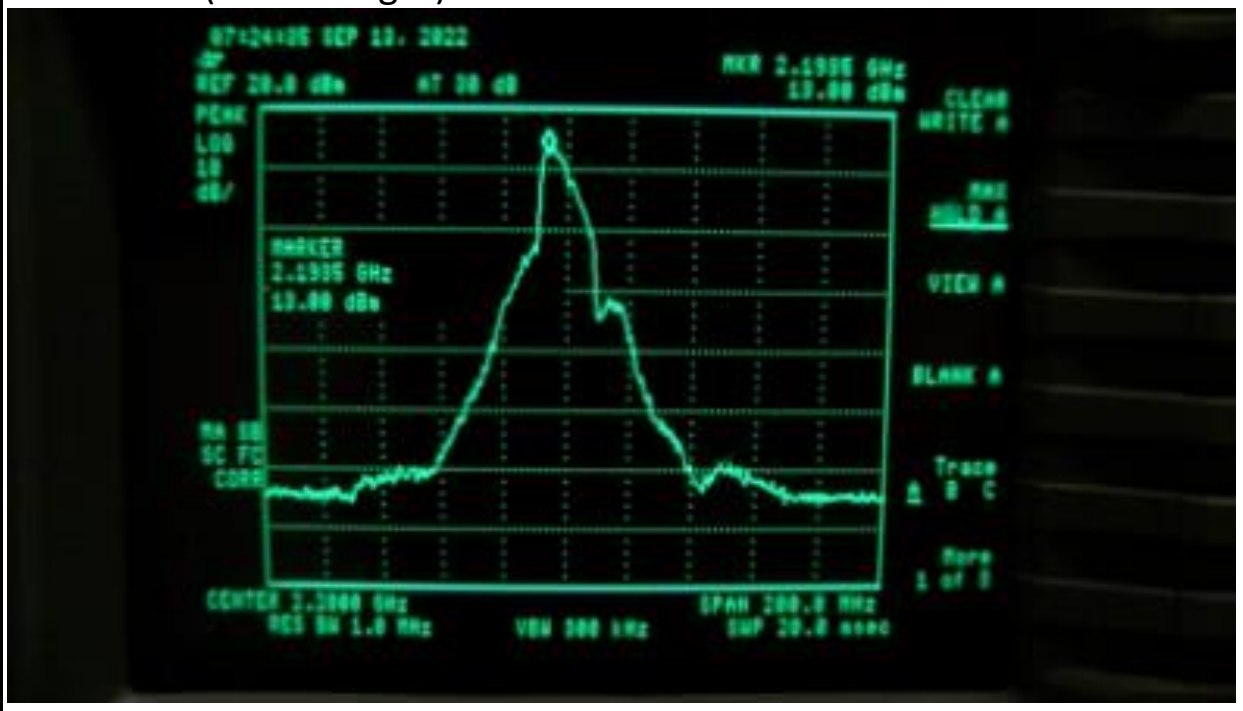 |
| | (note: frequency span is twice that of the low power image) |

## Performance data

No best in class criteria is specified.

| sUAS and Target Frequency | | Ambient Signal | | | sUAS Signal | | |
|---|---|---|---|---|---|---|---|
| Cleo Robotics Dronut X1P 2.435 GHz | **Image** | 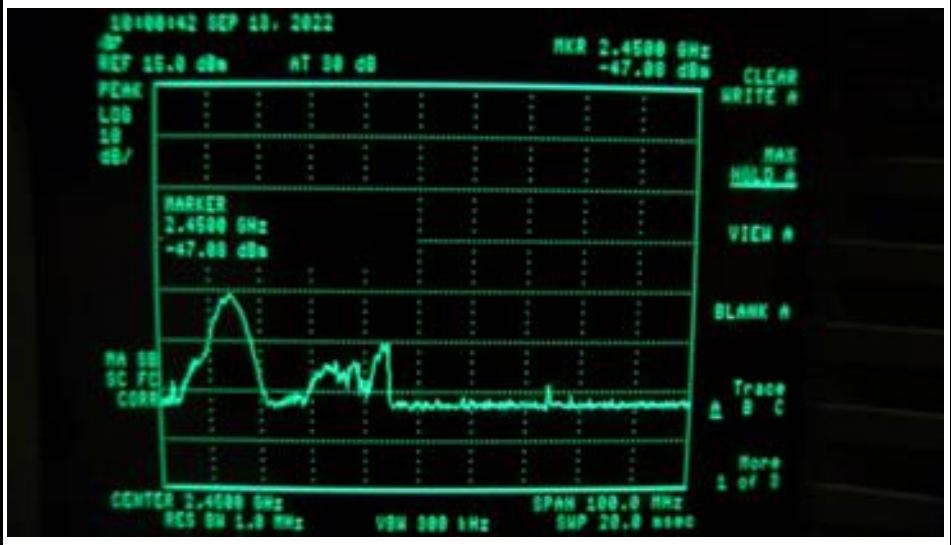 | | | 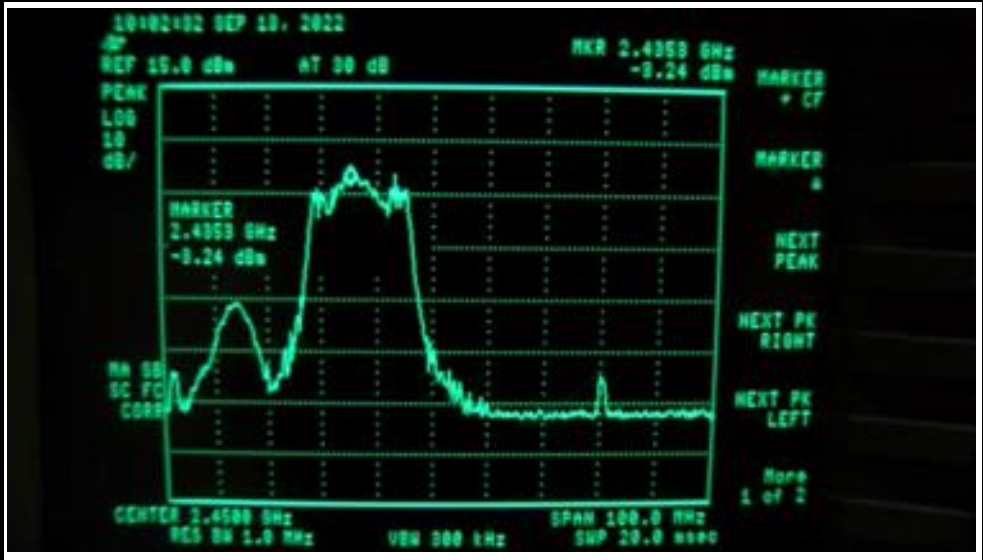 | | |
| | **Freq Start (GHz)** | 2.4 | | | 2.4 | | |
| | **Freq End (GHz)** | 2.5 | | | 2.5 | | |
| | **Peaks (GHz)** | - | | | 2.435 | | |
| | | **sUAS with Low Power Interference** | | | **sUAS with High Power Interference** | | |
| | **Image** | 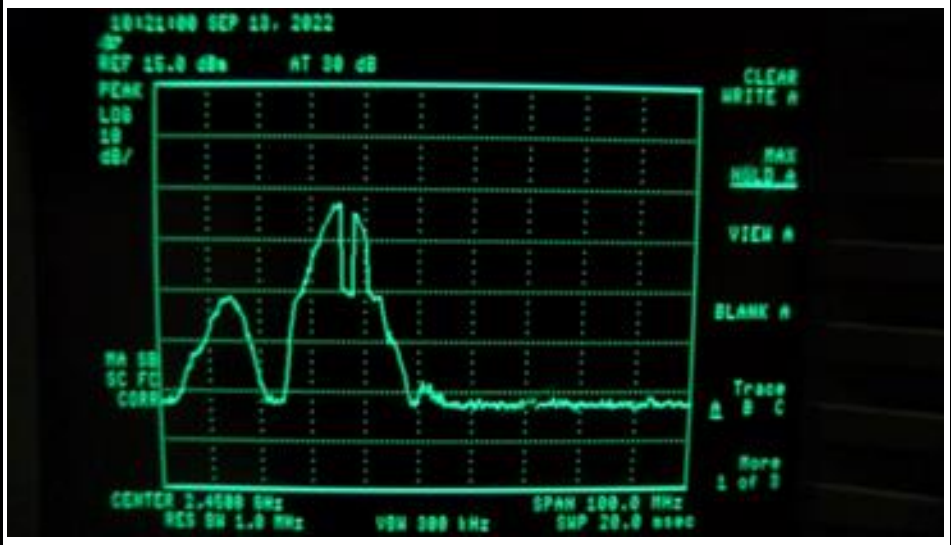 | | | 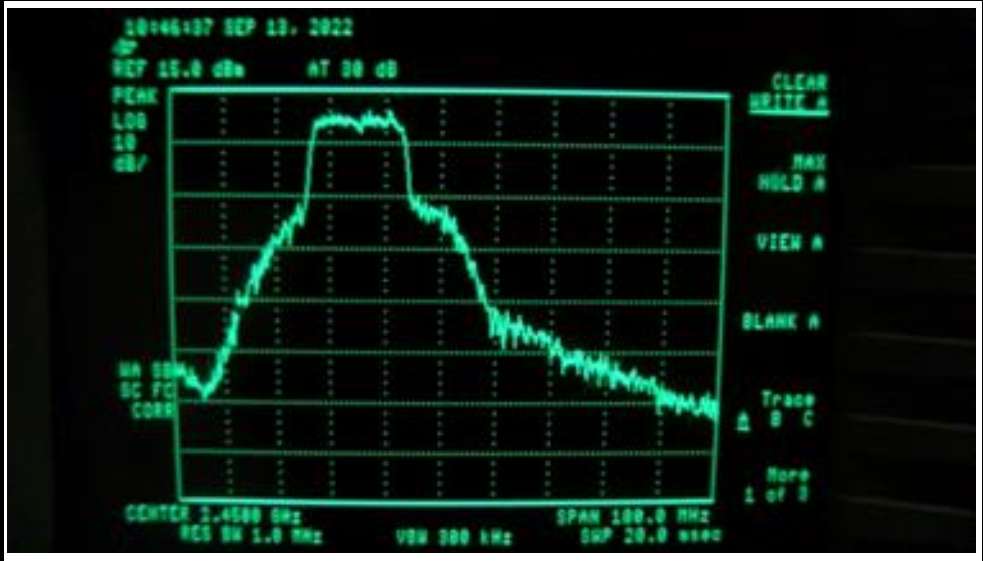 | | |
| | **Freq Start (GHz)** | 2.4 | | | 2.4 | | |
| | **Freq End (GHz)** | 2.5 | | | 2.5 | | |
| | **Peaks (GHz)** | - | | | - | | |
| | **Metrics** | **Grounded** | **Hover** | **Yaw in Place** | **Grounded** | **Hover** | **Yaw in Place** |
| | **Takeoff** | X | - | - | X | - | - |
| | **Video Link** | X | ✓ | // | X | X | X |
| | **Flight Control** | X | ✓ | ✓ | X | X | X |
| | **Auto Land** | - | X | X | - | ✓ | ✓ |
| | **Channel Hop** | X | X | X | X | X | X |
| | **Input Persist** | - | - | N/A | - | - | ✓ |

| sUAS and Target Frequency | | Ambient Signal | | | sUAS Signal | | |
|---|---|---|---|---|---|---|---|
| FLIR Black Hornet PRS<br><br>364 MHz | Image | 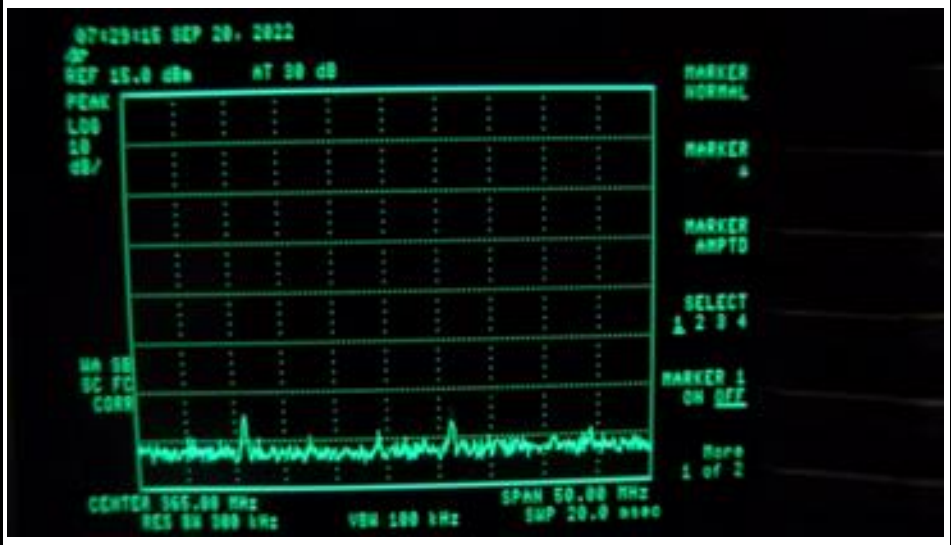 | | | 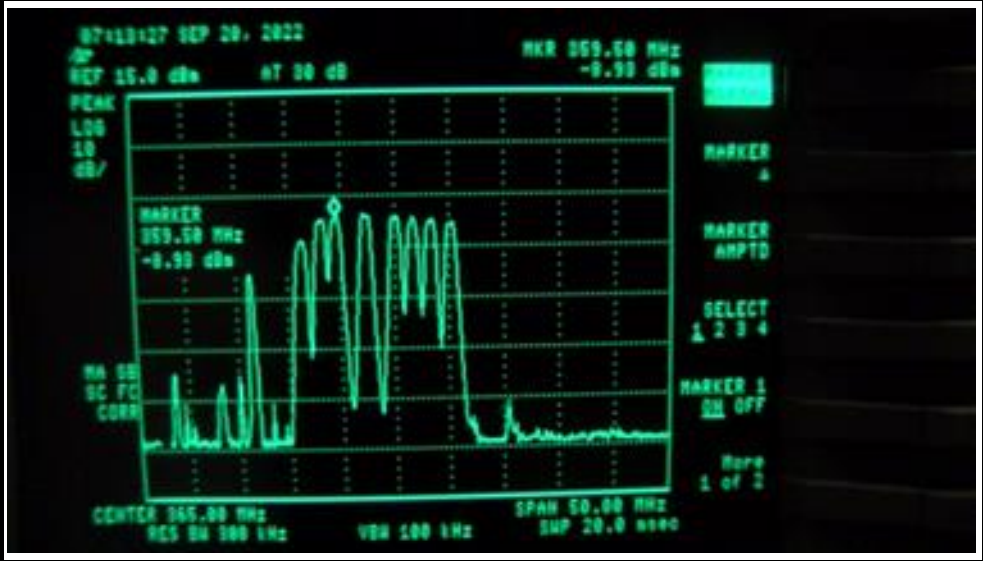 | | |
| | Freq Start (MHz) | 340 | | | 340 | | |
| | Freq End (MHz) | 390 | | | 390 | | |
| | Peaks (MHz) | - | | | 355 - 372 | | |
| | | sUAS with Low Power Interference | | | sUAS with High Power Interference | | |
| | Image | 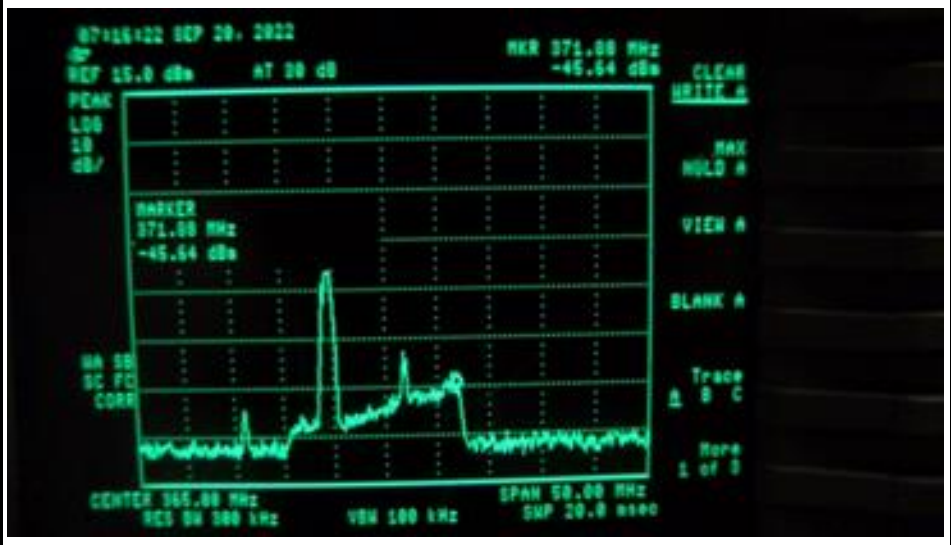 | | | 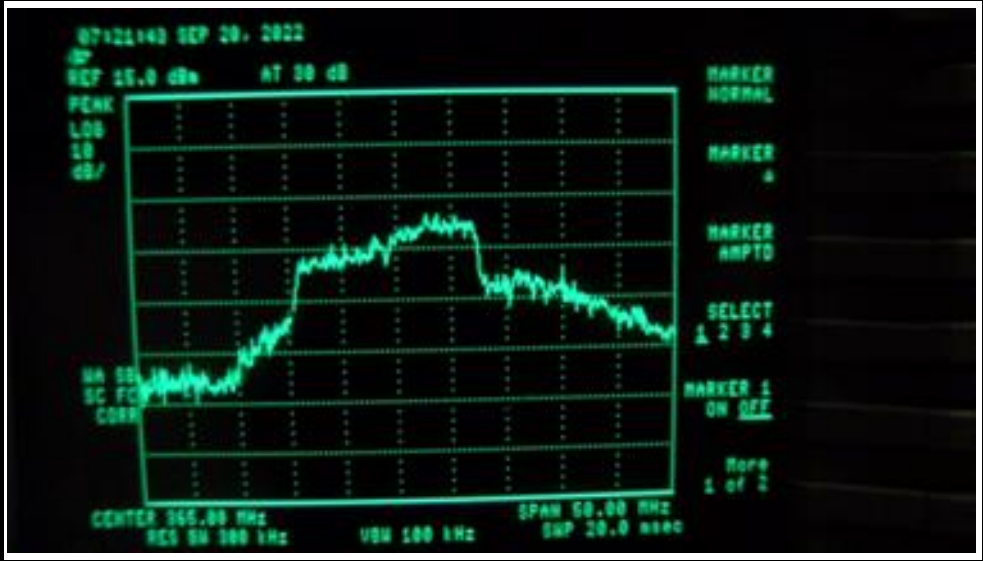 | | |
| | Freq Start (MHz) | 340 | | | 340 | | |
| | Freq End (MHz) | 390 | | | 390 | | |
| | Peaks (MHz) | - | | | - | | |
| | Metrics | Grounded | Hover | Yaw in Place | Grounded | Hover | Yaw in Place |
| | Takeoff | ✓ | - | - | ✓ | - | - |
| | Video Link | ✓ | ✓ | ✓ | ✓ | X | X |
| | Flight Control | ✓ | ✓ | ✓ | // | X | X |
| | Auto Land | - | X | X | - | ✓ | ✓ |
| | Channel Hop | ✓ | ✓ | ✓ | ✓ | ✓ | ✓ |
| | Input Persist | - | - | N/A | - | - | X |

| sUAS and Target Frequency | | Ambient Signal | | | sUAS Signal | | |
|---|---|---|---|---|---|---|---|
| Flyability Elios 2 GOV<br><br>2.457 GHz | Image | 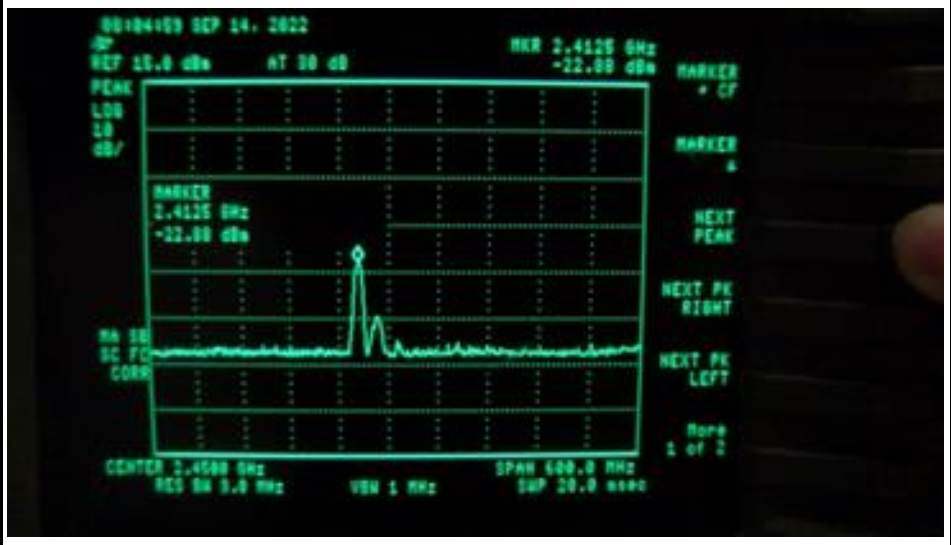 | | | 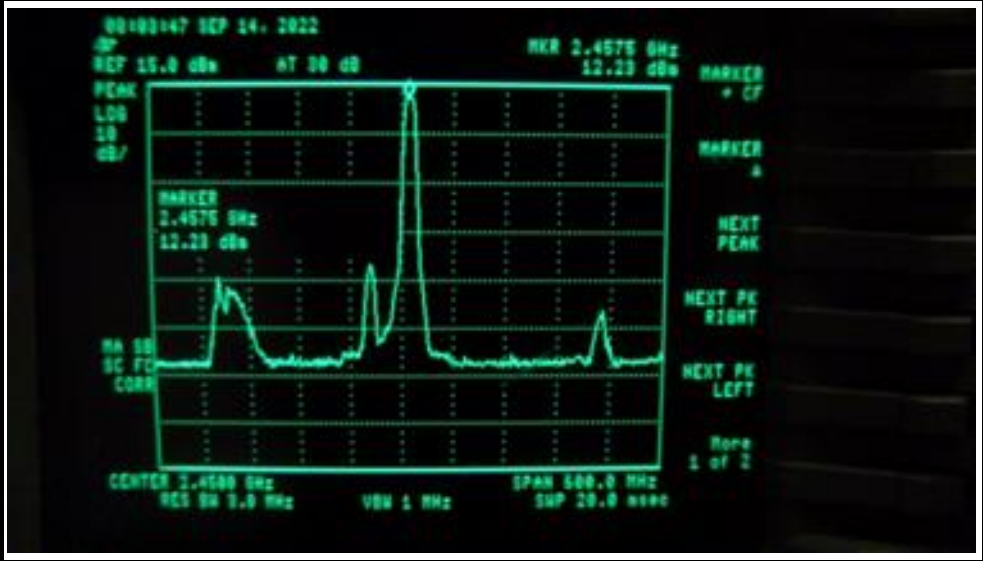 | | |
| | Freq Start (GHz) | 2.15 | | | 2.15 | | |
| | Freq End (GHz) | 2.75 | | | 2.75 | | |
| | Peaks (GHz) | 2.4125 | | | 2.231, 2.4575, 2.681 | | |
| | | sUAS with Low Power Interference | | | sUAS with High Power Interference | | |
| | Image | 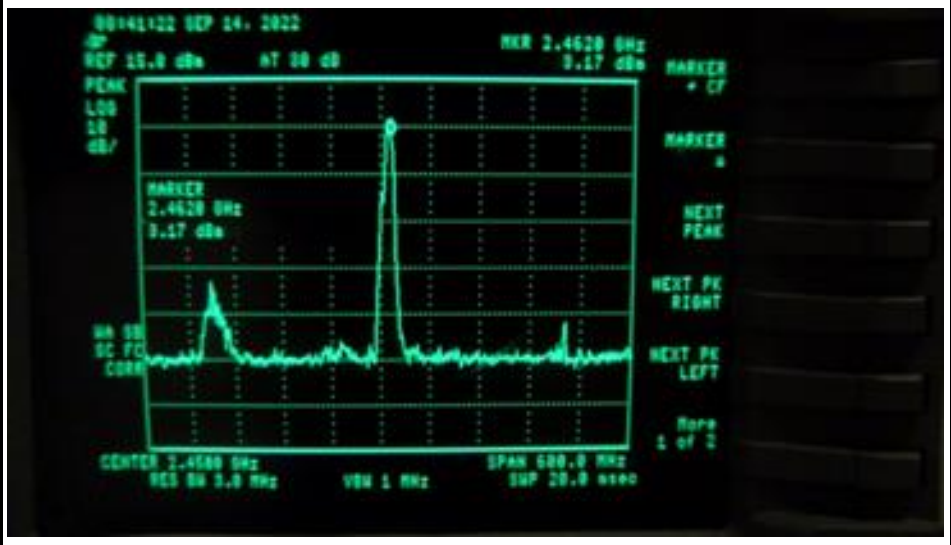 | | | 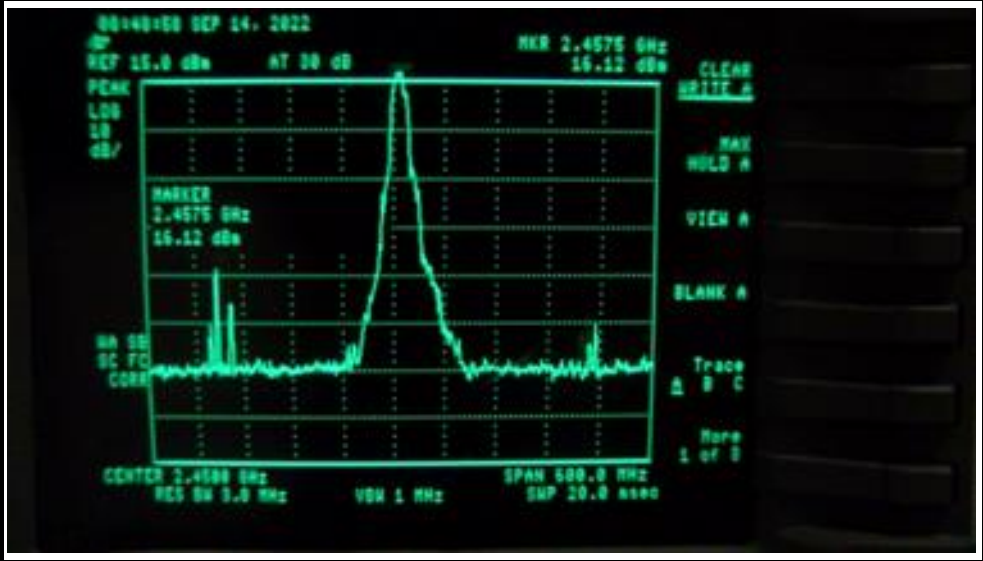 | | |
| | Freq Start (GHz) | 2.15 | | | 2.15 | | |
| | Freq End (GHz) | 2.75 | | | 2.75 | | |
| | Peaks (GHz) | - | | | - | | |
| | Metrics | Grounded | Hover | Yaw in Place | Grounded | Hover | Yaw in Place |
| | Takeoff | ✓ | - | - | X | - | - |
| | Video Link | X | X | X | X | X | X |
| | Flight Control | ✓ | ✓ | ✓ | X | X | X |
| | Auto Land | - | X | X | - | ✓ | ✓ |
| | Channel Hop | X | X | X | X | X | X |
| | Input Persist | - | - | N/A | - | - | X |

| sUAS and Target Frequency | | Ambient Signal | | | sUAS Signal | | |
|---|---|---|---|---|---|---|---|
| Lumenier Nighthawk V3<br><br>2.443 GHz | Image | 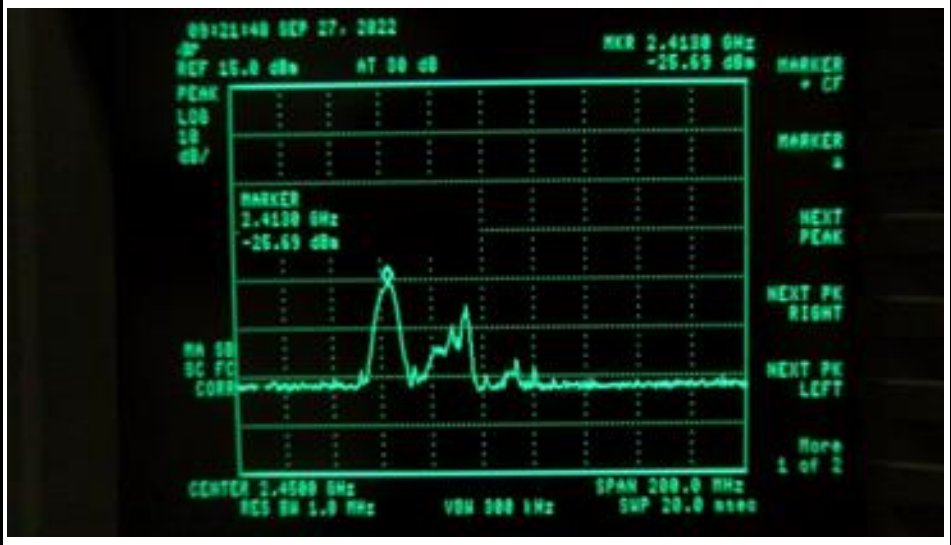 | | | 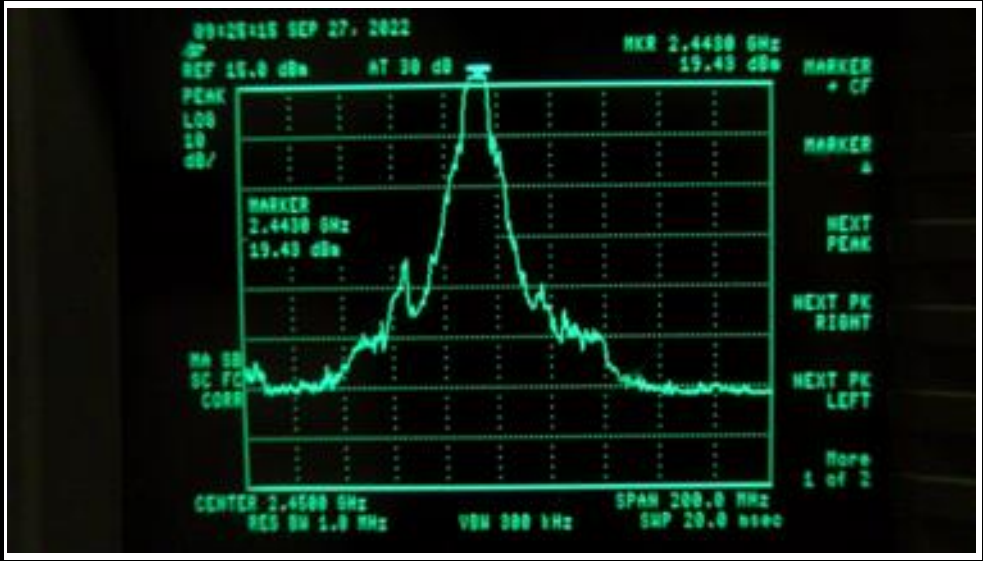 | | |
| | Freq Start (GHz) | 2.35 | | | 2.35 | | |
| | Freq End (GHz) | 2.55 | | | 2.55 | | |
| | Peaks (GHz) | 2.413 | | | 2.443 | | |
| | | **sUAS with Low Power Interference** | | | **sUAS with High Power Interference** | | |
| | Image | 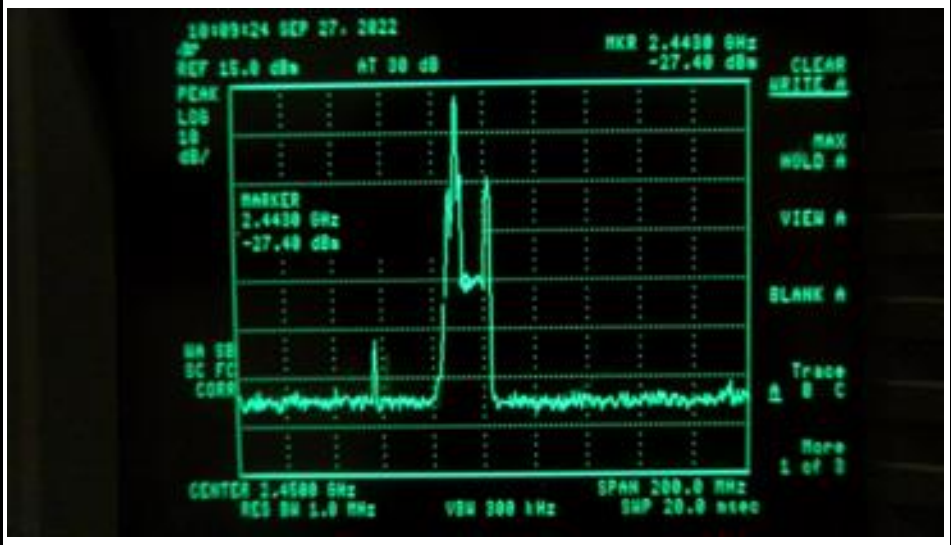 | | | 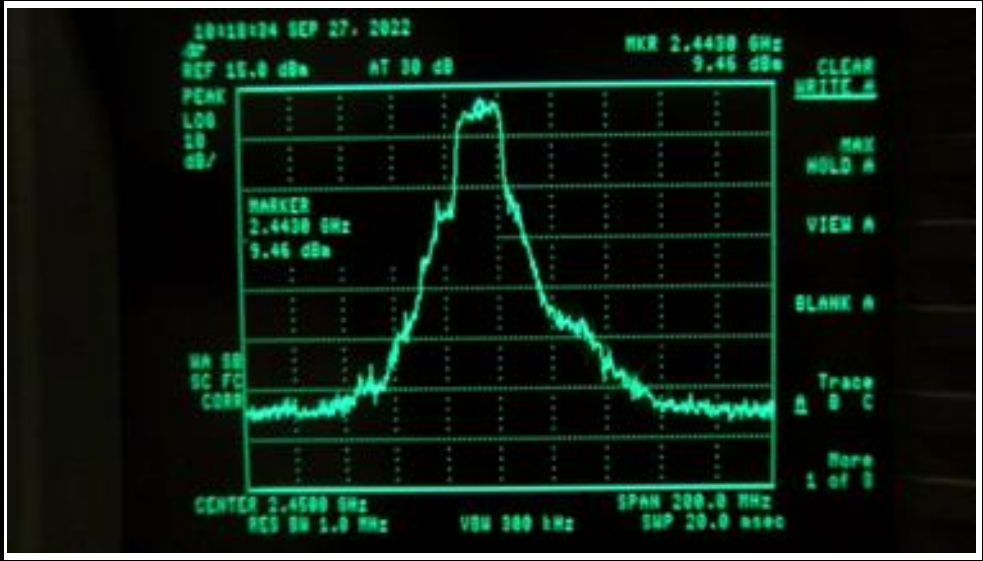 | | |
| | Freq Start (GHz) | 2.35 | | | 2.35 | | |
| | Freq End (GHz) | 2.55 | | | 2.55 | | |
| | Peaks (GHz) | - | | | - | | |
| | **Metrics** | **Grounded** | **Hover** | **Yaw in Place** | **Grounded** | **Hover** | **Yaw in Place** |
| | Takeoff | ✓ | - | - | X | - | - |
| | Video Link | ✓ | ✓ | ✓ | X | X | X |
| | Flight Control | ✓ | ✓ | ✓ | X | X | X |
| | Auto Land | - | X | X | - | ✓ | ✓ |
| | Channel Hop | X | X | X | X | X | X |
| | Input Persist | - | - | ✓ | - | - | X |

| sUAS and Target Frequency | | Ambient Signal | | | sUAS Signal | | |
|---|---|---|---|---|---|---|---|
| Parrot ANAFI USA GOV<br><br>2.455 GHz | Image | 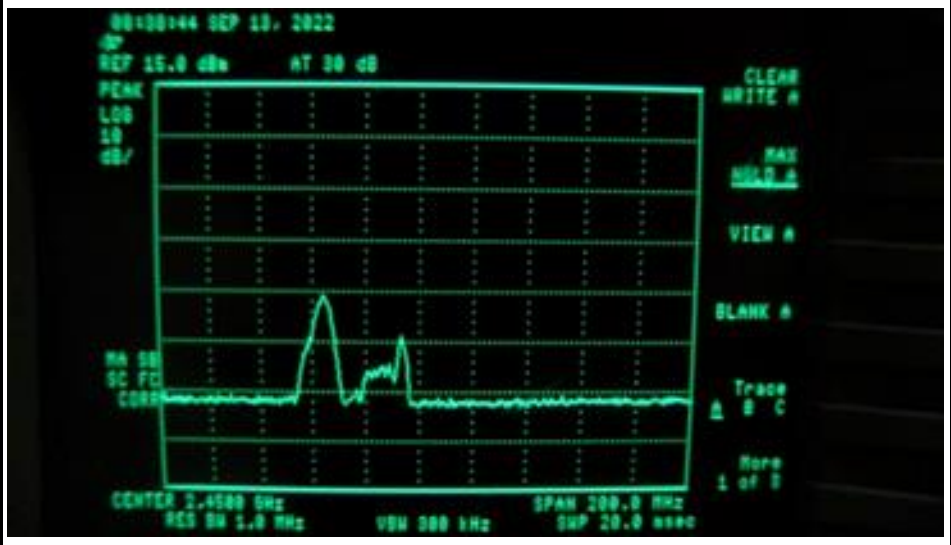 | | | 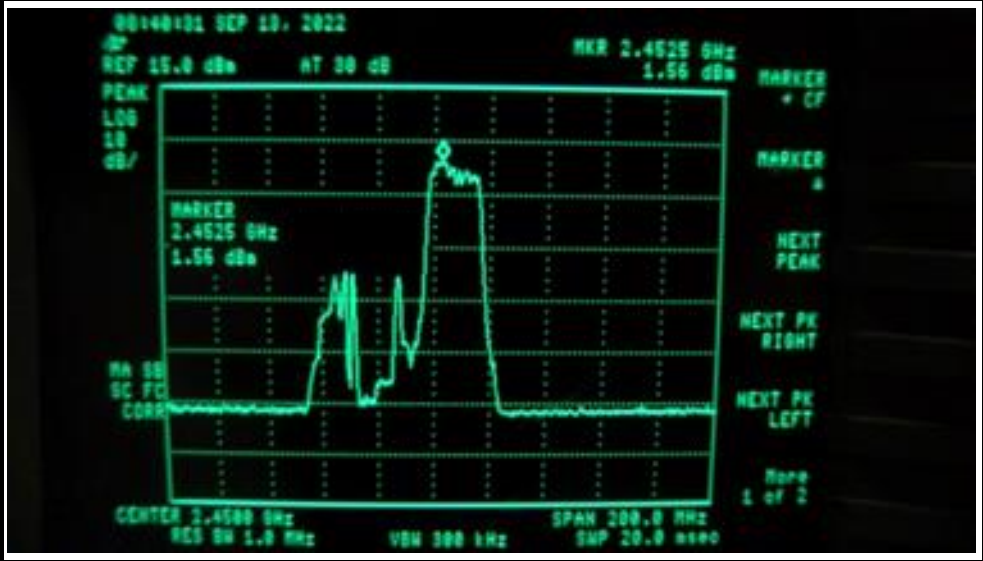 | | |
| | Freq Start (GHz) | 2.35 | | | 2.35 | | |
| | Freq End (GHz) | 2.55 | | | 2.55 | | |
| | Peaks (GHz) | - | | | 2.4525 | | |
| | | sUAS with Low Power Interference | | | sUAS with High Power Interference | | |
| | Image | 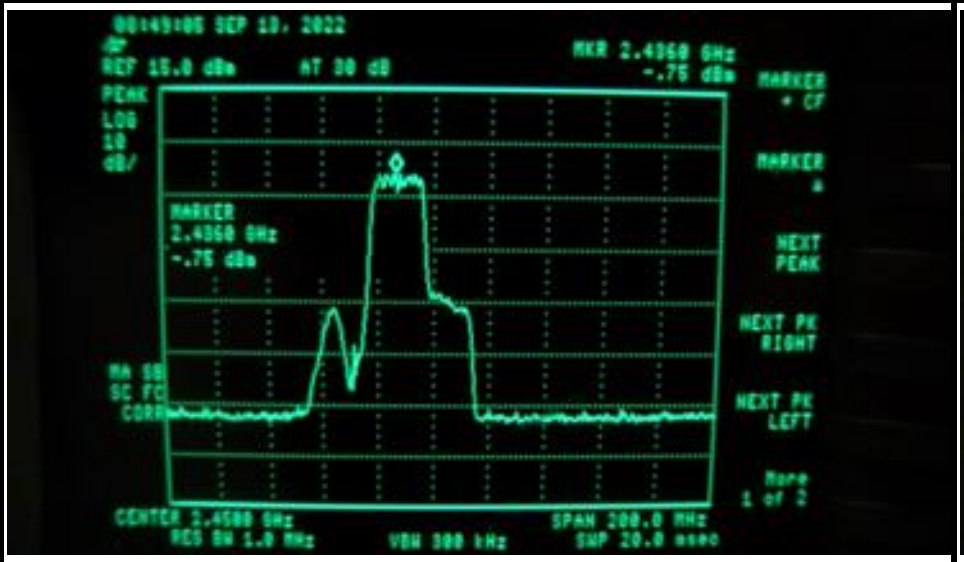 | | | 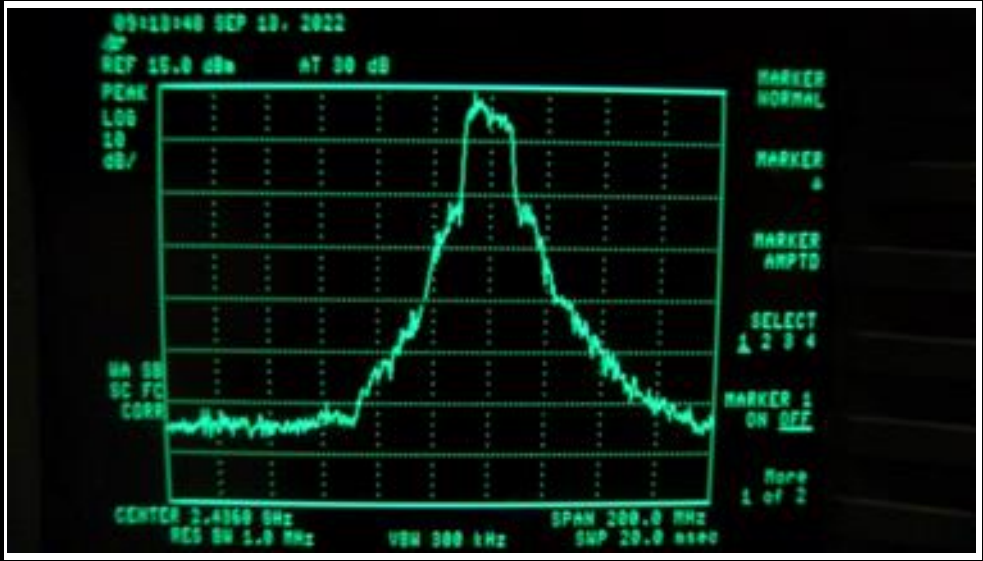 | | |
| | Freq Start (GHz) | 2.35 | | | 2.35 | | |
| | Freq End (GHz) | 2.55 | | | 2.55 | | |
| | Peaks (GHz) | 2.436 | | | 2.455 | | |
| | Metrics | Grounded | Hover | Yaw in Place | Grounded | Hover | Yaw in Place |
| | Takeoff | ✓ | - | - | X | - | - |
| | Video Link | ✓ | ✓ | // | X | X | X |
| | Flight Control | ✓ | ✓ | ✓ | X | X | X |
| | Auto Land | - | X | X | - | X | X |
| | Channel Hop | ✓ | ✓ | ✓ | X | X | X |
| | Input Persist | - | - | N/A | - | - | X |

| sUAS and Target Frequency | | Ambient Signal | | | sUAS Signal | | |
|---|---|---|---|---|---|---|---|
| Skydio X2D<br><br>1.864 GHz | Image | 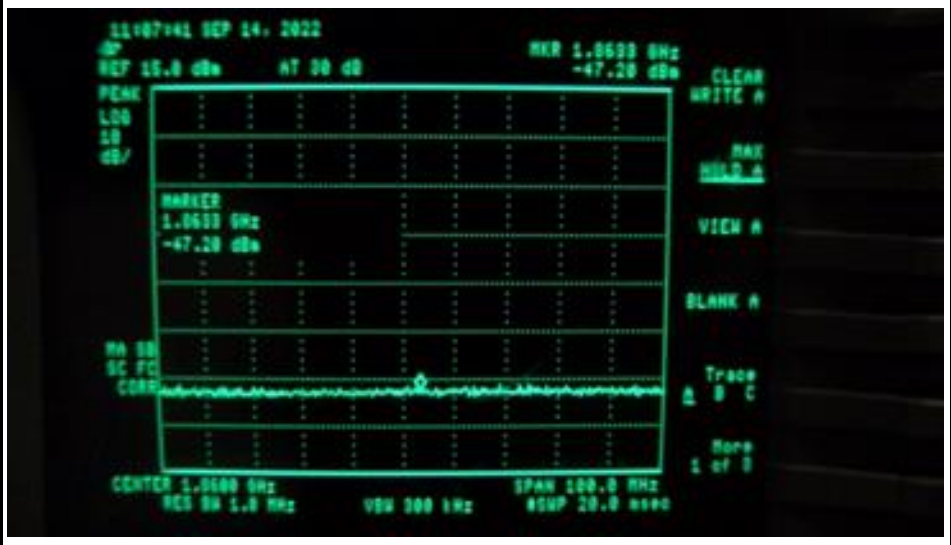 | | | 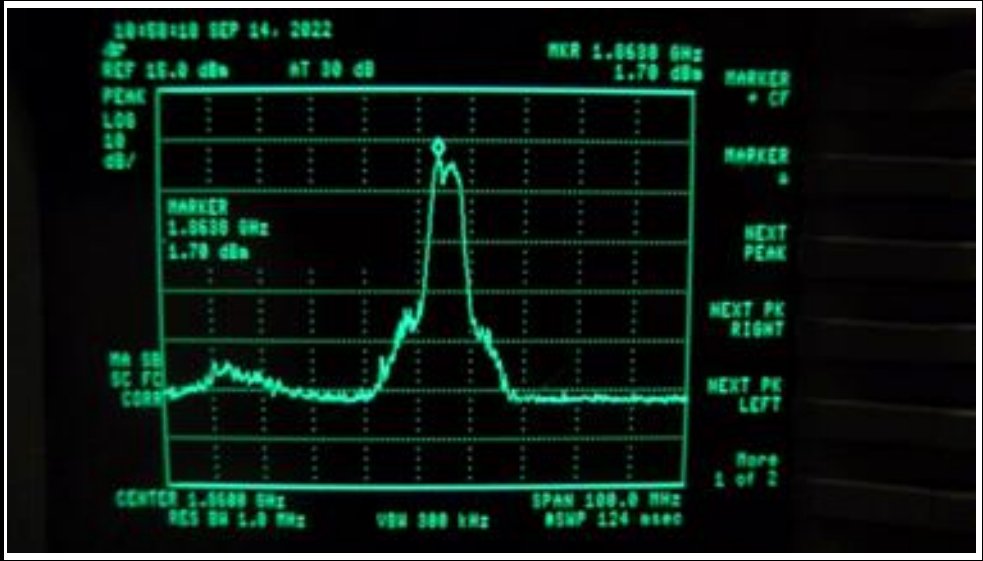 | | |
| | Freq Start (GHz) | 1.81 | | | 1.81 | | |
| | Freq End (GHz) | 1.91 | | | 1.91 | | |
| | Peaks (GHz) | - | | | 1.8638 | | |
| | | sUAS with Low Power Interference | | | sUAS with High Power Interference | | |
| | Image | 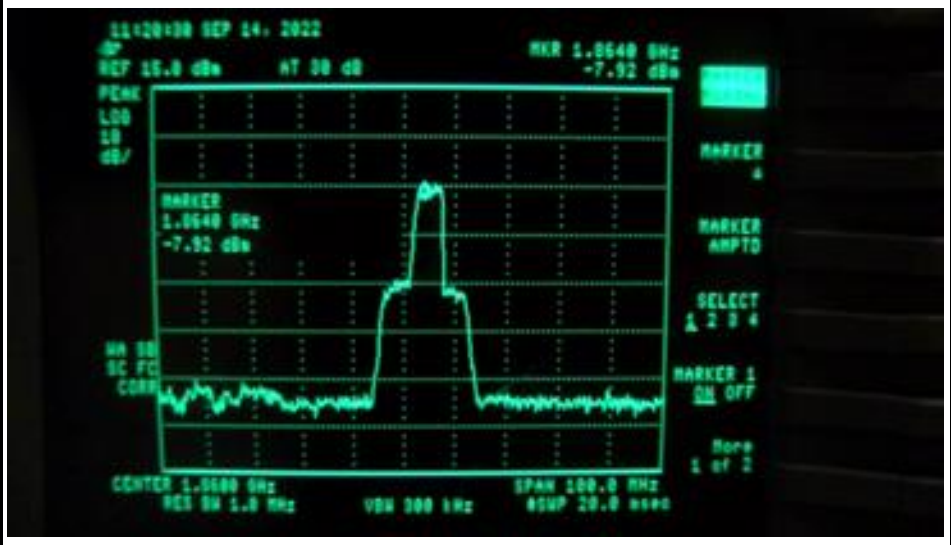 | | | 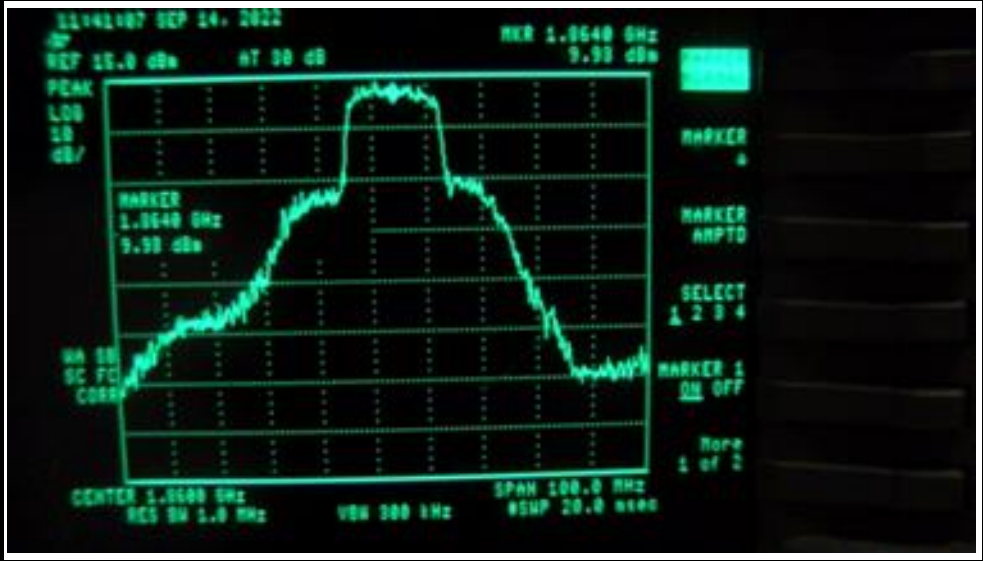 | | |
| | Freq Start (GHz) | 1.81 | | | 1.81 | | |
| | Freq End (GHz) | 1.91 | | | 1.91 | | |
| | Peaks (GHz) | - | | | - | | |
| | Metrics | Grounded | Hover | Yaw in Place | Grounded | Hover | Yaw in Place |
| | Takeoff | ✓ | - | - | X | - | - |
| | Video Link | ✓ | ✓ | ✓ | X | X | X |
| | Flight Control | ✓ | ✓ | ✓ | X | X | X |
| | Auto Land | - | X | X | - | X | X |
| | Channel Hop | X | X | X | X | X | X |
| | Input Persist | - | - | N/A | - | - | X |

| sUAS and Target Frequency | | Ambient Signal | | | sUAS Signal | | |
|---|---|---|---|---|---|---|---|
| Teal Golden Eagle*<br><br>2.469 GHz | Image | 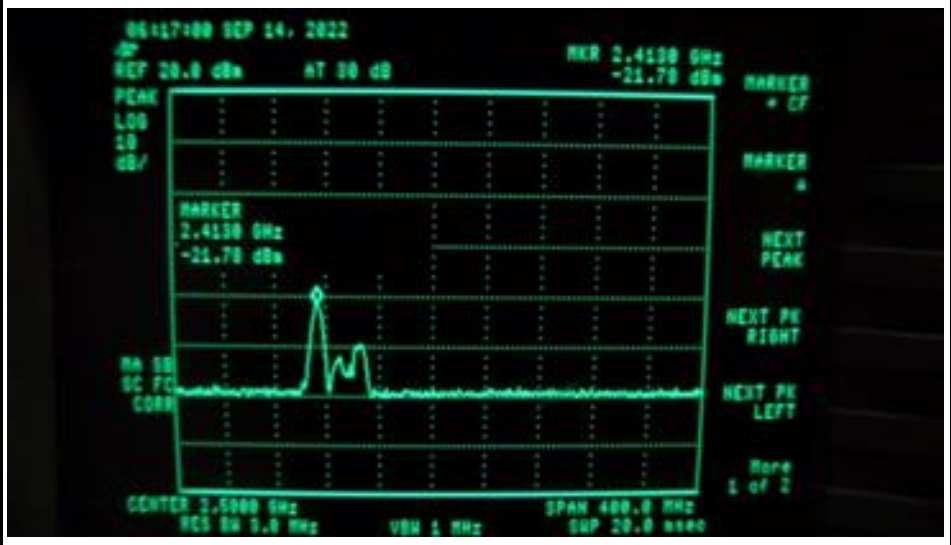 | | | 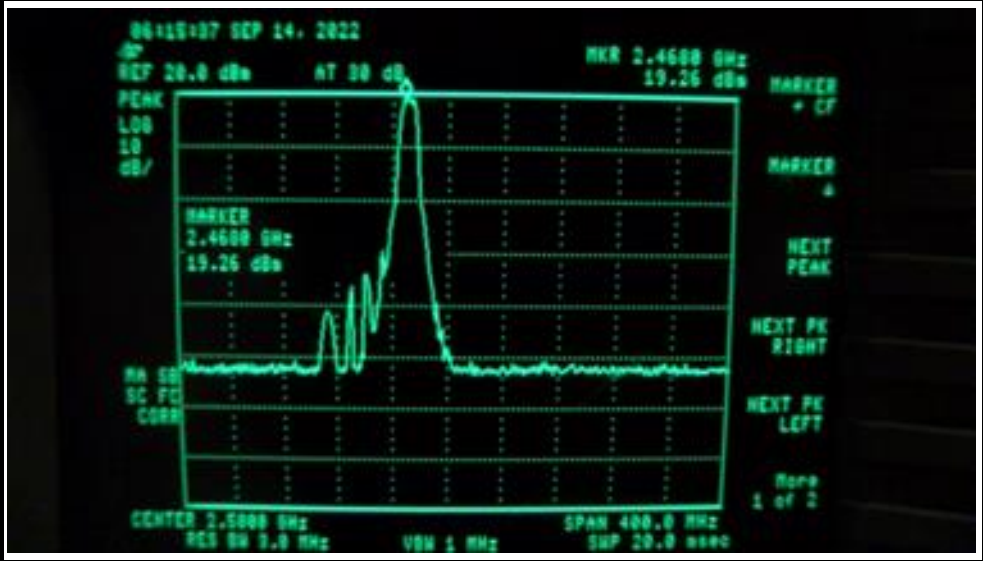 | | |
| | Freq Start (GHz) | 2.3 | | | 2.3 | | |
| | Freq End (GHz) | 2.7 | | | 2.7 | | |
| | Peaks (GHz) | 2.413 | | | 2.468 | | |
| | | sUAS with Low Power Interference | | | sUAS with High Power Interference | | |
| | Image | 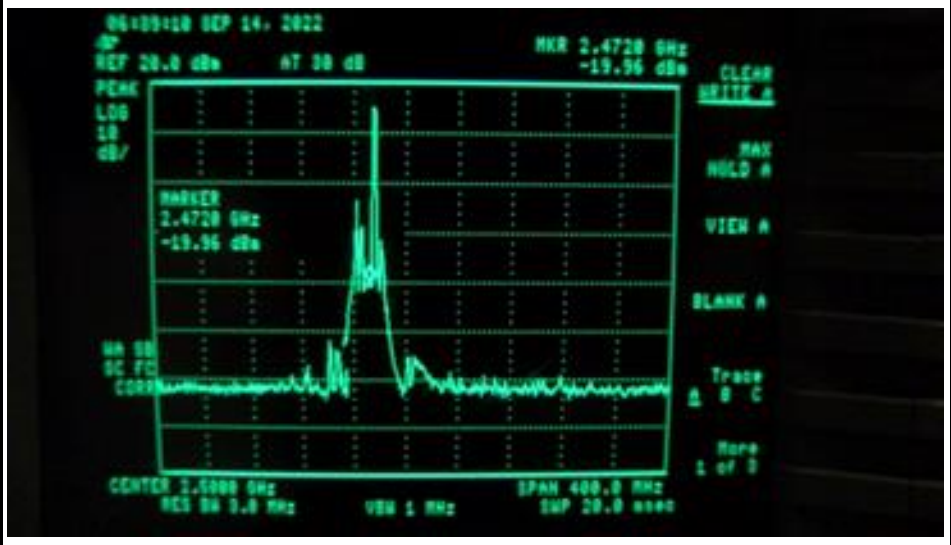 | | | 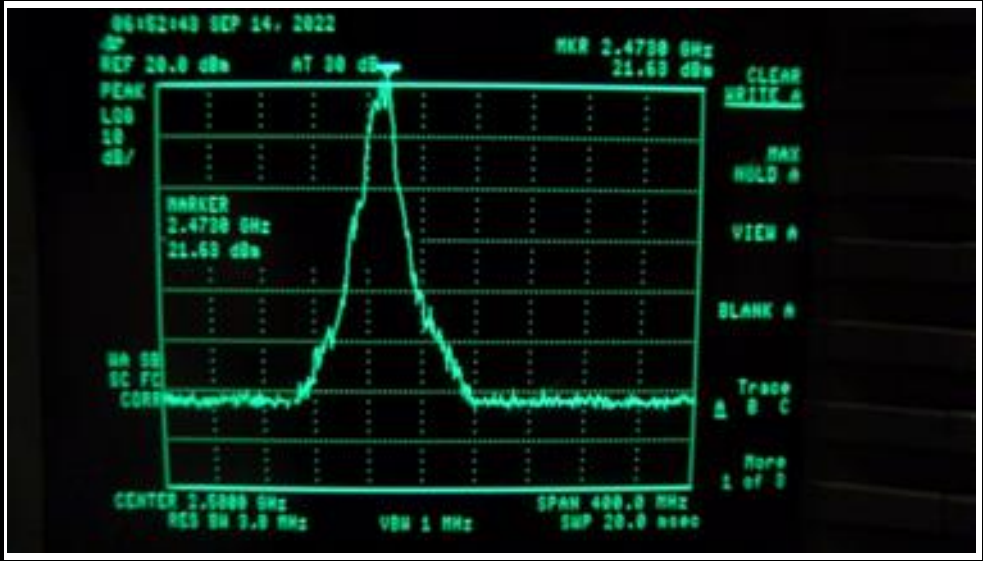 | | |
| | Freq Start (GHz) | 2.3 | | | 2.3 | | |
| | Freq End (GHz) | 2.7 | | | 2.7 | | |
| | Peaks (GHz) | - | | | - | | |
| | Metrics | Grounded | Hover | Yaw in Place | Grounded | Hover | Yaw in Place |
| | Takeoff | ✓ | X | X | X | X | X |
| | Video Link | // | X | X | X | X | X |
| | Flight Control | // | X | X | X | X | X |
| | Auto Land | - | X | X | - | X | X |
| | Channel Hop | X | X | X | X | X | X |
| | Input Persist | - | X | X | - | X | X |

*Note: Due to instability of the Teal Golden Eagle when flying indoors, we did not attempt to confirm its ability to Hover and Yaw in Place during low or high power interference.

| sUAS and Target Frequency | | Ambient Signal | | | sUAS Signal | | |
|---|---|---|---|---|---|---|---|
| Vantage Robotics Vesper† 1.853 GHz | Image | 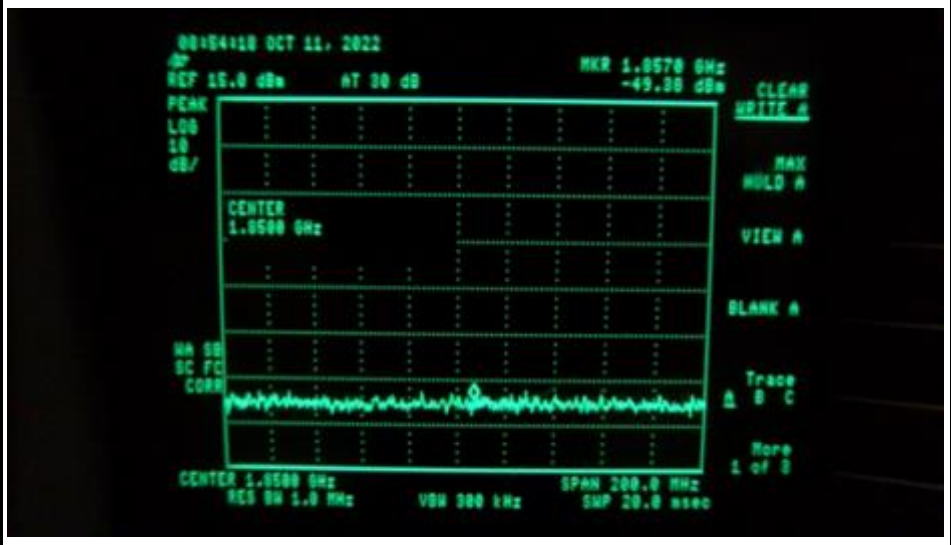 | | | 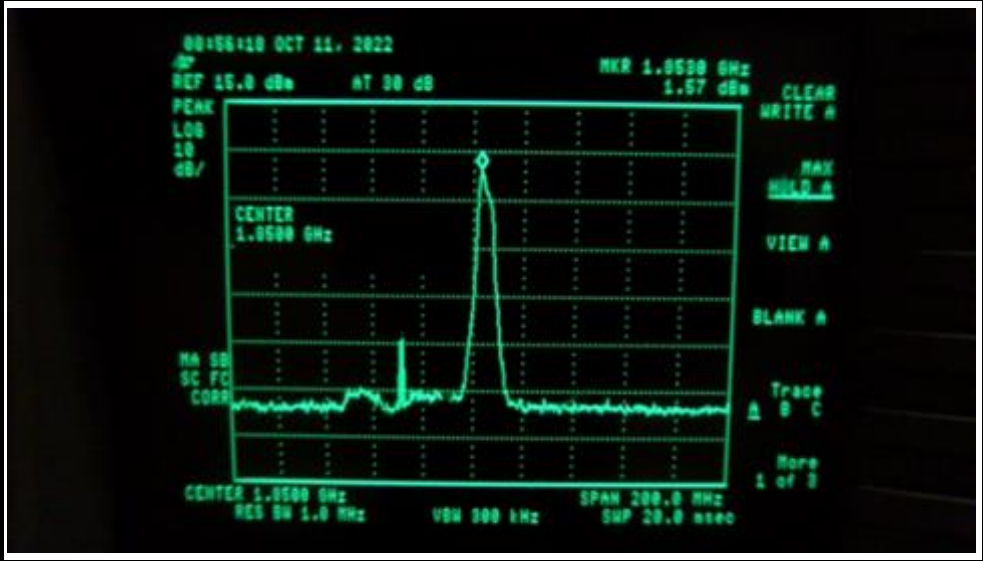 | | |
| | Freq Start (GHz) | 1.75 | | | 1.75 | | |
| | Freq End (GHz) | 1.95 | | | 1.95 | | |
| | Peaks (GHz) | - | | | 1.853 GHz, 1.82 GHz | | |
| | | **sUAS with Low Power Interference** | | | **sUAS with High Power Interference** | | |
| | Image | 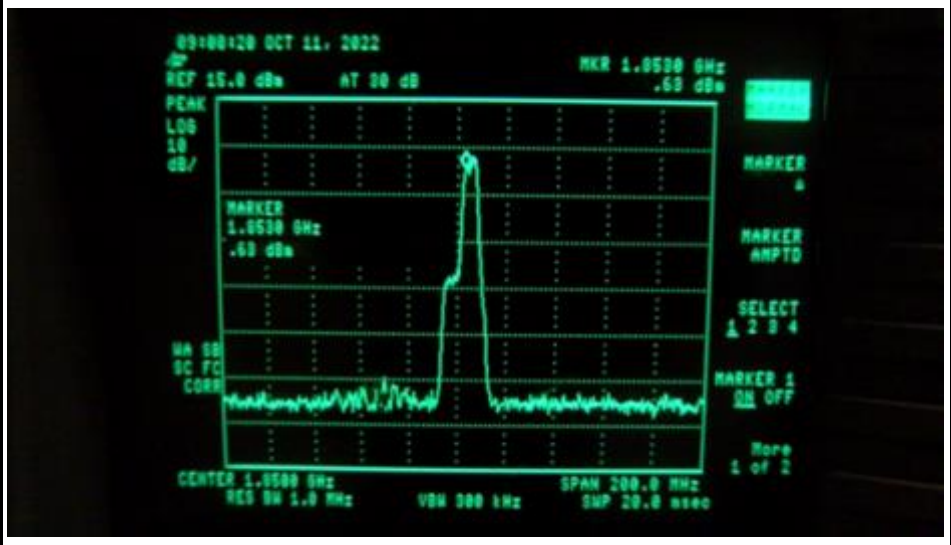 | | | 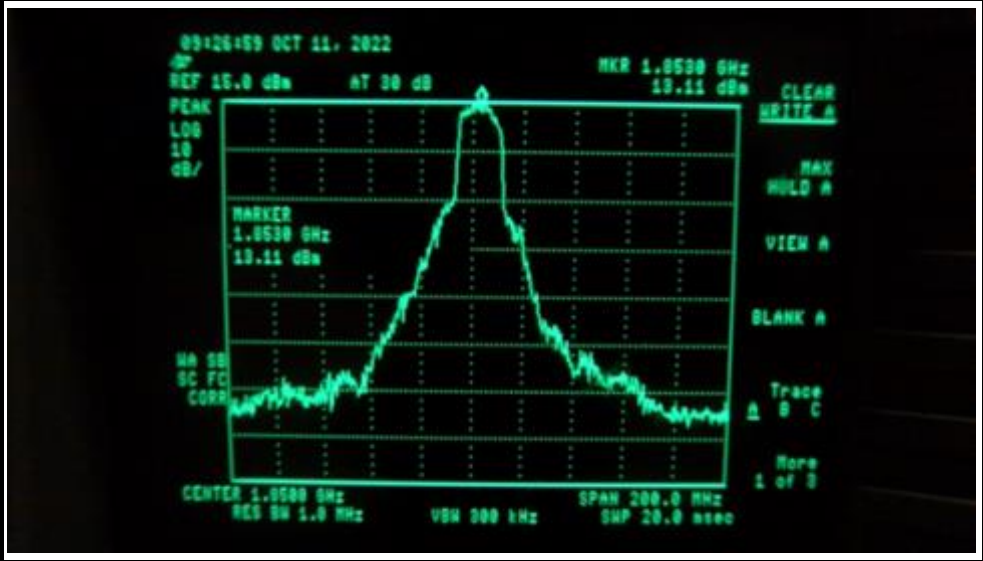 | | |
| | Freq Start (GHz) | 1.75 | | | 1,75 | | |
| | Freq End (GHz) | 1.95 | | | 1.95 | | |
| | Peaks (GHz) | - | | | - | | |
| | **Metrics** | **Grounded** | **Hover** | **Yaw in Place** | **Grounded** | **Hover** | **Yaw in Place** |
| | Takeoff | ✓ | - | - | X | - | - |
| | Video Link | ✓ | ✓ | ✓ | X | X | N/A |
| | Flight Control | ✓ | ✓ | ✓ | X | // | |
| | Auto Land | - | X | X | - | X | |
| | Channel Hop | X | X | X | X | X | |
| | Input Persist | - | - | N/A | - | - | |

†Note: During the Hover test, the Vantage Robotics Vesper was receiving very few inputs from the pilot, got stuck ascending, and then veered left, then right, colliding with a wall and disassembling itself. It sustained significant damage so was not able to perform the Yaw in Place test.

# Field Readiness

## Runtime Endurance

### Summary of Test Method

The sUAS is continuously maneuvered either for flight throughout an environment or camera movement when stationary to inspect an environment. Three types of Runtime Endurance tests are specified; for all three tests, the operator maneuvers the sUAS as described until either (a) the battery life is exhausted, or (b) the OCU warns the operator of low battery, requiring the sUAS to be flown back to the launch point. All tests can be run in lighted (100 lux or greater) or dark (less than 1 lux) conditions. The three tests are:

<u>Indoor Movement</u>: In an indoor environment, the sUAS navigates within a series of poles and two apertures to form a figure-8 path with elevation changes. The obstacle avoidance settings on the sUAS can be varied (e.g., 1 m clearance to obstacles, no obstacle avoidance, etc.).

<u>Hover and Stare Activities</u>: In either an outdoor or indoor environment, the sUAS launches and hovers at a waypoint, then the operator maneuvers the camera to predefined position and zoom settings, idles for 2 minutes, and then maneuvers to the next camera position, idles again for 2 minutes, and repeats this process. The sUAS can be configured as GPS-enabled or VIO-enabled.

<u>Perch and Stare Activities</u>: In either an outdoor or indoor environment, the sUAS launches, lands on a platform, and the operator maneuvers the camera to predefined position and zoom settings, idles for 2 minutes, and then maneuvers to the next camera position, idles again for 2 minutes, and repeats this process. The sUAS can be configured as GPS-enabled or VIO-enabled.

Metrics from all four tests should be reported in order to demonstrate a spread of expected performance duration, pending the type of mission being performed.

See Figure 1. Metrics from all tests should be reported in order to demonstrate a spread of expected performance duration, pending the type of mission being performed.

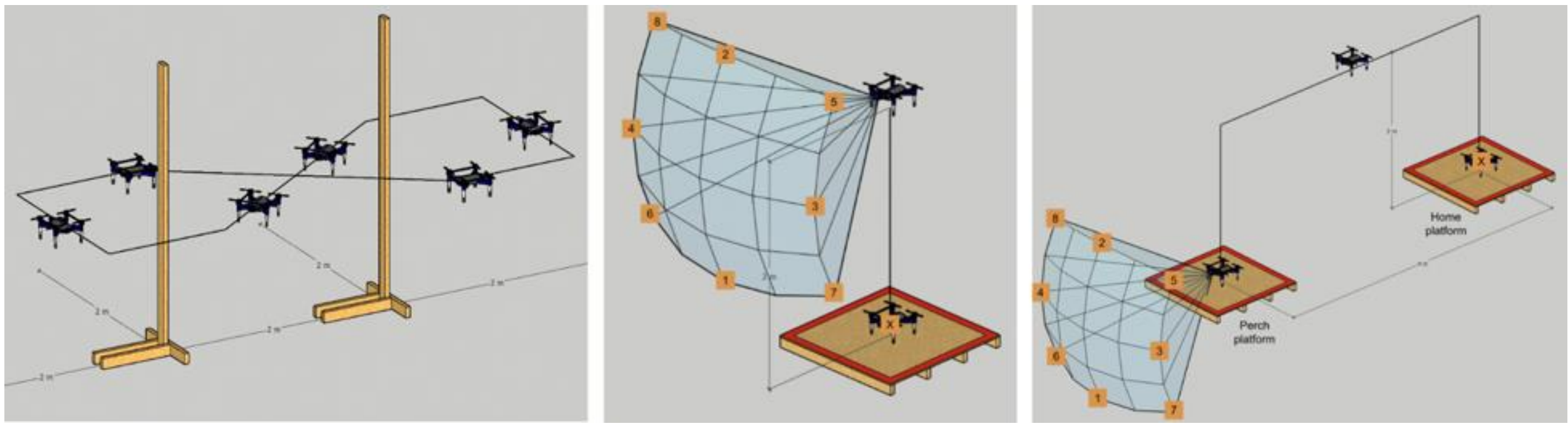

*Figure 1. Apparatuses for the Runtime Endurance test method: indoor movement (left), hover and stare activities (middle), perch and stare activities (right).*

## Benchmarking Results

Indoor Movement: 1 figure-8 lap = nominal 13 m

Best in class:

- Indoor Movement = duration of approximately 30 minutes or higher
- Hover and Stare = duration of approximately 30 minutes or higher
- Perch and Stare = duration of 300 minutes (5 hours) or higher

| sUAS | Test | Indoor Movement | | | Hover and Stare | Perch and Stare |
|---|---|---|---|---|---|---|
| | Metric | Duration (min) | Distance (m) | Average speed (m/s) | Duration (min) | Duration (min) |
| Cleo Robotics Dronut X1P | | 8 | 260 | 0.5 | 9 | 182 |
| FLIR Black Hornet PRS | | 13 | 390 | 0.5 | 13 | - |
| Flyability Elios 2 GOV | | 10 | 442 | 0.7 | 9 | - |
| Lumenier Nighthawk V3* | | X | X | X | 10 | X |
| Parrot ANAFI USA GOV | | 32 | 299 | 0.2 | 30 | 315 |
| Skydio X2D† | | 31 | 754 | 0.4 | 29 | 222 |
| Teal Golden Eagle# | | X | X | X | 12 | 44 |
| Vantage Robotics VesperΔ | | - | - | - | 23 | X |
| **Best in class** | | **Indoor Movement** | | | **Hover and Stare** | **Perch and Stare** |
| | | Parrot ANAFI USA GOV<br>Skydio X2D | | | Parrot ANAFI USA GOV<br>Skydio X2D | Parrot ANAFI USA GOV |
| | | **Runtime Endurance, overall** | | | | |
| | | Parrot ANAFI USA GOV | | | | |

*Note: The Lumenier Nighthawk V3 has a tendency to overheat over long periods of time, causing it to disconnect from the controller. For this reason, neither the Indoor Movement or Perch and Stare tests were successfully evaluated.

†Note: The Skydio X2D will reset itself approximately every hour in order to clear its available RAM space for recording data. This is alerted to the operator on the OCU and the operator must reconnect at range when this happens.

#Note: Due to instability of the Teal Golden Eagle when flying indoors, we did not attempt to evaluate its Indoor Movement endurance. The controller has a tendency to disconnect from the system around 30% battery. During the Perch and Stare test, the camera unit overheated causing the camera signal to cut out, triggering the end of the test, despite still having some amount of battery remaining.

ΔNote: The Vantage Robotics Vesper was unavailable for testing due to being out with the vendor for repairs.

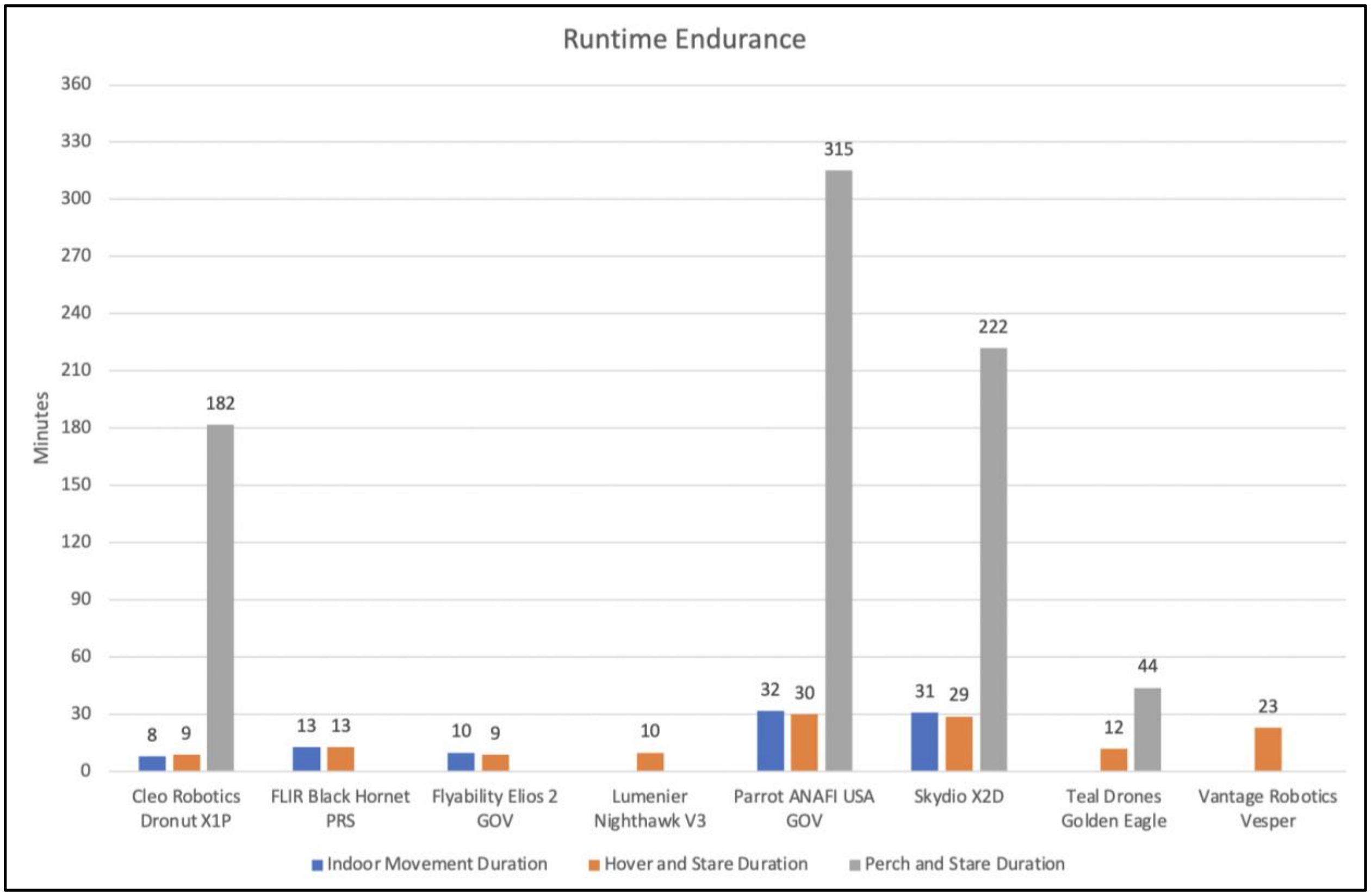

Runtime Endurance
Minutes
360
330
300
270
240
210
180
150
120
90
60
30
0
8 9 182
13 13
10 9
10
32 30 315
31 29 222
12 44
23
Cleo Robotics Dronut X1P
FLIR Black Hornet PRS
Flyability Elios 2 GOV
Lumenier Nighthawk V3
Parrot ANAFI USA GOV
Skydio X2D
Teal Drones Golden Eagle
Vantage Robotics Vesper
Indoor Movement Duration
Hover and Stare Duration
Perch and Stare Duration

# Takeoff and Land/Perch

## Summary of Test Method

A series of conditions are specified that define variations in the ground plane (pitch/roll angle, sensor interference due to material/proximity to external electronics) and nearby obstructions (overhead or lateral obstructions). These variations in the environment can impact a system's ability to takeoff and/or land/perch as it is common for sUAS to have built-in functionality that checks for level ground and/or the presence of obstructions nearby. Such functionality may not allow for the system to launch due to safety concerns (e.g., sUAS behavior may require ascension to a certain height upon takeoff before continuing to operate), or systems without such safety precautions may allow them to attempt takeoff and land/perch regardless of the environment, which may cause collisions or rollovers. Ten conditions are specified in terms of ground plane material, angle, and obstructions:

| Condition | Ground plane material | Ground plane angle | Obstructions | Image |
|---|---|---|---|---|
| 1 | Wood | 0° | None | 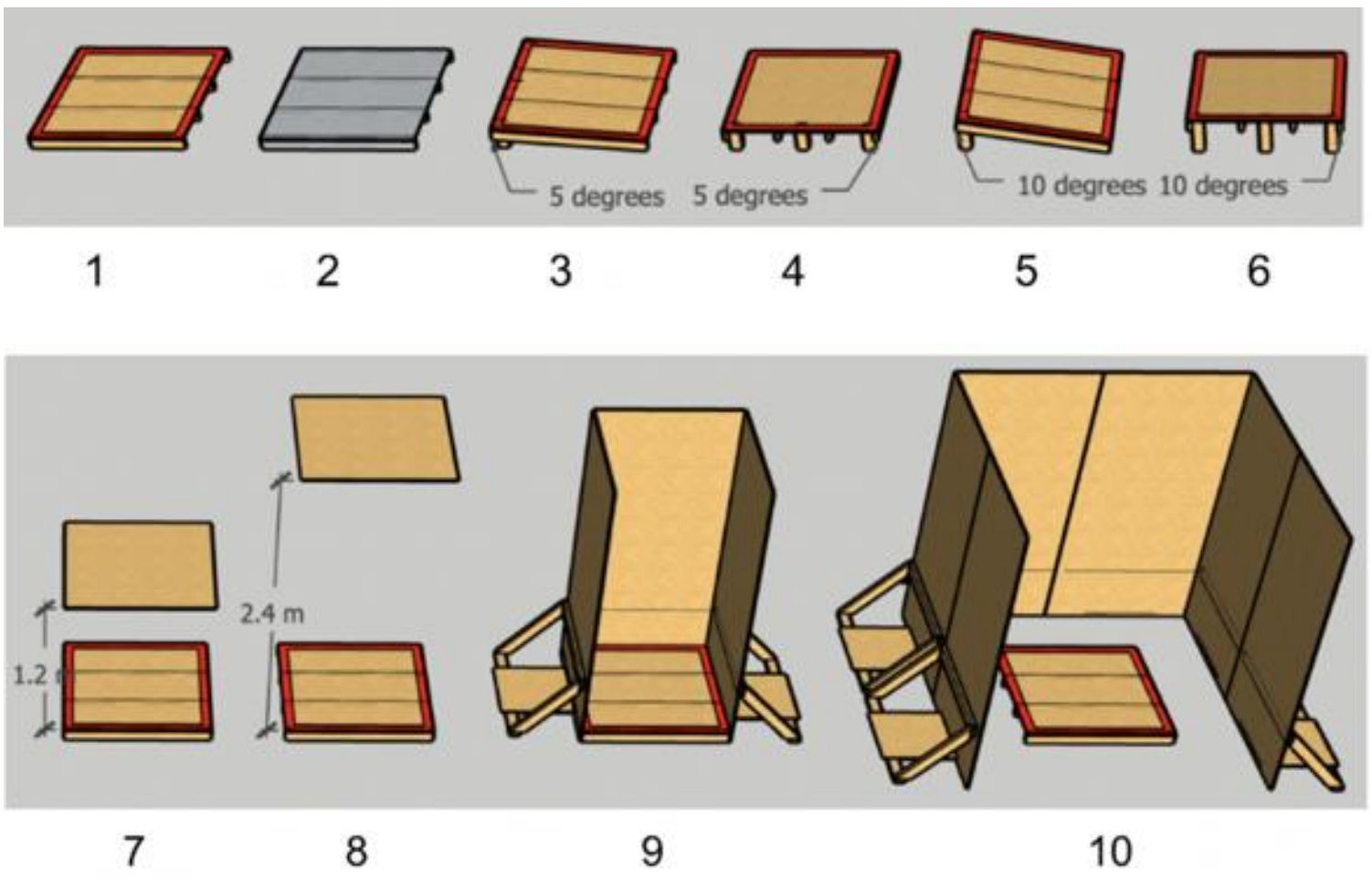  |
| 2 | Metal with embedded electronics | 0° | None | |
| 3 | Wood | 5° roll | None | |
| 4 | Wood | 5° pitch | None | |
| 5 | Wood | 10° roll | None | |
| 6 | Wood | 10° pitch | None | |
| 7 | Wood | 0° | 1.2 m (4 ft) overhead | |
| 8 | Wood | 0° | 2.4 m (8 ft) overhead | |
| 9 | Wood | 0° | 1.2 m (4 ft) lateral | |
| 10 | Wood | 0° | 2.4 m (8 ft) lateral | |

Data for conditions 1 and 3-10 are shown in this report. Condition 2 was not run.

Two tests are specified:

<u>Takeoff</u>: Starting from a ground position on top of a platform, the sUAS attempts to launch into the air, navigate forward at least 3 m (10 ft).

<u>Land/Perch</u>: Starting hovering in the air at a position 3 m (10 ft) away from the landing platform, the sUAS attempts to navigate toward the platform and land on it.

Both tests can be performed back-to-back so long as takeoff is successful. Each test under each condition is performed multiple times to establish statistical significance and the associated probability of success and confidence levels based on the number of successes and failures (see the metrics section). Tests can be attempted in lighted (100 lux or greater) or dark (less than 1 lux) conditions.

## Benchmarking Results

All tests were conducted in lighted conditions, except for a variation on condition 1 (flat) that was evaluated in darkness. It is presented in a separate subsection below under "Dark Performance."

### Takeoff

For each condition, 10 repetitions were attempted unless otherwise noted. If the first attempt resulted in a failure or presented other safety concerns, that condition was likely abandoned and moved onto the next condition.

Best in class = 90% success or higher across all conditions

| | | Condition | | | | | | | | |
|---|---|---|---|---|---|---|---|---|---|---|
| **sUAS** | **Metrics** | **Flat** | **5° roll** | **5° pitch** | **10° roll** | **10° pitch** | **1.2 m overhead** | **2.4 m overhead** | **1.2 m lateral** | **2.4 m lateral** |
| Cleo Robotics Dronut X1P | Completion | 100% | 100% | 80% | 80% | 90% | 0% | 100% | 100% | 100% |
| | Collisions | 0 | 0 | 0 | 0 | 0 | 1 | 0 | 0 | 0 |
| | Rollovers | 0 | 0 | 2 | 2 | 1 | 1 | 0 | 0 | 0 |
| FLIR Black Hornet PRS | Completion | - | - | - | - | - | - | - | - | - |
| | Collisions | - | - | - | - | - | - | - | - | - |
| | Rollovers | - | - | - | - | - | - | - | - | - |
| Flyability Elios 2 GOV | Completion | 100% | 100% | 100% | 100% | 100% | 100% | 100% | 100% | 100% |
| | Collisions | 0 | 0 | 0 | 0 | 0 | 0 | 0 | 0 | 0 |
| | Rollovers | 0 | 0 | 0 | 0 | 0 | 0 | 0 | 0 | 0 |
| Lumenier Nighthawk V3 | Completion | 100% | 100% | 100% | 100% | 100% | 100% | 100% | 100% | 100% |
| | Collisions | 0 | 0 | 0 | 0 | 0 | 0 | 0 | 0 | 0 |
| | Rollovers | 0 | 0 | 0 | 0 | 0 | 0 | 0 | 0 | 0 |
| Parrot ANAFI USA GOV | Completion | 100% | 100% | 100% | 100% | 100% | 100% | 100% | 100% | 100% |
| | Collisions | 0 | 0 | 0 | 0 | 0 | 0 | 0 | 0 | 0 |
| | Rollovers | 0 | 0 | 0 | 0 | 0 | 0 | 0 | 0 | 0 |
| Skydio X2D* | Completion | 100% | 100% | 100% | 100% | 100% | 0% | X | 0% | 100% |
| | Collisions | 0 | 0 | 0 | 0 | 0 | 0 | 1 | 0 | 0 |
| | Rollovers | 0 | 0 | 0 | 0 | 0 | 0 | 0 | 0 | 0 |
| Teal Golden Eagle† | Completion | 100% | 100% | 100% | 100% | 100% | 100% | 100% | 100% | 100% |
| | Collisions | 0 | 0 | 0 | 0 | 0 | 0 | 0 | 3 | 0 |
| | Rollovers | 0 | 0 | 0 | 0 | 0 | 0 | 0 | 0 | 0 |
| Vantage Robotics Vesper | Completion | 100% | 100% | 100% | 100% | 100% | 100% | 100% | 90% | 100% |
| | Collisions | 0 | 0 | 0 | 0 | 1 | 3 | 0 | 3 | 0 |
| | Rollovers | 0 | 0 | 0 | 0 | 0 | 0 | 0 | 0 | 0 |

**Best in class**

**Takeoff, uneven ground**
- Flyability Elios 2 GOV
- Lumenier Nighthawk V3
- Parrot ANAFI USA GOV
- Teal Golden Eagle
- Vantage Robotics Vesper

**Takeoff, confined space**
- Flyability Elios 2 GOV
- Lumenier Nighthawk V3
- Parrot ANAFI USA GOV
- Teal Golden Eagle

**Takeoff, overall**
- Flyability Elios 2 GOV
- Lumenier Nighthawk V3
- Parrot ANAFI USA GOV
- Teal Golden Eagle

*Note: When testing the Skydio X2D, the 10° pitch condition was only attempted 5 times due to the system sliding down the platform as it was taking off, presenting a potential safety concern. The 2.4 m overhead condition was only attempted once due to the system's antennae hitting the ceiling; the Skydio X2D attempts to elevate to 3 m (10 ft) upon takeoff, presenting a safety concern so the condition was abandoned. The Skydio X2D's will assess the area before taking off, which prevented it from taking off during the 1.2 m overhead and 1.2 m lateral obstruction conditions.

†Note: When the Teal Golden Eagle was evaluated, it was using an older version of its firmware that resulted in more stable indoor flight, which later (upon a mandatory update) appeared to have begun performing worse indoors.

## Land/Perch

For each condition, 10 repetitions were attempted unless otherwise noted. If the first attempt resulted in a failure or presented other safety concerns, that condition was likely abandoned and moved onto the next condition.

Best in class = 90% success or higher across all conditions

| sUAS | Metrics | Condition: Flat | 5° roll | 5° pitch | 10° roll | 10° pitch | 1.2 m overhead | 2.4 m overhead | 1.2 m lateral | 2.4 m lateral |
|---|---|---|---|---|---|---|---|---|---|---|
| Cleo Robotics Dronut X1P | Completion | 90% | 90% | 63% | 88% | 90% | 100% | 100% | 100% | 100% |
| | Collisions | 0 | 0 | 3 | 0 | 2 | 0 | 0 | 0 | 0 |
| | Rollovers | 1 | 1 | 0 | 1 | 1 | 0 | 0 | 0 | 0 |
| FLIR Black Hornet PRS | Completion | - | - | - | - | - | - | - | - | - |
| | Collisions | - | - | - | - | - | - | - | - | - |
| | Rollovers | - | - | - | - | - | - | - | - | - |
| Flyability Elios 2 GOV | Completion | 100% | 100% | 100% | 100% | 100% | 100% | 100% | 100% | 100% |
| | Collisions | 0 | 0 | 0 | 0 | 0 | 0 | 0 | 0 | 0 |
| | Rollovers | 0 | 0 | 0 | 0 | 0 | 0 | 0 | 0 | 0 |
| Lumenier Nighthawk V3 | Completion | 100% | 100% | 80% | 100% | 100% | 100% | 100% | 100% | 100% |
| | Collisions | 0 | 0 | 2 | 0 | 0 | 0 | 0 | 0 | 0 |
| | Rollovers | 0 | 0 | 0 | 0 | 0 | 0 | 0 | 0 | 0 |
| Parrot ANAFI USA GOV | Completion | 100% | 100% | 100% | 100% | 100% | 100% | 100% | 100% | 100% |
| | Collisions | 0 | 0 | 0 | 0 | 0 | 0 | 0 | 0 | 0 |
| | Rollovers | 0 | 0 | 0 | 0 | 0 | 0 | 0 | 0 | 0 |
| Skydio X2D* | Completion | 100% | 100% | 100% | X | 0% | 100% | 100% | 0% | 100% |
| | Collisions | 0 | 0 | 0 | 1 | 2 | 0 | 0 | 0 | 0 |
| | Rollovers | 0 | 0 | 0 | 0 | 0 | 0 | 0 | 0 | 0 |
| Teal Golden Eagle | Completion | 100% | 100% | 100% | 100% | 100% | 100% | 100% | 100% | 90% |
| | Collisions | 1 | 0 | 1 | 0 | 0 | 0 | 0 | 10+ | 10+ |
| | Rollovers | 0 | 0 | 0 | 0 | 0 | 0 | 0 | 0 | 0 |
| Vantage Robotics Vesper | Completion | 100% | 100% | 100% | 90% | 90% | 90% | 90% | 100% | 70% |
| | Collisions | 0 | 0 | 0 | 1 | 0 | 1 | 2 | 0 | 1 |
| | Rollovers | 0 | 0 | 0 | 0 | 0 | 0 | 0 | 0 | 0 |
| | **Best in class** | **Land/Perch, uneven ground** | | | | | **Land/Perch, confined space** | | | |
| | | Flyability Elios 2 GOV<br>Parrot ANAFI USA GOV<br>Teal Golden Eagle | | | | | Flyability Elios 2 GOV<br>Lumenier Nighthawk V3<br>Parrot ANAFI USA GOV | | | |
| | | **Land/Perch, overall** | | | | | | | | |
| | | Flyability Elios 2 GOV<br>Parrot ANAFI USA GOV | | | | | | | | |

*Note: During the 10° roll condition, the Skydio X2D props slightly clipped the platform upon landing, and as the motors spun down, the drone slid down the platform. Only 1 attempt at this condition was conducted, then abandoned due to safety concerns. The system's obstacle avoidance functionality (even when configured to the "minimal" setting) would not allow it to enter the 1.2 m lateral obstruction condition.

## Dark Operations

This test only consisted of running a variation of condition 1 (flat), but in darkness. A set of 5 repetitions were attempted. If takeoff was not successful, then the lights were turned on to allow the system to takeoff, then the lights were turned off while hovering, and land/perch was attempted.

Best in class = 90% success or higher

| sUAS | Metrics | Flat, darkness | |
|---|---|---|---|
| | | Takeoff | Land/Perch |
| Cleo Robotics Dronut X1P | Completion | 20% | 100% |
| | Collisions | 0 | 0 |
| | Rollovers | 0 | 0 |
| FLIR Black Hornet PRS* | Completion | 0% | 0% |
| | Collisions | 0 | 2 |
| | Rollovers | 0 | 0 |
| Flyability Elios 2 GOV | Completion | 100% | 100% |
| | Collisions | 0 | 0 |
| | Rollovers | 0 | 0 |
| Lumenier Nighthawk V3† | Completion | - | - |
| | Collisions | - | - |
| | Rollovers | - | - |
| Parrot ANAFI USA GOV | Completion | 100% | 100% |
| | Collisions | 0 | 0 |
| | Rollovers | 0 | 0 |
| Skydio X2D | Completion | 0% | 0% |
| | Collisions | 0 | 0 |
| | Rollovers | 0 | 0 |
| Teal Golden Eagle# | Completion | X | X |
| | Collisions | X | X |
| | Rollovers | X | X |
| Vantage Robotics VesperΔ | Completion | - | - |
| | Collisions | - | - |
| | Rollovers | - | - |
| | **Best in class** | **Takeoff, in darkness** | **Land/Perch, in darkness** |
| | | Flyability Elios 2 GOV<br>Parrot ANAFI USA GOV | Cleo Robotics Dronut X1P<br>Flyability Elios 2 GOV<br>Parrot ANAFI USA GOV |
| | | **Takeoff and Land/Perch, in darkness** | |
| | | Flyability Elios 2 GOV<br>Parrot ANAFI USA GOV | |

*Note: The FLIR Black Hornet PRS must takeoff from and land in the operator's hand.

†Note: The Nighthawk has illuminators and is typically able to operate in the dark, but was not able to be tested due to availability (out for repairs).

#Note: Due to instability of the Teal Golden Eagle when flying indoors, we did not attempt to fly it to confirm communications performance at each location.

ΔNote: Due to a firmware issue, the vendor recommended the Vantage Robotics Vesper to be grounded, making it unavailable to perform this test.

# Room Clearing

## Summary of Test Method

While the environments in a mission-context where room clearing can be performed will vary in terms of room dimensions and types of obstructions in the room, a standard room is specified for this test method to be representative of a room clearing task. The nominally-sized room has a series of visual acuity targets on all surfaces and is without obstructions to provide a clear view of all targets for inspection. A future variant of this test method may be developed that includes one or more sets of standard obstruction layouts. The sUAS can takeoff inside of the room or enter from outside, whichever is preferred; either way, the actual test will not begin until after the sUAS is hovering in the center of the room. From there, the sUAS performs a visual inspection of the room. While not required, it is recommended that the sUAS remain in the center of the room and manipulate its gimbal camera to increase vertical field of view (FOV) for inspecting the floor and ceiling, while yawing in place.

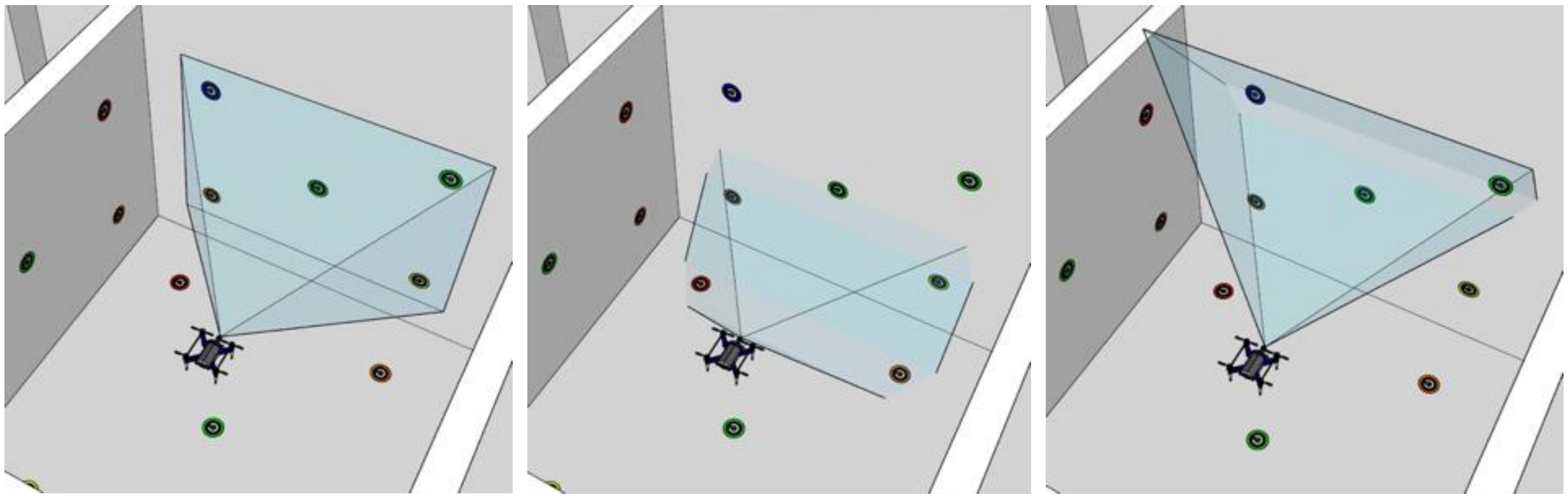

*Figure 1. Rendering of example sUAS field of view when inspecting a wall, floor, and ceiling by manipulating the gimbal camera. Note: ceiling is not shown, but is present during testing.*

The sUAS gimbal movement range and FOV will impact the number of visual acuity targets that can be inspected; e.g., some sUAS will not be able to see the targets on the floor or ceiling due to lack of gimbal capability, while others may be able to see multiple surfaces at once through the use of 360 degree cameras. Additionally, the control and stabilization of the sUAS is exercised by attempting to yaw in place to scan the room. The sUAS is free to move through the room as needed (e.g., navigate forward, back, ascend, descend, etc.), although the room is intentionally narrow to influence a more expedient scanning technique of yawing in place. Two variants of room clearing capability are exercised:

- Static camera: Without the use of camera zoom functionality, likely resulting in faster, coarser room clearing at reduced visual acuity.
- Zoom camera: Allowing for the use of camera zoom functionality if available, likely resulting in slower, finer room clearing at increased visual acuity.

During the test, the operator inspects the visual acuity targets and the test lasts until all visual acuity targets able to be inspected (i.e., those that the sUAS has the capability of inspecting; some targets may not be able to be inspected due to limitations in sUAS gimbal movement), have been successfully inspected.

Room clearing be run either as an elemental or operational test:

Elemental Room Clearing: The operator may maintain line-of-sight with the sUAS such as by following the system with the OCU and standing in the doorway to maintain communications link, allowing for room clearing to be evaluated in as close to an ideal setting as possible and reduce potential collisions with the boundaries.

Operational Room Clearing: The operator is positioned away from the room with their back to the doorway, unable to maintain line of sight throughout the test. This is similar to an actual operational mission, including all related situation awareness issues that may arise (e.g., collisions with the boundaries, misunderstanding which wall is being inspected, etc.).

## Benchmarking Results

Operational room clearing was performed for all systems.

Best in class = 90% coverage or higher with average acuity of 3 mm or higher (note: lower acuity measurements in millimeters equate to higher acuity)

| sUAS | Metrics | In-situ Static cam | In-situ Zoom cam* | Post-hoc Static cam | Post-hoc Zoom cam* | Post-hoc 360 Superzoom |
|---|---|---|---|---|---|---|
| Cleo Robotics Dronut X1P | Duration (min) | 3.4 | - | 1.4 | - | - |
| | Coverage | 93% | - | 93% | - | - |
| | Average acuity (mm) | 11.2 | - | 7.7 | - | - |
| FLIR Black Hornet PRS | Duration (min) | 5.2 | 5.2 | 5.2 | 5.2 | - |
| | Coverage | 83% | 83% | 83% | 83% | - |
| | Average acuity (mm) | 7.8 | 7.8 | 7.4 | 7.4 | - |
| Flyability Elios 2 GOV | Duration (min) | 5.0 | - | 2.0 | - | - |
| | Coverage | 100% | - | 100% | - | - |
| | Average acuity (mm) | 5.5 | - | 2.8 | - | - |
| Lumenier Nighthawk V3 | Duration (min) | 4.0 | - | 4.0 | - | - |
| | Coverage | 100% | - | 100% | - | - |
| | Average acuity (mm) | 7.1 | - | 6.9 | - | - |
| Parrot ANAFI USA GOV | Duration (min) | 4.6 | 12.1 | 2.1 | 7.8 | - |
| | Coverage | 100% | 100% | 100% | 100% | - |
| | Average acuity (mm) | 3.0 | 1.3 | 2.9 | 1.3 | - |
| Skydio X2D | Duration (min) | 6.6 | 6.6 | 3.6 | 10.7 | 0.4 |
| | Coverage | 100% | 100% | 100% | 100% | 100% |
| | Average acuity (mm) | 2.7 | 2.7 | 2.2 | 1.5 | 19.1 |
| Teal Golden Eagle | Duration (min) | - | - | - | - | - |
| | Coverage | - | - | - | - | - |
| | Average acuity (mm) | - | - | - | - | - |
| Vantage Robotics Vesper | Duration (min) | 3.6 | 3.6 | 2.4 | 2.4 | - |
| | Coverage | 96% | 96% | 96% | 96% | - |
| | Average acuity (mm) | 3.0 | 3.0 | 2.8 | 2.8 | - |

| | | In-situ Static cam | In-situ Zoom cam | Post-hoc Static cam | Post-hoc Zoom cam | Post-hoc Superzoom |
|---|---|---|---|---|---|---|
| | **Best in class** | Parrot ANAFI USA GOV<br>Skydio X2D | Parrot ANAFI USA GOV<br>Skydio X2D | Flyability Elios 2 GOV<br>Parrot ANAFI USA GOV<br>Skydio X2D | Parrot ANAFI USA GOV<br>Skydio X2D | N/A |
| | | **In-situ, overall** | | **Post-hoc, overall** | | |
| | | Parrot ANAFI USA GOV<br>Skydio X2D | | Parrot ANAFI USA GOV<br>Skydio X2D | | |

*Note: for systems whose digital zoom did not appear to result in higher acuity, the metrics from the static cam test was copied over rather than run separately. For systems without zoom (optical or digital), the zoom cam condition was not run at all.

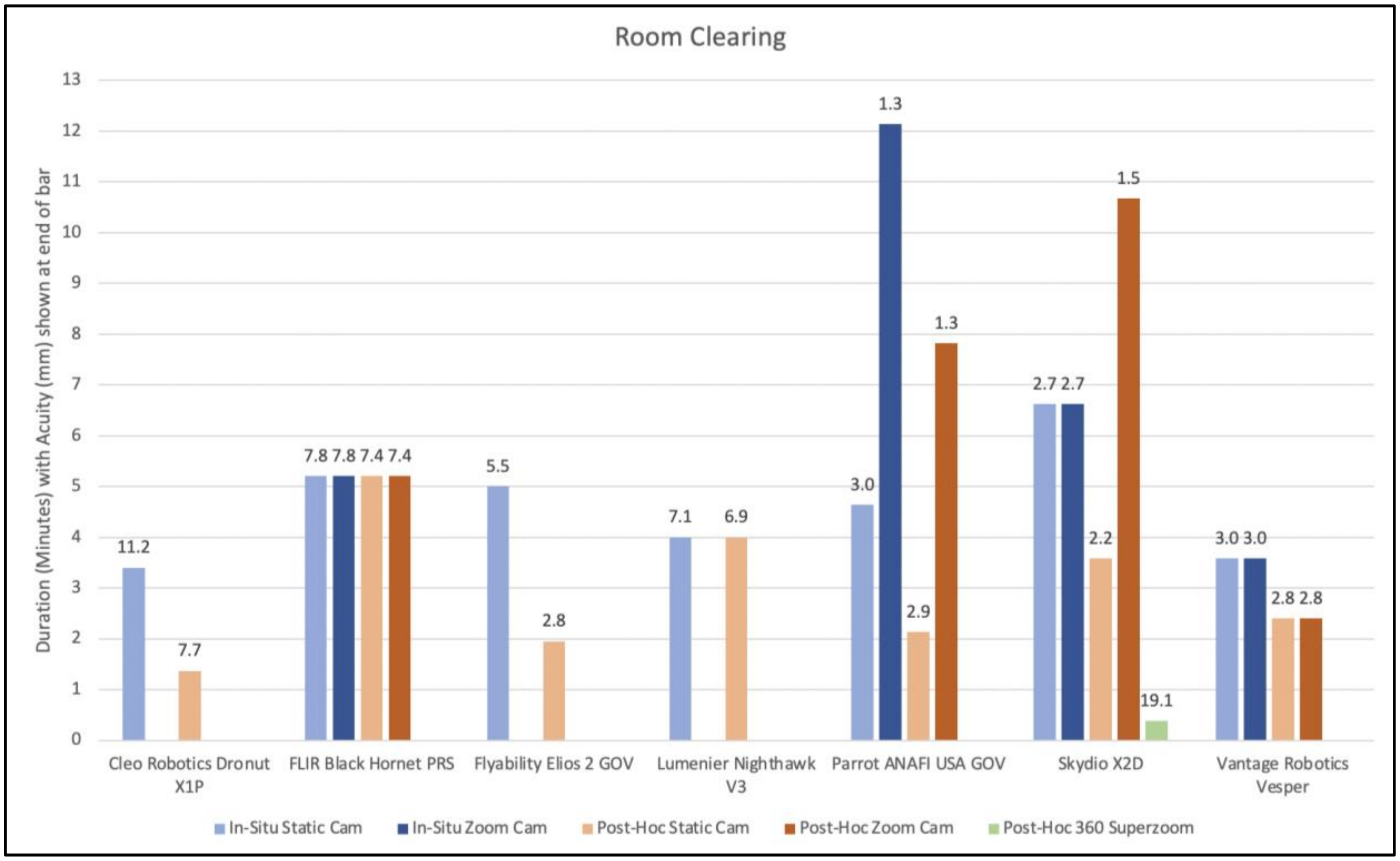

Room Clearing
Duration (Minutes) with Acuity (mm) shown at end of bar
0
1
2
3
4
5
6
7
8
9
10
11
12
13
11.2
7.7
7.8 7.8 7.4 7.4
5.5
2.8
7.1
6.9
3.0
1.3
2.9
1.3
2.7 2.7
2.2
1.5
19.1
3.0 3.0
2.8 2.8
Cleo Robotics Dronut X1P
FLIR Black Hornet PRS
Flyability Elios 2 GOV
Lumenier Nighthawk V3
Parrot ANAFI USA GOV
Skydio X2D
Vantage Robotics Vesper
In-Situ Static Cam
In-Situ Zoom Cam
Post-Hoc Static Cam
Post-Hoc Zoom Cam
Post-Hoc 360 Superzoom

# Indoor Noise Level

## Summary of Test Method

Evaluating the noise generated by each sUAS is measured simply by using a decibel meter to record the volume in comparison to a baseline (i.e., the environment without the sUAS operating) when operating indoors and outdoors. The relevant environment characteristics of each environment are recorded (e.g., indoor room dimensions, obstructions between the decibel meter and sUAS, outdoor terrain) in order to contextualize the results. The sUAS performs basic tasks (takeoff, hovering in place, yawing in place, pitching forward and back, rolling left and right, ascending and descending, camera movement, and landing) in each environment while the decibel meter is placed at a near distance of 3 m (10 ft) and a far distance of 20 m (65 ft) or greater. Minimum, maximum, and average decibel levels are reported for each condition.

## Benchmarking Results

Metrics in decibels (dB)

Best in class = noise level is less than 1 standard deviation of the average noise level across all systems per position and task

| sUAS | 2.5 m Hover | 2.5 m Ascension | 5 m Hover | 5 m Ascension | 10 m Hover | 10 m Ascension |
|---|---|---|---|---|---|---|
| Cleo Robotics Dronut X1P | 90 | 92.9 | 88.2 | 91 | 82.2 | 84.6 |
| FLIR Black Hornet PRS | 54.5 | 55.8 | 53.5 | 54 | 49.5 | 51 |
| Flyability Elios 2 GOV | 103 | 106.1 | 102 | 105.7 | 93 | 101.9 |
| Lumenier Nighthawk V3 | 80 | 85.5 | 79 | 81.9 | 70 | 73.7 |
| Parrot ANAFI USA GOV | 83 | 86.1 | 80 | 85 | 67 | 76.1 |
| Skydio X2D | 81.5 | 82.8 | 80 | 81.6 | 71 | 73.5 |
| Teal Golden Eagle* | X | X | X | X | X | X |
| Vantage Robotics Vesper | 85.5 | 88.6 | 84.2 | 86.7 | 80 | 97.6 |
| **Best in class** | **Indoor Noise Level, overall** | | | | | |
| | FLIR Black Hornet PRS | | | | | |

*Note: Due to instability of the Teal Golden Eagle when operating indoors, it was not evaluated.

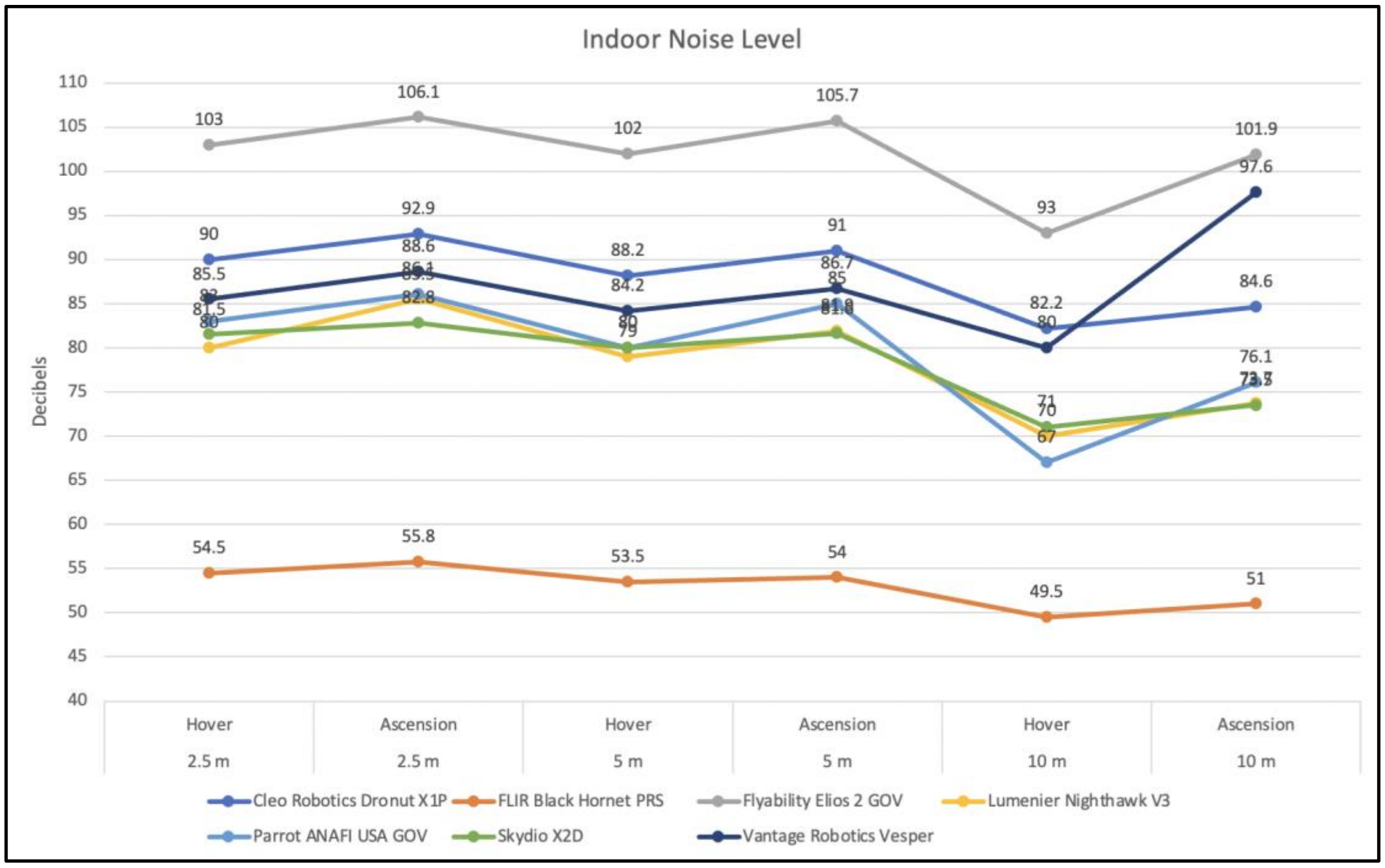
Indoor Noise Level
Decibels
Hover 2.5 m
Ascension 2.5 m
Hover 5 m
Ascension 5 m
Hover 10 m
Ascension 10 m
Cleo Robotics Dronut X1P
FLIR Black Hornet PRS
Flyability Elios 2 GOV
Lumenier Nighthawk V3
Parrot ANAFI USA GOV
Skydio X2D
Vantage Robotics Vesper

## Logistics Characterization

### Summary of Test Method

A series of characteristics concerning the logistics of operating, maintaining, and collecting data are outlined across seven categories that are to be filled with the relevant information for each sUAS platform being evaluated: physical measurements, power, heat dissipation, safety precautions, body/frame and maintenance, data collection and access, and system survivability. In each category, several fields are posed as prompts/questions for the user to respond to based on empirical evidence and experience of operating and maintaining the system. Some data may be able to be initially derived from vendor-provided specification sheets, but should be verified empirically. All fields are open response, although some may require a specific format of response (e.g., yes/no, dimension units, etc.). The information captured under each field is as follows:

- Physical measurements:
  - Deployed size, Collapsed size, Controller size, Table / Screen size, Dimensions of drone carry case, Weight of drone carry case w/o drone, Dimensions of controller carry case,
  - Weight of frame without battery, Battery weight, Weight of controller, Weight of controller carry case w/o drone, Dimensions of charger carry case, Weight of controller carry case w/o controller, Weight total
- Power:
  - Battery type, Battery charge time, Average flight with full battery, Controller battery type, Controller charge time, Can the power be switched on/off
  - Is battery level displayed to user, Is battery remaining time indicated, Is flight time remaining indicated, Is the user prompted about damaged battery, Are actions prevented at critical battery levels, Are there failsafe lockouts, Can user override failsafe lockouts, What behavior is exhibited at critical power Is battery connection easily accessible
- Heat dissipation and consideration:
  - Operation Temp range, Can the drone idle without overheating, Does it have internal or passive cooling, Is the user prompted on critical heat levels, What happens on overheat, Do batteries need to cool down after use
- Safety precautions:
  - Does operation require hearing protection, Does operation require eye protection, Does operation require head protection, Does operation suggest a respirator
- Frame and maintenance:
  - Are parts serviceable, Are parts reinforced or weatherized, Are parts custom made or off shelf, Is drone 3d printed or mass produced, Are parts interchangeable, Is drone stored fully assembled, Are propellers protected, Can prop guards be attached, Are tools provided with drone
- Data collection and access:
  - Does the drone start recording when armed, Is data stored onboard or in controller, How is data accessed, What format is data stored in, Is software required to process data
- System survivability:
  - Visual detectability, Audible signature, Cybersecurity/encryption
- Capabilities:
  - Does the drone have obstacle avoidance, Does the drone prevent hitting objects, Can obstacle avoidance be disabled, Does the drone have auto takeoff, Does the drone have an auto land, Does the drone have an emergency stop, Is the drone able to carry a payload

Across a set of sUAS platforms, their logistics characteristics can be compared. Additionally, criteria can be set for each characteristic to determine if a system meets the relevant requirements set forth by another entity. For example, soldier feedback provided to the Soldier-Borne Sensor (SBS) program indicated that a minimum of 2 hours of HD video be able to be recorded. This threshold of acceptable performance can be compared to the information provided for candidate sUAS in the *data collection & access* category. For a coarse representation of requirement matching, a percentage can be calculated of the number of fields that match a given set of criteria.

## Benchmarking Results

While the test method specification does allow for comparison against a set of requirements, no comprehensive set of criteria was provided to compare the sUAS characteristics against. So, no evaluations of the resulting characterizations were performed, aside from comparison to the other sUAS.

No best in class criteria is specified.

| sUAS | | |
|---|---|---|
| Make | Cleo Robotics | FLIR |
| Model | Dronut X1P (P = prototype) | Black Hornet PRS |
| sUAS Image | 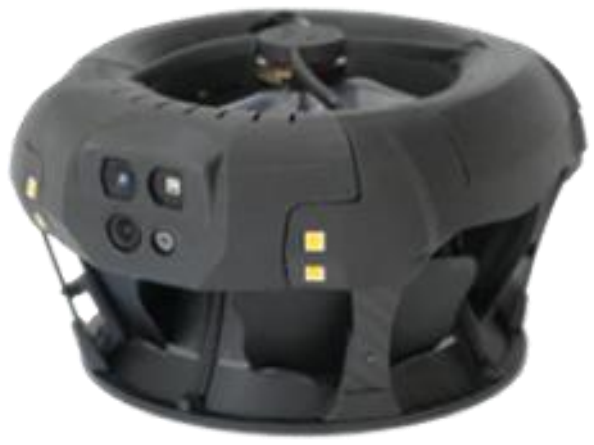 | 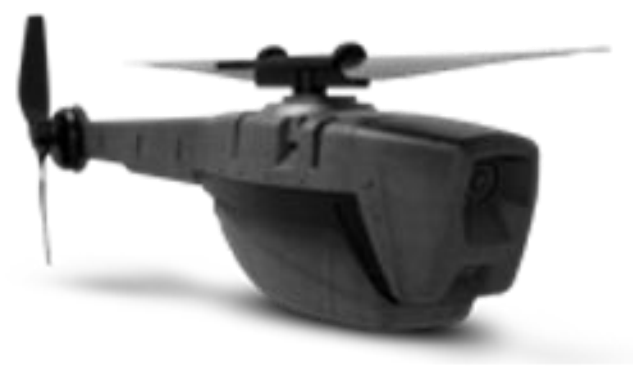 |
| **Physical measurements (inches and lbs)** | | |
| Deployed size | 6 x 6.75 x 4 | 7x 1.5 x 1 |
| Collapsed size | N/A | N/A |
| Weight of frame without battery lbs | 0.5 | 35 grams with battery |
| Battery weight lbs | 0.5 | 10 grams |
| Weight of controller lbs | 1 | 1 |
| Controller size | 6.5 x 8 x 4.5 | 8.5 x 4.5 x 1 |
| Tablet / Screen size | 6.25 x 3 | 3.5 x 6 |
| Dimensions of drone carry case | 14.5 x 10.5 x 6 | 7.5 x 3.5 x 2.25 (drone, controller, and battery charger carry cases are integrated as one unit) |
| Weight of drone carry case w/o drone lbs | 2 | 418 grams |
| Dimensions of controller carry case | N/A | 7.5 x 3.5 x 2.25 (drone, controller, and battery charger carry cases are integrated as one unit) |
| Weight of controller carry case w/o drone | N/A | N/A |
| Dimensions of charger carry case | N/A | 7.5 x 3.5 x 2.25 (drone, controller, and battery charger carry cases are integrated as one unit) |
| Weight of controller carry case w/o controller | N/A | N/A |
| Weight total | 4 | ~3 |
| **Power** | | |
| Battery type | Li-ion | UNKNOWN |
| Battery charge time | 1 hour | 20-30 min |
| Average flight with full battery | 9 min | 20 min |
| Controller battery type | Li-ion | UNKNOWN |
| Controller charge time | 1 Hour | N/A |
| Can the power be switched on/off | No | Yes |
| Are there failsafe lockouts | Yes | Yes |
| Can user override failsafe lockouts | No | Yes |
| What behavior is exhibited at critical power | Fails to maintain altitude | It will prompt the user to return. this can be overridden three times before autoland |
| Is battery connection easily accessible | No | Yes |
| **Heat dissipation and consideration** | | |
| Operation Temp range | Not Defined | Information unavailable |
| Can the drone idle without overheating | Yes | Yes |
| Does it have internal or passive cooling | Active | Passive |
| What happens on overheat | Dissociation from radio | Autoland or return to base, has not been tested much |
| Do batteries need to cool down after use | Yes | No |
| **Safety precautions** | | |
| Does the drone have an emergency stop | Yes | Yes |
| Does operation require hearing protection when operator is nearby | No | No |
| Does operation require eye protection when operator is nearby | No, props are covered but best to use anyway | No |
| Does operation require head protection when operator is nearby | Recommended | No |
| **Frame and maintenance** | | |
| Are parts serviceable | Yes* | Yes, but just rotors and battery |
| Are parts reinforced or weatherized | No | Yes |
| Are parts custom made or off shelf | Custom | Custom |
| Is drone 3D printed or mass produced | 3d printed parts | Mass produced |
| Are parts interchangeable | Yes* | Yes, but only rotors and batteries |
| Is drone stored fully assembled | Yes* | Yes |
| Are propellers protected | Yes | No |
| Can prop guards be attached | No | No |
| Are tools provided with drone | No, none needed for operation | N/A |
| Is the drone able to exert force on an object without crashing | Yes, conditional | No |
| **Data collection and access** | | |
| Does the drone start recording when armed | No | Yes, on controller |
| Is data stored onboard or in controller | No storage onboard | Only controller |
| How is data accessed | N/A | Must plug in external cable and download to computer from control station. |
| What format is data stored in | N/A | .ts |
| Is software required to process data | Yes | No, just permission to access |
| **System survivability** | | |
| Visual detectability - at what range does the drone become easily visible | It would be hard to detect at long range, but is intended for short range | The drone is very small and quiet, meaning that it is hard to detect at close range |
| Audible signature - at what range is the drone detectable | Short | Short |
| Cybersecurity - is the drone encrypted | No | No, does not store data |
| **Capabilities** | | |
| Does the drone have obstacle avoidance | No | No |
| Does the drone prevent hitting objects | No | No |
| Can obstacle avoidance be disabled | N/A | N/A |
| Does the drone have auto takeoff | Yes | Yes |
| Does the drone have an auto land | Yes | Yes |
| Does the drone have an emergency stop | Yes | No |
| Is the drone able to carry a payload | Yes | No, too small |

| sUAS | | |
|---|---|---|
| Make | Flyability | Lumenier |
| Model | Elios 2 GOV | Nighthawk V3 |
| sUAS Image | 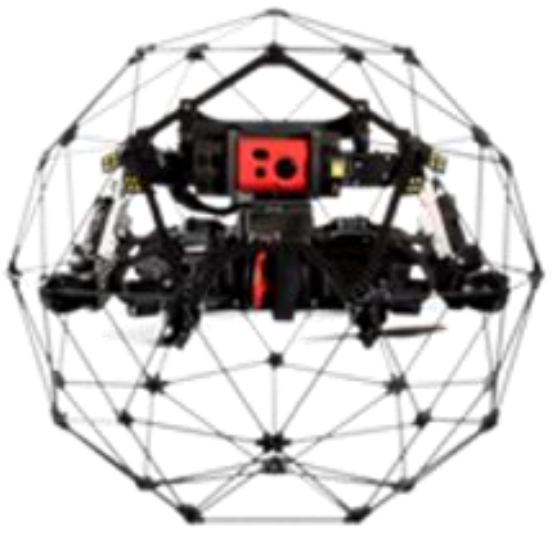 | 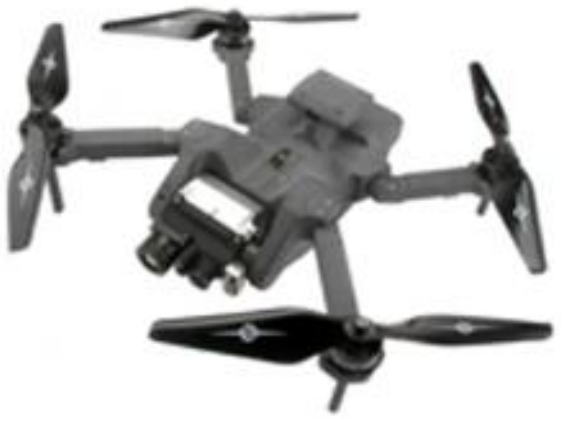 |
| **Physical measurements (inches and lbs)** | | |
| Deployed size | 15.5 x 15.5 x 14.5 | 20 x 17 x 3.5 |
| Collapsed size | N/A | 8.5 x 3.75 x 3.5 |
| Weight of frame without battery lbs | 2 | 1.5 |
| Battery weight lbs | 1 | 0.5 |
| Weight of controller lbs | 4.5 | 4.5 |
| Controller size | 8.25 x 8.25 x 5 | 10.2 x 5.25 x 2.31 |
| Tablet / Screen size | 9 x 8.5 x 3 | 6 x 3.5 |
| Dimensions of drone carry case | 22.75 x 16.75 x 19 | 18 x 21 x 9 |
| Weight of drone carry case w/o drone lbs | 40 | 10.5 |
| Dimensions of controller carry case | N/A | N/A |
| Weight of controller carry case w/o drone | N/A | N/A |
| Dimensions of charger carry case | N/A | 17 x 13 x 10 |
| Weight of controller carry case w/o controller | N/A | 10.5 |
| Weight total | 47.5 | 27.5 |
| **Power** | | |
| Battery type | Li-po | Li-ion |
| Battery charge time | 1 Hour | 1 hour |
| Average flight with full battery | 9 min | 10 min |
| Controller battery type | Li-ion | Li-ion |
| Controller charge time | 1 Hour | 1 Hour |
| Can the power be switched on/off | No | Yes |
| Are there failsafe lockouts | Yes | Yes |
| Can user override failsafe lockouts | Yes | Yes |
| What behavior is exhibited at critical power | Auto land | fails to maintain altitude |
| Is battery connection easily accessible | No | Yes |
| **Heat dissipation and consideration** | | |
| Operation Temp range | 50 to 86 F | Not Defined |
| Can the drone idle without overheating | No | Yes |
| Does it have internal or passive cooling | Passive | Passive |
| What happens on overheat | Motor and battery damage | Drone disconnects from radio |
| Do batteries need to cool down after use | Yes | Yes |
| **Safety precautions** | | |
| Does the drone have an emergency stop | Yes | Yes |
| Does operation require hearing protection when operator is nearby | Yes | No |
| Does operation require eye protection when operator is nearby | Yes | Yes |
| Does operation require head protection when operator is nearby | Recommended | Recommended |
| **Frame and maintenance** | | |
| Are parts serviceable | Yes | No |
| Are parts reinforced or weatherized | No | No |
| Are parts custom made or off shelf | Custom | Custom |
| Is drone 3D printed or mass produced | 3d printed parts | 3d printed parts |
| Are parts interchangeable | Yes | No |
| Is drone stored fully assembled | Yes | No, props must be removed |
| Are propellers protected | Yes | No |
| Can prop guards be attached | Not needed | No (?) |
| Are tools provided with drone | Yes | Yes |
| Is the drone able to exert force on an object without crashing | Yes | No |
| **Data collection and access** | | |
| Does the drone start recording when armed | Yes | Yes |
| Is data stored onboard or in controller | Both | Onboard |
| How is data accessed | SD card removal or direct connection to drone | SD card removal, tools required |
| What format is data stored in | .mov , .thm, .log | .ts |
| Is software required to process data | Yes | No |
| **System survivability** | | |
| Visual detectability - at what range does the drone become easily visible | The drone is easily detected due to its shape, size, lights, and the noise level of its propellers when in use | The drone would be hard to detect a medium to long range when at high ceiling |
| Audible signature - at what range is the drone detectable | Long | Medium |
| Cybersecurity - is the drone encrypted | No | No |
| **Capabilities** | | |
| Does the drone have obstacle avoidance | No | Yes, front and side avoidance |
| Does the drone prevent hitting objects | No | Yes |
| Can obstacle avoidance be disabled | N/A | Yes |
| Does the drone have auto takeoff | Yes, on critical state or comms loss | Yes |
| Does the drone have an auto land | Yes | No |
| Does the drone have an emergency stop | Yes | Yes |
| Is the drone able to carry a payload | Yes | Yes |

| sUAS | | |
|---|---|---|
| Make | Parrot | Skydio |
| Model | ANAFI USA GOV | X2D |
| sUAS Image | 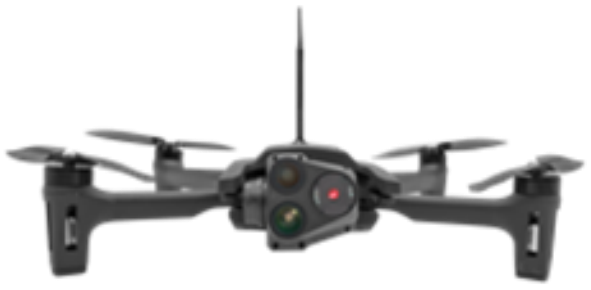 | 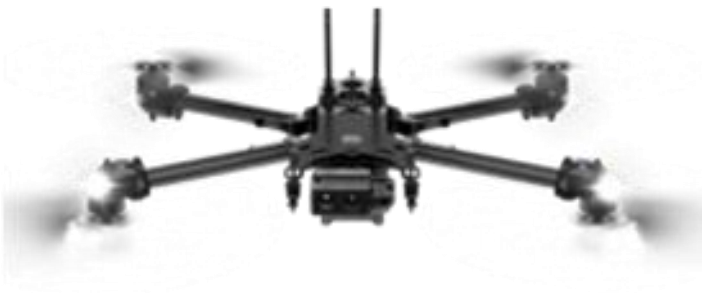 |
| **Physical measurements (inches and lbs)** | | |
| Deployed size | 15.5 x 12 x 3 | 25 x 22 x 8 |
| Collapsed size | 10 x 4 x 3 | 12 x 4.5 x 4.5 |
| Weight of frame without battery lbs | 0.5 | 2 |
| Battery weight lbs | 0.5 | 1 |
| Weight of controller lbs | 2.5 | 2.5 |
| Controller size | 8 x 12 x 2.25 | 11 x 5.5 x 3 |
| Tablet / Screen size | 4.5 x 7 | 6.25 x 3 |
| Dimensions of drone carry case | 17 x 14 x 7 | 14 x 22 x 9 |
| Weight of drone carry case w/o drone lbs | 9.5 | 10 |
| Dimensions of controller carry case | N/A | N/A |
| Weight of controller carry case w/o drone | N/A | N/A |
| Dimensions of charger carry case | N/A | N/A |
| Weight of controller carry case w/o controller | N/A | N/A |
| Weight total | 13 | 15.5 |
| **Power** | | |
| Battery type | Li-ion | Li-ion |
| Battery charge time | 1 hour | 1 hour |
| Average flight with full battery | 30 min | 35 min |
| Controller battery type | Li-ion | Li-ion |
| Controller charge time | 1 Hour | 1 Hour |
| Can the power be switched on/off | Yes | Yes |
| Are there failsafe lockouts | Yes | Yes |
| Can user override failsafe lockouts | Yes | No |
| What behavior is exhibited at critical power | Auto Land after timer | Auto land after timer |
| Is battery connection easily accessible | Yes | Yes |
| **Heat dissipation and consideration** | | |
| Operation Temp range | -32 to 120 F | -10 to 109 F |
| Can the drone idle without overheating | Yes | Yes |
| Does it have internal or passive cooling | Active | Active |
| What happens on overheat | Battery performance reduction, motor damage | Battery performance reduction, motor damage |
| Do batteries need to cool down after use | No | No |
| **Safety precautions** | | |
| Does the drone have an emergency stop | No | No |
| Does operation require hearing protection when operator is nearby | No, but noise is upsetting | No |
| Does operation require eye protection when operator is nearby | Yes | Yes |
| Does operation require head protection when operator is nearby | Recommended | Recommended |
| **Frame and maintenance** | | |
| Are parts serviceable | No | No |
| Are parts reinforced or weatherized | Yes | Yes |
| Are parts custom made or off shelf | Shelf | Shelf |
| Is drone 3D printed or mass produced | Mass produced | Mass produced |
| Are parts interchangeable | No, RMA needed | No, RMA needed |
| Is drone stored fully assembled | Yes, folded | Yes, folded |
| Are propellers protected | No | No |
| Can prop guards be attached | Yes | No |
| Are tools provided with drone | No, not needed | Yes |
| Is the drone able to exert force on an object without crashing | No | No |
| **Data collection and access** | | |
| Does the drone start recording when armed | Yes | Yes, but LVR only. 4k on request |
| Is data stored onboard or in controller | Onboard | Controller |
| How is data accessed | SD card removal or direct connection to drone | SD card removal or direct connection to drone |
| What format is data stored in | .Mp4 | .Lev, .Mov, .Thm |
| Is software required to process data | No | Yes |
| **System survivability** | | |
| Visual detectability - at what range does the drone become easily visible | Due to drones size and color it would be hard to detect a medium to long ranges close to the ground | Due to drones size and color it would be easy to detect at close to medium range, even at flight ceiling. if runner lights are on, it is easily detected |
| Audible signature - at what range is the drone detectable | Medium | Long |
| Cybersecurity - is the drone encrypted | Can be | Can Be |
| **Capabilities** | | |
| Does the drone have obstacle avoidance | No | Yes, 360 avoidance |
| Does the drone prevent hitting objects | No | Yes |
| Can obstacle avoidance be disabled | N/A | Yes, if you pay for functionality |
| Does the drone have auto takeoff | Yes | Yes |
| Does the drone have an auto land | Yes | Yes |
| Does the drone have an emergency stop | No | No |
| Is the drone able to carry a payload | Yes | Yes |

| sUAS | | |
|---|---|---|
| Make | Teal | Vantage Robotics |
| Model | Golden Eagle | Vesper |
| sUAS Image | 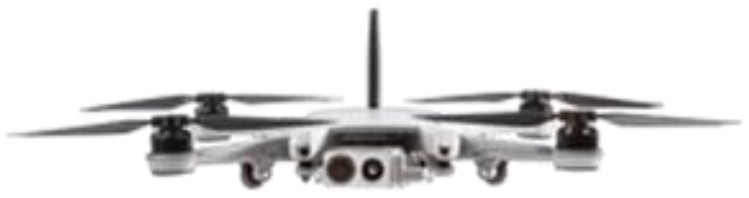 | 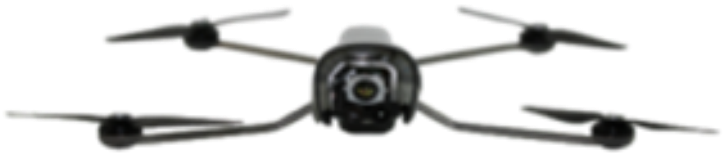 |
| **Physical measurements (inches and lbs)** | | |
| Deployed size | 19 x 17.5 x 2 | 14 x 13 x 3 |
| Collapsed size | 9.75 x 7.5 x 3.25 | 12.5 x 3 x 2.5 main body module |
| Weight of frame without battery lbs | 1.5 | 1 |
| Battery weight lbs | 1 | 0.5 |
| Weight of controller lbs | 3.5 | 2 |
| Controller size | 13.75 x 8 x 2.75 in | 9.75 x 5.5 x 1.5 |
| Tablet / Screen size | 8.75 x 5.5 | 6 x 3.75 |
| Dimensions of drone carry case | 16.25 x 13.25 x 7 | 21.5 x 17 x 10 |
| Weight of drone carry case w/o drone lbs | 11 | 15 |
| Dimensions of controller carry case | 16.25 x 13.25 x 5.5 in | N/A |
| Weight of controller carry case w/o drone | 10 | N/A |
| Dimensions of charger carry case | N/A | N/A |
| Weight of controller carry case w/o controller | N/A | N/A |
| Weight total | 27 | 18.5 |
| **Power** | | |
| Battery type | Li-ion | Li-ion |
| Battery charge time | 1 hour | 1 hour |
| Average flight with full battery | 15 min | 23 min |
| Controller battery type | Li-ion | Li-ion |
| Controller charge time | 1 hour | 1 Hour |
| Can the power be switched on/off | Yes | Yes |
| Are there failsafe lockouts | No | Yes |
| Can user override failsafe lockouts | No | No |
| What behavior is exhibited at critical power | Radio disconnection | Auto land or radio disconnection |
| Is battery connection easily accessible | Yes | Yes |
| **Heat dissipation and consideration** | | |
| Operation Temp range | -32 to 110 F | -32 to 95 F On Ground |
| Can the drone idle without overheating | No | Yes, but does heat up a lot |
| Does it have internal or passive cooling | Active | Passive |
| What happens on overheat | Disconnection from radio | Battery performance reduction, motor damage |
| Do batteries need to cool down after use | No | Yes |
| **Safety precautions** | | |
| Does the drone have an emergency stop | Yes | Yes |
| Does operation require hearing protection when operator is nearby | No | No |
| Does operation require eye protection when operator is nearby | Yes | Yes |
| Does operation require head protection when operator is nearby | Yes | Recommended |
| **Frame and maintenance** | | |
| Are parts serviceable | Yes | Yes |
| Are parts reinforced or weatherized | Yes | Yes |
| Are parts custom made or off shelf | Shelf | Shelf |
| Is drone 3D printed or mass produced | Mass produced | Mass produced |
| Are parts interchangeable | Yes | Yes |
| Is drone stored fully assembled | Yes, folded | Yes, but can be modularly taken apart |
| Are propellers protected | No, but have added prop guards | Yes, with shrouded prop module |
| Can prop guards be attached | Yes | No |
| Are tools provided with drone | Yes | No, none needed for operation |
| Is the drone able to exert force on an object without crashing | No | Yes |
| **Data collection and access** | | |
| Does the drone start recording when armed | No | No |
| Is data stored onboard or in controller | Onboard | Onboard |
| How is data accessed | SD card removal or direct connection to drone | Sd card removal |
| What format is data stored in | .ts | .Mp4 |
| Is software required to process data | No | No |
| **System survivability** | | |
| Visual detectability - at what range does the drone become easily visible | could be easily detected at close range, but would be harder to detect at long range or at its height ceiling. | The drone would be hard to detect a medium to long range when at high ceiling |
| Audible signature - at what range is the drone detectable | Medium | Medium |
| Cybersecurity - is the drone encrypted | Can be | No |
| **Capabilities** | | |
| Does the drone have obstacle avoidance | No | No |
| Does the drone prevent hitting objects | No | No |
| Can obstacle avoidance be disabled | N/A | N/A |
| Does the drone have auto takeoff | Yes | Yes |
| Does the drone have an auto land | No | Yes |
| Does the drone have an emergency stop | Yes | Yes |
| Is the drone able to carry a payload | Yes | Yes |

# Interface

## Operator Control Unit (OCU) Characterization

### Summary of Test Method

A series of characteristics concerning the OCU, its input functionalities through the controller, and the output provided via display modalities on the interface are outlined across five categories that are to be filled with the relevant information for each sUAS platform being evaluated: controller and UI, power, navigation, camera, and additional functionality and accessories. In each category, several fields are posed as prompts/questions for the user to respond to based on empirical evidence and experience of operating and maintaining the system. Some data may be able to be initially derived from vendor-provided specification sheets, but should be verified empirically. All fields are open response, although some may require a specific format of response (e.g., yes/no, dimension units, etc.). The information captured under each field is as follows:

- Controller and UI:
    - Is the controller labeled, How many non virtual buttons are there
    - Do flight modes change the configuration of how functionality is mapped to the controller inputs
    - How is the user alerted to critical states, Is flight information fused with nav page, Do settings reset on power cycle
    - Does the drone have obstacle avoidance, Does the drone prevent hitting objects, Can obstacle avoidance be disabled, Are obstacle avoidance notifications shown
    - Does the drone have an auto land, Does the drone have an emergency stop, Are some features disabled in specific modes, Controller display lighting
- Power:
    - Is battery remaining time indicated, Is flight time remaining indicated, Is the user prompted about damaged battery, Are actions prevented at critical battery levels
- Communications link:
    - Does the interface display the current comms link connection level Is the user prompted about reduced comms link Is the user prompted about loss of comms link What happens on comms loss Does the drone alert the user to magnetic interference
- Navigation:
    - What type of navigation system is used, Is the drone GPS capable, Is mapping data displayed during flight, Is the drone GPS Denied compatible, Does its behavior change without GPS
    - Is touch screen used during flight
    - Maximum wind resistance, Do flight modes change wind resistance
    - Do flight modes change obstacle avoidance, Does the drone switch modes automatically, Are critical environmental conditions alerted, Is the drone able to hover in place w/o input
- Camera:
    - Are cameras usable when not armed, Is there a thermal camera
    - Is information fused on nav footage, Is nav cam always visible in menus in flight
    - Is zoom digital or physical
- Additional functionality and accessories:
    - Does the drone have illuminators, Does the drone have IR sensors / emitters, Does the drone have a laser or pointer, Is the drone able to carry a payload

Across a set of sUAS platforms, their OCU characteristics can be compared. Additionally, criteria can be set for each characteristic to determine if a system meets the relevant requirements set forth by another entity. For example, soldier feedback provided to the Soldier-Borne Sensor (SBS) program indicated that, ideally, a system's OCU functionality should not change when operating with GPS or when GPS-denied. This threshold of acceptable performance can be compared to the information provided for candidate sUAS in the *navigation* category. For a coarse representation of requirement matching, a percentage can be calculated of the number of fields that match a given set of criteria.

## Benchmarking Results

While the test method specification does allow for comparison against a set of requirements, no comprehensive set of criteria was provided to compare the sUAS characteristics against. So, no evaluations of the resulting characterizations were performed, aside from comparison to the other sUAS.

No best in class criteria is specified.

| **sUAS info** | | |
|---|---|---|
| Make | Cleo Robotics | FLIR |
| Model | Dronut X1P | Black Hornet PRS |
| GCS/OCU Image | 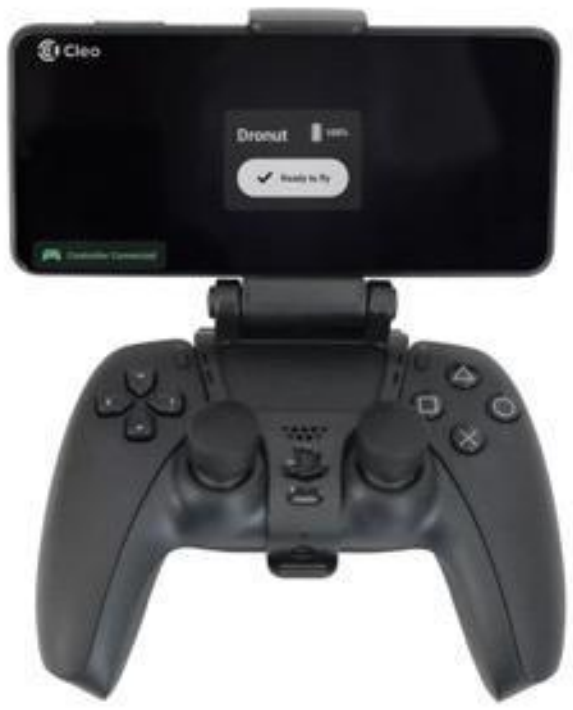  | 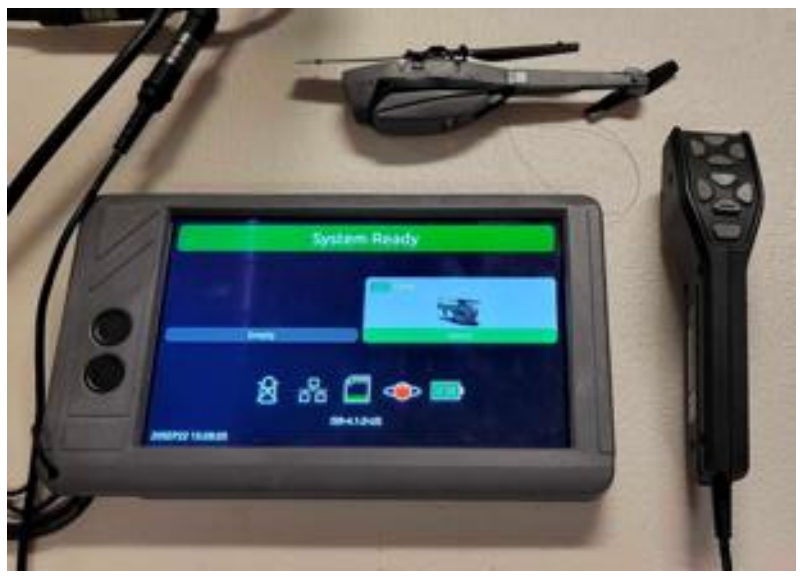  |
| | 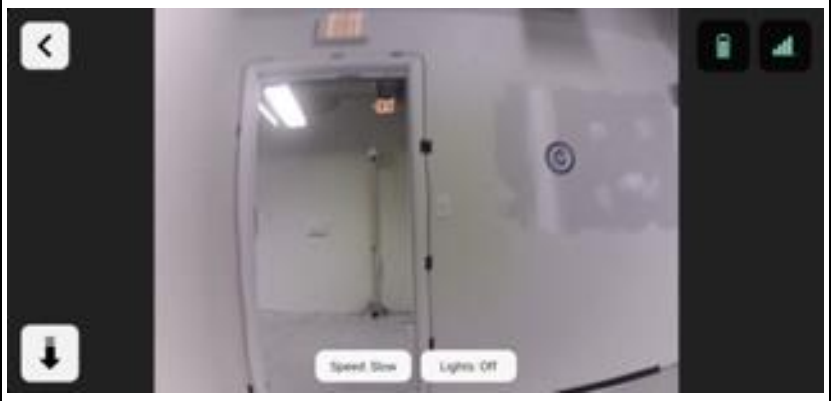 | 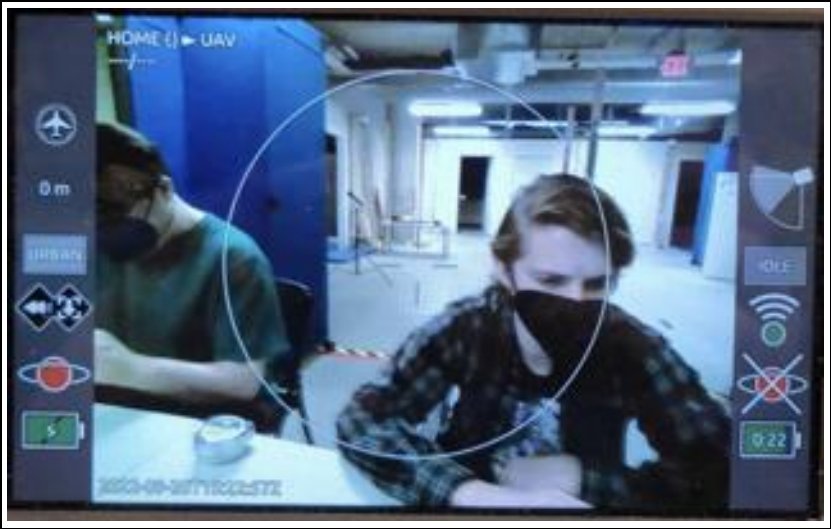  |
| **Controller and UI** | | |
| Is the controller labeled | No | No |
| Is there an onboard manual / control guide | No | Yes |
| How many physical buttons are there | 2 | 13 |
| Does the controller use a touch screen? | Yes | No |
| Does the controller use a touch screen to start flight | Yes | N/A |
| Is use of the touch screen required during flight | Yes | N/A |
| Is the touch screen easily responsive | Yes | N/A |
| Do flight modes change the configuration of how functionality is mapped to the controller inputs | No | No |
| How is the user alerted to critical states | Prompts in center screen 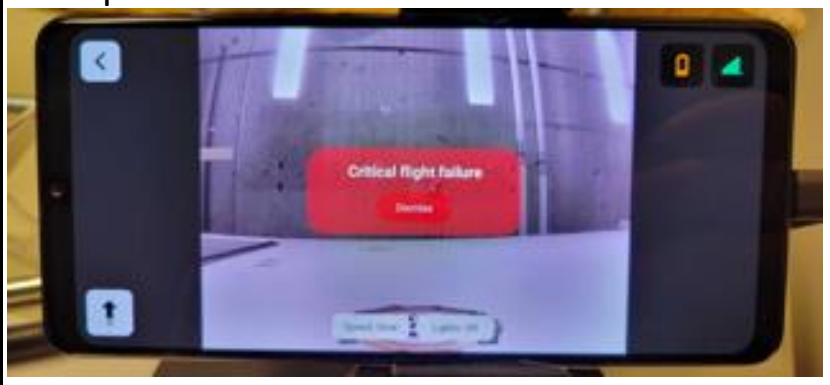  | Pop ups that can be overridden, sides of UI have indicators |
| Do settings reset on power cycle | Yes | Yes, on drone. all data on controller |
| Are obstacle avoidance notifications shown | No | N/A |
| If the drone has an auto-land, how/when is it engaged | Yes | Yes, at low battery but has override |
| If the drone has auto takeoff, is it a physical or a virtual button? | Virtual, drone ascends on arming command | Physical |
| If the drone has an emergency stop that can be initiated by the user, is it a physical or a virtual button? | Physical, pressing down both control sticks Estops drone | Physical, button combo disarmed when held |
| Are some features disabled in specific modes | No | Yes |
| Controller display lighting | Yes | Yes |
| **Power** | | |
| Is battery remaining time indicated | Yes, Bar 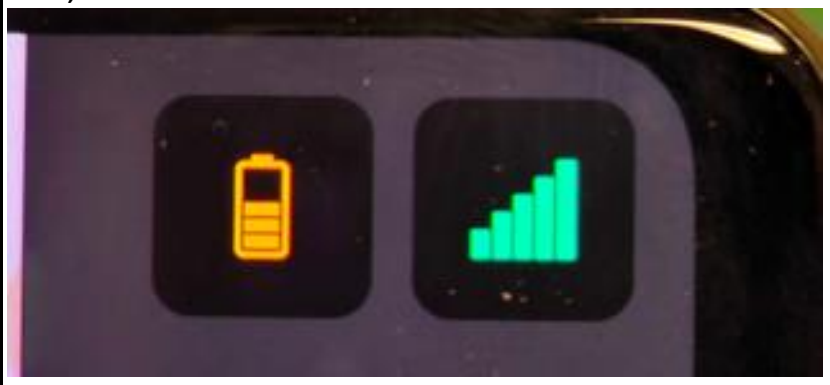 | Yes, bar and time 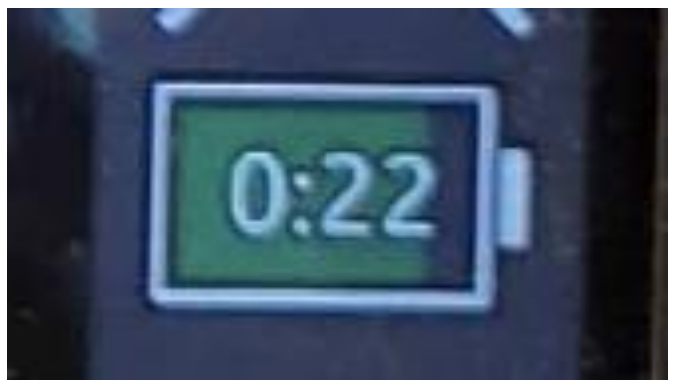  |
| Is flight time remaining indicated | No | Yes, Time is indicated and will give warnings at low battery levels |
| Is the user prompted about damaged battery | No | No |
| Are actions prevented at critical battery levels | No | Yes |
| **Communications link** | | |
| Does the interface display the current comms link connection level | Yes, bar 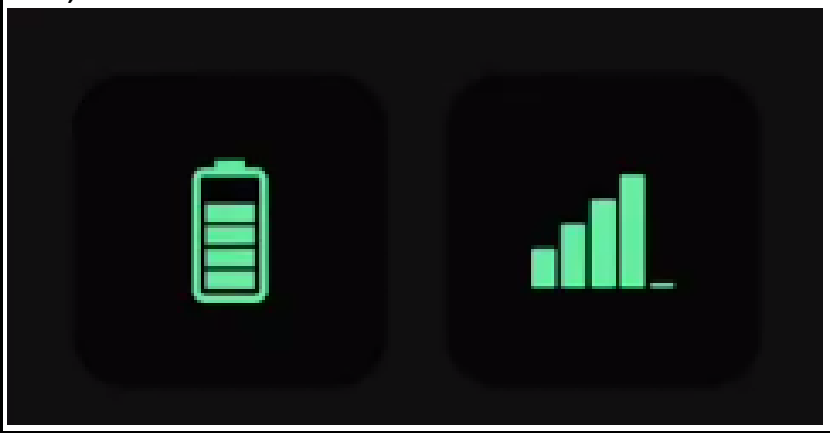 | Yes, 3 rounded bar 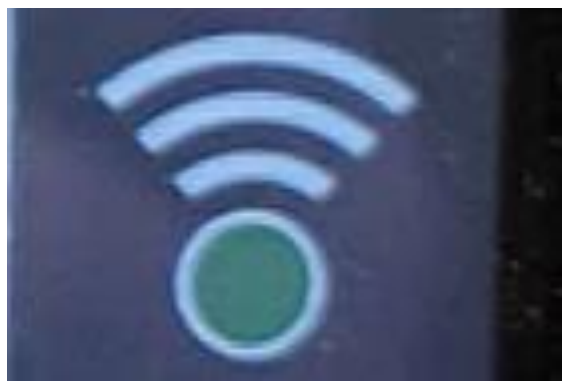 |
| Is the user prompted about reduced comms link | No | Yes, overlay text |
| Is the user prompted about loss of comms link | Yes, will blank out UI with overlay text that says disconnected 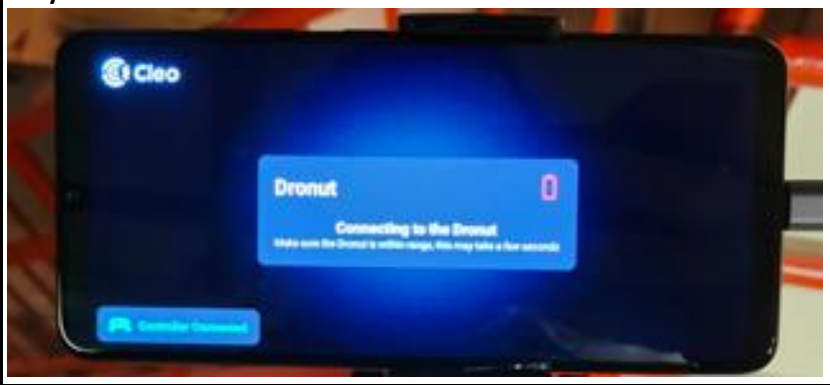  | Yes, overlay text notification |
| What happens on comms loss | Follows last given command until connection regained | Return home, or auto land |
| Does the drone alert the user to magnetic interference | No | No |
| **Navigation** | | |
| Is 3D mapping data displayed during flight | No | No |
| Does its behavior change without GPS | No | Yes, VIO more susceptible to lower light and drone will move slower |
| Is the user prompted about loss of GPS | N/A | Yes, UI gps indicator on side of screen |
| Is touch screen used during flight | Yes, to land or change settings | N/A |
| Do flight modes change obstacle avoidance | No | N/A |
| Does the drone switch modes automatically | No, user requested | Yes |
| Are critical environmental conditions alerted | No | Yes, UI popup notification |

| Camera | | |
|---|---|---|
| Are cameras usable when not armed | Yes | Yes |
| Do the navigation camera streams also display UI elements (e.g., map, system status) | No | N/A - no data stored on drone, all footage is image of controller |
| Is nav cam always visible in menus during flight | Yes | No |
| can the camera zoom, is it digital or optical | N/A | Digital |
| Are there multiple cameras used for varied zoom levels | No | Yes |
| Expected Horizontal Field of View of main camera model Degrees | 100 | Information unavailable |
| Expected Horizontal Field of View of thermal camera model Degrees | N/A | Information unavailable |
| Horizontal Field of View of main nav camera Degrees | 80 | 85 |
| Vertical Field of View of main nav camera Degrees | 59 | 69 |
| Gimbal Range Up Degrees | N/A | N/A |
| Gimbal Range Down Degrees | N/A | N/A |
| Field of Regard of main nav camera Up Degrees | 80 | 85 |
| Field of Regard of main nav camera Down Degrees | 59 | 69 |
| Thermal Horizontal Field of View Degrees | N/A | N/A |
| Thermal Vertical Field of View Degrees | N/A | N/A |
| Field of Regard of Thermal camera UP Degrees | N/A | N/A |
| Field of Regard of Thermal camera Down Degrees | N/A | N/A |
| **Additional functionality and accessories** | | |
| Does the drone have illuminators | Yes | No |
| Does the drone have a laser or pointer | No | No |

| sUAS info | | |
|---|---|---|
| Make | Flyability | Lumenier |
| Model | Elios 2 GOV | Nighthawk V3 |
| GCS/OCU Image | 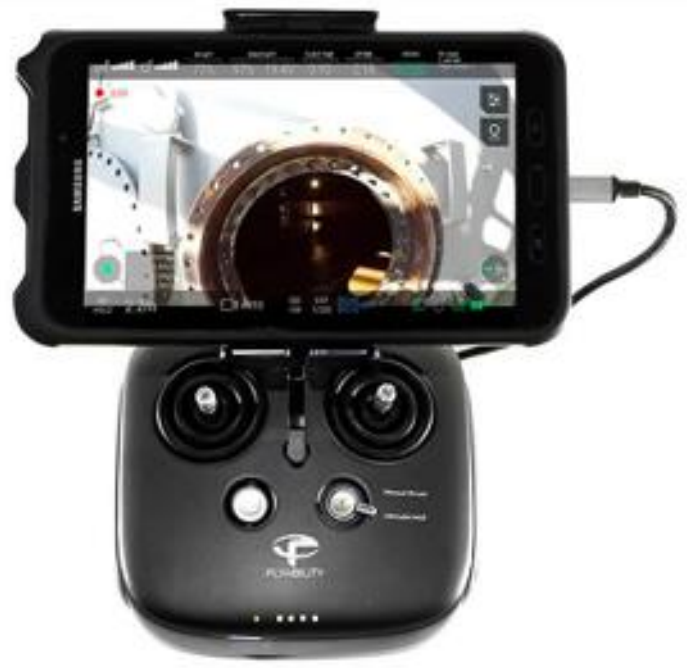 | 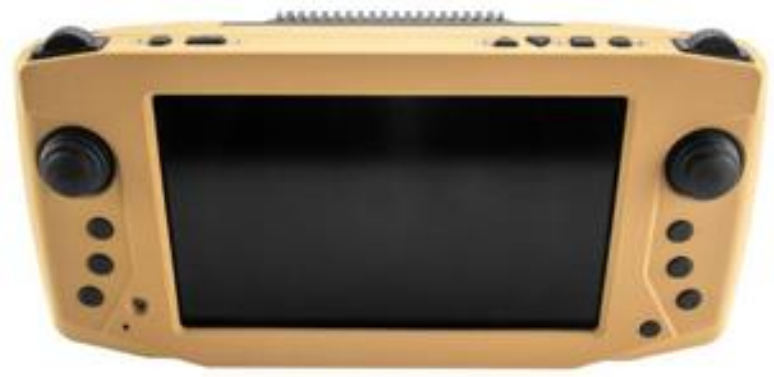 |
| | 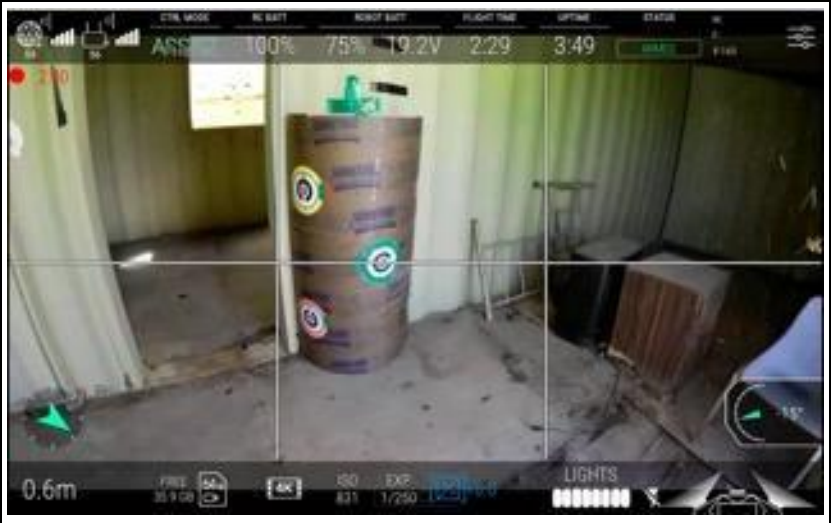 | 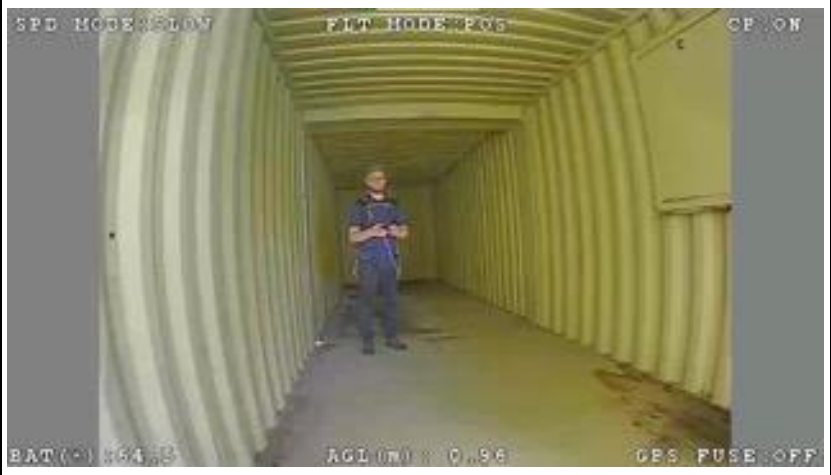 |
| **Controller and UI** | | |
| Is the controller labeled | Partially | No |
| Is there an onboard manual / control guide | No | No |
| How many physical buttons are there | 16 | 12 |
| Does the controller use a touch screen? | Yes | Yes |
| Does the controller use a touch screen to start flight | Yes | Yes |
| Is use of the touch screen required during flight | No | No |
| Is the touch screen easily responsive | Yes | No |
| Do flight modes change the configuration of how functionality is mapped to the controller inputs | Yes | No |
| How is the user alerted to critical states | Push prompts on screen and color changes of the screen border | QGC command line, but not visible during flight |
| Do settings reset on power cycle | Yes | Yes |
| Are obstacle avoidance notifications shown | N/A | No |
| If the drone has an auto-land, how/when is it engaged | Yes, but only at critical battery | No, but can detect landing if on ground and throttle is pulled down |
| If the drone has auto takeoff, is it a physical or a virtual button? | N/A | N/A |
| If the drone has an emergency stop that can be initiated by the user, is it a physical or a virtual button? | Physical, pattern on control sticks | Physical, arming button is also used to Estop and disarm |
| Are some features disabled in specific modes | Yes | Yes, OA turns off at higher speeds |
| Controller display lighting | Yes | Yes |
| **Power** | | |
| Is battery remaining time indicated | Yes, % 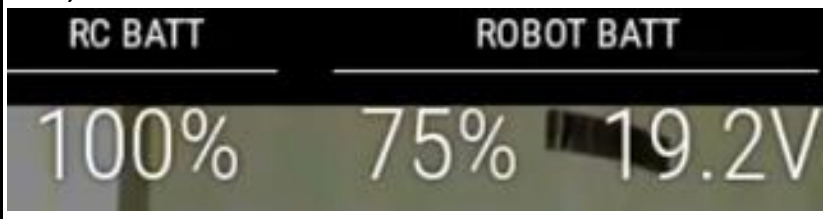  | Yes, % 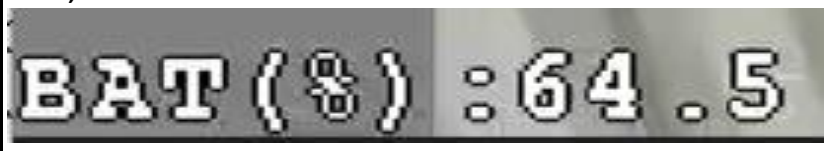  |
| Is flight time remaining indicated | Yes, timer 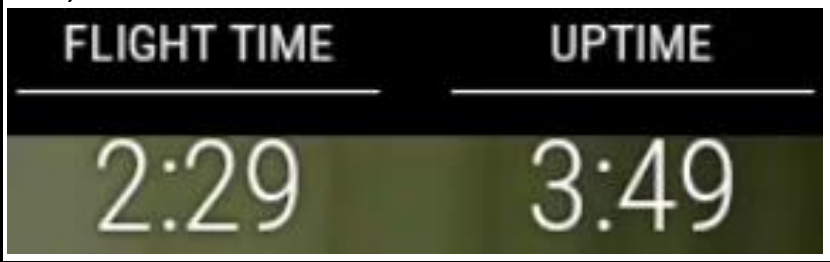  | No |
| Is the user prompted about damaged battery | Yes, batteries track flights, usage, and expiration 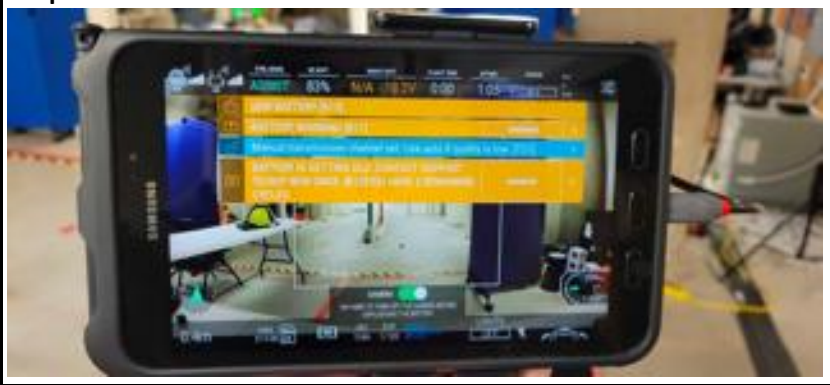 | No |
| Are actions prevented at critical battery levels | Yes | No |
| **Communications link** | | |
| Does the interface display the current comms link connection level | Yes, bar 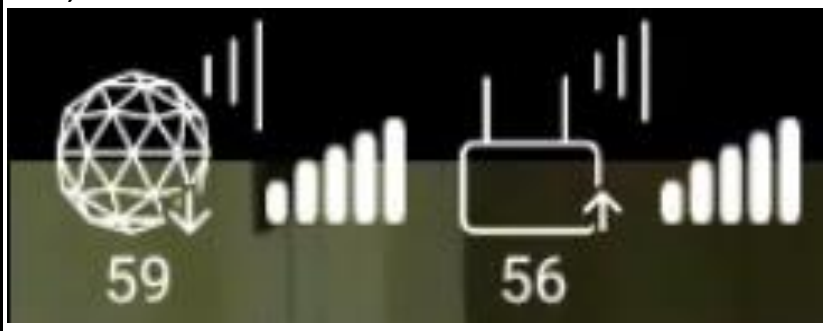  | No |
| Is the user prompted about reduced comms link | Yes, push notification about strong or weak signal 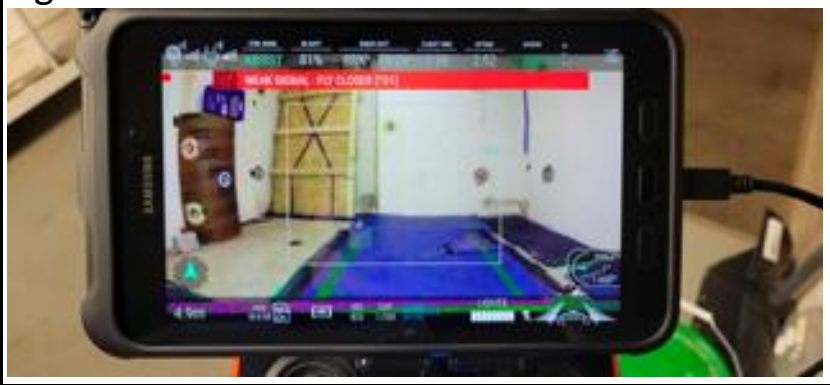 | No |
| Is the user prompted about loss of comms link | Yes, will say communications loss and show UI but blank, loads back to splash page | Yes, after reconnect the controller supplies a popup saying comms lost 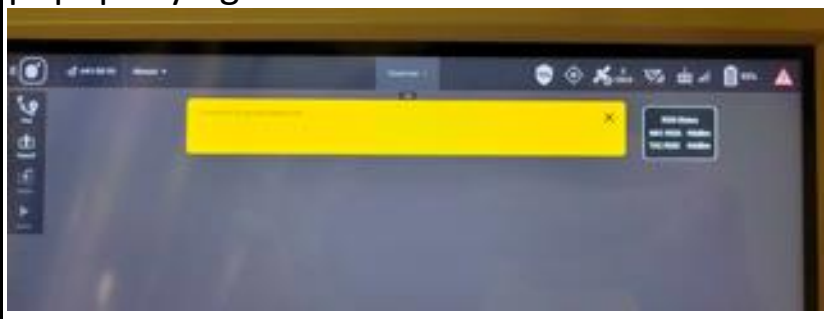 |
| What happens on comms loss | Auto land on full coms loss, idles props to cool system | Lands and disarms |
| Does the drone alert the user to magnetic interference | No* | No |
| **Navigation** | | |
| Is 3D mapping data displayed during flight | No | No |
| Does its behavior change without GPS | No | Yes, harder to hold position due to drift |
| Is the user prompted about loss of GPS | N/A | Yes, UI gps indicator on side of screen |
| Is touch screen used during flight | No, after initialization, unless changing settings | No |
| Do flight modes change obstacle avoidance | No | Yes, at different speeds the drone will increase its avoidance distance but turn it off at highest speed |
| Does the drone switch modes automatically | No, user requested | No, user requested |
| Are critical environmental conditions alerted | Yes, popup notification | No |
| **Camera** | | |
| Are cameras usable when not armed | Yes | Yes |
| Do the navigation camera streams also display UI elements (e.g., map, system status) | Yes | No |
| Is nav cam always visible in menus during flight | Yes | No |
| can the camera zoom, is it digital or optical | Digital | Digital |
| Are there multiple cameras used for varied zoom levels | No | No |
| Expected Horizontal Field of View of main camera model Degrees | 114 | Undefined in documentation |
| Expected Horizontal Field of View of thermal camera model Degrees | 56 | Undefined in documentation |
| Horizontal Field of View of main nav camera Degrees | 123 | 102 |
| Vertical Field of View of main nav camera Degrees | 77 | 74 |
| Gimbal Range Up Degrees | 90 | 75 |
| Gimbal Range Down Degrees | 90 | 75 |
| Field of Regard of main nav camera Up Degrees | 128 | 112 |
| Field of Regard of main nav camera Down Degrees | 128 | 112 |
| Thermal Horizontal Field of View Degrees | 60 | 40 |
| Thermal Vertical Field of View Degrees | 43 | 35 |
| Field of Regard of Thermal camera UP Degrees | 122 | 92 |
| Field of Regard of Thermal camera Down Degrees | 122 | 92 |
| **Additional functionality and accessories** | | |
| Does the drone have illuminators | Yes | Yes |
| Does the drone have a laser or pointer | No | Yes |

| sUAS info | | |
|---|---|---|
| Make | Parrot | Skydio |
| Model | ANAFI USA GOV | X2D |
| GCS/OCU Image | 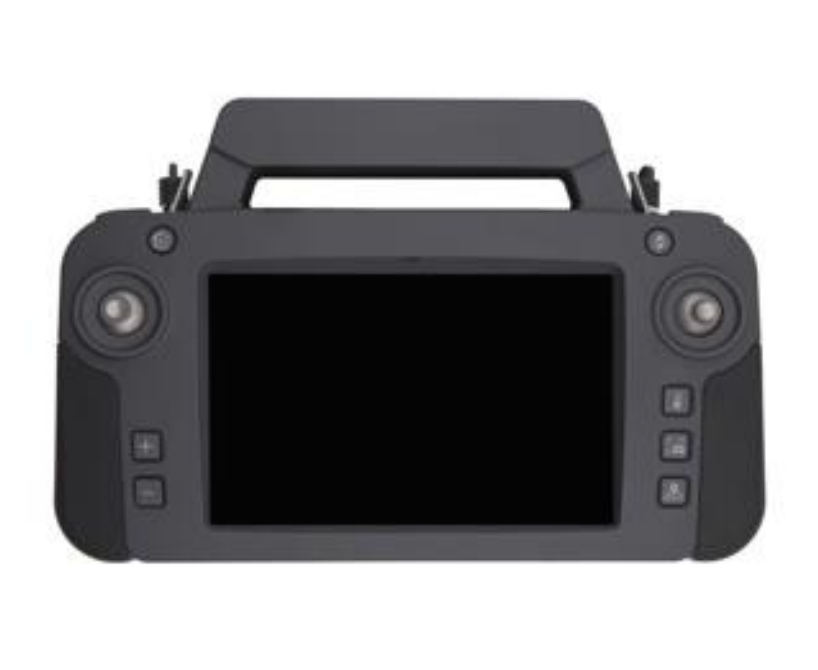 | 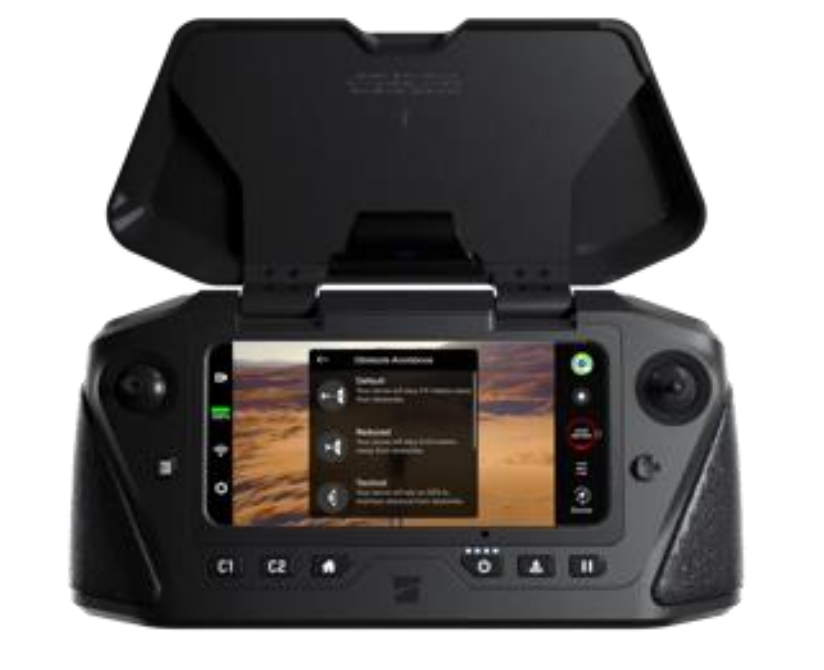 |
| | 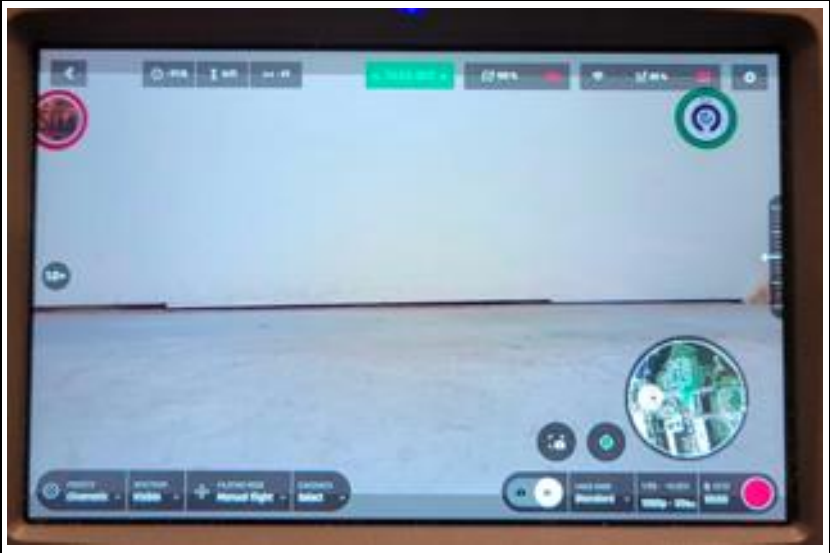 | 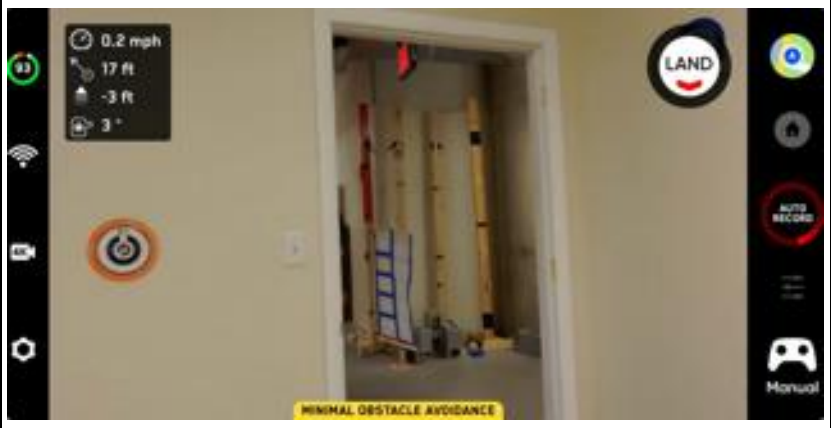 |
| **Controller and UI** | | |
| Is the controller labeled | Yes | Yes |
| Is there an onboard manual / control guide | Yes | No |
| How many physical buttons are there | 12 | 16 |
| Does the controller use a touch screen? | Yes | Yes |
| Does the controller use a touch screen to start flight | Yes | Yes |
| Is use of the touch screen required during flight | No | No |
| Is the touch screen easily responsive | Yes | Yes |
| Do flight modes change the configuration of how functionality is mapped to the controller inputs | No | No |
| How is the user alerted to critical states | Prompts at the top and bottom of the screen depending on the type of notification | Prompts on screen at top or center of screen 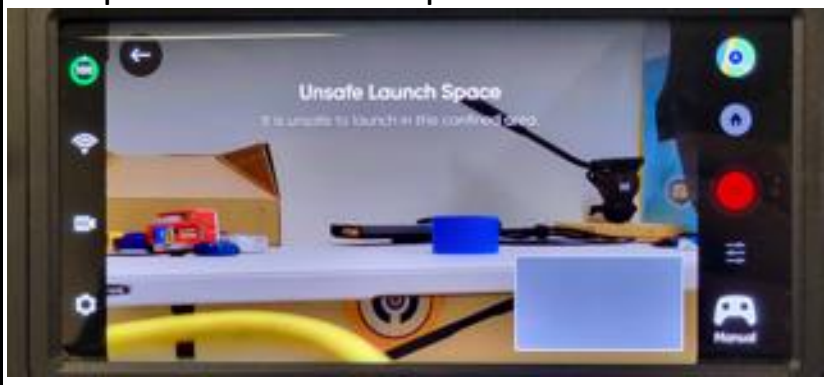 |
| Do settings reset on power cycle | Yes | Yes |
| Are obstacle avoidance notifications shown | N/A | Yes, mode is also displayed at bottom of screen 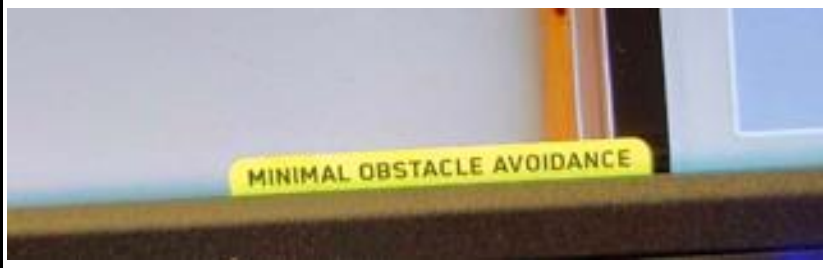 |
| If the drone has an auto-land, how/when is it engaged | Yes, and will auto land if close to ground and throttle is pulled down | Yes |
| If the drone has auto takeoff, is it a physical or a virtual button? | Both, Drone forces takeoff on arming and must ascend | Both, Drone forces takeoff on arming and must ascend |
| If the drone has an emergency stop that can be initiated by the user, is it a physical or a virtual button? | N/A | N/A |
| Are some features disabled in specific modes | Yes | Yes, night mode turns off OA |
| Controller display lighting | Yes | Yes |
| **Power** | | |
| Is battery remaining time indicated | Yes, % 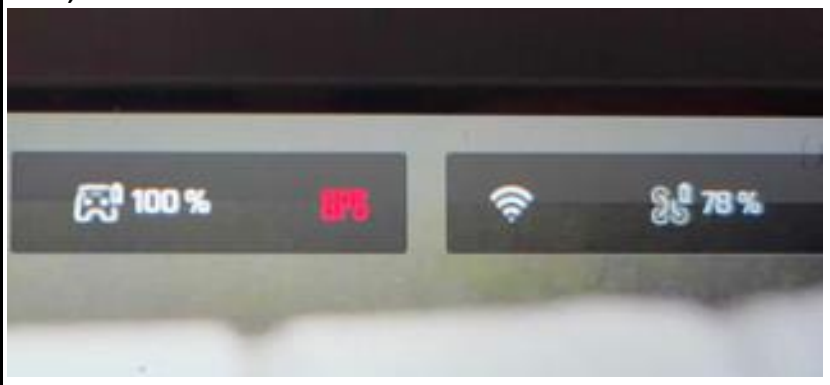  | Yes, % and circle graph 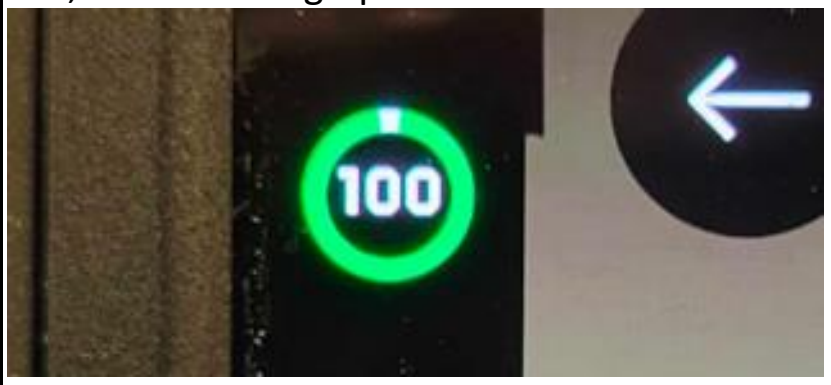  |
| Is flight time remaining indicated | No, but a timer is given at critical battery until autoland | Yes, once at low battery, system displays time over battery gauge |
| Is the user prompted about damaged battery | Yes, if critical damage, will display on calibration | Yes, if critical damage, will display on calibration |
| Are actions prevented at critical battery levels | Yes | Yes |
| **Communications link** | | |
| Does the interface display the current comms link connection level | Yes, rounded bar 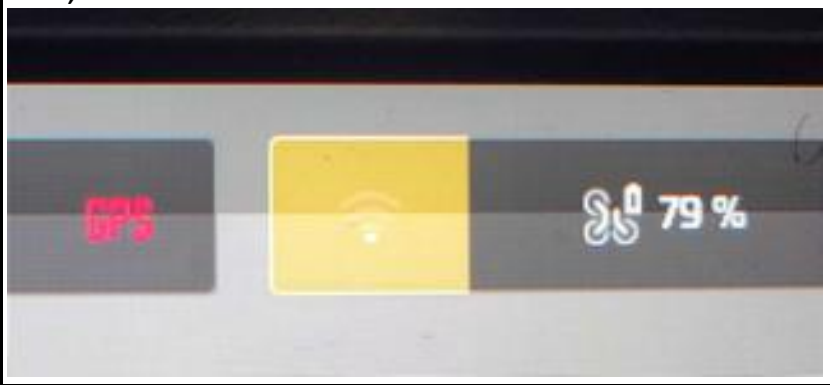  | Yes, rounded bar 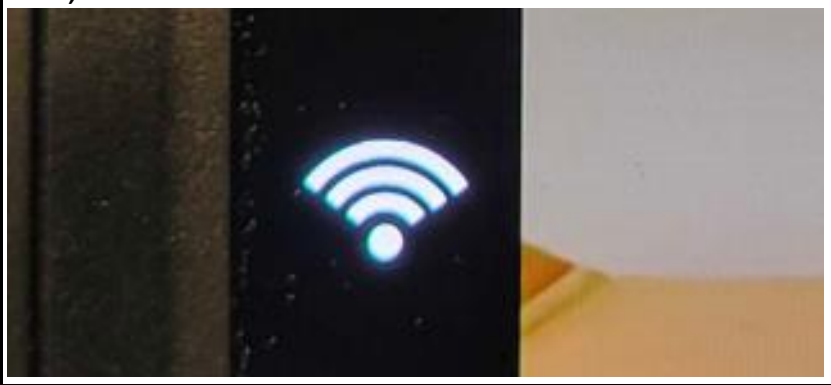 |
| Is the user prompted about reduced comms link | Yes, border notification about reduced communications | Yes, overlay text notification |
| Is the user prompted about loss of comms link | Yes, will show communication loss prompt at top and UI will go to static 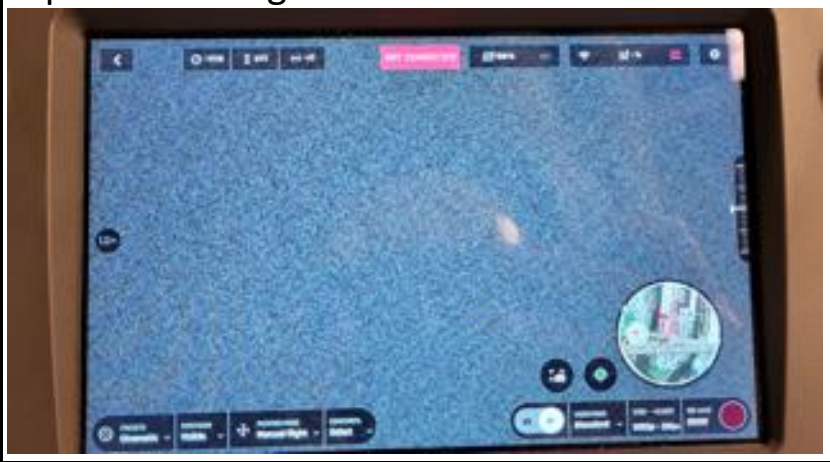 | Yes, will say the drone is too far away and to move closer, or goes to a full disconnect screen |
| What happens on comms loss | Lands and disarms | Lands and disarms |
| Does the drone alert the user to magnetic interference | No* | Yes* |
| **Navigation** | | |
| Is 3D mapping data displayed during flight | No | No, but preview of 3d scan is available |
| Does its behavior change without GPS | Yes, some mods disabled | Yes, some mods disabled |
| Is the user prompted about loss of GPS | Yes, border notification | Yes, if gps flight mode is on, the drone will Estop or land |
| Is touch screen used during flight | No, after initialization, unless changing settings | Yes, can be to direct camera or change settings |
| Do flight modes change obstacle avoidance | No | Yes, night made turns if off |
| Does the drone switch modes automatically | No, user requested | Yes, for night mode |
| Are critical environmental conditions alerted | Yes, border notification | Yes, overlay text notifications |

| Camera | | |
|---|---|---|
| Are cameras usable when not armed | Yes | No |
| Do the navigation camera streams also display UI elements (e.g., map, system status) | Yes | Yes |
| Is nav cam always visible in menus during flight | Yes | No |
| can the camera zoom, is it digital or optical | Digital | Digital |
| Are there multiple cameras used for varied zoom levels | Yes | Yes |
| Expected Horizontal Field of View of main camera model Degrees | 84 | 46 |
| Expected Horizontal Field of View of thermal camera model Degrees | 50 | Undefined in documentation |
| Horizontal Field of View of main nav camera Degrees | 72 | 34 |
| Vertical Field of View of main nav camera Degrees | 43 | 23 |
| Gimbal Range Up Degrees | 90 | 91 |
| Gimbal Range Down Degrees | 90 | 89 |
| Field of Regard of main nav camera Up Degrees | 112 | 103 |
| Field of Regard of main nav camera Down Degrees | 112 | 103 |
| Thermal Horizontal Field of View Degrees | 47 | 23 |
| Thermal Vertical Field of View Degrees | 40 | 13 |
| Field of Regard of Thermal camera UP Degrees | 110 | 98 |
| Field of Regard of Thermal camera Down Degrees | 110 | 96 |
| **Additional functionality and accessories** | | |
| Does the drone have illuminators | No | No, but does have runner lights |
| Does the drone have a laser or pointer | No | No |

| sUAS info | | |
|---|---|---|
| Make | Teal | Vantage Robotics |
| Model | Golden Eagle | Vesper |
| GCS/OCU Image | 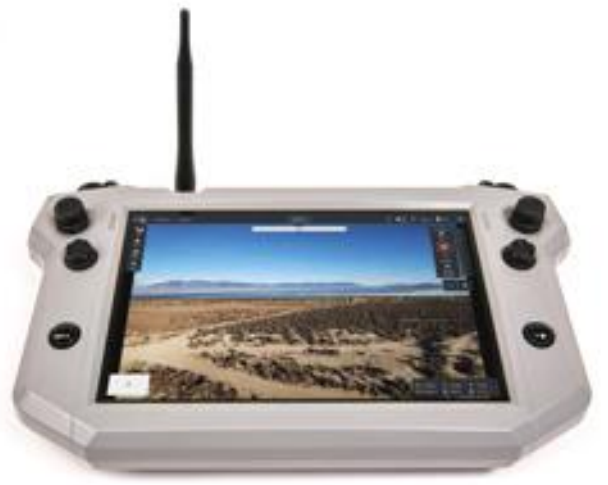 | 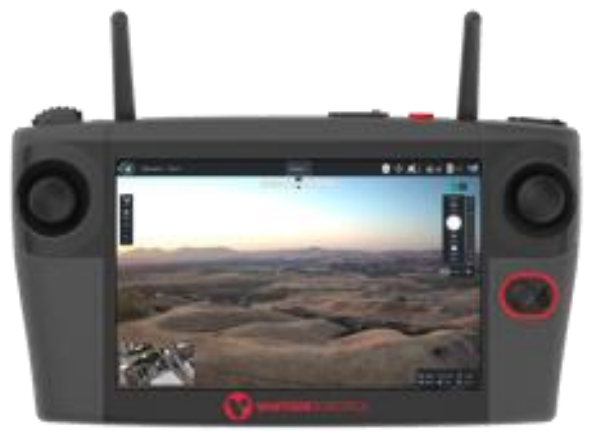 |
| | 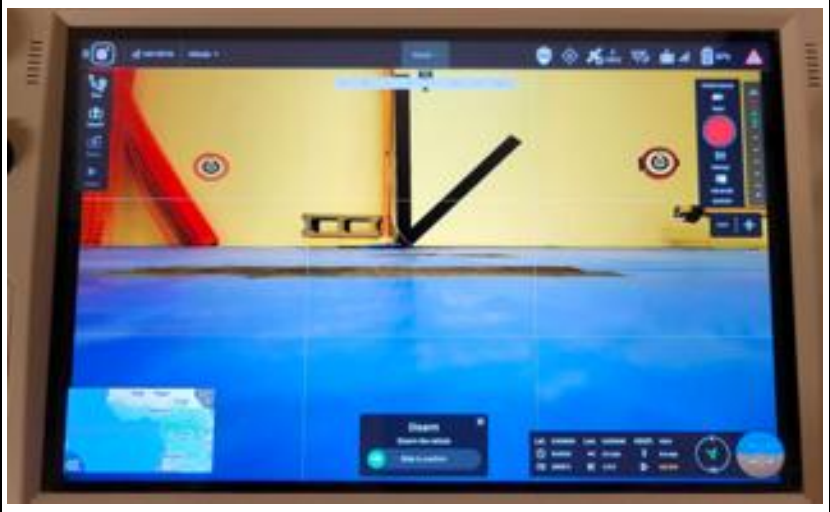 | 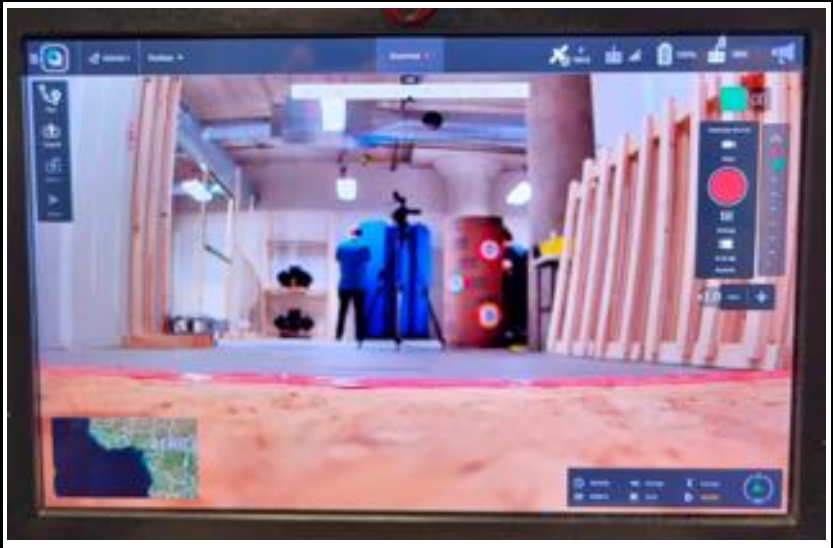 |
| **Controller and UI** | | |
| Is the controller labeled | No | No |
| Is there an onboard manual / control guide | No | No |
| How many physical buttons are there | 10 | 13 |
| Does the controller use a touch screen? | Yes | Yes |
| Does the controller use a touch screen to start flight | Yes | Yes |
| Is use of the touch screen required during flight | Yes* | Yes |
| Is the touch screen easily responsive | No | No |
| Do flight modes change the configuration of how functionality is mapped to the controller inputs | Yes | No |
| How is the user alerted to critical states | Prompts at top of screen and QGC command line 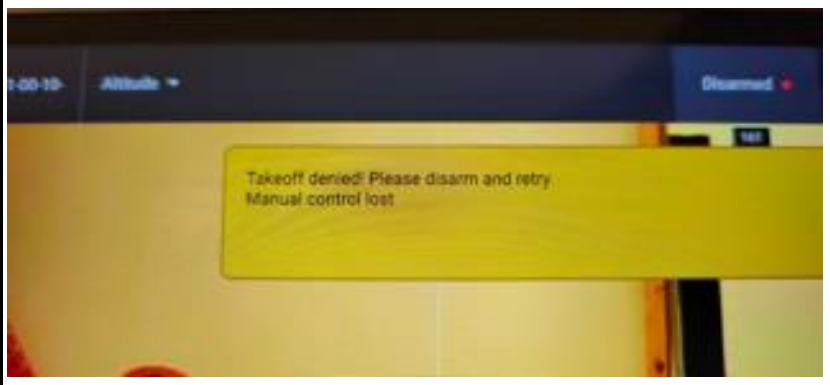  | Prompts at top of screen and QGC command line 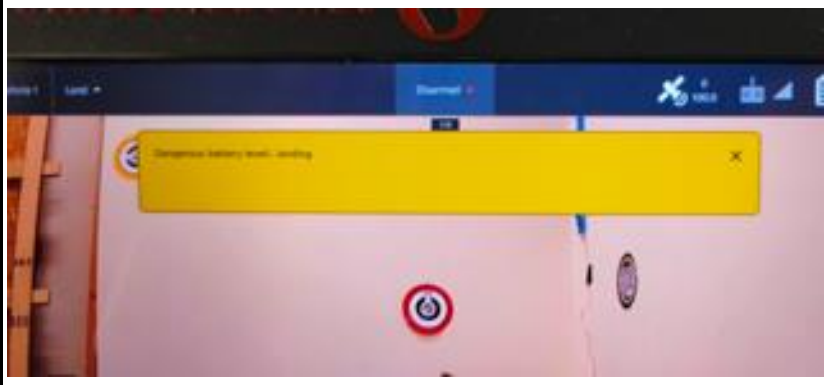 |
| Do settings reset on power cycle | Yes | Yes |
| Are obstacle avoidance notifications shown | No | N/A |
| If the drone has an auto-land, how/when is it engaged | No, but can detect landing if on ground and throttle is pulled down | Yes, and will auto land if on the ground and throttle is pulled down |
| If the drone has auto takeoff, is it a physical or a virtual button? | Both, drone can take off or be armed | Virtual, drone ascends to determine height on request. can arm without auto takeoff |
| If the drone has an emergency stop that can be initiated by the user, is it a physical or a virtual button? | Physical, dedicated cover switch | Virtual, when pressing the arming button again, a slide on screen can be used to Stop the drone |
| Are some features disabled in specific modes | Yes | Yes |
| Controller display lighting | Yes | Yes |
| **Power** | | |
| Is battery remaining time indicated | Yes, %<br>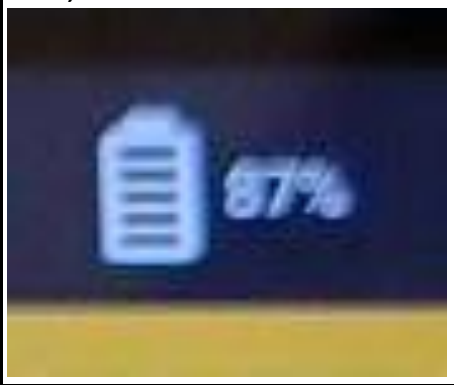 | Yes, %<br>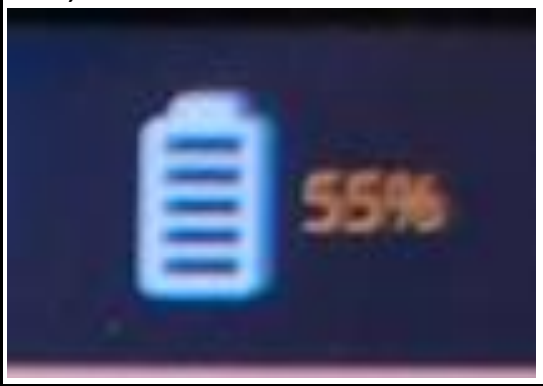 |
| Is flight time remaining indicated | No | No |
| Is the user prompted about damaged battery | No | Yes, if battery is not inserted correctly, will display popup error |
| Are actions prevented at critical battery levels | No | No |
| **Communications link** | | |
| Does the interface display the current comms link connection level | Yes, bar<br>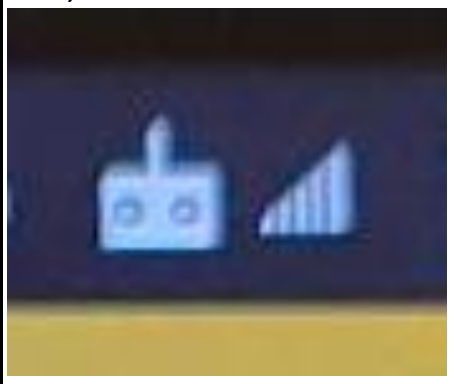 | Yes, bar<br>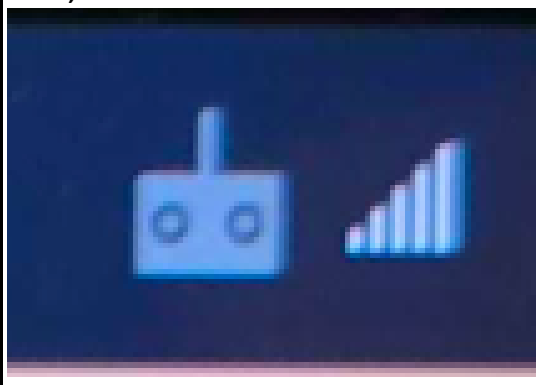 |
| Is the user prompted about reduced comms link | No | No |
| Is the user prompted about loss of comms link | Yes, after reconnect the controller supplies a popup saying comms lost, and audio cue | Yes, after reconnect the controller supplies a popup saying comms lost |
| What happens on comms loss | Broadcasts SOS audio signal, drifts uncontrolled until comes to landing | Lands and Disarms |
| Does the drone alert the user to magnetic interference | Yes | Yes |
| **Navigation** | | |
| Is 3D mapping data displayed during flight | No | No |
| Does its behavior change without GPS | Yes | Yes, harder to hold position due to drift |
| Is the user prompted about loss of GPS | Yes, popup notification | Yes, popup notification |
| Is touch screen used during flight | Yes, to move camera, change settings, or reconnect | Yes, to land |
| Do flight modes change obstacle avoidance | No | No |
| Does the drone switch modes automatically | No, user requested | No, user requested |
| Are critical environmental conditions alerted | Yes, UI popup notification | Yes, UI popup notification |
| **Camera** | | |
| Are cameras usable when not armed | Yes | Yes |
| Do the navigation camera streams also display UI elements (e.g., map, system status) | Yes | Yes |
| Is nav cam always visible in menus during flight | No | No |
| can the camera zoom, is it digital or optical | Digital | Digital |
| Are there multiple cameras used for varied zoom levels | No | Yes |
| Expected Horizontal Field of View of main camera model Degrees | 80 | 63 |
| Expected Horizontal Field of View of thermal camera model Degrees | 34 | 24 |
| Horizontal Field of View of main nav camera Degrees | 85 | 71 |
| Vertical Field of View of main nav camera Degrees | 65.5 | 38 |
| Gimbal Range Up Degrees | 35 | 35 |
| Gimbal Range Down Degrees | 66 | 90 |
| Field of Regard of main nav camera Up Degrees | 68 | 54 |
| Field of Regard of main nav camera Down Degrees | 50 | 109 |
| Thermal Horizontal Field of View Degrees | 33 | 26.5 |
| Thermal Vertical Field of View Degrees | 28 | 13 |
| Field of Regard of Thermal camera UP Degrees | 49 | 42 |
| Field of Regard of Thermal camera Down Degrees | 80 | 97 |
| **Additional functionality and accessories** | | |
| Does the drone have illuminators | No | No |
| Does the drone have a laser or pointer | No | No |

# Obstacle Avoidance

## Obstacle Avoidance and Collision Resilience

### Summary of Test Method

The test method consists of flying an sUAS directly towards different types of obstacles and recording the response. The test method is categorized into two classes based on the fundamental capabilities of the sUAS being evaluated. Such systems can usually be classified as having (a) an active collision avoidance system based on a full (or shared) autonomy mode, or (b) a passive collision resilience system (e.g., propeller guards or cage) that limit the impact of collisions with obstacles when they do occur. The different classes of sUAS obstacle capabilities each require their own test methodology. sUAS without any obstacle avoidance or collision resilience capabilities cannot be tested using this method. Metrics such as stopping times and maximum deceleration experienced will be evaluated. The appropriate tests will be performed for three scenarios: (a) head-on-collision course with obstacle (i.e. flight direction is perpendicular to plane of obstacle), (b) collision path that is angled at 45-degrees from the plane of the obstacle, and (c) sideways collision with impact on sUAS starboard or portside. Tests will be performed for the following obstacle elements: walls, chain link fences, mesh materials, and doors.

Obstacle Avoidance: The obstacle avoidance test methodologies pertain to any sUAS systems that possess autonomous obstacle avoidance capabilities, i.e. the sUAS should be able to perceive the presence of an obstacle and take corrective actions to avoid collisions. This test methodology does not cover human-piloted sUAS. The test method evaluates various metrics such as minimum time to collision, minimum distance to collision, and number of collisions. The tests show not only if the system is able to avoid the obstacle but it also assesses sUAS performance for different materials. Some of the materials used in these tests (such as chain link fence and meshes) are significantly more difficult to perceive by sUAS systems as compared to others (such as doors and walls). The tests seek to assess the obstacle avoidance performance in these different scenarios.

Collision Resilience: Collision resilience test methodologies apply to all sUAS, including human-piloted and autonomous systems. These tests evaluate the ability of the sUAS to be resilient to collisions, by analyzing numerical metrics such as maximum deceleration experienced during a collision event, and pre-post collision change in velocity, as well as newly-devised categorical resilience metrics, as discussed in the Metrics section. These test methods are especially useful for analyzing the efficacy of sUAS platforms with additional protection such as propeller guards or cages. The tests with wall, chain link fence, and mesh, have similar methodology which includes flying towards the obstacle in various configurations (forward flight, sideways flight) and at various angles (trajectory is perpendicular to obstacle or at 45 degrees). The collision resilience tests for doors have slightly different setups which include flying towards a door obstacle that is closed, partially open, or open. There are three different types of Collision Resilience (CR) metrics: Modified Acceleration Severity Index (MASI), Maximum Delta-V, and Categorical metrics for success/failure outcomes.

## Benchmarking Results

Note: due to the lack of reliable indoor obstacle avoidance behaviors available on several sUAS platforms and the risks associated with failed obstacle avoidance tests due to the fragility of said platforms, no benchmarking data is available the Obstacle Avoidance tests. Thus, only data from the Collision Resilience test is shown in this report.

### Collision Resilience

Four types of obstacles were used for evaluating Collision Resilience: wall, mesh, chain link fence, and door. The first three obstacles were evaluated with five collision tests (head-on, port side, starboard, port side at 45° incidence, and starboard at 45° incidence), while the door was evaluated in three conditions (open, closed, and partially open). All means and standard deviations were calculated from 5 runs.

Only the sUAS with protective hardware are included in this evaluation (Cleo Robotics Dronut X1P, Flyability Elios 2 GOV, Vantage Robotics Vesper), as all others without propeller guards are not able to withstand collisions.

| 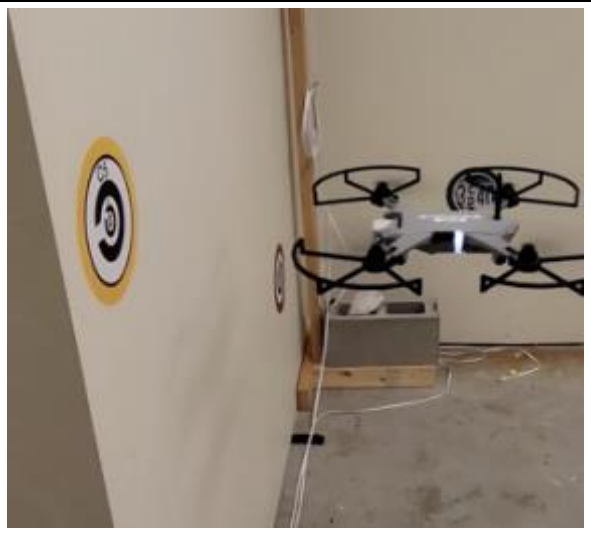 | 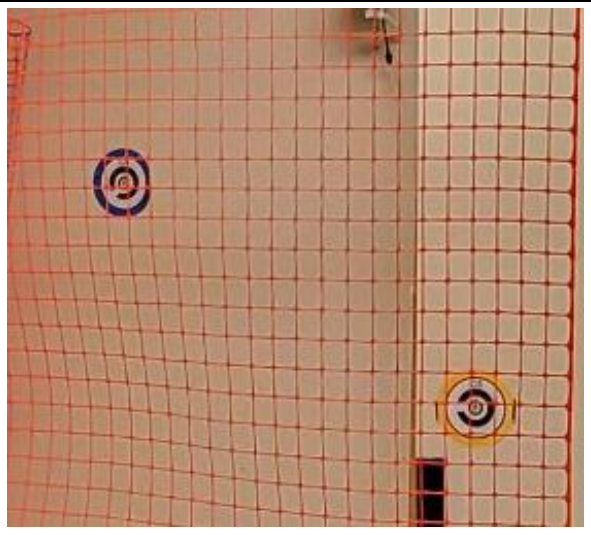 | 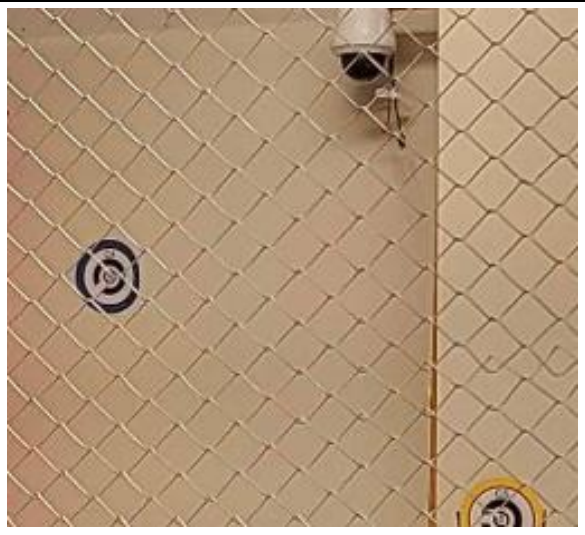 | 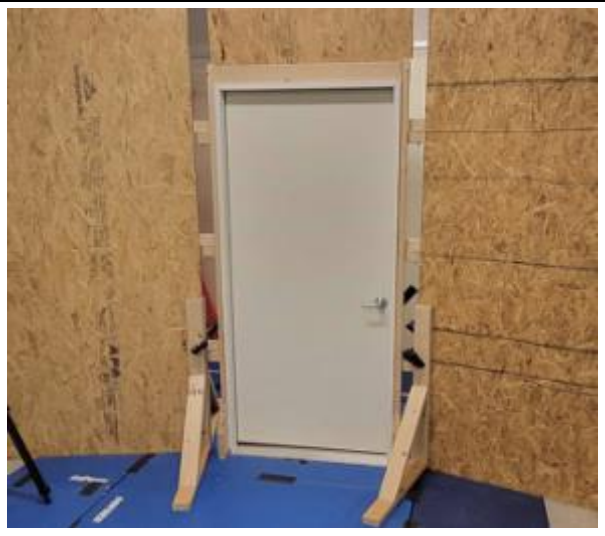 |
|---|---|---|---|
| **Wall** | **Mesh** | **Chain link fence** | **Door** |

**Modified Acceleration Severity Index (MASI)**

The dimensionless Modified Acceleration Severity Index (MASI) metric is defined as:

$$MASI = \frac{1}{g}\sqrt{a_x^2 + a_y^2 + a_z^2}$$

where $a_x$ represents the longitudinal acceleration of the sUAS, $a_y$ represents the lateral acceleration of the sUAS, $a_z$ represents the vertical acceleration of the sUAS, and g represents the acceleration due to gravity (9.8 $m/s^2$). All quantities are in units of $m/s^2$. For the performed tests, sUAS flight only occurred in the horizontal plane, so then we assume that $a_z = 0\ m/s^2$.

No best in class criteria is specified due to the Vantage Robotics Vesper only performing a little over half of the available tests (10 of 18), leaving only two remaining platforms to evaluate, whose performance, on average, is very similar to each other.

| Obstacle | Test | Metrics (m/s²) | sUAS | | | | | |
|---|---|---|---|---|---|---|---|---|
| | | | Cleo Robotics Dronut X1P | | Flyability Elios 2 GOV | | Vantage Robotics Vesper* | |
| | | | Mean | Stdev | Mean | Stdev | Mean | Stdev |
| Wall | **Head-on Collision** | X-axis | -0.03 | 0.01 | -0.12 | 0.15 | -0.01 | 0.00 |
| | | Y-axis | -0.01 | 0.04 | -0.03 | 0.03 | -0.03 | 0.05 |
| | **Port Side Collision** | X-axis | -0.04 | 0.02 | -0.02 | 0.01 | X | |
| | | Y-axis | -0.05 | 0.01 | -0.02 | 0.01 | | |
| | **Starboard Collision** | X-axis | -0.05 | 0.01 | -0.02 | 0.01 | X | |
| | | Y-axis | -0.05 | 0.00 | -0.01 | 0.01 | | |
| | **Port Side Collision 45° Incidence** | X-axis | -0.05 | 0.03 | -0.06 | 0.08 | X | |
| | | Y-axis | -0.08 | 0.05 | -0.03 | 0.03 | | |
| | **Starboard Collision 45° Incidence** | X-axis | -0.14 | 0.14 | -0.12 | 0.19 | X | |
| | | Y-axis | -0.06 | 0.01 | -0.14 | 0.19 | | |
| Mesh | **Head-on Collision** | X-axis | -0.078 | 0.010 | -0.084 | 0.010 | -0.087 | 0.010 |
| | | Y-axis | -0.029 | 0.007 | -0.021 | 0.007 | -0.017 | 0.005 |
| | **Port Side Collision** | X-axis | -0.033 | 0.003 | -0.018 | 0.001 | X | |
| | | Y-axis | -0.067 | 0.013 | -0.061 | 0.051 | | |
| | **Starboard Collision** | X-axis | -0.039 | 0.020 | -0.019 | 0.003 | X | |
| | | Y-axis | -0.016 | 0.022 | -0.033 | 0.023 | | |
| | **Port Side Collision 45° Incidence** | X-axis | -0.084 | 0.008 | -0.076 | 0.015 | -0.112 | 0.048 |
| | | Y-axis | -0.051 | 0.010 | -0.043 | 0.015 | -0.088 | 0.092 |
| | **Starboard Collision 45° Incidence** | X-axis | -0.068 | 0.020 | -0.079 | 0.020 | -0.054 | 0.044 |
| | | Y-axis | -0.017 | 0.002 | -0.025 | 0.010 | -0.041 | 0.037 |
| Chain link fence | **Head-on Collision** | X-axis | -0.100 | 0.016 | -0.099 | 0.035 | -0.106 | 0.026 |
| | | Y-axis | -0.027 | 0.005 | -0.043 | 0.014 | -0.025 | 0.006 |
| | **Port Side Collision** | X-axis | -0.057 | 0.028 | -0.017 | 0.012 | X | |
| | | Y-axis | -0.087 | 0.072 | -0.064 | 0.046 | | |
| | **Starboard Collision** | X-axis | -0.062 | 0.027 | -0.067 | 0.033 | X | |
| | | Y-axis | -0.026 | 0.005 | -0.009 | 0.012 | | |
| | **Port Side Collision 45° Incidence** | X-axis | -0.092 | 0.029 | -0.105 | 0.024 | -0.046 | 0.027 |
| | | Y-axis | -0.037 | 0.009 | -0.055 | 0.016 | -0.020 | 0.014 |
| | **Starboard Collision 45° Incidence** | X-axis | -0.051 | 0.011 | -0.010 | 0.012 | 0.359 | 0.069 |
| | | Y-axis | -0.037 | 0.039 | -0.026 | 0.008 | -0.043 | 0.013 |
| Door | **Open** | X-axis | -0.044 | 0.023 | -0.023 | 0.012 | -0.015 | 0.007 |
| | | Y-axis | -0.044 | 0.048 | -0.021 | 0.012 | -0.015 | 0.009 |
| | **Closed** | X-axis | -0.192 | 0.237 | -0.157 | 0.114 | -0.062 | 0.081 |
| | | Y-axis | -0.032 | 0.046 | -0.092 | 0.124 | -0.067 | 0.078 |
| | **Partially Open** | X-axis | -0.220 | 0.244 | -0.110 | 0.103 | -0.040 | 0.023 |
| | | Y-axis | -0.096 | 0.138 | -0.058 | 0.024 | -0.049 | 0.022 |
| **Average MASI Across All Tests** | | | -0.067 | - | -0.064 | - | -0.028 | - |

*Note: The Vantage Robotics Vesper was not able to be evaluated in all conditions due to firmware issues that raised safety concerns.

**Maximum Delta-V**

The Maximum Delta-V metric calculates the maximum value of the change in velocity before and after collision over a given time window. For the performed sUAS tests, the Maximum Delta-V is evaluated over a 0.3s window that begins at the time instant of sUAS' collision with the obstacle. A caveat for this test is that the sensing apparatus must record position and velocity information at a frequency of at least 10 Hz to obtain a good estimate of the metric. All values are in m/s.

No best in class criteria is specified due to the Vantage Robotics Vesper only performing a little over half of the available tests (11 of 18), leaving only two remaining platforms to evaluate, whose performance, on average, is very similar to each other.

| Obstacle | Test | sUAS | | | | | |
|---|---|---|---|---|---|---|---|
| | | Cleo Robotics Dronut X1P | | Flyability Elios 2 GOV | | Vantage Robotics Vesper* | |
| | | Mean | Stdev | Mean | Stdev | Mean | Stdev |
| Wall | Head-on Collision | 1.352 | 0.587 | 1.667 | 0.279 | 0.698 | 0.320 |
| | Port Side Collision | 1.910 | 0.440 | 1.783 | 0.304 | X | |
| | Starboard Collision | 1.735 | 0.210 | 1.584 | 0.259 | X | |
| | Port Side Collision 45° Incidence | 10.469 | 4.090 | 1.950 | 0.431 | 2.911 | 0.624 |
| | Starboard Collision 45° Incidence | 2.916 | 0.996 | 2.275 | 0.327 | X | |
| Mesh | Head-on Collision | 3.422 | 0.514 | 1.716 | 0.203 | 3.670 | 0.936 |
| | Port Side Collision | 4.676 | 0.935 | 2.462 | 0.409 | X | |
| | Starboard Collision | 2.723 | 0.480 | 1.749 | 0.356 | X | |
| | Port Side Collision 45° Incidence | 0.079 | 0.007 | 0.078 | 0.016 | 0.063 | 0.054 |
| | Starboard Collision 45° Incidence | 3.843 | 0.500 | 1.582 | 0.275 | 4.105 | 3.765 |
| Chain link fence | Head-on Collision | 4.783 | 0.306 | 1.988 | 0.453 | 3.454 | 0.351 |
| | Port Side Collision | 3.555 | 1.402 | 2.052 | 0.508 | X | |
| | Starboard Collision | 2.259 | 0.334 | 1.627 | 0.125 | X | |
| | Port Side Collision 45° Incidence | 2.158 | 0.715 | 1.825 | 0.179 | 4.110 | 0.260 |
| | Starboard Collision 45° Incidence | 2.653 | 0.901 | 1.781 | 0.312 | 1.503 | 0.539 |
| Door | Open | 1.113 | 0.248 | 0.895 | 0.144 | 0.870 | 0.257 |
| | Closed | 1.337 | 0.635 | 1.919 | 1.224 | 3.980 | 6.366 |
| | Partially Open | 2.486 | 1.671 | 0.930 | 0.113 | 1.714 | 0.913 |
| Average Maximum Delta-V Across All Tests | | 2.971 | - | 1.659 | - | 2.462 | - |

*Note: The Vantage Robotics Vesper was not able to be evaluated in all conditions due to firmware issues that raised safety concerns.

### Categorical Metrics

The Collision Resiliency (CR) Categorical metric is used to determine the various success or failure scenarios of the sUAS operation in indoor or subterranean environments. Broadly, the three categories represent successful test (A), failed test (B), and test abandonment (C). A lower alphanumeric is better, i.e. CR-A1 indicates better performance than CR-B2. The categories are described in the table below:

| Categorical CR metric | Description of observed behavior during test |
|---|---|
| **CR-A1** | **A: Resilient:** The category level A represents that the sUAS passed the resiliency test with no or little degradation in operation. Specifically, Category CR-A1 corresponds to the scenario that **that the sUAS platform collided with the obstacle, but did not suffer any failure, and was able to continue operations. This represents perfect collision resilience properties.** |
| **CR-A2** | **Category CR-A2** is similar to CR-A1 except for the fact that the sUAS temporarily loses operational continuity. For example, after a collision with an obstacle has occurred, the sUAS may retreat to a fail-safe mode, such as executing a safe landing. However, this action does not imply lack of resilience, as the sUAS can return to its operational capacity after the fail-safe mode is disabled (such as return to flight after a safe landing). |
| **CR-A3** | **Category CR-A3** is similar to CR-A2 except for the fact that in this scenario the sUAS fails to enter a fail-safe mode after the collision event, but is still able to return to operation after the event. Thus, the sUAS suffers only a temporary loss in operational continuity. For example, the sUAS may suffer an uncontrolled descent (i.e., crash) in this scenario as opposed to the scenario in CR-A2 where, for example, the descent was a programmed landing activated by a fail-safe mode. |
| **CR-B1** | **B: Lack of resiliency:** The levels in this category correspond to the scenario where the sUAS failed to resolve a collision gracefully. |
| **CR-B2** | **Category CR-B2** indicates further degradation of behavior as compared to CR-B1. Specifically, in this scenario, the sUAS may execute a successful landing in fail-safe mode, but communication drop-out prevents a return to operation. Without communication, teleoperation of the sUAS cannot be carried out, i.e. sUAS integrity is maintained, but flight capabilities are lost. In this scenario, the sUAS control system may or may not be operational, but since control commands cannot be communicated. **This failure mode does not apply to the collision resilience of autonomous sUAS platforms.** |
| **CR-B3** | **Category CR-B3** is one level of further degradation as compared to CR-B2. In this scenario, the sUAS' attempt to execute a safe landing after the collision is unsuccessful. Specifically, the sUAS may have landed on with a tilt or flipped over, i.e. not in its usual take-off configuration. In this scenario, resume flight operations after a take-off event can not be guaranteed. In this scenario, the control system and/or the communication channel may or may not be operational. |
| **CR-B4** | **Category CR-B4** indicates the final level of performance degradation as it includes loss of structural integrity of the sUAS platform, presenting a *potential permanent loss of operational continuity*. It is possible to provide a minor distinction between actual structural damage and disintegration of some components (such as protective propeller guards), but they both indicate a lack of collision resilience of the sUAS structural frame. |
| **CR-C1** | **Category CR-C1** corresponds to a scenario where the evaluation team had to terminate the test to ensure safety and structural integrity of the sUAS platform. |

Best in class = A1 (i.e., resilient) performance across all tests

| | Head-on collision | | | | | Port side collision | | | | | Starboard collision | | | | | Port side collision 45° incidence | | | | | Starboard collision 45° incidence | | | | |
|---|---|---|---|---|---|---|---|---|---|---|---|---|---|---|---|---|---|---|---|---|---|---|---|---|---|
| **Wall** | **1** | **2** | **3** | **4** | **5** | **1** | **2** | **3** | **4** | **5** | **1** | **2** | **3** | **4** | **5** | **1** | **2** | **3** | **4** | **5** | **1** | **2** | **3** | **4** | **5** |
| Cleo Robotics Dronut X1P | B1 | B1 | B1 | B1 | B1 | A1 | A1 | A1 | A1 | A1 | A1 | A1 | A1 | A1 | A1 | A1 | A1 | A1 | A1 | A1 | A1 | A1 | A1 | A1 | A1 |
| Flyability Elios 2 GOV | A1 | A1 | A1 | A1 | A1 | A1 | A1 | A1 | A1 | A1 | A1 | A1 | A1 | A1 | A1 | A1 | A1 | A1 | A1 | A1 | A1 | A1 | A1 | A1 | A1 |
| Vantage Robotics Vesper | B4 | B4 | B4 | B4 | B4 | B4 | C1 | C1 | C1 | C1 | B4 | C1 | C1 | C1 | C1 | A1 | B4 | C1 | C1 | C1 | B4 | C1 | C1 | C1 | C1 |
| **Chain link fence** | **1** | **2** | **3** | **4** | **5** | **1** | **2** | **3** | **4** | **5** | **1** | **2** | **3** | **4** | **5** | **1** | **2** | **3** | **4** | **5** | **1** | **2** | **3** | **4** | **5** |
| Cleo Robotics Dronut X1P | A1 | A1 | A1 | A1 | A1 | B3 | A1 | A1 | A1 | A1 | A1 | A1 | B3 | A1 | B3 | A1 | A1 | A1 | A1 | A1 | A1 | A1 | A1 | A1 | A1 |
| Flyability Elios 2 GOV | A1 | A1 | A1 | A1 | A1 | A1 | A1 | A1 | A1 | A1 | A1 | A1 | A1 | A1 | A1 | A1 | A1 | A1 | A1 | A1 | A1 | A1 | A1 | A1 | A1 |
| Vantage Robotics Vesper | A1 | A1 | A1 | A1 | A1 | A1 | B4 | C1 | C1 | C1 | B4 | C1 | C1 | C1 | C1 | A1 | A1 | A1 | A1 | A1 | A1 | A1 | A1 | A1 | A1 |
| **Mesh** | **1** | **2** | **3** | **4** | **5** | **1** | **2** | **3** | **4** | **5** | **1** | **2** | **3** | **4** | **5** | **1** | **2** | **3** | **4** | **5** | **1** | **2** | **3** | **4** | **5** |
| Cleo Robotics Dronut X1P | A1 | A1 | A1 | A1 | A1 | A1 | A1 | A1 | A1 | A1 | A1 | A1 | A1 | A1 | A1 | A1 | A1 | A1 | A1 | A1 | A1 | A1 | A1 | A1 | A1 |
| Flyability Elios 2 GOV | A1 | A1 | A1 | A1 | A1 | A1 | A1 | A1 | A1 | A1 | A1 | A1 | A1 | A1 | A1 | A1 | A1 | A1 | A1 | A1 | A1 | A1 | A1 | A1 | A1 |
| Vantage Robotics Vesper | A1 | A1 | A1 | A1 | A1 | B4 | C1 | C1 | C1 | C1 | B4 | C1 | C1 | C1 | C1 | B4 | A1 | A1 | A1 | A1 | B4 | A1 | A1 | A1 | A1 |
| **Door** | **Open** | | | | | **Closed** | | | | | **Partially open** | | | | | | | | | | | | | | |
| | **1** | **2** | **3** | **4** | **5** | **1** | **2** | **3** | **4** | **5** | **1** | **2** | **3** | **4** | **5** | | | | | | | | | | |
| Cleo Robotics Dronut X1P | A1 | A1 | A1 | A1 | A1 | B3 | A2 | B3 | B3 | A2 | A2 | A2 | B3 | B3 | B3 | | | | | | | | | | |
| Flyability Elios 2 GOV | A1 | A1 | A1 | A1 | A1 | A1 | A1 | A1 | A1 | A1 | A1 | A1 | A1 | A1 | A1 | | | | | | | | | | |
| Vantage Robotics Vesper | A1 | A1 | A1 | A1 | A1 | A1 | A1 | B4 | A1 | A1 | A1 | A1 | A1 | A1 | A1 | | | | | | | | | | |
| **Best in class** | **Collision Resilience: Categorical Metrics** | | | | | | | | | | | | | | | | | | | | | | | | |
| | Flyability Elios 2 GOV | | | | | | | | | | | | | | | | | | | | | | | | |

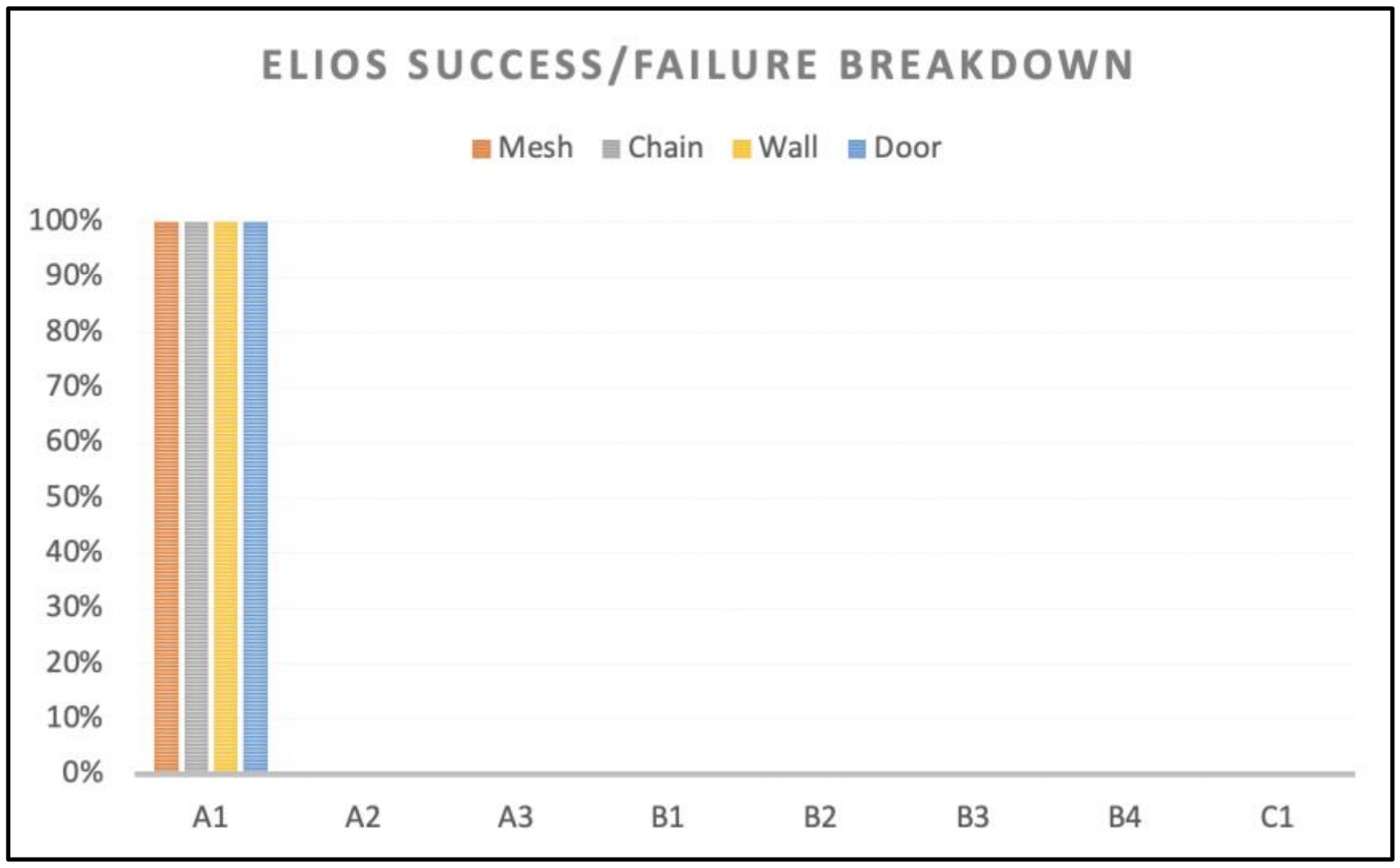
ELIOS SUCCESS/FAILURE BREAKDOWN
Mesh
Chain
Wall
Door
100%
90%
80%
70%
60%
50%
40%
30%
20%
10%
0%
A1
A2
A3
B1
B2
B3
B4
C1

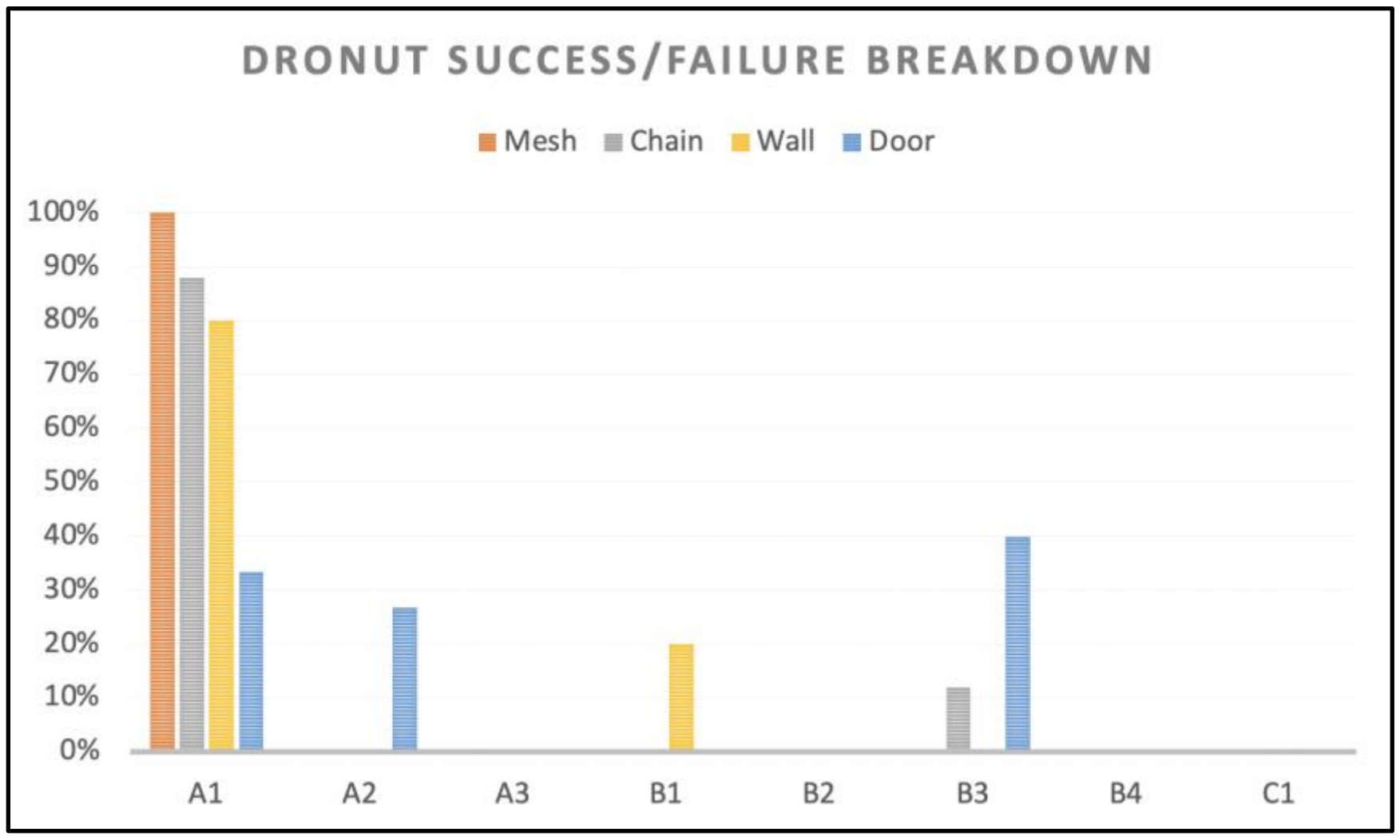
DRONUT SUCCESS/FAILURE BREAKDOWN
Mesh
Chain
Wall
Door
100%
90%
80%
70%
60%
50%
40%
30%
20%
10%
0%
A1
A2
A3
B1
B2
B3
B4
C1

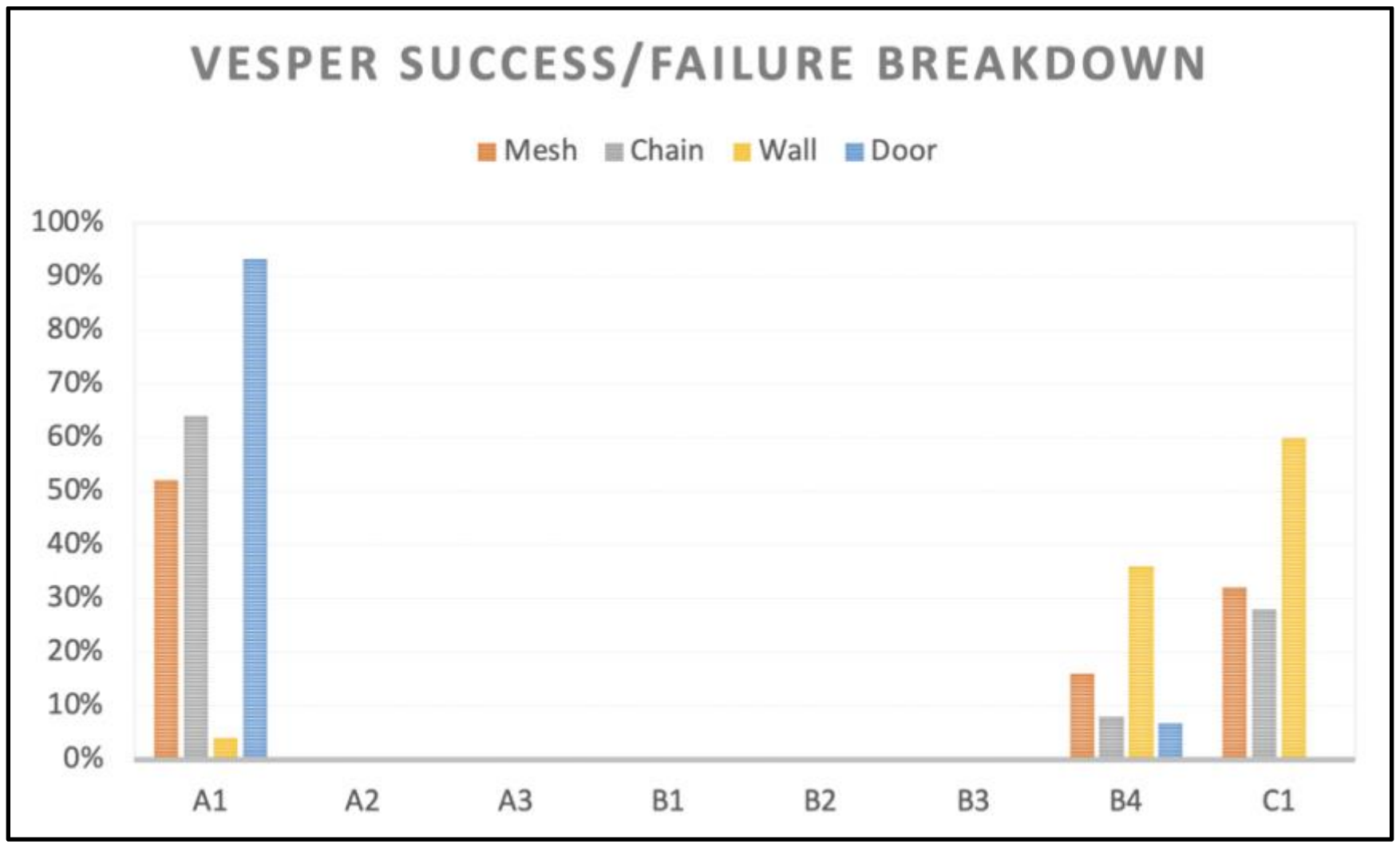
VESPER SUCCESS/FAILURE BREAKDOWN
Mesh
Chain
Wall
Door
100%
90%
80%
70%
60%
50%
40%
30%
20%
10%
0%
A1
A2
A3
B1
B2
B3
B4
C1

# Navigation

## Position and Traversal Accuracy

Affiliated publications: [Meriaux and Jerath, 2022]

### Summary of Test Method

The test method consists of four different tests: (a) wall following, (b) waypoint navigation, (c) straight line path traversals, (d) corner navigation, and (e) aperture navigation. Each test is conducted five times for each sUAS being evaluated and the metrics are aggregated across tests. All tests require the availability of telemetry data either via on-board vendor provided data streams, or external tracking systems. All tests can be performed with minimal additional apparatus (beyond the tracking system) if appropriate subterranean or indoor environments are available. If existing environments do not meet specifications, they can be constructed using readily available materials. All evaluation flights should be performed using line-of-sight operation as remote FPV operation may confound navigation capabilities of the sUAS.

Wall Following: The wall following test examines the ability of the sUAS to navigate a specific traversal path while operating in the vicinity of a wall at both 1 m (close) and 2 m (far) from the wall. This is a common use case scenario in specific indoor and subterranean operations. This test is performed in two common sUAS orientations for such missions: parallel (i.e., sUAS camera/front is pointed parallel to the wall surface while moving along it, pitching to fly forward) and perpendicular (i.e., sUAS camera/front pointed perpendicular to the wall while moving along it, strafing right or left to fly sideways).

Waypoint Navigation: The waypoint navigation evaluation methodology determines the ability of the sUAS to land at the desired waypoint location. The accuracy and precision metric is for reaching the desired waypoint, defined using the difference between the desired waypoint location and the final landing position of the sUAS.

Linear Path Traversal: The straight line traversal will require the sUAS to fly in a rectangular pattern made of four (4) linear path traversals. Deviations from the rectangular path will be used to evaluate the ability of the sUAS to perform straight line traversals. If a limited flight testing area is available, a single linear path traversal may be used for evaluation (instead of rectangular path).

Hallway Navigation: This test seeks to examine the ability of the sUAS to navigate a confined space with turns (such as a corridor or hallway). To eliminate the confounding factors associated with piloting skills, the test requires a flight pattern such that the corner navigation is performed via an in-place 90-degree turn, rather than smooth turning curves that expert pilots might execute in confined spaces. This test examines the effects of wind eddy currents in cases where there are walls on both sides of the sUAS (such as hallways). Hallway-induced wind eddy currents are expected to generate higher turbulence than the other navigation tests discussed here.

Corner Navigation: This test is similar to the hallway navigation test, but with corner partitions only on one side of the sUAS' flight path

Aperture Navigation: This test evaluates the ability of the sUAS platform to successfully navigate through an aperture. In subterranean environments, sometimes drones need to fly in such cases. They must be able to do so without contact with the surrounding material but if it does it must be able to withstand collision. This is why this test is not numerically evaluated like the other navigation tests but has a tiered result table listed below.

| Result | Condition of test | Explanation of result |
|---|---|---|
| A1 | Pass through no contact | Drone went through aperture did not touch any sides |
| A2 | Pass through contact no rip | Drone went through aperture did touch sides but no tears |
| A3 | Pass through contact ripped | Drone went through aperture did touch sides tear occurred |
| B1 | Failed pass through due to contact | Drone was unable to go through and land properly due to contact with aperture |

## Path Deviation

Path deviation: Deviation of the sUAS platform's actual trajectory from a defined straight line path.

1. For the Wall Following test, the path is a constant distance away from a wall, parallel to it (i.e., along the edge of the testing zone).
2. For the Linear Path Traversal test, the path is one or more defined straight line(s) within the testing zone.
3. For the Corner Navigation test, the path is two straight line(s) that constitute a 90-degree turn trajectory.

The evaluation of navigation abilities of the sUAS is defined as the deviation of the sUAS from the desired path. We evaluate both, the mean value of the deviation from the desired path (indicating the accuracy of navigation), as well as the standard deviation of deviation (indicating the precision of navigation).

Best in class = average deviations across all tests run for all platforms (i.e., metrics from all tests except Wall Following 1 m, Hallway Corner Navigation, Aperture 1.5x, Aperture 2x, and Through Door) that are less than the aggregate average of those data points (0.122)

C = data recorded by the external tracking system become corrupted, so no performance metrics are available

| Test | Metrics (m) | sUAS | | | | | | | | | | | | | |
|---|---|---|---|---|---|---|---|---|---|---|---|---|---|---|---|
| | | Cleo Robotics Dronut X1P | | Flyability Elios 2 GOV | | Lumenier Nighthawk V3* | | Parrot ANAFI USA GOV | | Skydio X2D | | Teal Golden Eagle† | | Vantage Robotics Vesper | |
| | | Mean | Stdev | Mean | Stdev | Mean | Stdev | Mean | Stdev | Mean | Stdev | Mean | Stdev | Mean | Stdev |
| **Wall Following 1 m** | Deviation in x direction | 0.425 | 0.050 | 0.004 | 0.001 | 0.164 | 0.053 | 0.188 | 0.032 | C | | 0.219 | 0.041 | 0.263 | 0.069 |
| | Stdev of deviation in x | 0.176 | - | 0.004 | - | 0.117 | - | 0.134 | - | | | 0.115 | - | 0.176 | - |
| | Deviation in y direction | 0.009 | 0.001 | 0.217 | 0.051 | 0.006 | 0.001 | 0.013 | 0.002 | | | 0.008 | 0.003 | 0.006 | 0.002 |
| | Stdev of deviation in y | 0.010 | - | 0.090 | - | 0.006 | - | 0.014 | - | | | 0.010 | - | 0.006 | - |
| **Wall Following 2 m** | Deviation in x direction | 0.159 | 0.020 | 0.012 | 0.005 | 0.065 | 0.018 | 0.179 | 0.073 | 0.207 | 0.073 | 0.123 | 0.058 | 0.202 | 0.047 |
| | Stdev of deviation in x | 0.104 | - | 0.014 | - | 0.046 | - | 0.088 | - | 0.152 | - | 0.078 | - | 0.074 | - |
| | Deviation in y direction | 0.009 | 0.001 | 0.057 | 0.011 | 0.006 | 0.001 | 0.011 | 0.002 | 0.009 | 0.001 | 0.011 | 0.003 | 0.007 | 0.001 |
| | Stdev of deviation in y | 0.009 | - | 0.037 | - | 0.005 | - | 0.012 | - | 0.008 | - | 0.020 | - | 0.007 | - |
| **Corner Navigation** | Deviation in x direction | 0.157 | 0.041 | 0.097 | 0.020 | 0.130 | 0.025 | 0.126 | 0.030 | 0.200 | 0.050 | 0.082 | 0.027 | 0.095 | 0.034 |
| | Stdev of deviation in x | 0.184 | - | 0.105 | - | 0.162 | - | 0.145 | - | 0.227 | - | 0.099 | - | 0.140 | - |
| | Deviation in y direction | 0.251 | 0.046 | 0.085 | 0.036 | 0.086 | 0.010 | 0.101 | 0.024 | 0.092 | 0.058 | 0.108 | 0.057 | 0.120 | 0.066 |
| | Stdev of deviation in y | 0.257 | - | 0.144 | - | 0.168 | - | 0.165 | - | 0.123 | - | 0.140 | - | 0.184 | - |
| **Hallway Corner Navigation** | Deviation in x direction | 0.097 | 0.025 | 0.178 | 0.217 | - | | 0.140 | 0.035 | 0.078 | 0.012 | X | | 0.133 | 0.017 |
| | Stdev of deviation in x | 0.109 | - | 0.156 | - | | | 0.173 | - | 0.089 | - | | | 0.174 | - |
| | Deviation in y direction | 0.213 | 0.021 | 0.266 | 0.093 | | | 0.277 | 0.023 | 0.078 | 0.006 | | | 0.224 | 0.018 |
| | Stdev of deviation in y | 0.203 | - | 0.180 | - | | | 0.275 | - | 0.091 | - | | | 0.213 | - |
| **Aperture 1.5x** | Deviation in x direction | 0.185 | 0.116 | 0.156 | 0.033 | - | | 0.150 | 0.036 | 0.252 | 0.042 | X | | 0.258 | 0.047 |
| | Stdev of deviation in x | 0.133 | - | 0.093 | - | | | 0.101 | - | 0.160 | - | | | 0.152 | - |
| | Deviation in y direction | 0.076 | 0.074 | 0.036 | 0.025 | | | 0.036 | 0.011 | 0.019 | 0.008 | | | 0.061 | 0.026 |
| | Stdev of deviation in y | 0.151 | - | 0.089 | - | | | 0.090 | - | 0.052 | - | | | 0.155 | - |
| **Aperture 2x** | Deviation in x direction | 0.165 | 0.039 | 0.142 | 0.051 | - | | 0.126 | 0.026 | 0.314 | 0.085 | X | | 0.380 | 0.067 |
| | Stdev of deviation in x | 0.147 | - | 0.059 | - | | | 0.082 | - | 0.148 | - | | | 0.175 | - |
| | Deviation in y direction | 0.009 | 0.001 | 0.011 | 0.002 | | | 0.009 | 0.002 | 0.010 | 0.001 | | | 0.009 | 0.002 |
| | Stdev of deviation in y | 0.010 | - | 0.012 | - | | | 0.011 | - | 0.010 | - | | | 0.010 | - |

| Test | Metrics (m) | sUAS | | | | | | | | | | | | | |
|---|---|---|---|---|---|---|---|---|---|---|---|---|---|---|---|
| | | Cleo Robotics Dronut X1P | | Flyability Elios 2 GOV | | Lumenier Nighthawk V3* | | Parrot ANAFI USA GOV | | Skydio X2D | | Teal Golden Eagle† | | Vantage Robotics Vesper | |
| | | Mean | Stdev | Mean | Stdev | Mean | Stdev | Mean | Stdev | Mean | Stdev | Mean | Stdev | Mean | Stdev |
| **Forward Facing Forward Motion** | Deviation in x direction | 0.238 | 0.019 | 0.359 | 0.197 | 0.241 | 0.054 | 0.136 | 0.009 | 0.271 | 0.067 | 0.201 | 0.040 | 0.220 | 0.093 |
| | Stdev of deviation in x | 0.167 | - | 0.239 | - | 0.143 | - | 0.094 | - | 0.119 | - | 0.133 | - | 0.140 | - |
| | Deviation in y direction | 0.018 | 0.002 | 0.010 | 0.002 | 0.024 | 0.011 | 0.034 | 0.016 | 0.024 | 0.013 | 0.008 | 0.001 | 0.031 | 0.011 |
| | Stdev of deviation in y | 0.015 | - | 0.009 | - | 0.062 | - | 0.068 | - | 0.038 | - | 0.008 | - | 0.089 | - |
| **Forward Facing Sideways Motion** | Deviation in x direction | 0.026 | 0.004 | 0.007 | 0.001 | 0.004 | 0.002 | 0.011 | 0.002 | 0.020 | 0.006 | 0.007 | 0.001 | 0.004 | 0.000 |
| | Stdev of deviation in x | 0.020 | - | 0.012 | - | 0.006 | - | 0.011 | - | 0.033 | - | 0.007 | - | 0.003 | - |
| | Deviation in y direction | 0.281 | 0.156 | 0.053 | 0.015 | 0.196 | 0.031 | 0.149 | 0.018 | 0.373 | 0.198 | 0.262 | 0.040 | 0.163 | 0.027 |
| | Stdev of deviation in y | 0.123 | - | 0.036 | - | 0.094 | - | 0.103 | - | 0.176 | - | 0.153 | - | 0.107 | - |
| **Forward Facing Diagonal Motion** | Deviation in x direction | 0.165 | 0.015 | 0.182 | 0.054 | 0.133 | 0.022 | 0.274 | 0.084 | 0.251 | 0.031 | 0.381 | 0.056 | 0.134 | 0.038 |
| | Stdev of deviation in x | 0.117 | - | 0.104 | - | 0.086 | - | 0.218 | - | 0.118 | - | 0.233 | - | 0.086 | - |
| | Deviation in y direction | 0.100 | 0.010 | 0.111 | 0.033 | 0.103 | 0.019 | 0.146 | 0.046 | 0.152 | 0.019 | 0.202 | 0.028 | 0.094 | 0.015 |
| | Stdev of deviation in y | 0.072 | - | 0.062 | - | 0.105 | - | 0.110 | - | 0.084 | - | 0.128 | - | 0.086 | - |
| **Square** | Deviation in x direction | 0.228 | 0.019 | 0.228 | 0.017 | 0.228 | 0.024 | 0.228 | 0.024 | 0.228 | 0.042 | 0.228 | 0.018 | 0.228 | 0.027 |
| | Stdev of deviation in x | 0.090 | - | 0.090 | - | 0.090 | - | 0.090 | - | 0.090 | - | 0.090 | - | 0.090 | - |
| | Deviation in y direction | 0.013 | 0.027 | 0.013 | 0.025 | 0.013 | 0.039 | 0.013 | 0.032 | 0.013 | 0.031 | 0.013 | 0.016 | 0.013 | 0.026 |
| | Stdev of deviation in y | 0.011 | - | 0.011 | - | 0.011 | - | 0.011 | - | 0.011 | - | 0.011 | - | 0.011 | - |
| **Through Door** | Deviation in x direction | 0.228 | 0.056 | 0.416 | 0.476 | - | | C | | C | | X | | 0.423 | 0.055 |
| | Stdev of deviation in x | 0.090 | - | 0.413 | - | | | | | | | | | 0.162 | - |
| | Deviation in y direction | 0.013 | 0.001 | 0.020 | 0.013 | | | | | | | | | 0.009 | 0.001 |
| | Stdev of deviation in y | 0.011 | - | 0.032 | - | | | | | | | | | 0.006 | - |
| **Average Deviation Across All Tests Run For All Platforms** | | 0.137 | - | 0.101 | - | 0.102 | - | 0.117 | - | 0.153 | - | 0.136 | - | 0.109 | - |
| **Best in class** | | **Position and Traversal Accuracy: Path Deviation** | | | | | | | | | | | | | |
| | | Flyability Elios 2 GOV<br>Lumenier Nighthawk V3<br>Parrot ANAFI USA GOV<br>Vantage Robotics Vesper | | | | | | | | | | | | | |

*Note: The Lumenier Nighthawk V3 was not able to be evaluated in several tests due to being out for repair.

†Note: Due to instability of the Teal Golden Eagle when flying indoors, it was not evaluated in several tests for safety concerns.

## Waypoint Navigation

Metric shown is for waypoint accuracy (m).

Best in class = average waypoint accuracy across all tests run for all platforms (i.e., metrics from all tests except Wall Following 1 m, Hallway Corner Navigation, Aperture 1.5x, Aperture 2x, and Through Door) that are less than the aggregate average of those data points (0.234)

| Test | sUAS | | | | | | | | | | | | | |
|---|---|---|---|---|---|---|---|---|---|---|---|---|---|---|
| | Cleo Robotics Dronut X1P | | Flyability Elios 2 GOV | | Lumenier Nighthawk V3* | | Parrot ANAFI USA GOV | | Skydio X2D | | Teal Golden Eagle† | | Vantage Robotics Vesper | |
| | Mean | Stdev | Mean | Stdev | Mean | Stdev | Mean | Stdev | Mean | Stdev | Mean | Stdev | Mean | Stdev |
| **Wall Following 1 m** | 1.149 | 0.208 | 0.321 | 0.129 | 0.289 | 0.065 | 0.188 | 0.045 | C | | 0.132 | 0.051 | 0.530 | 0.233 |
| **Wall Following 2 m** | 0.586 | 0.166 | 0.196 | 0.044 | 0.096 | 0.060 | 0.126 | 0.091 | 0.126 | 0.091 | 0.197 | 0.102 | 0.161 | 0.075 |
| **Corner Navigation** | 0.135 | 0.086 | 0.204 | 0.054 | 0.338 | 0.083 | 0.293 | 0.037 | 0.688 | 0.125 | 0.167 | 0.111 | 0.380 | 0.110 |
| **Hallway Corner Navigation** | 0.256 | 0.197 | 0.814 | 1.316 | - | | 0.319 | 0.176 | 0.232 | 0.038 | X | | 0.285 | 0.163 |
| **Aperture 1.5x** | 1.718 | 0.606 | 2.147 | 0.249 | - | | 1.998 | 0.102 | 2.119 | 0.267 | X | | 1.810 | 0.160 |
| **Aperture 2x** | 1.497 | 1.312 | 0.868 | 0.850 | - | | 0.550 | 0.085 | 0.753 | 0.130 | X | | 0.804 | 0.158 |
| **Forward Motion** | 0.382 | 0.146 | 0.107 | 0.068 | 0.297 | 0.230 | 0.220 | 0.074 | 0.221 | 0.062 | 0.229 | 0.056 | 0.232 | 0.120 |
| **Side Motion** | 0.232 | 0.040 | 0.092 | 0.072 | 0.263 | 0.165 | 0.182 | 0.082 | 0.382 | 0.143 | 0.273 | 0.147 | 0.145 | 0.055 |
| **Diagonal Motion** | 0.274 | 0.101 | 0.072 | 0.025 | 0.144 | 0.072 | 0.102 | 0.047 | 0.613 | 0.196 | 0.184 | 0.095 | 0.173 | 0.068 |
| **Square** | 0.313 | 0.118 | 0.029 | 0.010 | 0.121 | 0.039 | 0.132 | 0.041 | 0.236 | 0.122 | 0.275 | 0.124 | 0.212 | 0.047 |
| **Through Door** | 0.684 | 0.200 | 0.436 | 0.355 | - | | C | | C | | X | | 0.707 | 0.100 |
| **Average Deviation Across All Tests Run For All Platforms** | 0.320 | - | 0.117 | - | 0.210 | - | 0.176 | - | 0.378 | - | 0.221 | - | 0.217 | - |
| **Best in class** | **Position and Traversal Accuracy: Waypoint Navigation** | | | | | | | | | | | | | |
| | Flyability Elios 2 GOV<br>Lumenier Nighthawk V3<br>Parrot ANAFI USA GOV<br>Teal Golden Eagle<br>Vantage Robotics Vesper | | | | | | | | | | | | | |

*Note: The Lumenier Nighthawk V3 was not able to be evaluated in several tests due to being out for repair.

†Note: Due to instability of the Teal Golden Eagle when flying indoors, it was not evaluated in several tests for safety concerns.

## Navigation Through Apertures

### Summary of Test Method

The operator commands the sUAS to navigate through an aperture that either exists already in a real-world environment (e.g., a doorway or window in a building) or a fabricated apparatus that matches the relevant dimensions and shapes for each type of aperture. Three types of apertures are defined for navigation tests, each of which require horizontal or vertical traversal through spaces that are horizontally and/or vertically confined: doorway, window, and manhole. Navigation is performed multiple times to establish statistical significance and the associated probability of success and confidence levels based on the number of successes and failures (see the metrics section). For each trial, the sUAS begins from a starting location that requires it to traverse in a direction not parallel to the navigation route through the aperture, which may also require it to turn. Similarly, the end location for each trial also requires the sUAS to traverse in a direction not parallel to the navigation route. More simply, a single trial constitutes the sUAS traversing from the A side of the apparatus to the B side, navigating through the aperture, then traversing back over to the A side. See Figure 1. Each navigation test can be run either as elemental or operational navigation.

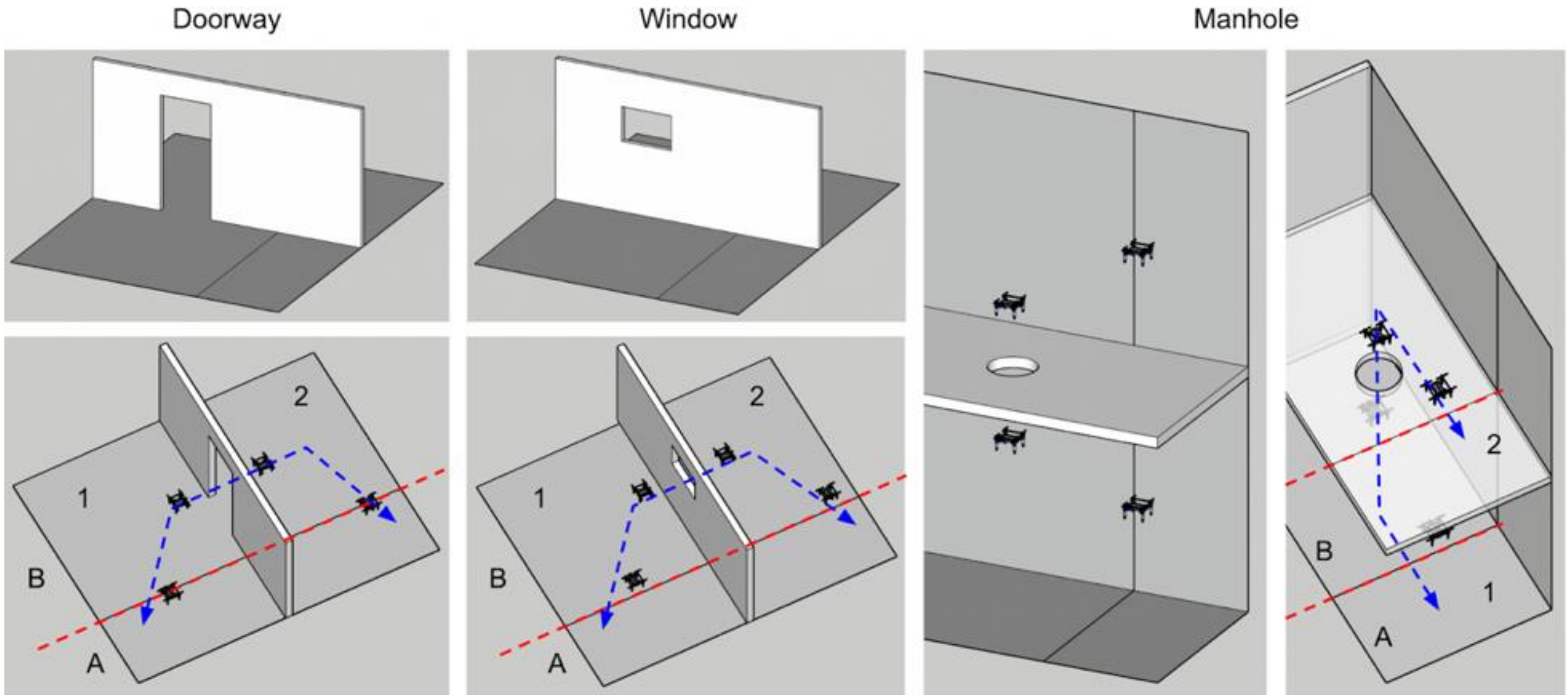

*Figure 1. Each type of aperture navigation test, left to right: doorway, window, and manhole.*

The areas on either side of the aperture should measure 3 m (118 in) square or larger to allow for much less obstructed flight than when navigating through the aperture. These areas may contain walls perpendicular to the wall/floor containing the aperture; for example, it is common for doors to be justified to one side of a room. The presence of these obstructions may be problematic for sUAS navigation due to airflow issues when a system flies too close to a wall and/or due to obstacle avoidance functionality (e.g., sUAS may attempt to maintain X distance between it and obstacles for safety, causing it to not be able to navigate through the aperture). If a wall is present on either area outside of the aperture, within 20 cm (8 in) of the edge of the aperture opening, then that area is considered obstructed.

## Benchmarking Results

Tests were conducted in two of the defined apertures: Doorway and Window. Notes:

- Average speed of the Flyability Elios 2 GOV during most navigation testing is not known because dedicated Navigation Through Apertures testing was not conducted with the system. Rather, the system demonstrated multiple successful navigation runs while conducting other tests that took place in the same environments (e.g., Indoor Mapping Accuracy).
- Two flight configurations of the Skydio X2D were tested: using standard joystick teleoperation and using its touch screen to place visual waypoints for autonomous flight to these waypoints. The corresponding data points are labeled “teleoperation” and “visual waypoints,” respectively.

### Doorway

No dedicated doorway testing was conducted. However, through a combination of the data recorded for window testing (which is a smaller aperture than the doorway) and doorway navigation performed in the NERVE Center test course included as part of several other tests, evaluations of sUAS capability to navigate through doorways can be derived.

#### Environment characterization

| NERVE Center Test Course | | | |
|---|---|---|---|
| **Environment** | **Outside aperture, area 1** | **Aperture** | **Outside aperture, area 2** |
| **Dimensions** | - | W x H: 91 x 203 cm (36 x 80 in) | - |
| **Lighting** | Well lit | - | Well lit |
| **Walls** | Drywall | - | Drywall |
| **Floor** | Concrete | - | Concrete |
| **Obstruction** | Wall on one side | - | Wall on one side |
| **Type** | Indoor | - | Indoor |
| **Image** | 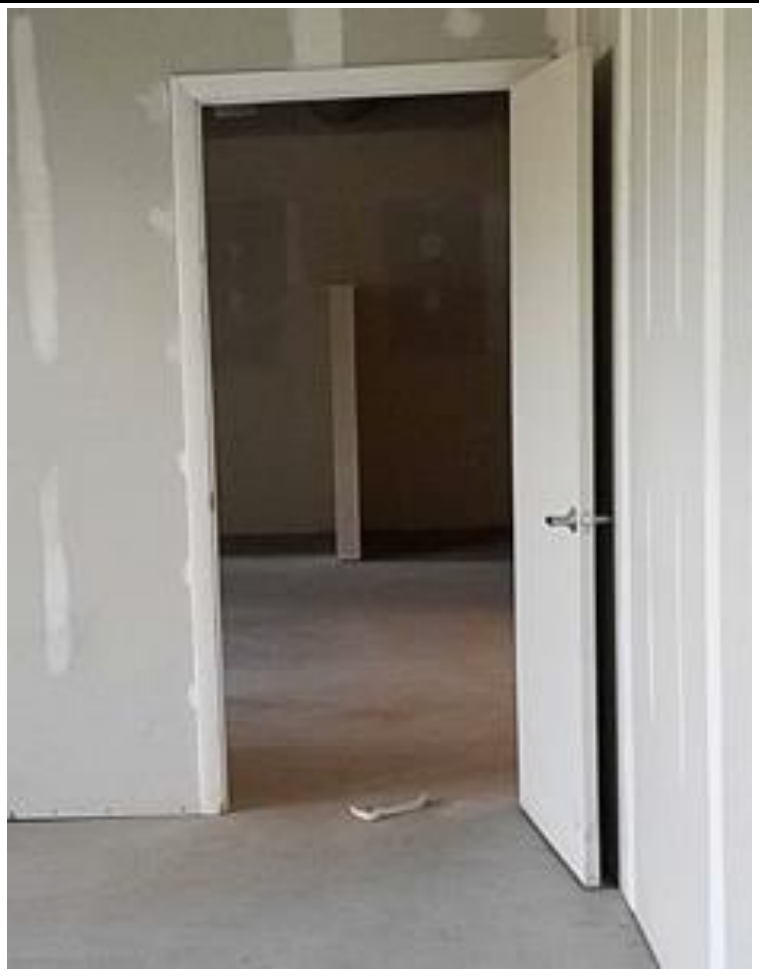 | | 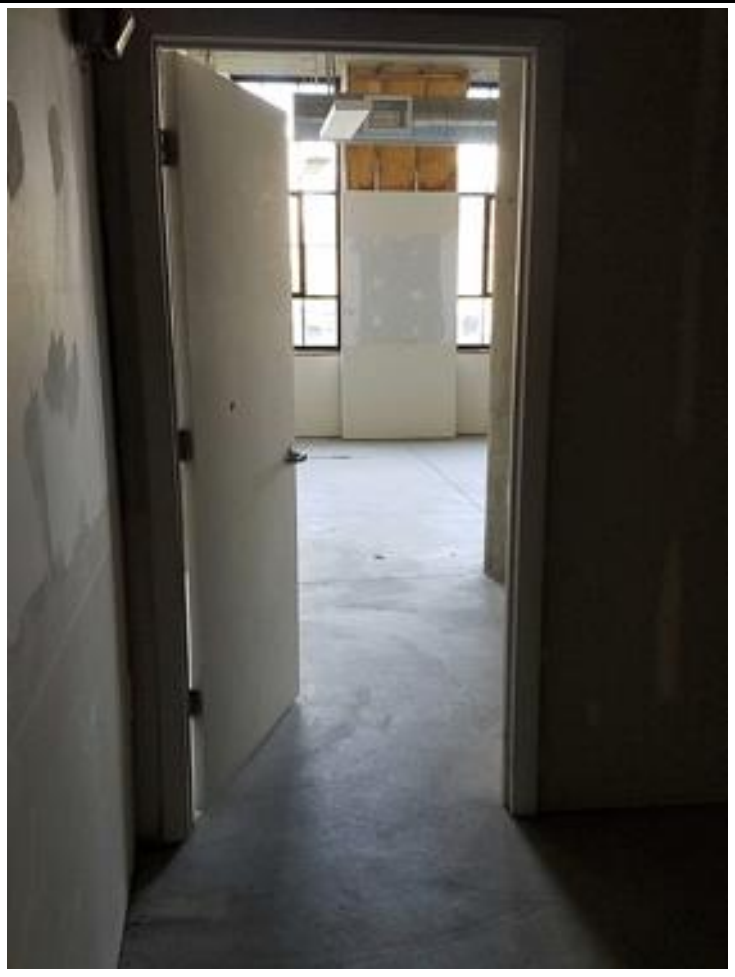 |

## Performance data

Testing was conducted as operational navigation unless only elemental navigation was possible due to NLOS communications issues or perceived operational risks.

A bar chart is not shown given that no average speed metrics were recorded for all evaluated systems.

Best in class = 100% completion for all navigation tests conducted

| sUAS | Metrics | NERVE Doorway |
|---|---|---|
| Cleo Robotics Dronut X1P | Navigation type | Operational |
| | Completion | 100% |
| | Average speed (m/s) | unknown |
| FLIR Black Hornet PRS | Navigation type | Operational |
| | Completion | 100% |
| | Average speed (m/s) | unknown |
| Flyability Elios 2 GOV | Navigation type | Operational |
| | Completion | 100% |
| | Average speed (m/s) | unknown |
| Lumenier Nighthawk V3 | Navigation type | Operational |
| | Completion | 100% |
| | Average speed (m/s) | unknown |
| Parrot ANAFI USA GOV | Navigation type | Elemental |
| | Completion | 100% |
| | Average speed (m/s) | unknown |
| Skydio X2D, teleoperation | Navigation type | Operational |
| | Completion | 100% |
| | Average speed (m/s) | unknown |
| Skydio X2D, visual waypoints | Navigation type | Operational |
| | Completion | 100% |
| | Average speed (m/s) | unknown |
| Teal Golden Eagle* | Navigation type | X |
| | Completion | X |
| | Average speed (m/s) | X |
| Vantage Robotics Vesper | Navigation type | Operational |
| | Completion | 100% |
| | Average speed (m/s) | unknown |
| | **Best in class** | **Doorway Navigation** |
| | | Cleo Robotics Dronut X1P<br>FLIR Black Hornet PRS<br>Flyability Elios 2 GOV<br>Lumenier Nighthawk V3<br>Parrot ANAFI USA GOV<br>Skydio X2D<br>Vantage Robotics Vesper |

*Note: Due to instability of the Teal Golden Eagle when flying indoors, it was not evaluated in this test.

## Window

### Environment characterization

Three instances of window navigation testing have been conducted: MUTC window, MUTC shaft entrance/exit, and a variable sized window at the NERVE Center.

| MUTC Window | | | |
|---|---|---|---|
| **Environment** | **Outside aperture, area 1** | **Aperture** | **Outside aperture, area 2** |
| **Dimensions** | - | W x H: 89 x 89 cm (35 x 35 in) | - |
| **Lighting** | Shaded Sun | - | Dark, Sun beam through aperture |
| **Walls** | Corrugated Steel, Conex Container | - | Corrugated Steel, Conex Container |
| **Floor** | Grass, Dirt | - | Wood |
| **Obstruction** | Vertical Scaffold Beams | - | Conex Container, Filing Cabinet |
| **Type** | Outdoor | - | Indoor |
| **Image** | 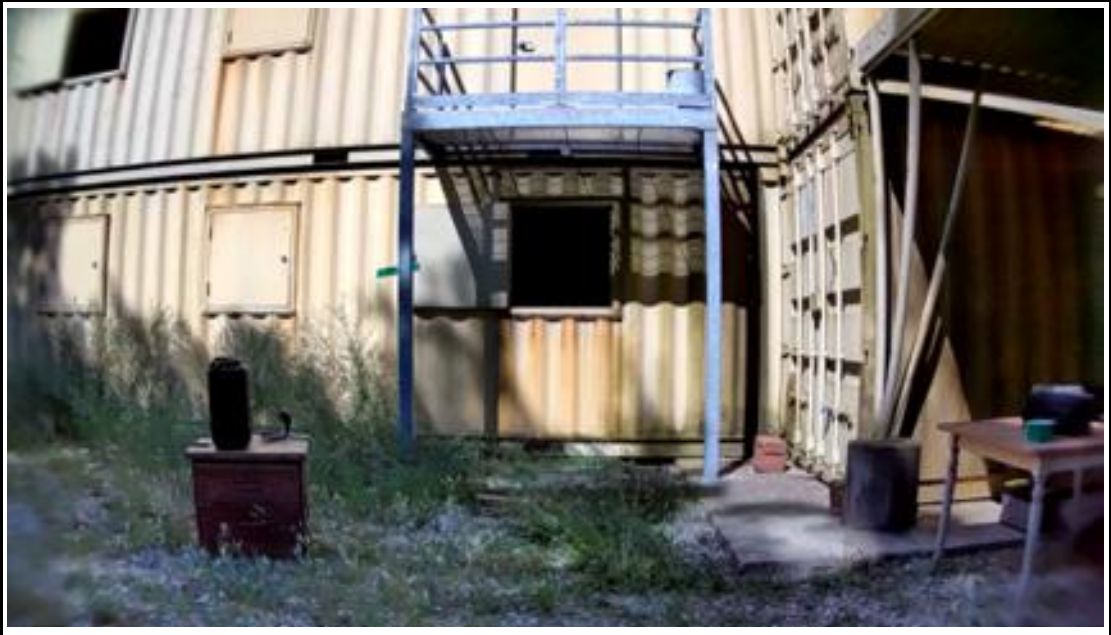 | | 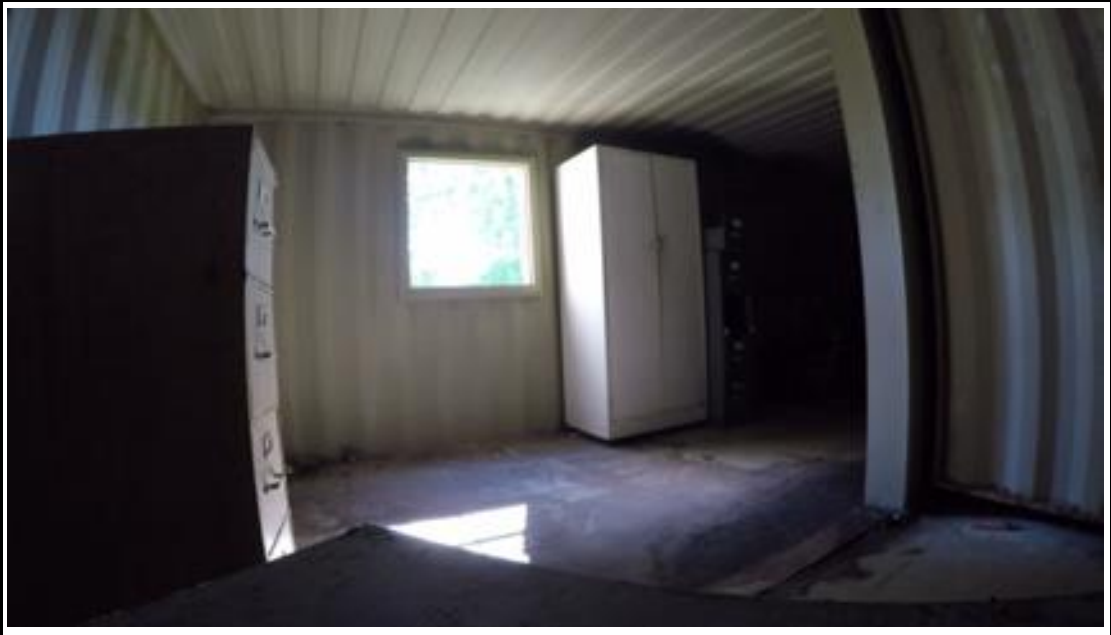 |

| MUTC Shaft Entrance/Exit | | | |
|---|---|---|---|
| **Environment** | **Outside aperture, area 1** | **Aperture** | **Outside aperture, area 2** |
| **Dimensions** | - | W x H: 132 x 97 cm (52 x 38 in)<br>Note: the ladder edge is considered the outer edge of the aperture | - |
| **Lighting** | Dark (<100 LUX) | - | Indirect Sunlight (100-200 LUX) |
| **Walls** | Corrugated Steel, Conex Container | - | Corrugated Steel, Conex Container |
| **Floor** | Wood | - | Safety Tarp |
| **Obstruction** | N/A | - | Ladder |
| **Type** | Indoor | - | "Outdoor" (Open roof) |
| **Image** | 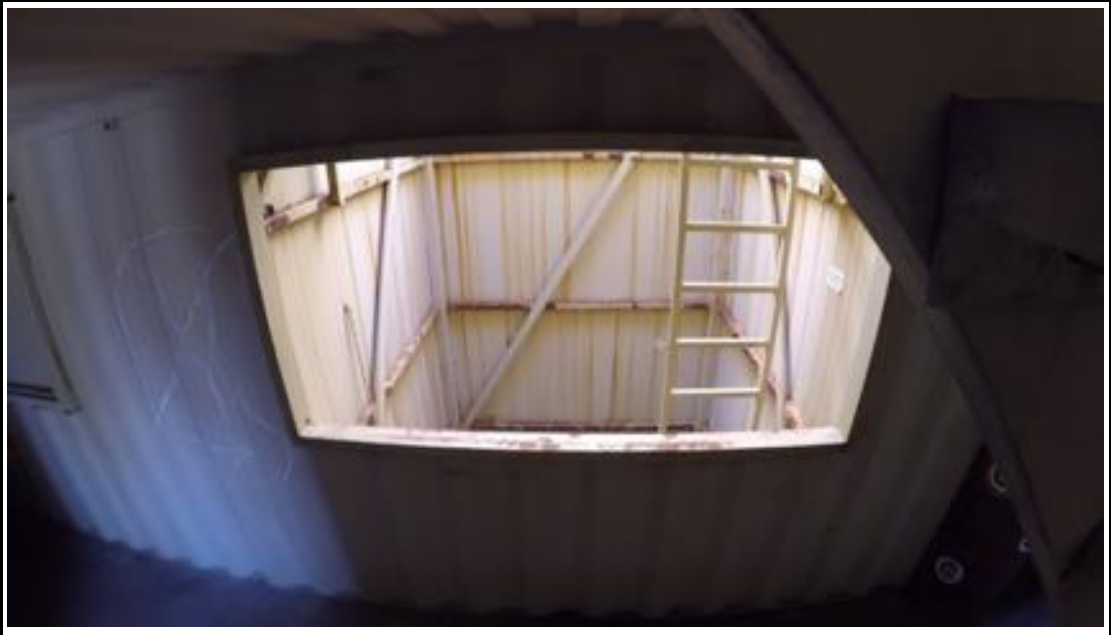 | | 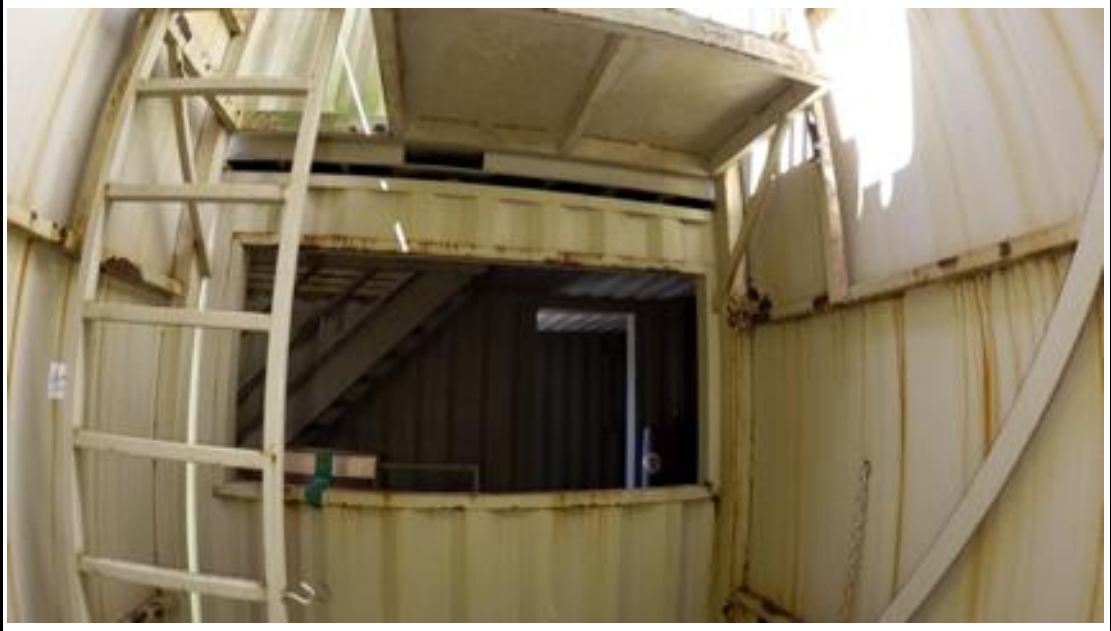 |

| NERVE Variable Window | | | |
|---|---|---|---|
| **Environment** | **Outside aperture, area 1** | **Aperture** | **Outside aperture, area 2** |
| **Dimensions** | - | W x H: Varies per system (see results table) | - |
| **Lighting** | Well lit | - | Well lit |
| **Walls** | Drywall, blackout curtain | - | Drywall, blackout curtain |
| **Floor** | Concrete, padding | - | Concrete, padding |
| **Obstruction** | N/A | - | N/A |
| **Type** | Indoor | - | Indoor |
| **Image**<br><br>(images shown are only one variation of the window) | 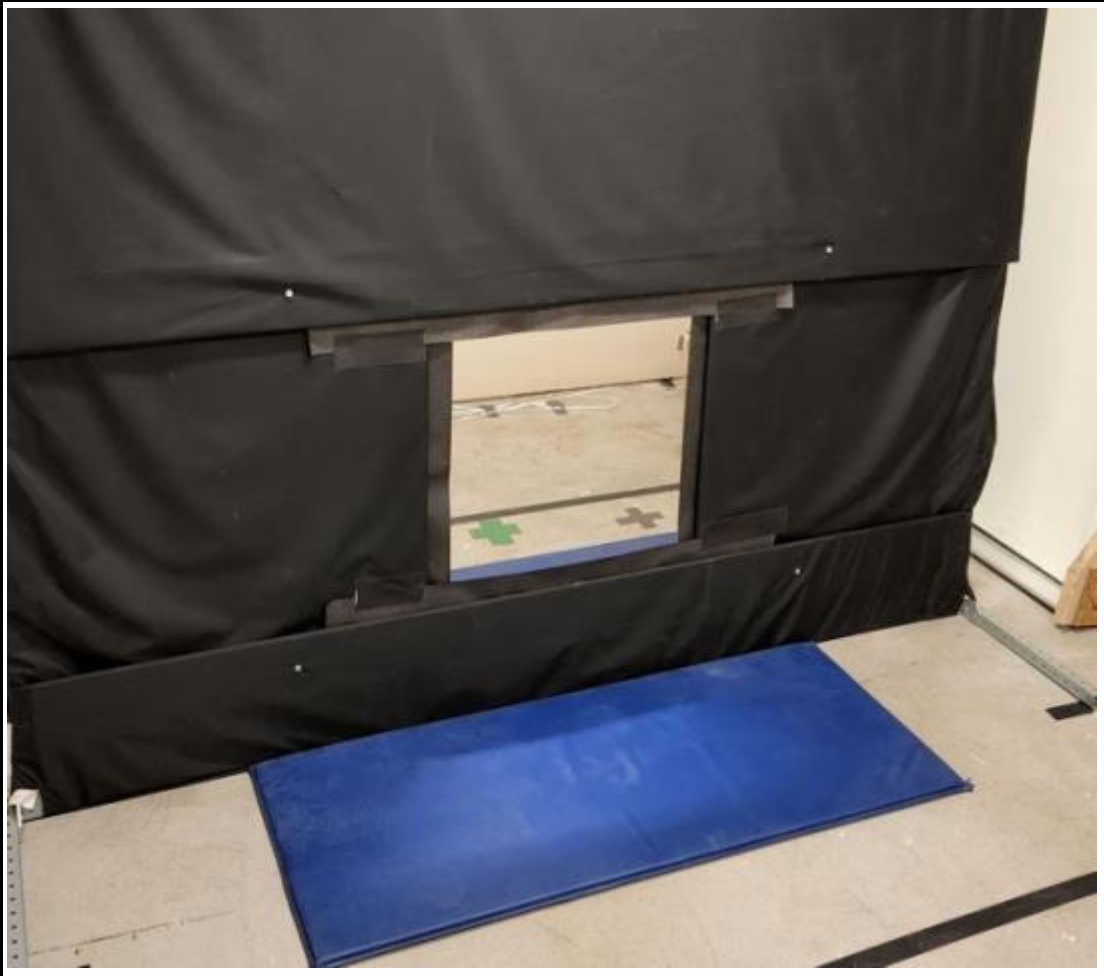 | 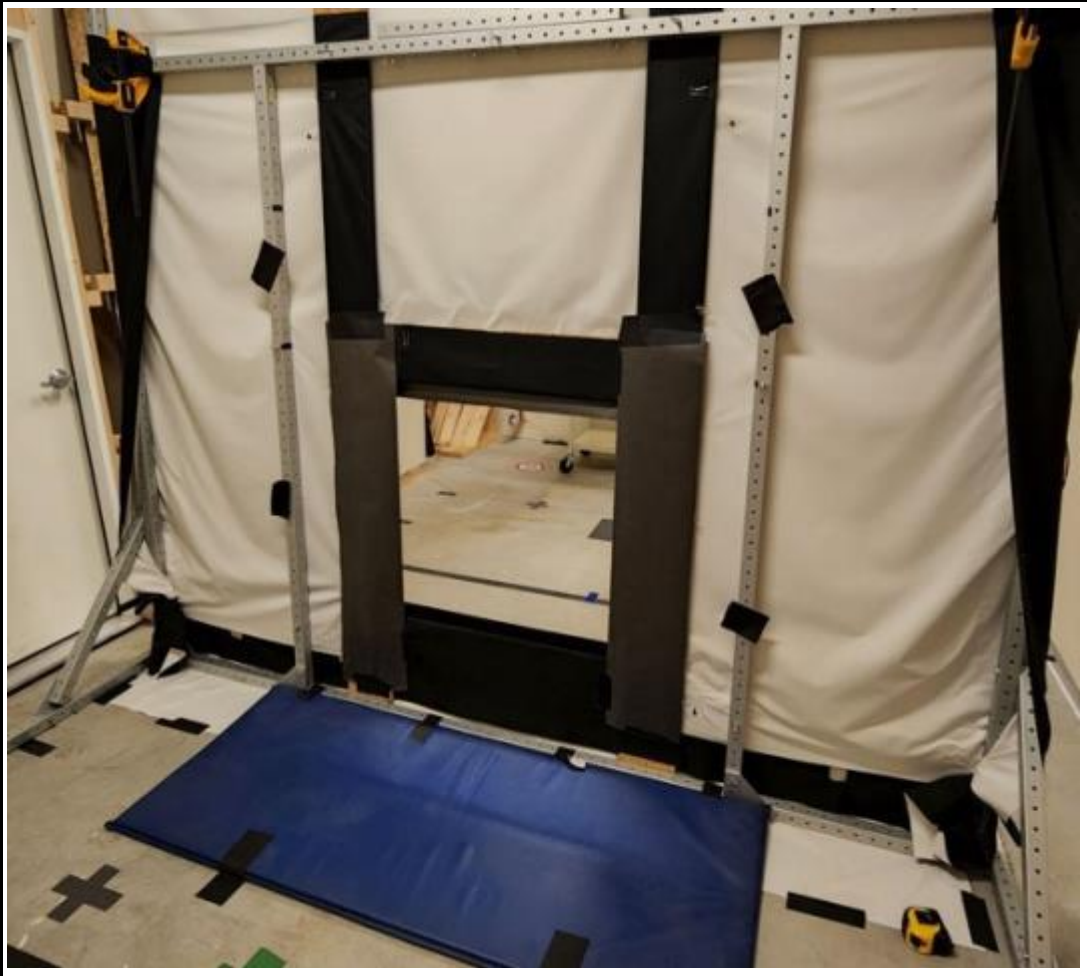 | |

## Performance data

Testing was conducted as operational navigation unless only elemental navigation was possible due to NLOS communications issues or perceived operational risks.

Only test results with 100% completion are shown in the chart below. The data from the MUTC shaft entrance was extracted from the Navigation Through Confined Spaces: Shaft testing, and the data from the NERVE variable window was extracted from the Position and Traversal Accuracy testing, so average speed metrics for those tests are unknown.

The variable window size used is 1.5 times the longest diagonal dimension (i.e., prop-tip to prop-tip) for each sUAS, with a lower bound of 30 cm (12 in).

Best in class = 2 or more navigation tests conducted with 100% completion for all navigation tests conducted

| sUAS | Metrics | MUTC Window | MUTC Shaft Entrance/Exit | NERVE Variable Window | |
|---|---|---|---|---|---|
| Cleo Robotics Dronut X1P | Navigation type | Elemental | Operational | Elemental | W x H:<br>30 x 30 cm<br>(12 x 12 in) |
| | Completion | 100% | 100% | 80% | |
| | Average speed (m/s) | 0.15 | unknown | unknown | |
| FLIR Black Hornet PRS | Navigation type | Elemental | Operational | Elemental | W x H:<br>30 x 30 cm<br>(12 x 12 in) |
| | Completion | 100% | 100% | 100% | |
| | Average speed (m/s) | 0.08 | unknown | 0.07 | |
| Flyability Elios 2 GOV | Navigation type | Operational | Operational | Operational | W x H:<br>59 x 59 cm<br>(23 x 23 in) |
| | Completion | 100% | 100% | 100% | |
| | Average speed (m/s) | unknown | unknown | unknown | |
| Lumenier Nighthawk V3 | Navigation type | - | Operational | - | - |
| | Completion | - | 100% | - | |
| | Average speed (m/s) | - | unknown | - | |
| Parrot ANAFI USA GOV | Navigation type | Elemental | Operational | Elemental | W x H:<br>63 x 63<br>(25 x 25 in) |
| | Completion | 33% | 100% | 100% | |
| | Average speed (m/s) | 0.10 | unknown | unknown | |
| Skydio X2D, teleoperation | Navigation type | - | - | Elemental | W x H:<br>112 x 112 cm<br>(44 x 44 in) |
| | Completion | - | - | 100% | |
| | Average speed (m/s) | - | - | unknown | |
| Skydio X2D, visual waypoints | Navigation type | - | - | - | - |
| | Completion | - | - | - | |
| | Average speed (m/s) | - | - | - | |
| Teal Golden Eagle* | Navigation type | X | X | X | X |
| | Completion | X | X | X | |
| | Average speed (m/s) | X | X | X | |
| Vantage Robotics Vesper | Navigation type | Elemental | - | Elemental | W x H:<br>61 x 61 cm<br>(24 x 24 in) |
| | Completion | 100% | - | 100% | |
| | Average speed (m/s) | 0.20 | - | unknown | |
| | **Best in class** | **Window Navigation** | | | |
| | | FLIR Black Hornet PRS<br>Flyability Elios 2 GOV<br>Vantage Robotics Vesper | | | |

*Note: Due to instability of the Teal Golden Eagle when flying indoors, it was not evaluated in this test.

Only test results with 100% completion are shown in the chart below.

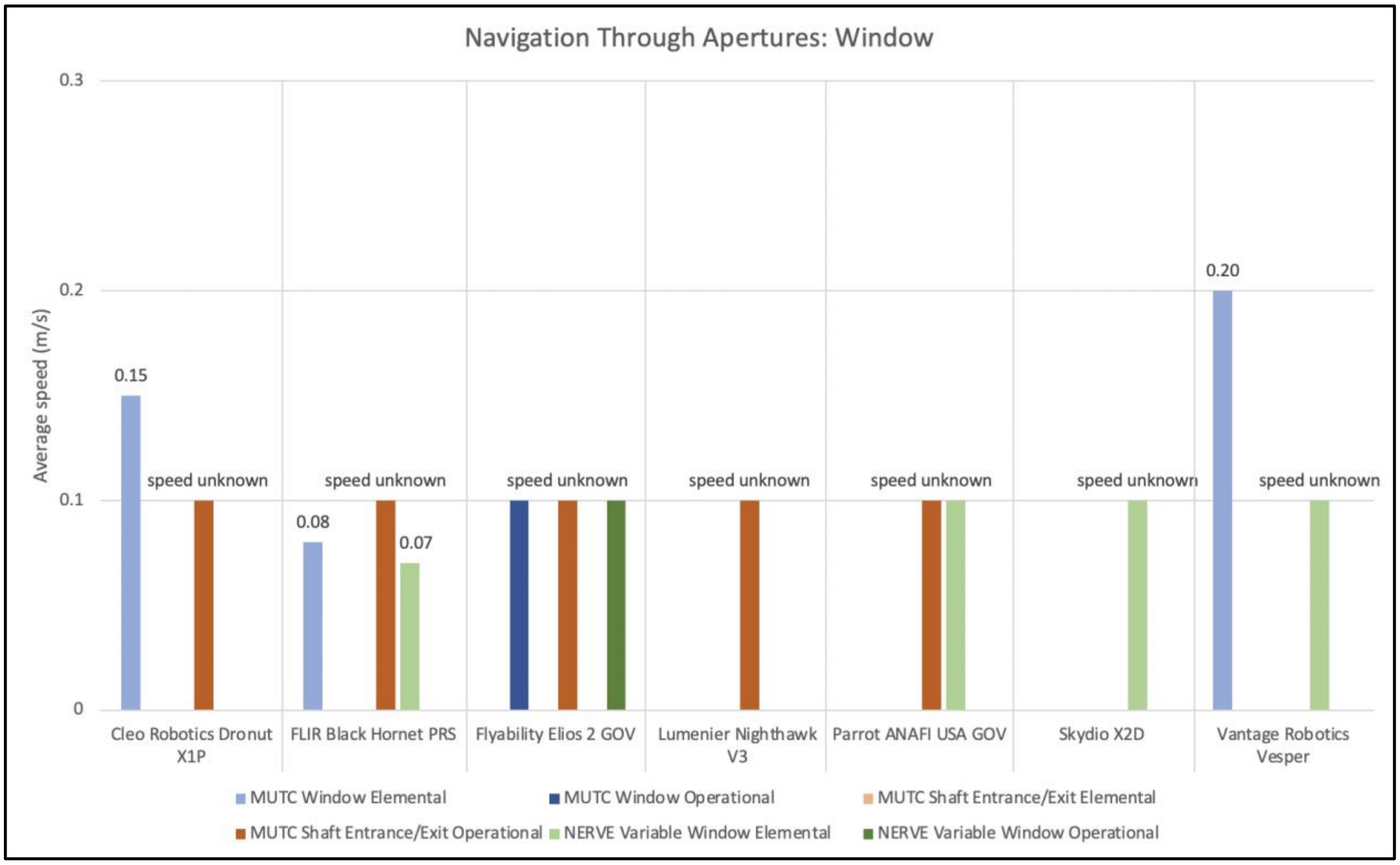

# Navigation Through Confined Spaces

## Summary of Test Method

The operator commands the sUAS to navigate through a confined space that either exists already in a real-world environment (e.g., a hallway or stairwell in a building) or a fabricated apparatus that matches the relevant dimensions and shapes for each type of confined space. Four types of confined spaces are defined for navigation tests, each of which require horizontal and/or vertical traversal through spaces that are horizontally and/or vertically confined: hallway, tunnel, stairwell/incline, and shaft. Navigation is performed multiple times to establish statistical significance and the associated probability of success and confidence levels based on the number of successes and failures. For each trial, the sUAS begins from a starting location that requires it to traverse in a direction not parallel to the navigation route through the confined space, which may also require it to turn. Similarly, the end location for each trial also requires the sUAS to traverse in a direction not parallel to the navigation route. More simply, a single trial constitutes the sUAS traversing from the A side of the apparatus to the B side, navigating through the confined space, then traversing back over to the A side. See Figure 1 and Figure 2. Each navigation test can be run either as elemental or operational navigation.

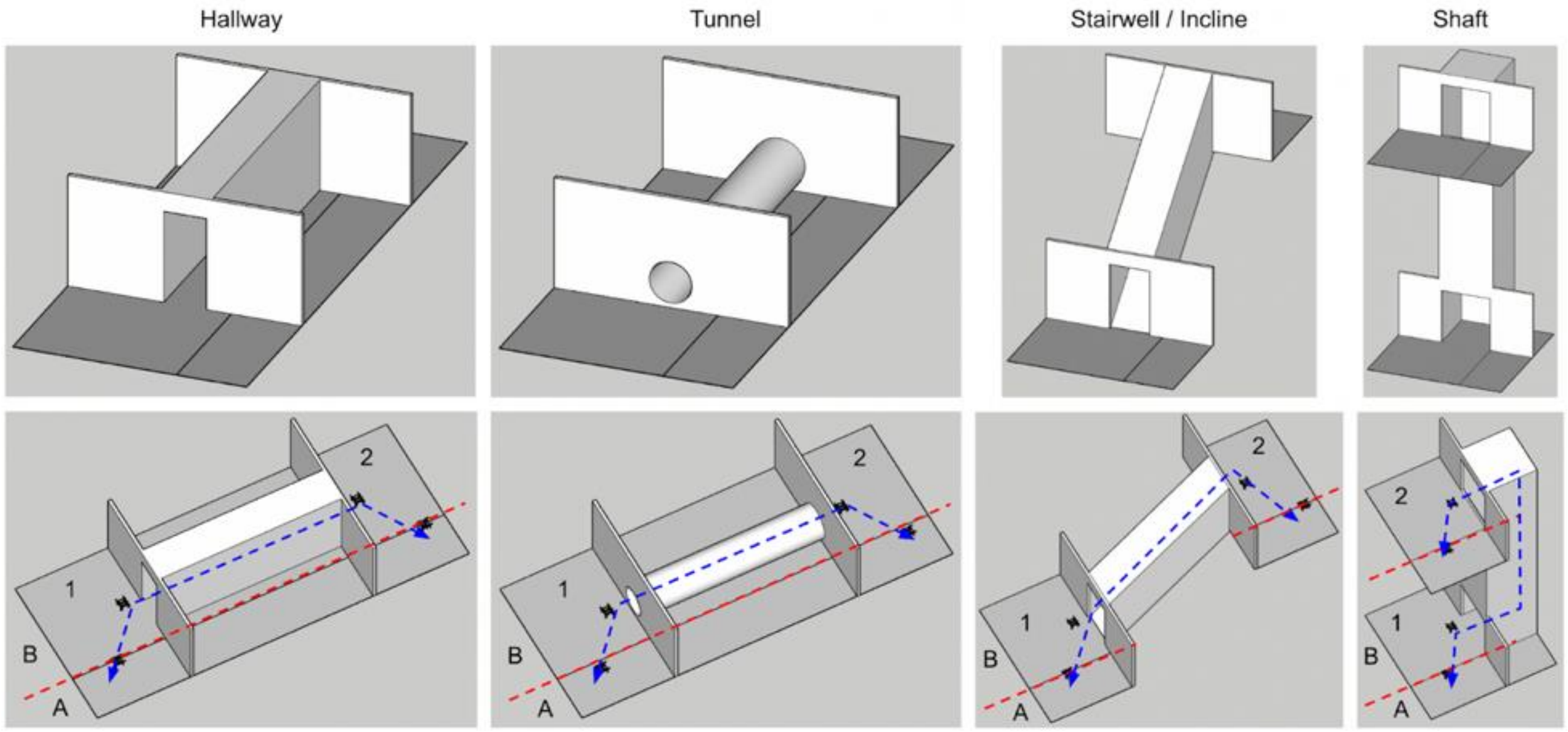

*Figure 1. Each type of confined space navigation test, left to right: hallway, tunnel, stairwell/incline, and shaft.*

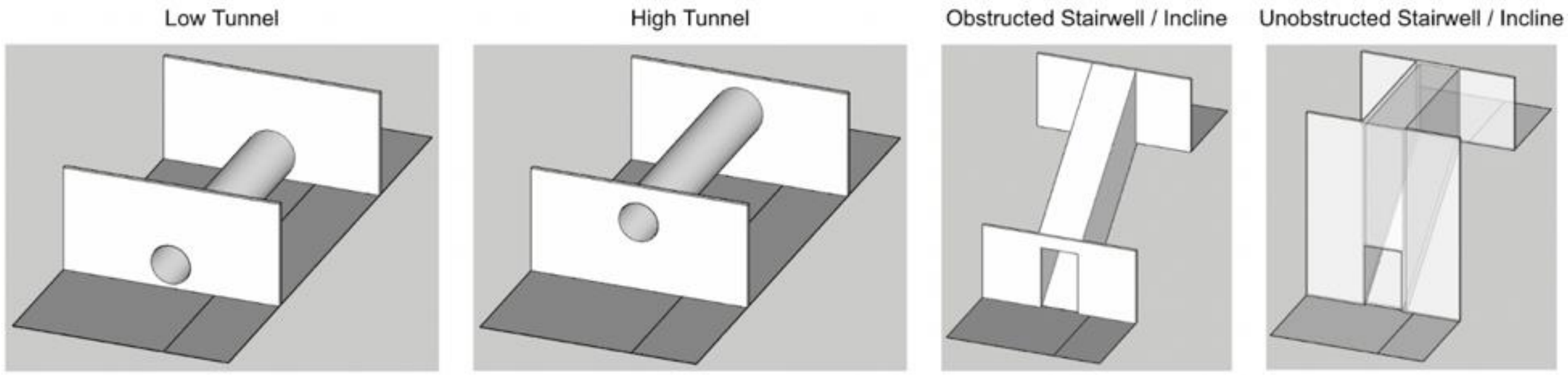

*Figure 2. Variations of the tunnel and stairwell/incline confined space navigation tests.*

## Benchmarking Results

Tests were conducted in each of the four defined confined spaces: Hallway, Tunnel, Stairwell, and Shaft. Notes:

- Average speed of the Flyability Elios 2 GOV during most navigation testing is not known because dedicated Navigation Through Confined Spaces testing was not conducted with the system. Rather, the system demonstrated multiple successful navigation runs while conducting other tests that took place in the same environments (e.g., Indoor Mapping Accuracy).
- Two flight configurations of the Skydio X2D were tested: using standard joystick teleoperation and using its touch screen to place visual waypoints for autonomous flight to these waypoints. The corresponding data points are labeled “teleoperation” and “visual waypoints,” respectively.

### Hallway

#### Environment characterization

Three instances of hallway navigation testing have been conducted, all of which took place at Muscatatuck Urban Training Center (MUTC): outdoor corridor, Conex hallway, and prison hallway.

| **MUTC Outdoor Corridor** | | | |
|---|---|---|---|
| **Environment** | **Outside confined space, area 1** | **Inside confined space** | **Outside confined space, area 2** |
| **Dimensions** | - | L x W x H: 2.4 x 1.5 x 5.5 m (8 x 5 x 18 ft) | - |
| **Lighting** | Well Lit, Shaded | Well Lit, Shaded | Well Lit, Shaded |
| **Walls** | Corrugated Steel, Conex Container | Corrugated Steel, Conex Container | Corrugated Steel, Conex Container |
| **Floor** | Dirt, Grass | Dirt, Grass | Dirt, Grass |
| **Obstruction** | Corrugated Steel, Conex Container | N/A | Corrugated Steel, Conex Container |
| **Type** | Outdoor | Outdoor | Outdoor |
| **Image** | 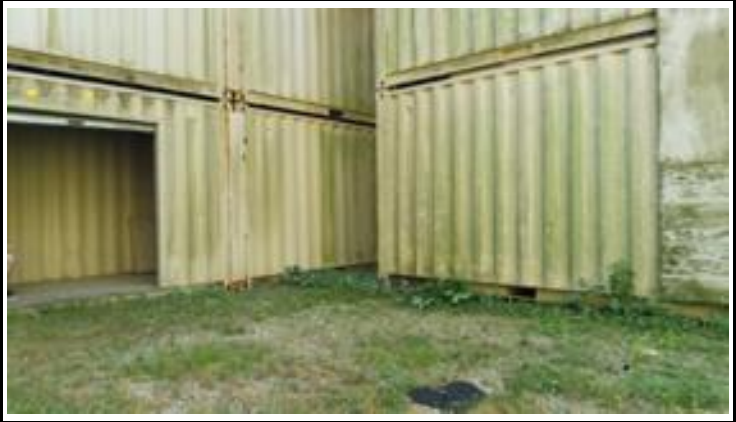 | 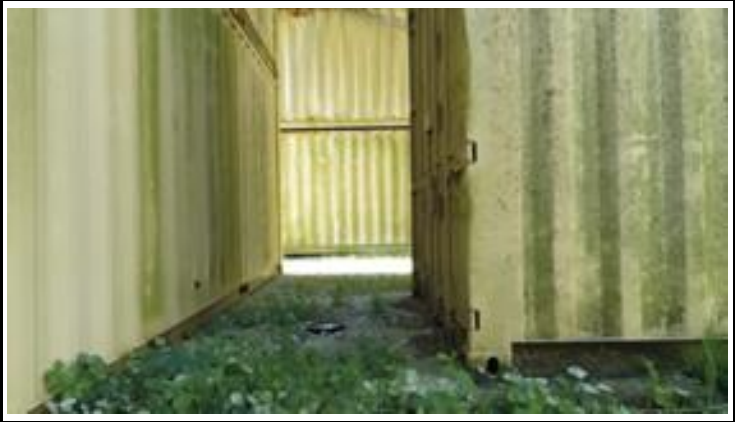 | 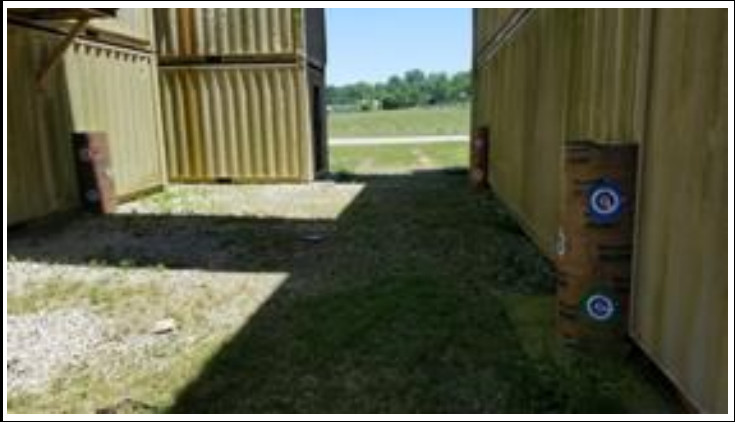 |

| **MUTC Conex Hallway** | | | |
|---|---|---|---|
| **Environment** | **Outside confined space, area 1** | **Inside confined space** | **Outside confined space, area 2** |
| **Dimensions** | - | L x W x H: 12.2 x 2.4 x 2.7 m (40 x 8 x 9 ft) | - |
| **Lighting** | Dim Indirect Sunlight | Indirect Sunlight, Sunbeams through windows | Dim Indirect Sunlight |
| **Walls** | Corrugated Steel, Conex Container | Corrugated Steel, Conex Container | Corrugated Steel, Conex Container |
| **Floor** | Wood | Wood | Wood |
| **Obstruction** | Corrugated Steel, Conex Container | N/A | Corrugated Steel, Conex Container |
| **Type** | Indoor | Indoor | Indoor |
| **Image** | 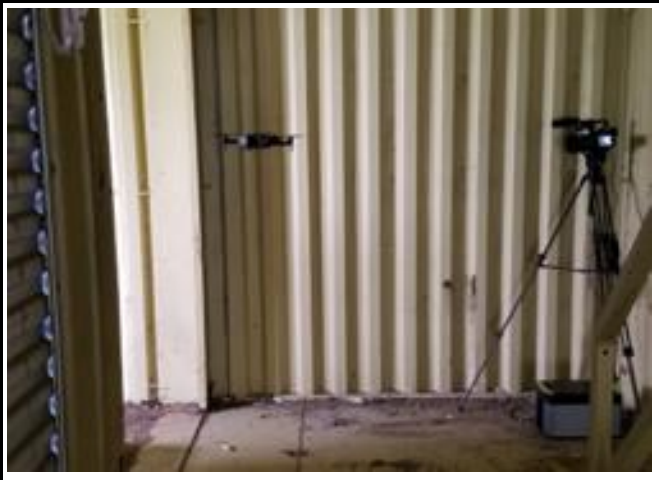 | 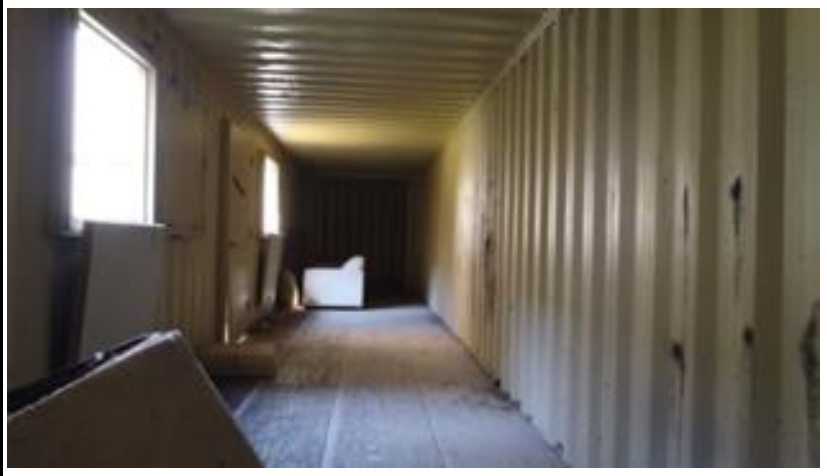 | 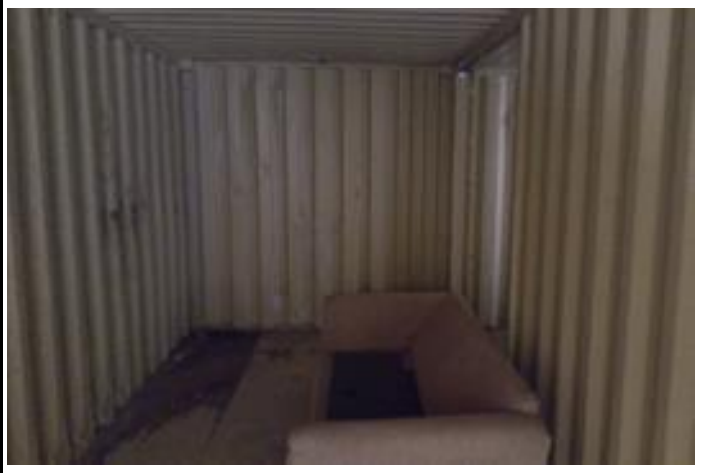 |

| MUTC Prison Hallway | | | |
|---|---|---|---|
| **Environment** | **Outside confined space, area 1** | **Inside confined space** | **Outside confined space, area 2** |
| **Dimensions** | - | L x W x H: 45.7 x 2.4 x 2.4 m (150 x 8 x 8 ft) | - |
| **Lighting** | Dark (1-10 LUX) | Mostly well lit (5-200 LUX) | Dark (1-10 LUX) |
| **Walls** | Drywall, Metal Bars | Drywall | Drywall |
| **Floor** | Concrete | Concrete | Concrete |
| **Obstruction** | Metal Bars (Jail Cell) | N/A | Bookshelf, Desk |
| **Type** | Indoor | Indoor | Indoor |
| **Image** | 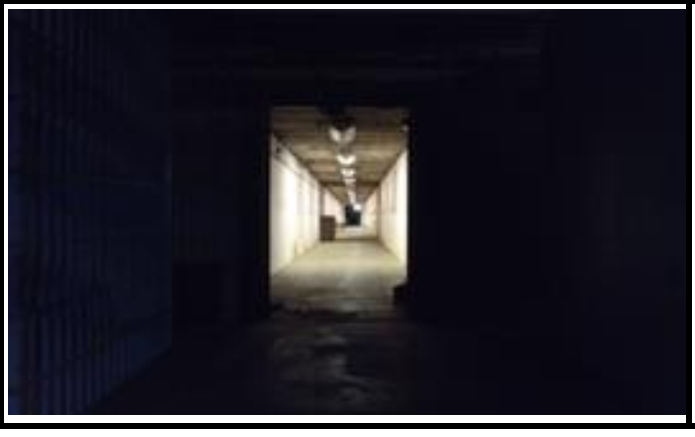 | 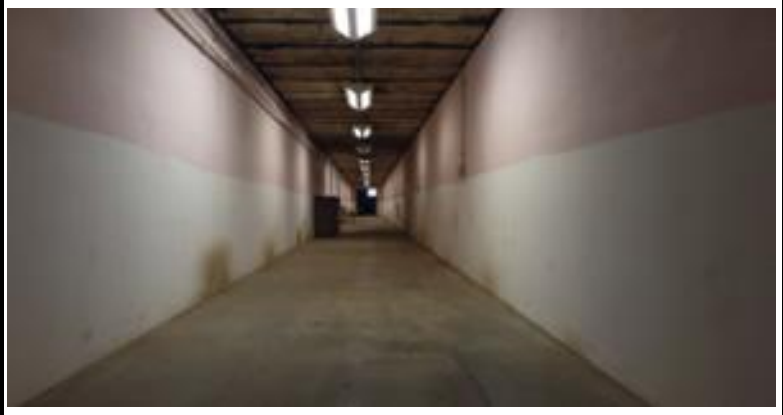 | 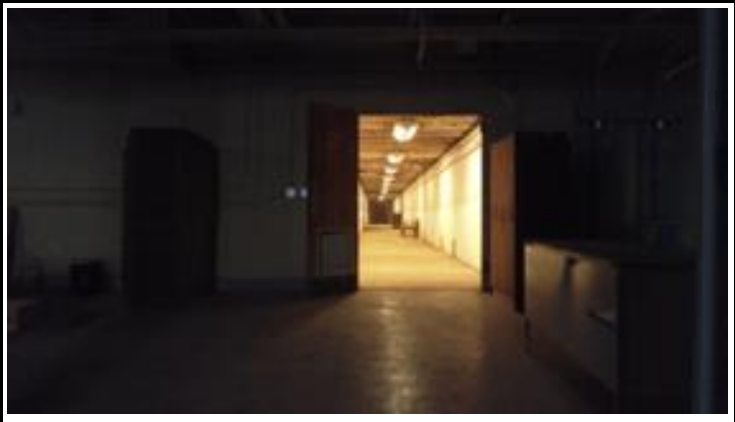 |

## Performance data

Testing was conducted as operational navigation unless only elemental navigation was possible due to NLOS communications issues or perceived operational risks.

Best in class = 100% completion for all operational navigation tests conducted

| sUAS | Metrics | MUTC Outdoor Corridor | MUTC Conex Hallway | MUTC Prison Hallway |
|---|---|---|---|---|
| Cleo Robotics Dronut X1P | Navigation type | - | - | Operational |
| | Completion | - | - | 100% |
| | Average speed (m/s) | - | - | 0.56 |
| FLIR Black Hornet PRS | Navigation type | - | - | Operational |
| | Completion | - | - | 100% |
| | Average speed (m/s) | - | - | 0.28 |
| Flyability Elios 2 GOV | Navigation type | Operational | Operational | Operational |
| | Completion | 100% | 100% | 100% |
| | Average speed (m/s) | unknown | unknown | unknown |
| Lumenier Nighthawk V3 | Navigation type | Operational | - | Operational |
| | Completion | 100% | - | 100% |
| | Average speed (m/s) | 0.39 | - | 0.42 |
| Parrot ANAFI USA GOV | Navigation type | Elemental | Operational | Operational |
| | Completion | 100% | 100% | 100% |
| | Average speed (m/s) | 0.52 | 0.28 | 0.56 |
| Skydio X2D, teleoperation | Navigation type | Operational | - | Operational |
| | Completion | 100% | - | 100% |
| | Average speed (m/s) | 0.39 | - | 0.83 |
| Skydio X2D, visual waypoints | Navigation type | Operational | - | Operational |
| | Completion | 100% | - | 25% |
| | Average speed (m/s) | 0.26 | - | N/A |
| Teal Golden Eagle* | Navigation type | X | X | X |
| | Completion | X | X | X |
| | Average speed (m/s) | X | X | X |
| Vantage Robotics Vesper | Navigation type | Elemental | Operational | Operational |
| | Completion | 100% | 25% | 25% |
| | Average speed (m/s) | 0.39 | 0.14 | 0.83 |
| | **Best in class** | **Hallway Navigation** | | |
| | | Cleo Robotics Dronut X1P<br>FLIR Black Hornet PRS<br>Flyability Elios 2 GOV<br>Lumenier Nighthawk V3<br>Parrot ANAFI USA GOV<br>Skydio X2D | | |

*Note: Due to instability of the Teal Golden Eagle when flying indoors, it was not evaluated in this test.

Only test results with 100% completion are shown in the chart below.

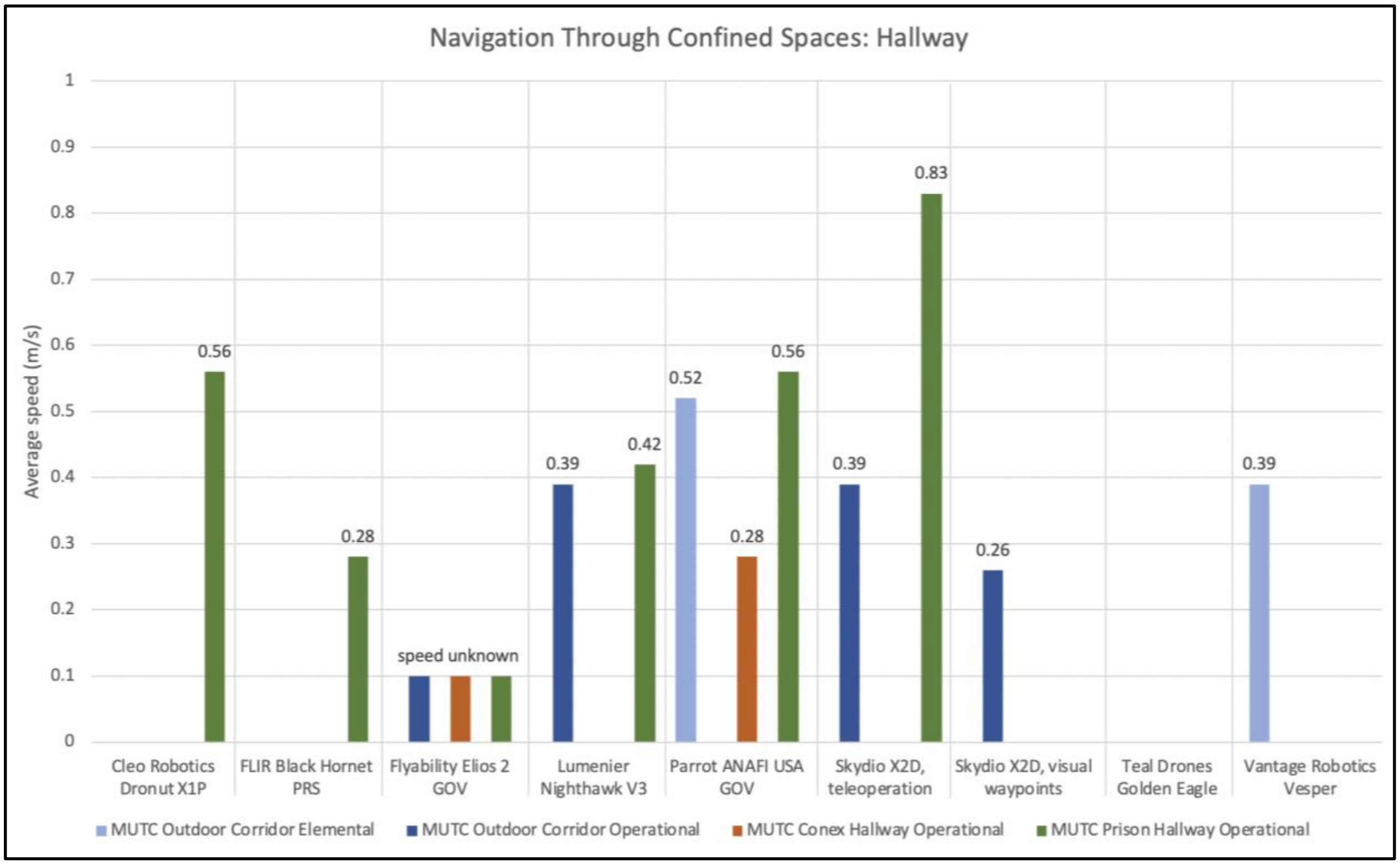

## Tunnel

### Environment characterization

Only two one-off tests of tunnel navigation were conducted: MUTC tunnel and NERVE tunnel.

| MUTC Tunnel | | | |
|---|---|---|---|
| **Environment** | **Outside confined space, area 1** | **Inside confined space** | **Outside confined space, area 2** |
| **Dimensions** | - | W x H: 1.3 x 1.3 m (4.4 x 4.2 ft), L unknown | - |
| **Lighting** | Shaded Sunlight | Dark | Sunlight |
| **Walls** | N/A | Dirt | N/A |
| **Floor** | Rocks, Dirt, Grass | Dirt, Mud | Grass |
| **Obstruction** | Metal Door | N/A | N/A |
| **Type** | Outdoor | Indoor | Outdoor |
| **Image** | 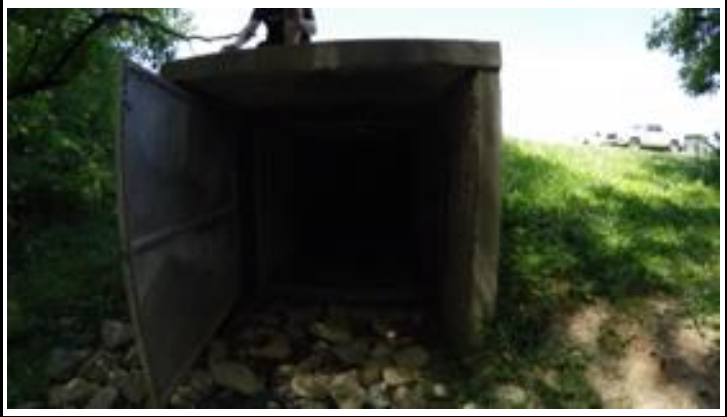 | 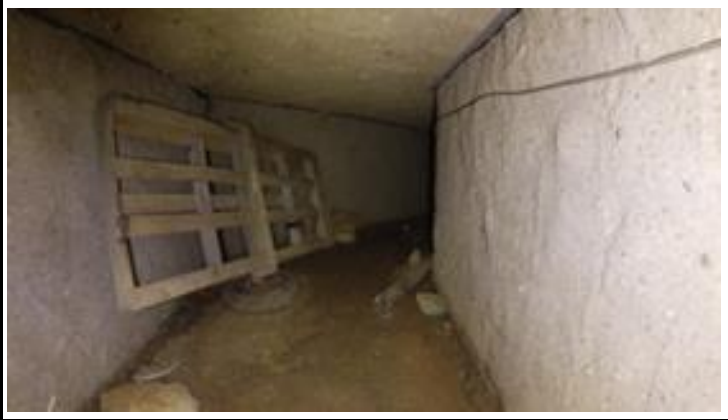 | 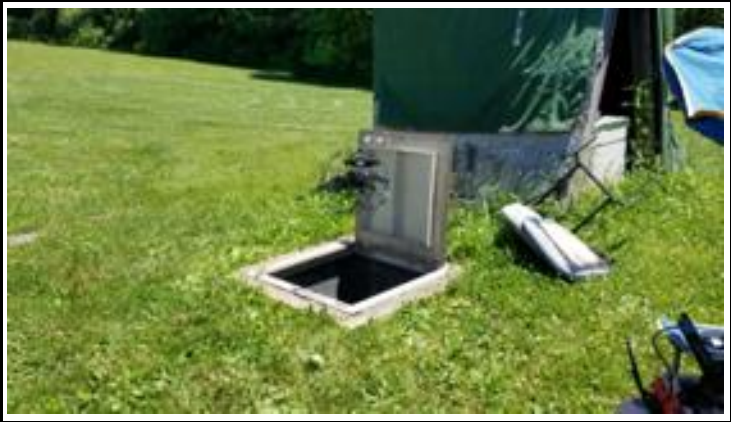 |

| NERVE Tunnel | | | |
|---|---|---|---|
| **Environment** | **Outside confined space, area 1** | **Inside confined space** | **Outside confined space, area 2** |
| **Dimensions** | - | W x H: 1.2 x 1.2 x 4.6 m (4 x 4 x 15 ft) | - |
| **Lighting** | Well lit | Well lit | Well lit |
| **Walls** | Drywall | Vinyl | Drywall |
| **Floor** | Concrete | Vinyl | Concrete |
| **Obstruction** | N/A | N/A | N/A |
| **Type** | Indoor | Indoor | Indoor |
| **Image** | 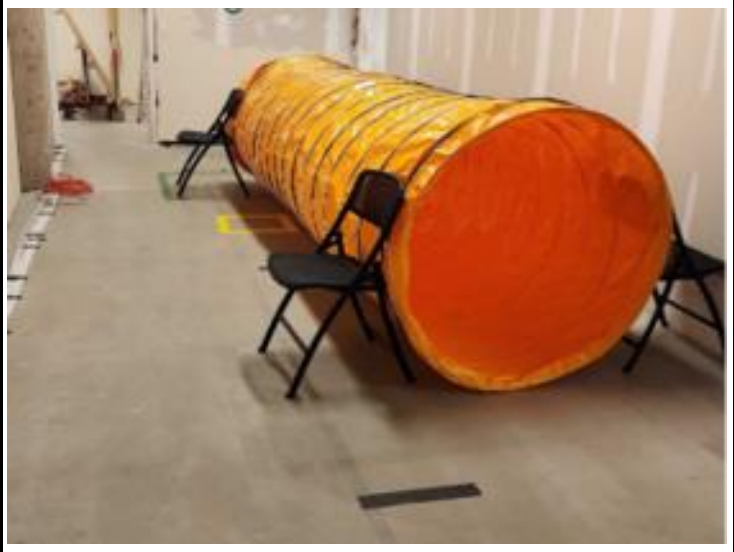 | 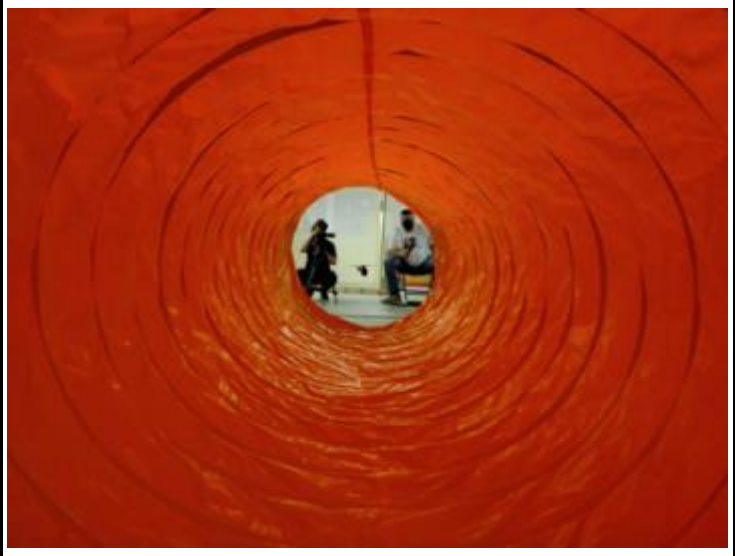 | 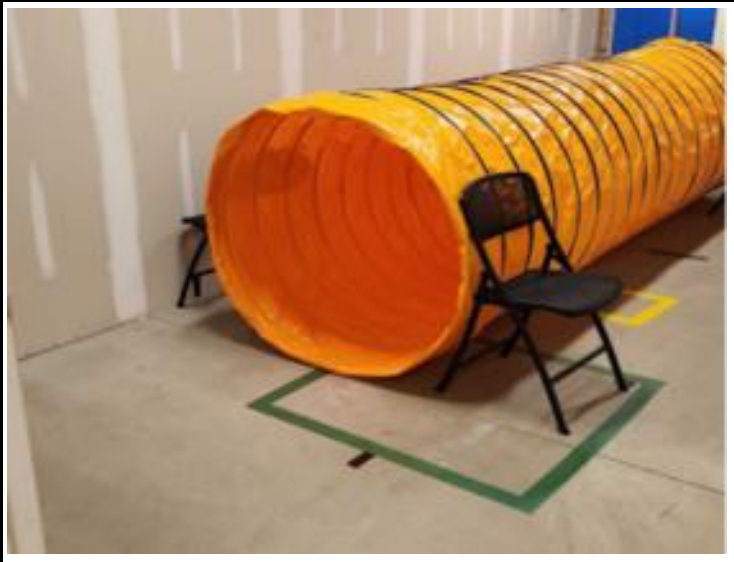 |

## Performance data

Testing was conducted as operational navigation unless only elemental navigation was possible due to NLOS communications issues or perceived operational risks.

Due to the lack of tunnel testing conducted, no best in class criteria is specified.

| sUAS | Metrics | MUTC Tunnel | NERVE Tunnel |
|---|---|---|---|
| Cleo Robotics Dronut X1P | Navigation type | - | - |
| | Completion | - | - |
| | Average speed (m/s) | - | - |
| FLIR Black Hornet PRS | Navigation type | - | Elemental |
| | Completion | - | 90% |
| | Average speed (m/s) | - | 0.2 |
| Flyability Elios 2 GOV | Navigation type | Elemental | - |
| | Completion | 100% | - |
| | Average speed (m/s) | unknown | - |
| Lumenier Nighthawk V3 | Navigation type | - | - |
| | Completion | - | - |
| | Average speed (m/s) | - | - |
| Parrot ANAFI USA GOV | Navigation type | - | - |
| | Completion | - | - |
| | Average speed (m/s) | - | - |
| Skydio X2D, teleoperation | Navigation type | - | - |
| | Completion | - | - |
| | Average speed (m/s) | - | - |
| Skydio X2D, visual waypoints | Navigation type | - | - |
| | Completion | - | - |
| | Average speed (m/s) | - | - |
| Teal Golden Eagle | Navigation type | - | - |
| | Completion | - | - |
| | Average speed (m/s) | - | - |
| Vantage Robotics Vesper | Navigation type | - | - |
| | Completion | - | - |
| | Average speed (m/s) | - | - |

## Stairwell

### Environment characterization

One instance of stairwell navigation testing was conducted at the MUTC subway platform.

| MUTC Subway Platform | | | |
|---|---|---|---|
| **Environment** | **Outside confined space, area 1** | **Inside confined space** | **Outside confined space, area 2** |
| **Dimensions** | - | L x W x H: 11.3 x 2.4 x 4.6 m (37 x 8 x 15 ft) | - |
| **Lighting** | Sunlight | Indirect Sunlight | Dim Indirect Sunlight |
| **Walls** | N/A | Concrete | Concrete, Metal |
| **Floor** | Grass, Concrete | Concrete | Concrete |
| **Obstruction** | None | Unobstructed Stairwell | Wall, Subway Car |
| **Type** | Outdoor | Indoor | Indoor |
| **Image** | 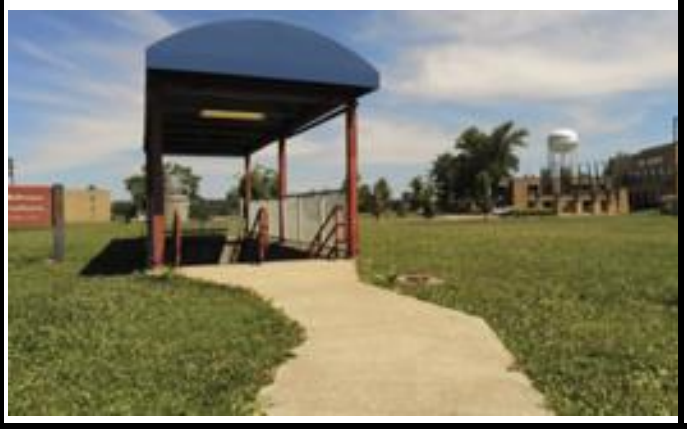 | 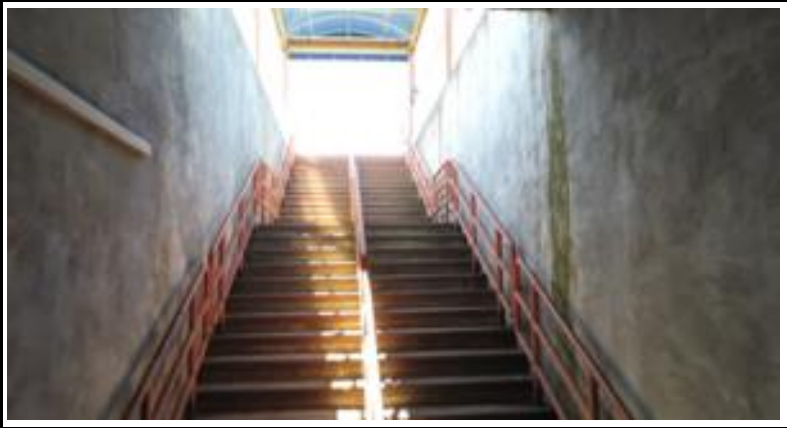 | 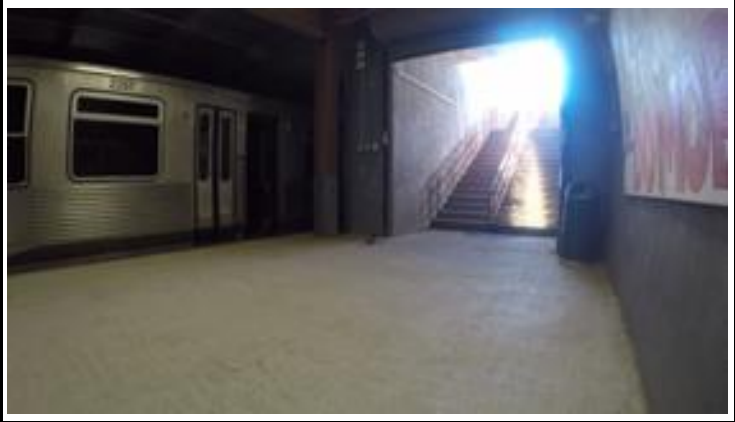 |

## Performance data

Testing was conducted as operational navigation unless only elemental navigation was possible due to NLOS communications issues or perceived operational risks.

Best in class = 100% completion operational navigation

| sUAS | Metrics | MUTC Subway Platform, Descension and Ascension |
|---|---|---|
| Cleo Robotics Dronut X1P | Navigation type | Elemental |
| | Completion | 100% |
| | Average speed (m/s) | 0.28 |
| FLIR Black Hornet PRS | Navigation type | Operational |
| | Completion | 67% |
| | Average speed (m/s) | N/A |
| Flyability Elios 2 GOV | Navigation type | Operational |
| | Completion | 100% |
| | Average speed (m/s) | 0.28 |
| Lumenier Nighthawk V3* | Navigation type | - |
| | Completion | - |
| | Average speed (m/s) | - |
| Parrot ANAFI USA GOV | Navigation type | Operational |
| | Completion | 100% |
| | Average speed (m/s) | 0.19 |
| Skydio X2D, teleoperation | Navigation type | Operational |
| | Completion | 100% |
| | Average speed (m/s) | 0.19 |
| Skydio X2D, visual waypoints | Navigation type | - |
| | Completion | - |
| | Average speed (m/s) | - |
| Teal Golden Eagle | Navigation type | Elemental |
| | Completion | 0% |
| | Average speed (m/s) | N/A |
| Vantage Robotics Vesper | Navigation type | Elemental |
| | Completion | 100% |
| | Average speed (m/s) | 0.28 |
| | **Best in class** | **Stairwell Navigation** |
| | | Flyability Elios 2 GOV<br>Parrot ANAFI USA GOV<br>Skydio X2D |

*Note: The Lumenier Nighthawk V3 was damaged at the time of testing, so it was not evaluated in this test.

Only test results with 100% completion are shown in the chart below.

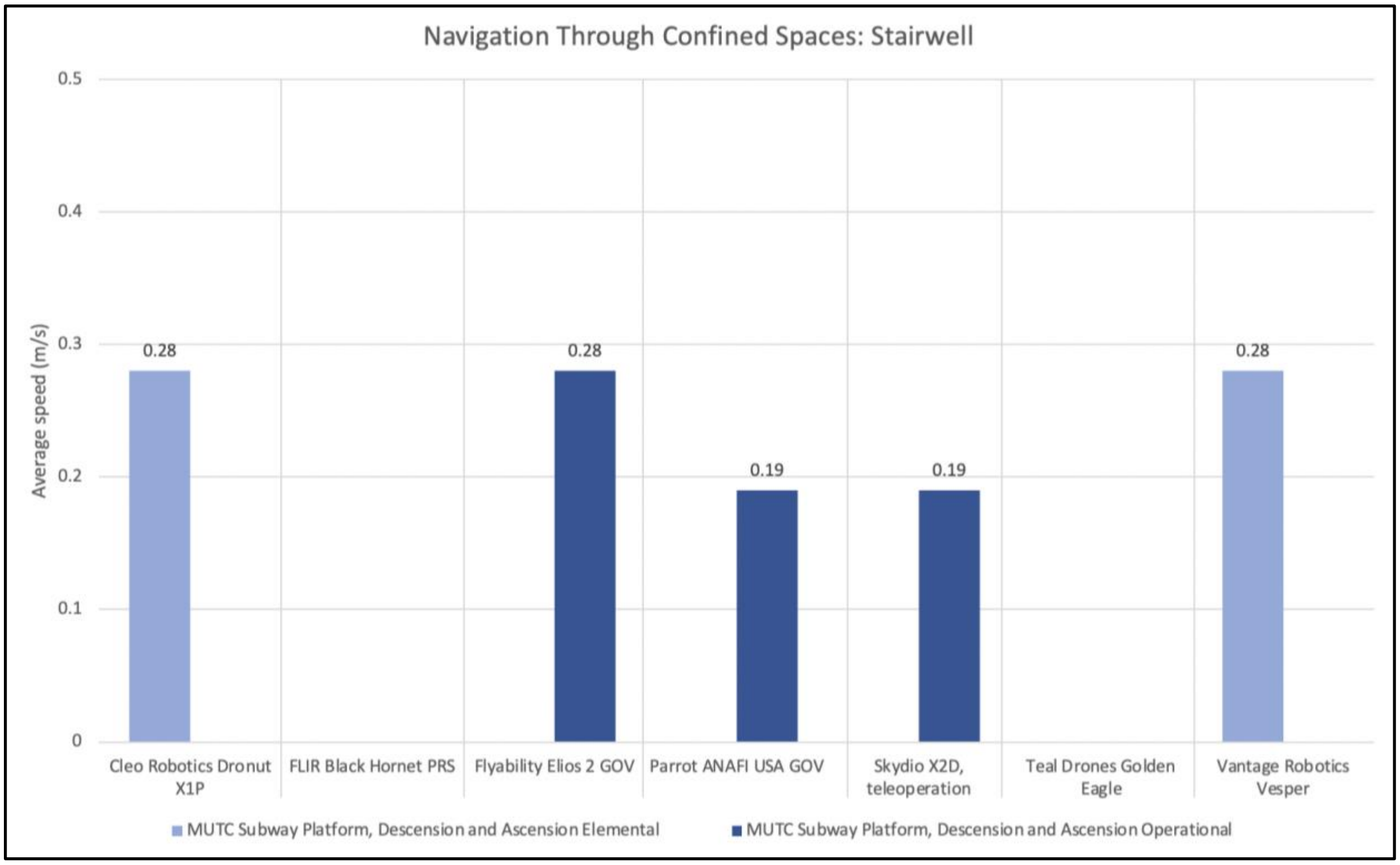

## Shaft

### Environment characterization

One instance of shaft navigation testing was conducted at the MUTC hotel trainer.

| MUTC Shaft | | | |
|---|---|---|---|
| **Environment** | **Outside confined space, area 1** | **Inside confined space** | **Outside confined space, area 2** |
| **Dimensions** | - | L x W x H: 1.2 x 2.4 x 6.1 m (4 x 8 x 20 ft) | - |
| **Lighting** | Dark (<100 LUX) | Indirect Sunlight (100-200 LUX) | Dark (<100 LUX) |
| **Walls** | Corrugated Steel, Conex Container | Corrugated Steel, Conex Container | Corrugated Steel, Conex Container |
| **Floor** | Wood | Safety Tarp | Wood |
| **Obstruction** | N/A | N/A | N/A |
| **Type** | Indoor | "Outdoor" (Open roof) | Indoor |
| **Image** | 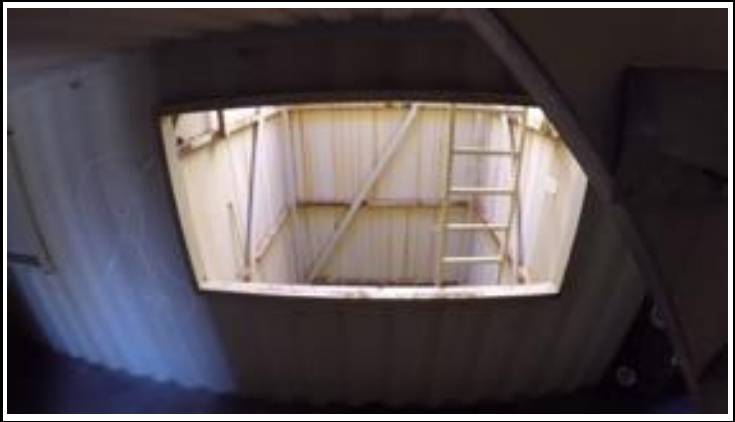 | 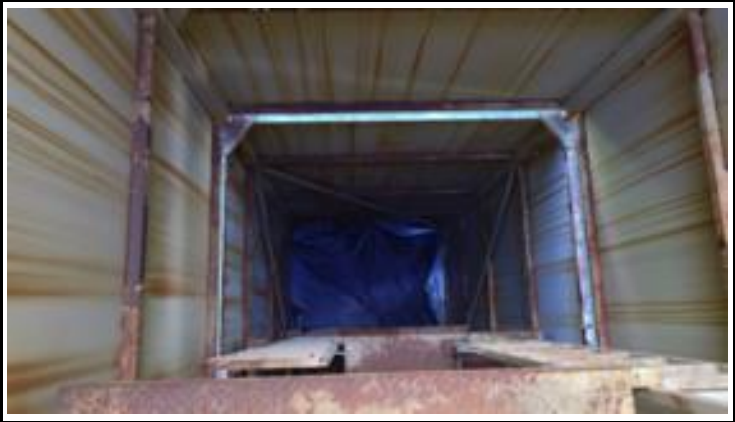 | 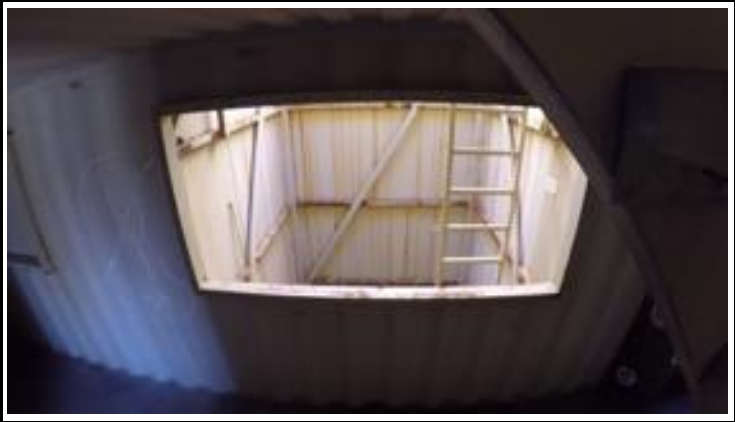 |

## Performance data

Testing was only conducted as operational navigation.

Best in class = 100% completion for all operational navigation tests conducted

| sUAS | Metrics | MUTC Shaft, Descension | MUTC Shaft, Ascension |
|---|---|---|---|
| Cleo Robotics Dronut X1P | Navigation type | Operational | Operational |
| | Completion | 100% | 100% |
| | Average speed (m/s) | 0.24 | 0.24 |
| FLIR Black Hornet PRS | Navigation type | Operational | Operational |
| | Completion | 100% | 100% |
| | Average speed (m/s) | 0.08 | 0.08 |
| Flyability Elios 2 GOV | Navigation type | Operational | Operational |
| | Completion | 100% | 100% |
| | Average speed (m/s) | unknown | unknown |
| Lumenier Nighthawk V3 | Navigation type | Operational | Operational |
| | Completion | 50% | 0% |
| | Average speed (m/s) | 0.06 | N/A |
| Parrot ANAFI USA GOV | Navigation type | Operational | Operational |
| | Completion | 50% | 0% |
| | Average speed (m/s) | 0.06 | N/A |
| Skydio X2D, teleoperation | Navigation type | - | - |
| | Completion | - | - |
| | Average speed (m/s) | - | - |
| Skydio X2D, visual waypoints | Navigation type | - | - |
| | Completion | - | - |
| | Average speed (m/s) | - | - |
| Teal Golden Eagle* | Navigation type | X | X |
| | Completion | X | X |
| | Average speed (m/s) | X | X |
| Vantage Robotics Vesper | Navigation type | - | - |
| | Completion | - | - |
| | Average speed (m/s) | - | - |
| | **Best in class** | **Shaft Navigation** | |
| | | Cleo Robotics Dronut X1P<br>FLIR Black Hornet PRS<br>Flyability Elios 2 GOV | |

*Note: Due to instability of the Teal Golden Eagle when flying indoors, it was not evaluated in this test.

Only test results with 100% completion are shown in the chart below.

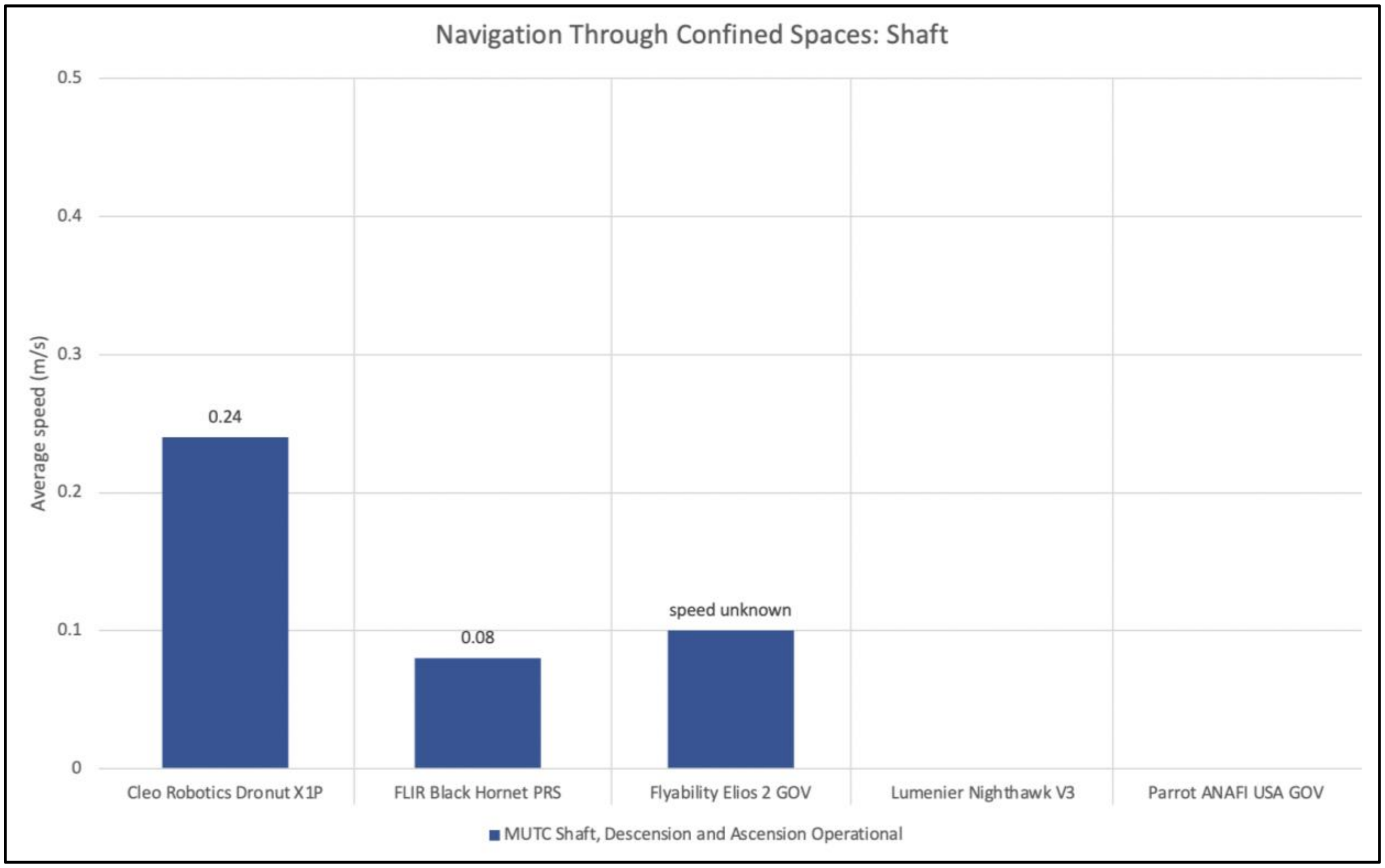

# Mapping

## Indoor Mapping Resolution

Affiliated publications: [Norton et al., 2021]

### Summary of Test Method

This test method is comprised of two separate tests for evaluating indoor mapping resolution:

Interior Boundaries: The operator maneuvers the sUAS along a specified approximate trajectory through a standard set of clearances which define a consistent standoff distance between the centroid of the sUAS and the walls and floor. While moving along the trajectory, the sUAS maps the environment while tilting and panning its mapping camera(s) and sensors as needed at each defined point, while maintaining a forward orientation (i.e., not yawing in place). Each point is navigated to sequentially and then back out in reserve order. If the sUAS is not able to reach a point due to confined space restrictions, it may reverse at any point in the sequence. The test can be run in lighted (100 lux or greater) or dark (less than 1 lux) conditions.

Shape Accuracy: The operator maneuvers the sUAS around a single split-cylinder fiducial mounted on two sides of a wall in order to map it, following manufacturer recommendations on effective flight and camera maneuvering techniques for mapping. There are no restrictions on the flight path taken. The collected data is downloaded to generate a map. The test can be run in lighted (100 lux or greater) or dark (less than 1 lux) conditions.

These tests should be run as a prerequisite to the Indoor Map Accuracy test.

### Benchmarking Results

Note: performance data is only shown for the Flyability Elios 2 GOV due to it being the only system with indoor mapping capabilities.

No best in class criteria is specified.

**Environment characterization**

| NERVE Indoor Mapping Resolution Apparatus | |
|---|---|
| **Lighting** | Dark (<100 LUX) (images below were taken while lit) |
| **Walls** | Drywall, Wood, Concrete |
| **Floor** | Concrete |
| **Type** | Indoor |
| **Image** | 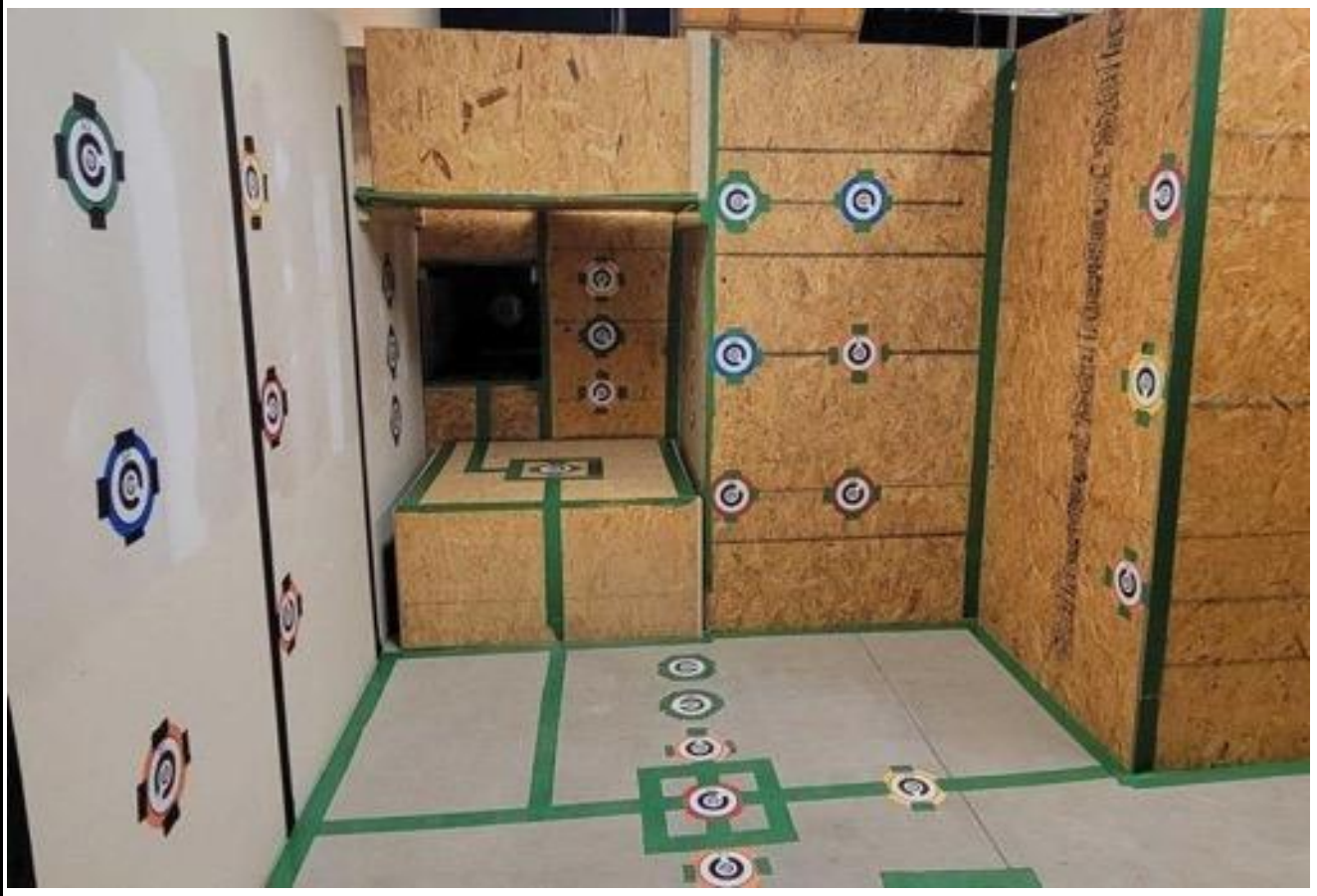 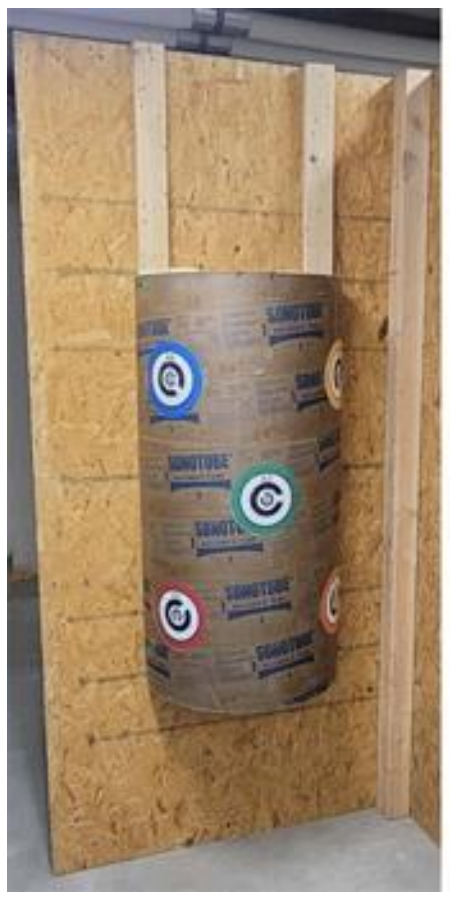 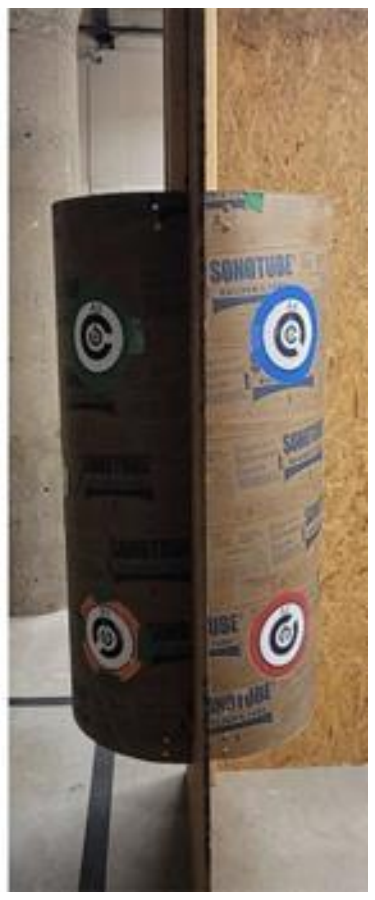 |

## Performance data

### Interior Boundaries

Point clouds and photogrammetric map with singulated visual acuity targets shown below.

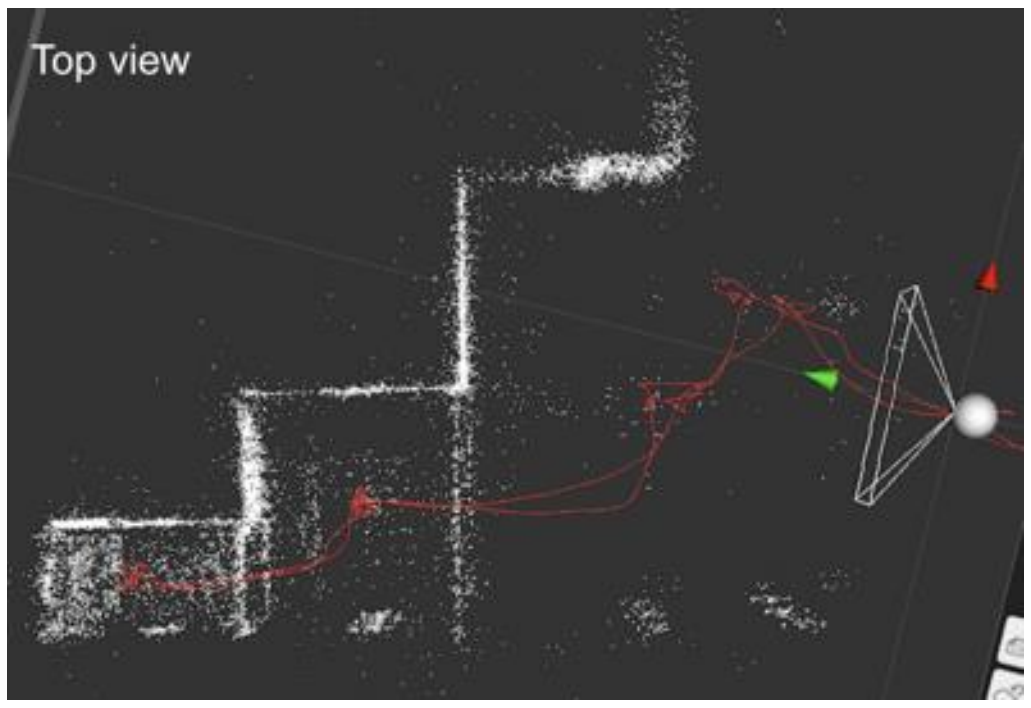

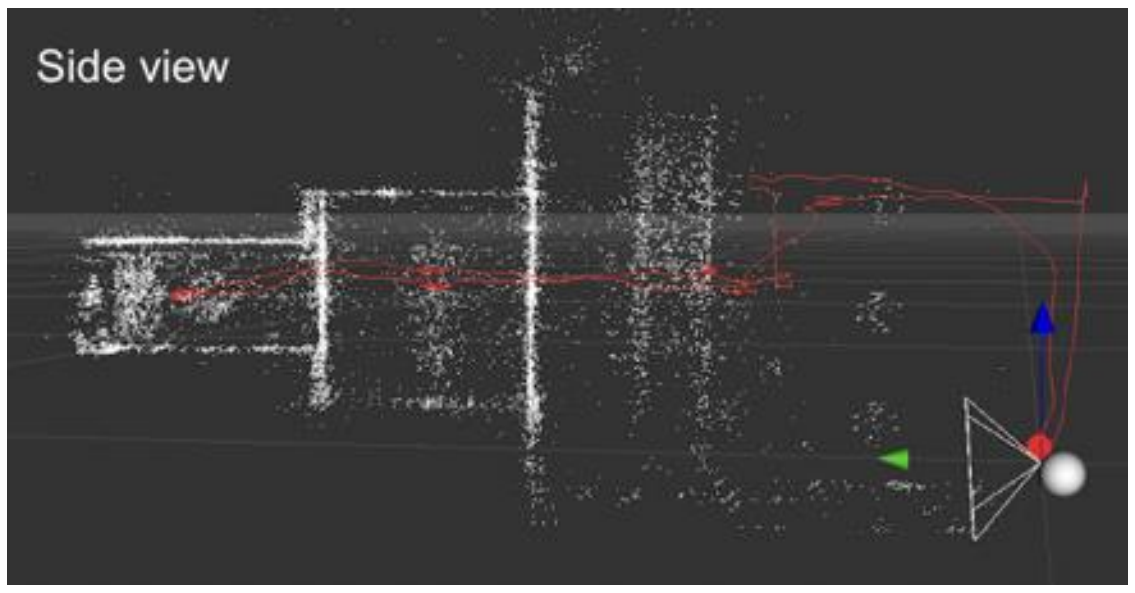

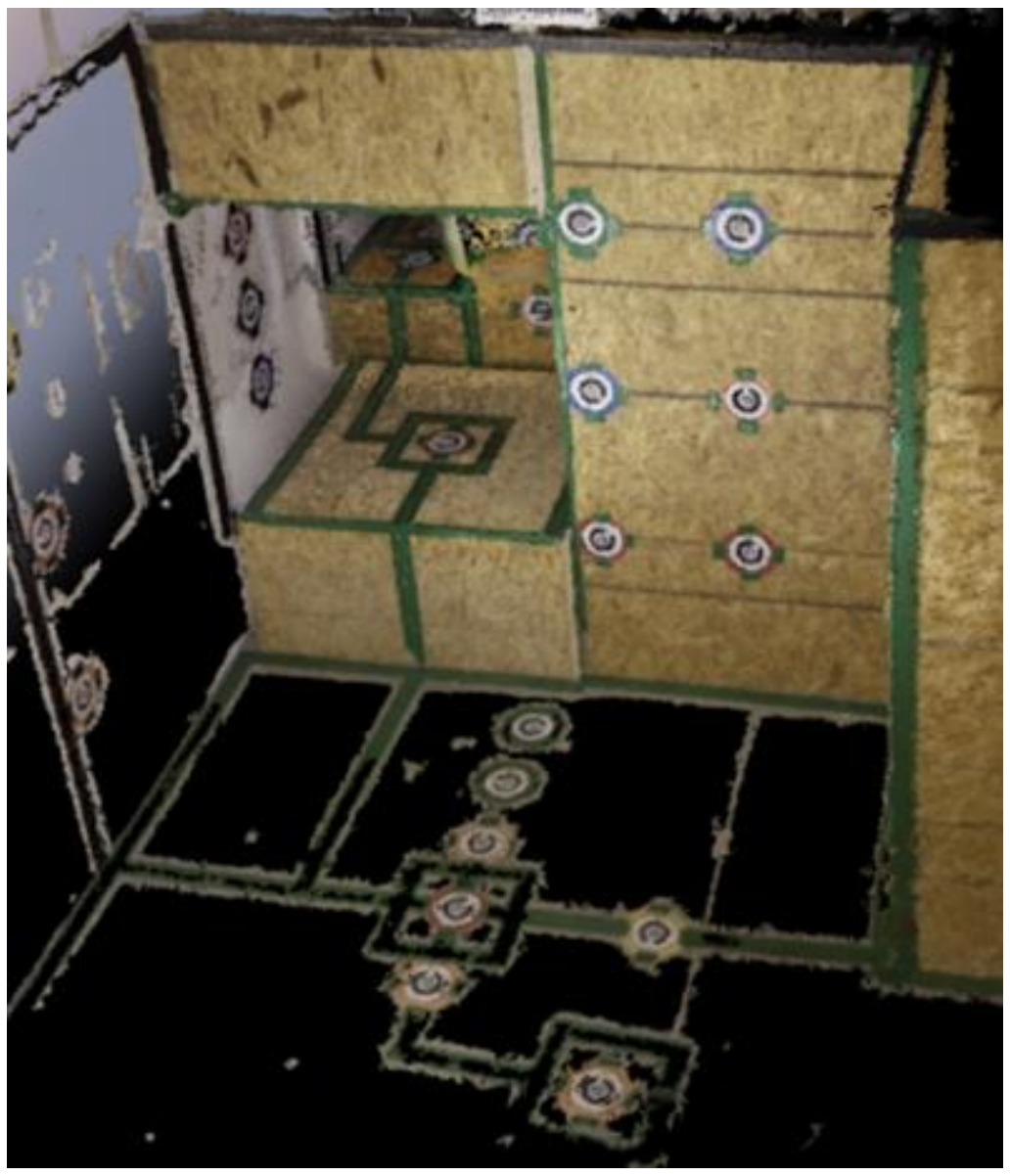

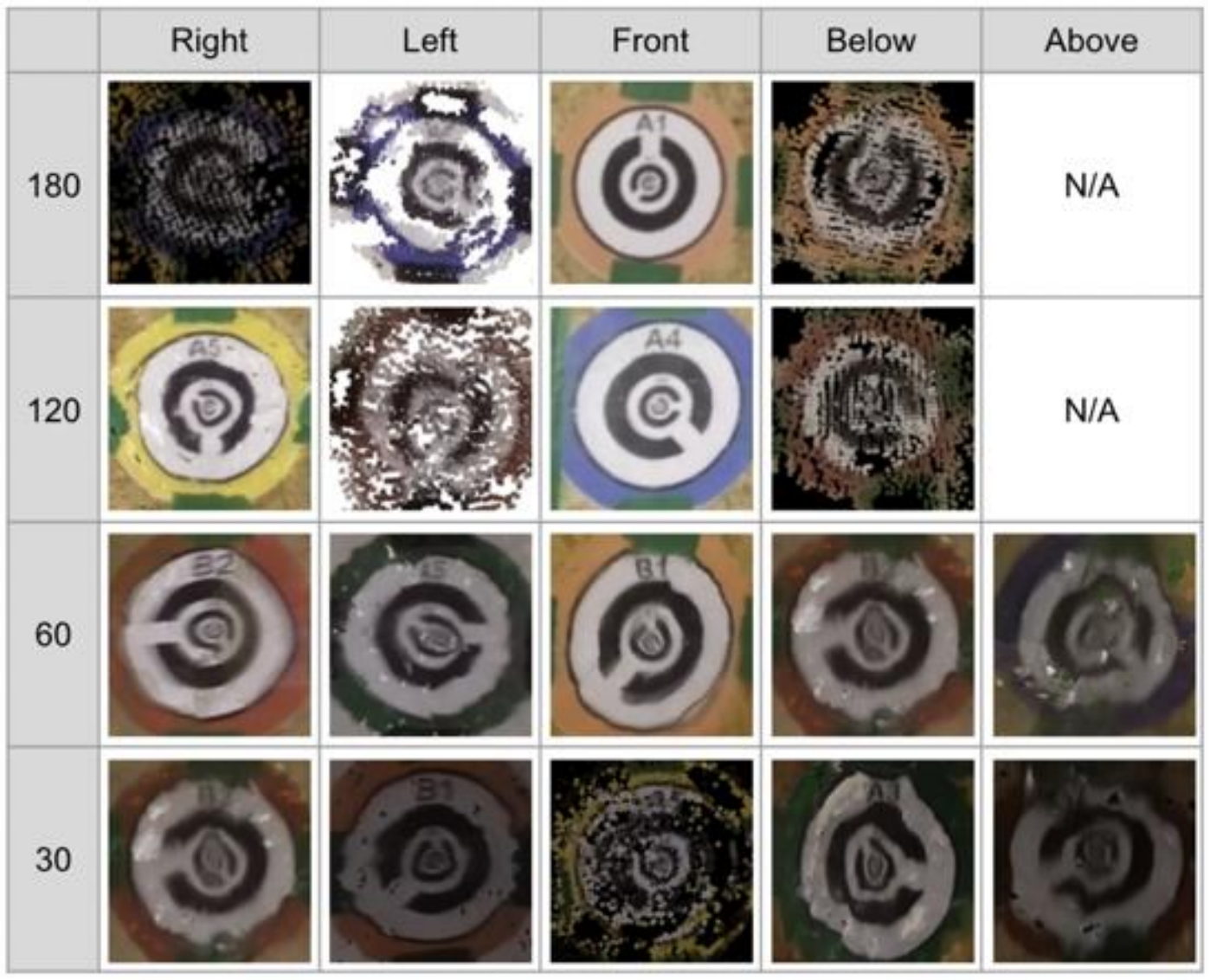

| sUAS | Stand off (cm) | Right | | | Left | | | Front | | | Below | | | Above | | | Average | | |
|---|---|---|---|---|---|---|---|---|---|---|---|---|---|---|---|---|---|---|---|
| | | Dim Acc (%) | FOV (%) | Acuity (mm) | Dim Acc (%) | FOV (%) | Acuity (mm) | Dim Acc (%) | FOV (%) | Acuity (mm) | Dim Acc (%) | FOV (%) | Acuity (mm) | Dim Acc (%) | FOV (%) | Acuity (mm) | Dim Acc (%) | FOV (%) | Acuity (mm) |
| Flyability Elios 2 GOV | 180 | 100 | 50 | 20 | 100 | 50 | 8 | 100 | 100 | 3 | 100 | 75 | 20 | n/a | n/a | n/a | 100 | 69 (+/- 24) | 13 (+/- 9) |
| | 120 | 100 | 100 | 3 | 100 | 100 | 20 | 100 | 100 | 3 | 100 | 100 | 20 | n/a | n/a | n/a | 100 | 100 | 12 (+/- 10) |
| | 60 | 100 | 100 | 3 | 100 | 100 | 8 | 100 | 100 | 3 | 100 | 100 | 8 | 100 | 100 | 8 | 100 | 100 | 6 (+/- 3) |
| | 30 | 100 | 100 | 8 | 100 | 100 | 3 | 100 | 100 | 8 | 100 | 100 | 3 | 100 | 100 | 8 | 100 | 100 | 6 (+/- 3) |

**Shape Accuracy**

Point cloud (overhead view) and photogrammetric map with singulated visual acuity target shown below.

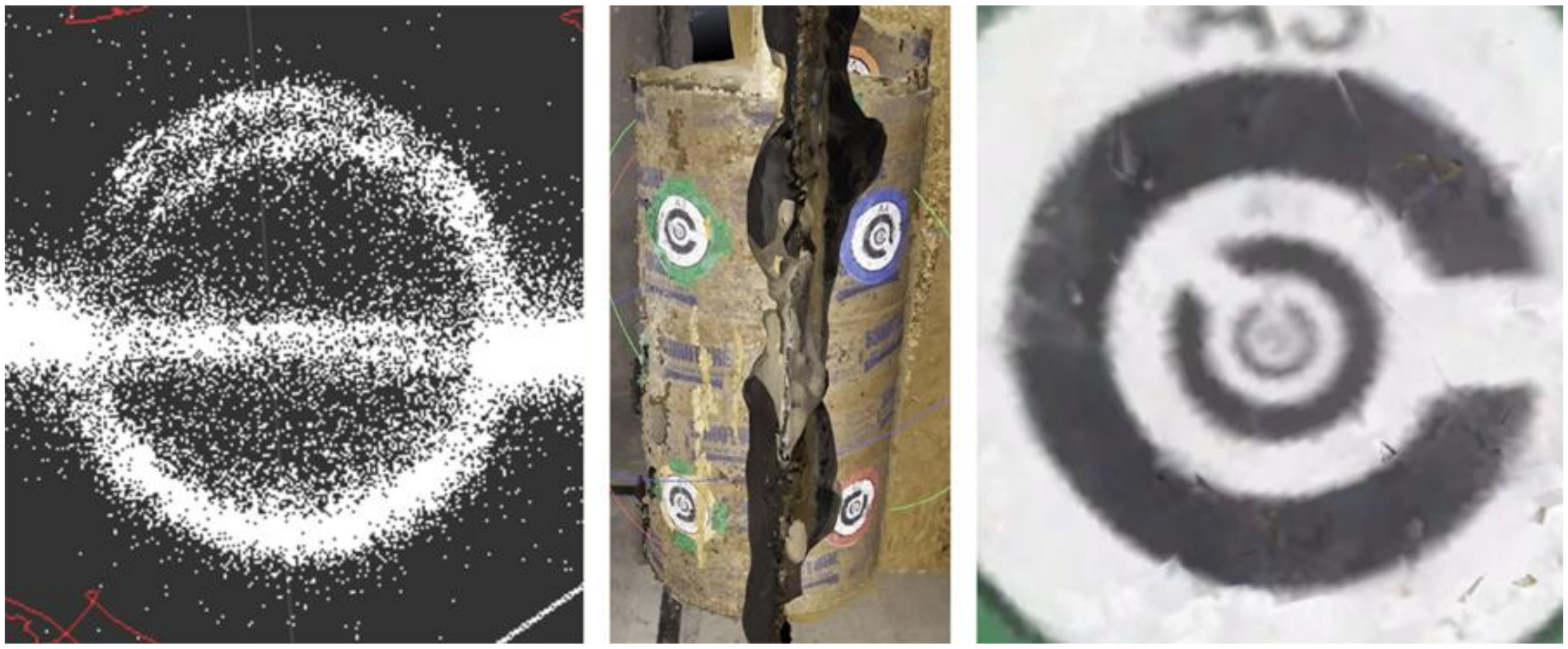

| sUAS | Shape accuracy | Acuity (mm) | Mapping time (min) | Processing time (min) |
|---|---|---|---|---|
| Flyability Elios 2 GOV | Complete | 3 | 7 | 7 (point cloud)<br>120 (photogrammetry) |

## Indoor Mapping Accuracy

Affiliated publications: [Norton et al., 2021]

### Summary of Test Method

The operator flies the sUAS through an environment to collect mapping data. A series of split-cylinder fiducials are positioned throughout the environment. An accurate ground truth of the environment must be available in order to compare the sUAS map. The ground truth for comparison is a map of the environment that is generated using a more precise method with high confidence of accuracy (e.g., industrial ground robot, handheld lidar system). An accurate 3D ground truth may be very expensive and/or difficult to generate, whereas a 2D map can be more easily gathered (e.g., architectural layout, dimensional measurement). Multiple flights may be conducted in order to change batteries. The collected data is downloaded to generate a map; if multiple flights were conducted, multiple maps will be generated for each incremental flight (e.g., map of flight 1, map or flights 1+2, etc.) and evaluated separately. For maps generated from multiple flights, evaluations should differentiate between automatic alignment of the individual maps and manual alignment performed by an operator. The test can be run in lighted (100 lux or greater) or dark (less than 1 lux) conditions, as either an elemental or operational test:

Elemental Mapping: The operator may maintain line-of-sight with the sUAS such as by following the system with the OCU throughout the environment to maintain communications link. This allows the system's map generation capability to be evaluated in as close to an ideal setting as possible.

Operational Mapping: The operator remains at the launch point during execution, unable to maintain line of sight throughout the test, without prior knowledge of the layout of the space. This is similar to an actual operational mission, including all related sUAS communications and operator situation awareness that may arise (e.g., losing comms link at range and/or through obstructions, monitoring battery life such that the sUAS can be flown back before it dies). If multiple flights are conducted, the sUAS must be flown back to the launch point where the operator is stationed in order to change batteries.

## Benchmarking Results

Note: performance data is only shown for the Flyability Elios 2 GOV due to it being the only system with indoor mapping capabilities.

No best in class criteria is specified.

# Horizontal Mapping

## Environment characterization

| NERVE Test Course | | | | | | | | | | | | |
|---|---|---|---|---|---|---|---|---|---|---|---|---|
| **Dimensions (L x W) and area** | 25.5 x 17.5 m = 446.3 $m^2$ (83.7 x 57.4 ft = 4804.4 $ft^2$) | | **Fiducials** | | | | | | | | | |
| **Number of rooms** | 12 | **Metrics** | **A** | **B** | **C** | **D** | **E** | **F** | **G** | **H** | **I** | **J** |
| **Number of floors** | 1 | **Minimum traversal distance (m)** | 11 | 8 | 35 | 5 | 12 | 7 | 27 | 7 | 16 | 10 |
| **Lighting** | Dark (<1 LUX) | **Minimum orientation changes** | 2 | 2 | 7 | 2 | 3 | 2 | 5 | 2 | 3 | 2 |
| **Walls** | Concrete, Drywall | **Difficulty rating** | M | L | H | L | M | L | H | L | M | L |
| **Floor** | Concrete | | | | | | | | | | | |
| **Type** | Indoor | | | | | | | | | | | |
| **Image** | 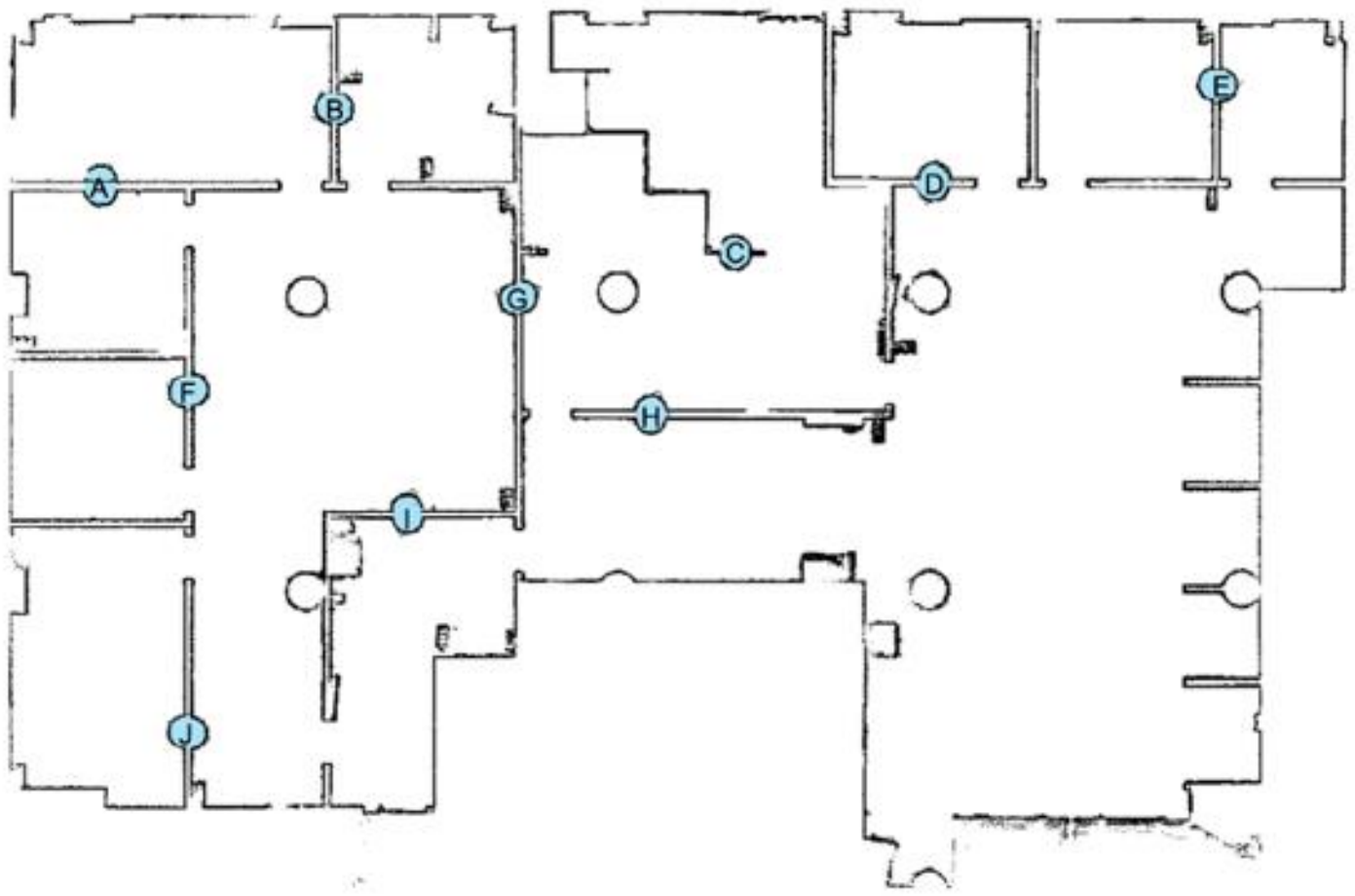 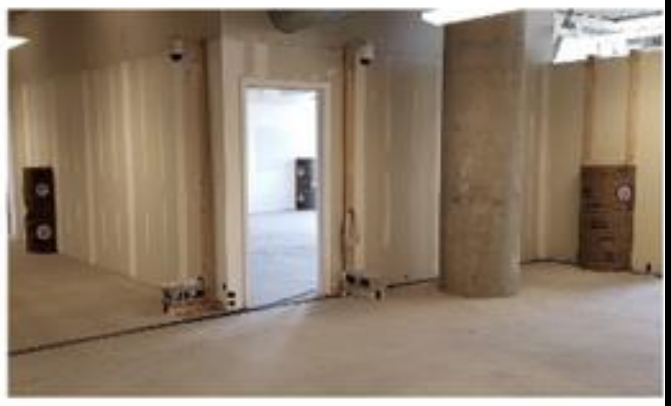 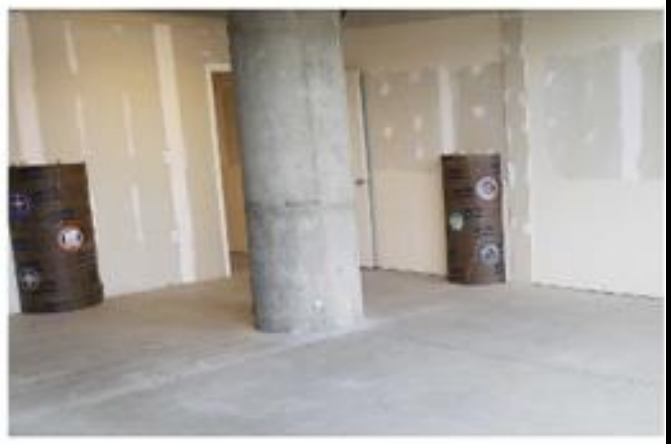 | | | | | | | | | | | |

## Performance data

### Two flights mapped with Flyability Inspector 3.0

Point cloud map generated from two flights (first flight in black, second flight in blue) merged automatically using CloudCompare shown below (overhead view) with singulated and labeled fiducials.

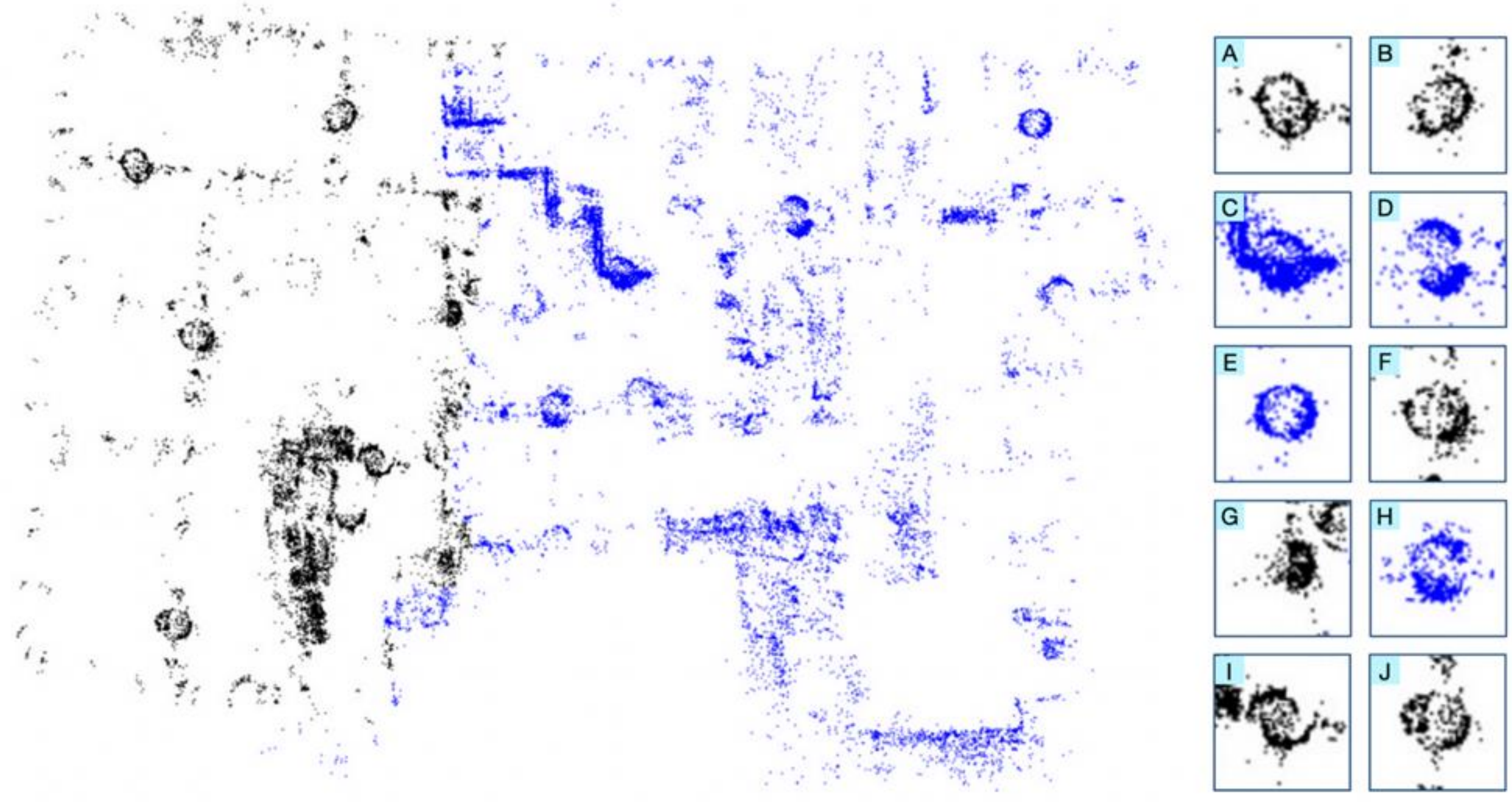

Performance data is provided for just the first flight and for the combined first and second flight in the table below.

| sUAS | Flight | Metrics | A | B | C | D | E | F | G | H | I | J | Map |
|---|---|---|---|---|---|---|---|---|---|---|---|---|---|
| Flyability Elios 2 GOV | 1 | Shape accuracy | C | C | | | | C | I | | S | | 60% |
| | | Coverage | ✓ | ✓ | X | X | X | ✓ | // | X | ✓ | ✓ | 55% |
| | | Mapping time (min) | | | | | | | | | | | 9 |
| | | Processing time (min) | | | | | | | | | | | 9 |
| | | Global error (cm) | | | | | | | | | | | 8 |
| | 1+2 | Shape accuracy | C | C | C | S | C | C | I | C | S | C | 70% |
| | | Coverage | ✓ | ✓ | ✓ | ✓ | ✓ | ✓ | // | ✓ | ✓ | ✓ | 95% |
| | | Mapping time (min) | | | | | | | | | | | 18 |
| | | Processing time (min) | | | | | | | | | | | 18 |
| | | Global error (cm) | | | | | | | | | | | 64 |

## Four flights mapped with Pix4Dmapper

Photogrammetric map generated from four flights merged manually using Pix4Dmapper shown below (overhead view) with singulated and labeled fiducials.

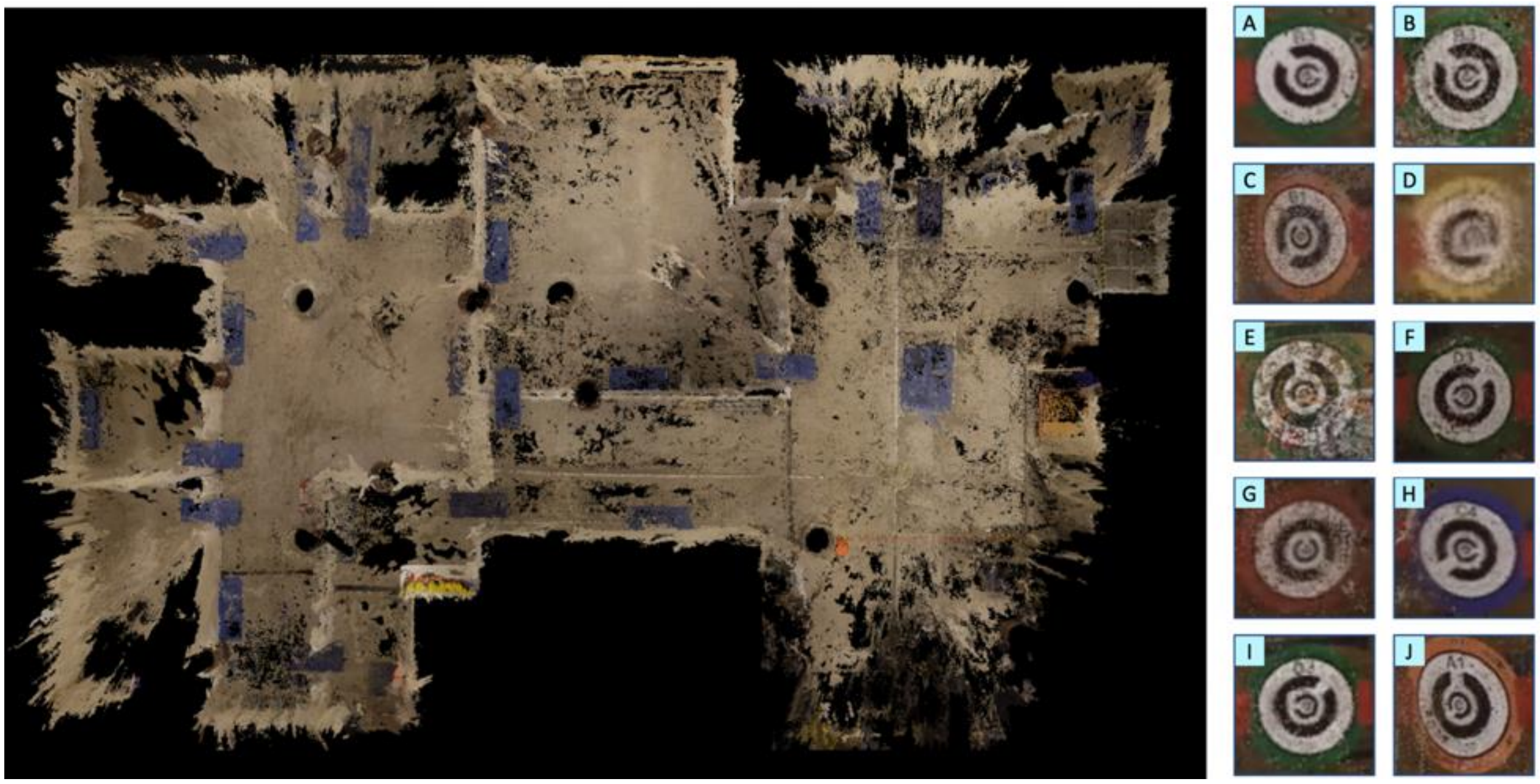

Performance data is only provided for the combined first, second, third, and four flight in the table below.

| sUAS | Flight | Metrics | Fiducials | | | | | | | | | | Map |
|---|---|---|---|---|---|---|---|---|---|---|---|---|---|
| | | | A | B | C | D | E | F | G | H | I | J | |
| Flyability Elios 2 GOV | 1+2+3+4 | Shape accuracy | C | C | C | C | I | I | C | C | | I | 67% |
| | | Visual acuity (mm) | 8 | 8 | 8 | 20 | 8 | 8 | 8 | 8 | | 3 | 8.8 (+/- 4.5) |
| | | Mapping time (min) | | | | | | | | | | | 32 |
| | | Processing time (min) | | | | | | | | | | | 720 |
| | | Global error (cm) | | | | | | | | | | | 5 |

# Vertical Mapping

## Environment characterization

| **NERVE Stairwell** | | | | |
|---|---|---|---|---|
| **Dimensions (L x W x H) and volume** | 6 x 3 x 10.2 m = 183.6 $m^3$ (19.7 x 9.8 x 33.5 ft = 6467.51 $ft^3$) | **Metrics** | **Fiducials** | |
| **Number of rooms** | 1 | | **A** | **B** |
| **Number of floors** | 3 | **Minimum traversal distance (m)** | 22 | 37 |
| **Lighting** | Dark (<1 LUX) | **Minimum orientation changes** | 6 | 10 |
| **Walls** | Concrete, Drywall | **Difficulty rating** | L | H |
| **Floor** | Concrete | | | |
| **Type** | Indoor | | | |
| **Image** | 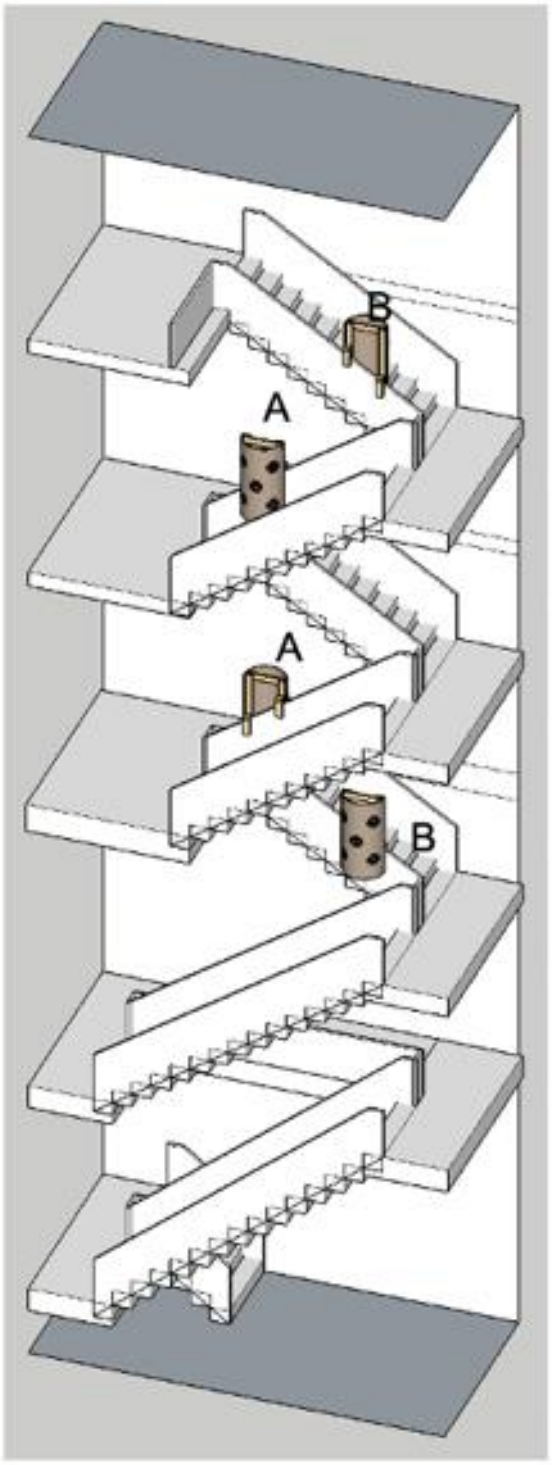  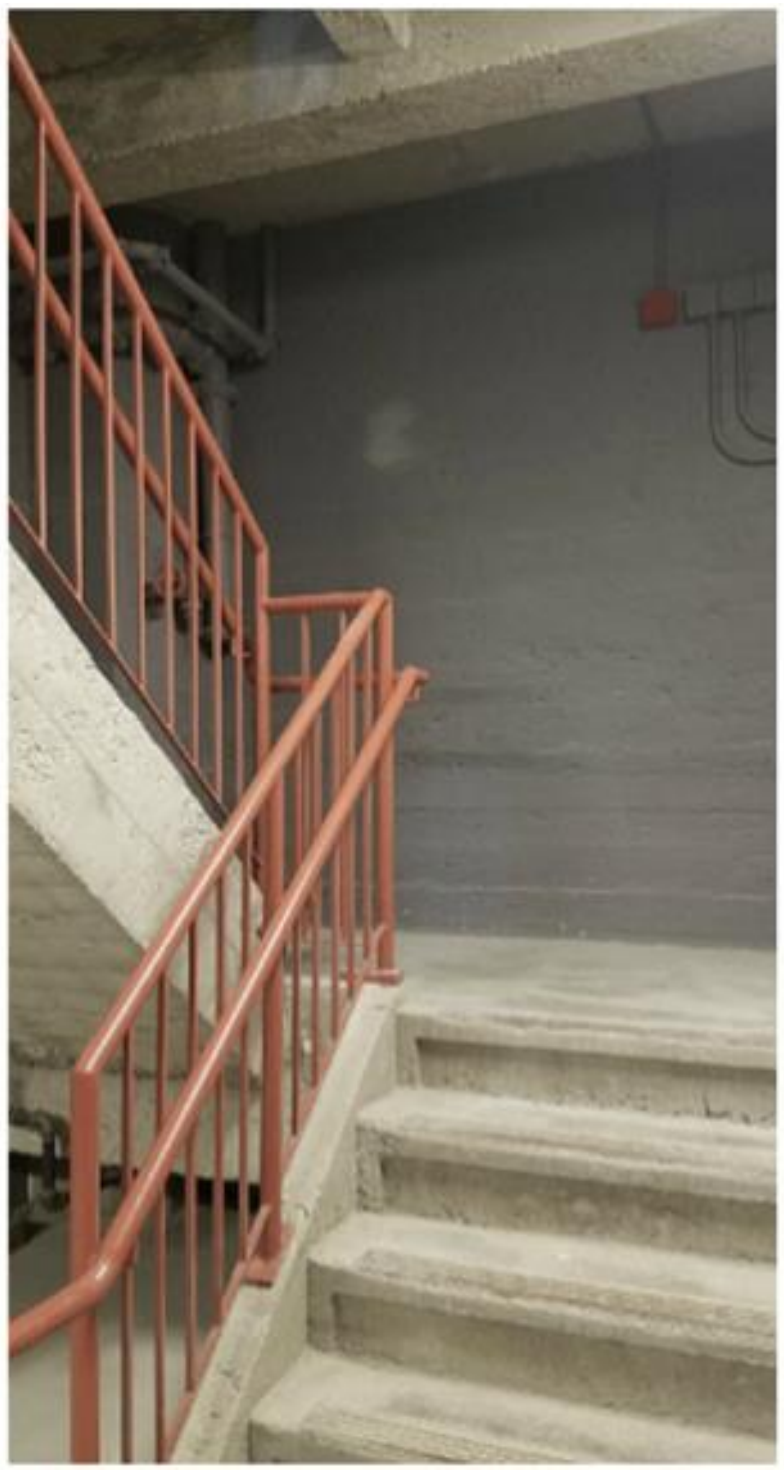 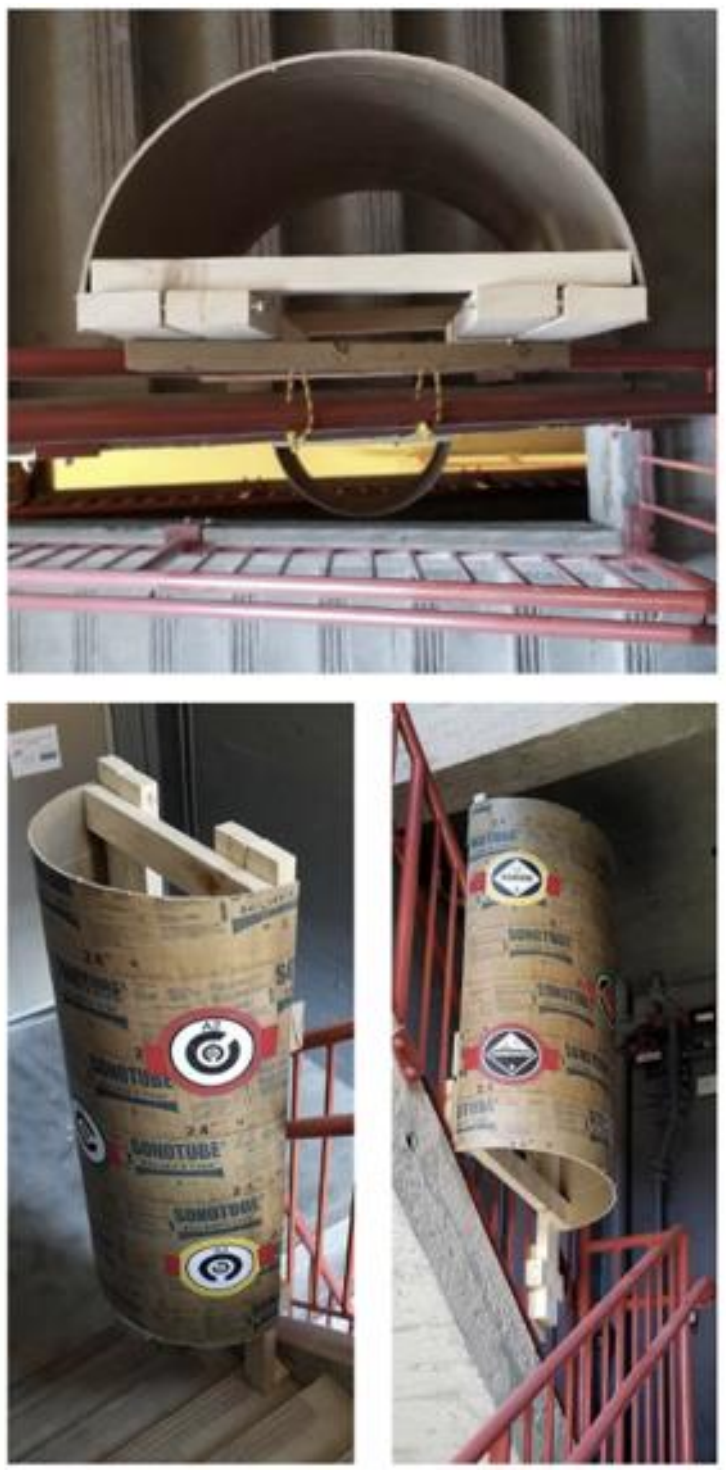 | | | |

## Performance data

### One flight mapped with Flyability Inspector 3.0

Point cloud map generated from one flights shown below (side view) with singulated and labeled fiducials.

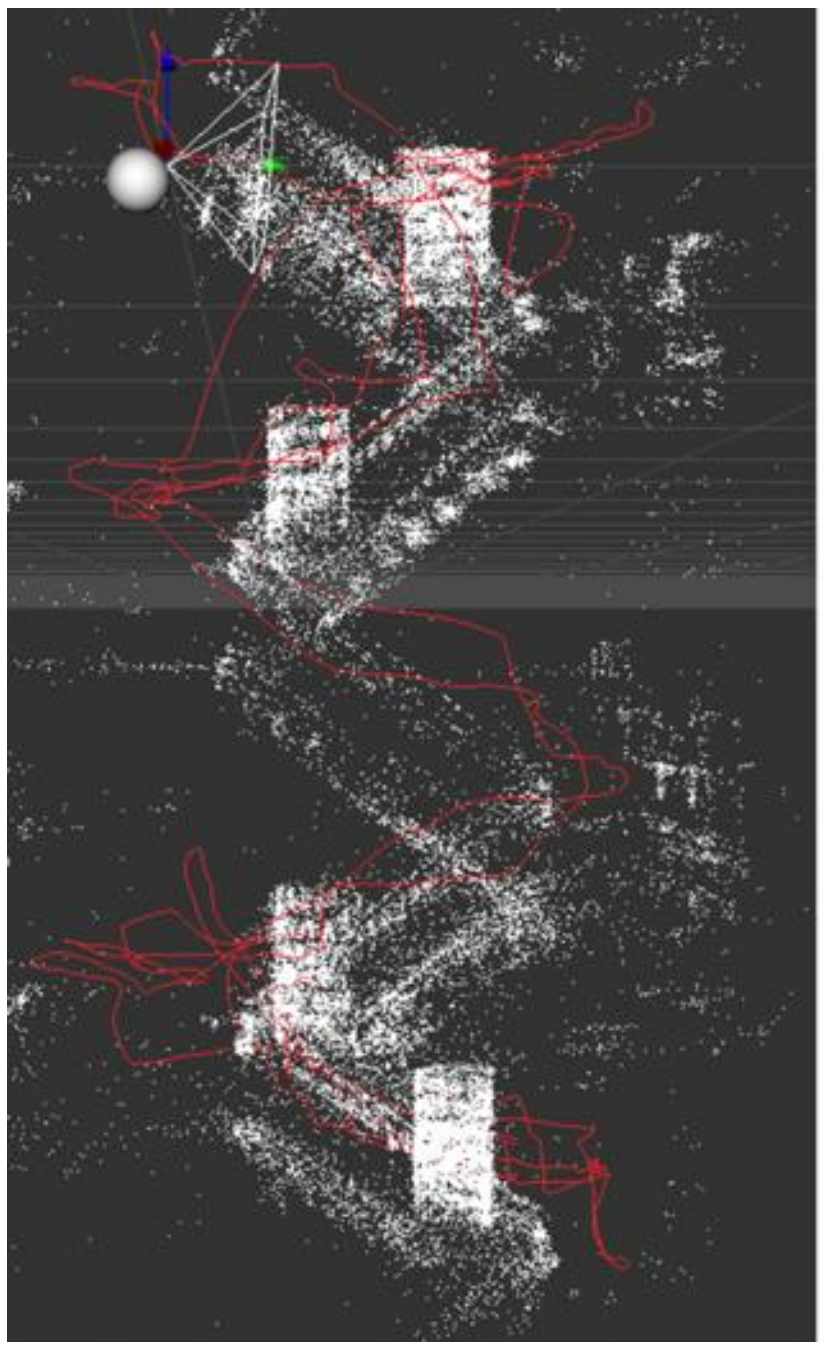

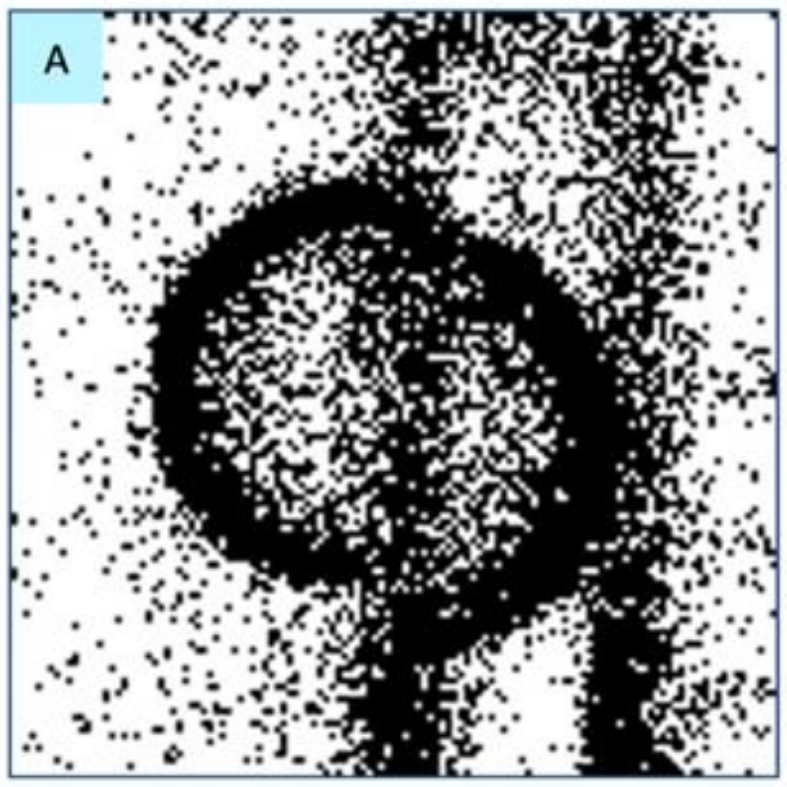

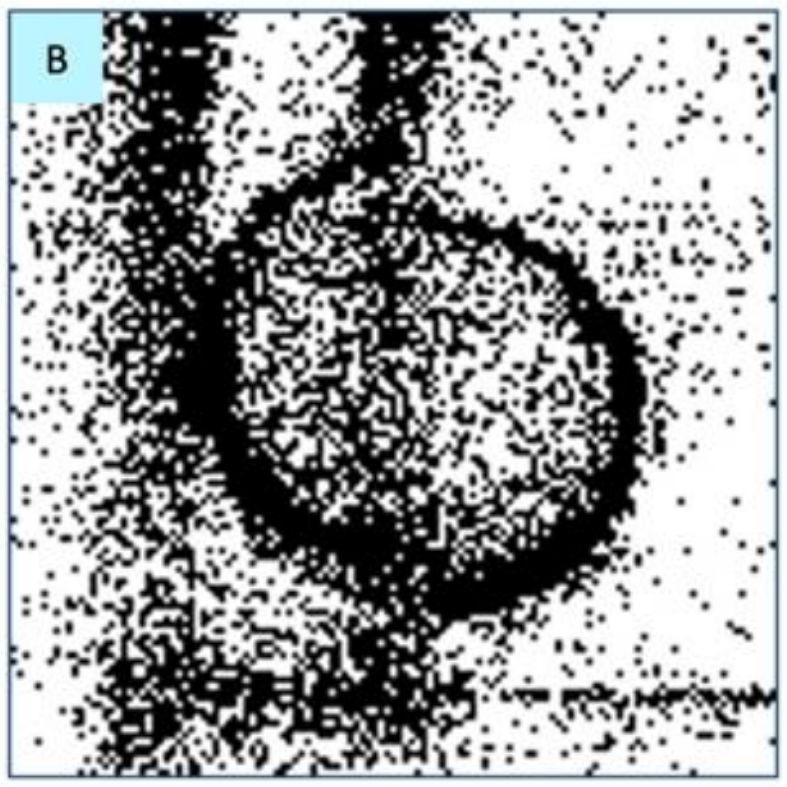

| sUAS | Flight | Metrics | Fiducials | | Map |
|---|---|---|---|---|---|
| | | | **A** | **B** | |
| Flyability Elios 2 GOV | 1 | Shape accuracy | C | C | 100% |
| | | Coverage | ✓ | ✓ | 100% |
| | | Mapping time (min) | | | 6 |
| | | Processing time (min) | | | 8 |
| | | Global error (cm) | | | N/A* |

*Global error can only be evaluated with a minimum of 3 fiducials, so it is not measured here.

## 3D Mapping

### Environment characterization

| MUTC Fire Trainer | | | | | | | | | | | | |
|---|---|---|---|---|---|---|---|---|---|---|---|---|
| **Dimensions (L x W x H) and volume** | 17.1 x 24.4 x 9.9 m = 4130.7 m$^3$ (56.1 x 80.1 x 32.5 ft = 146042.3 ft$^3$) | | **Fiducials** | | | | | | | | | |
| **Number of rooms** | 16 | **Metrics** | A | B | C | D | E | F | G | H | I | J |
| **Number of floors** | 4 | **Minimum traversal distance (m)** | Due to the variety of paths available to reach each fiducial half, difficulty metrics are not provided | | | | | | | | | |
| **Lighting** | 4-100K | **Minimum orientation changes** | | | | | | | | | | |
| **Walls** | Corrugated steel | **Difficulty rating** | | | | | | | | | | |
| **Floor** | Corrugated steel, grass | | | | | | | | | | | |
| **Type** | Indoor and Outdoor | | | | | | | | | | | |
| **Image** | 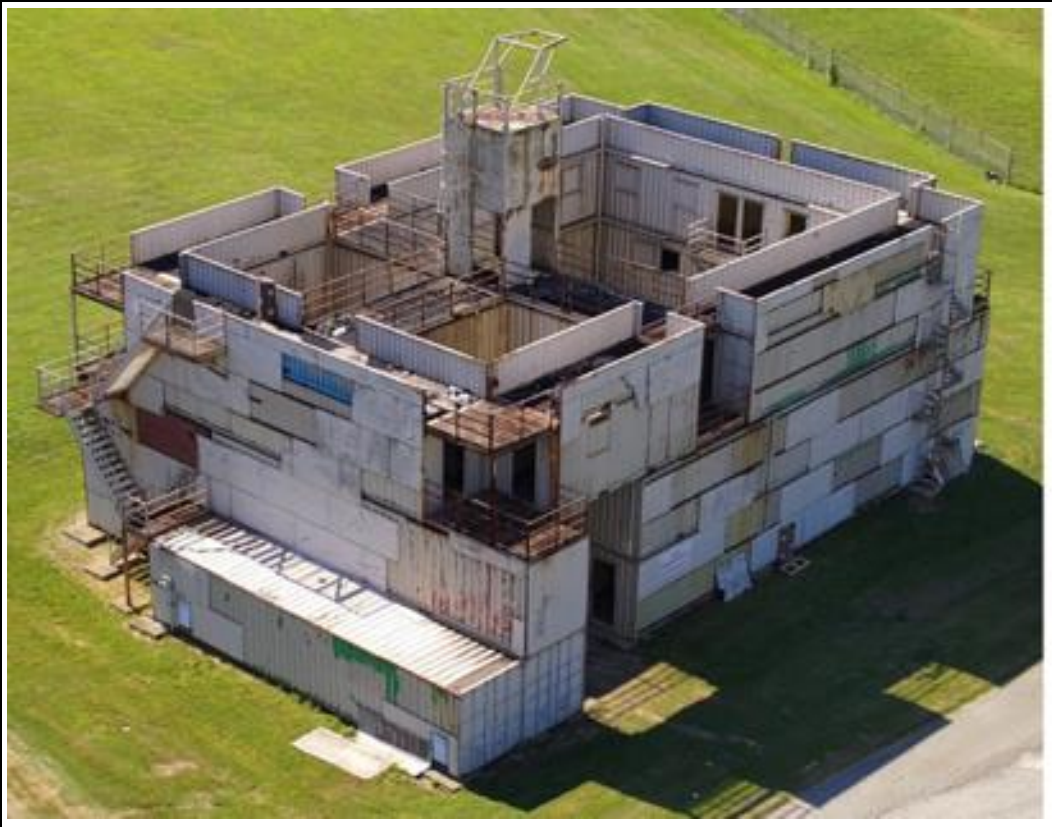 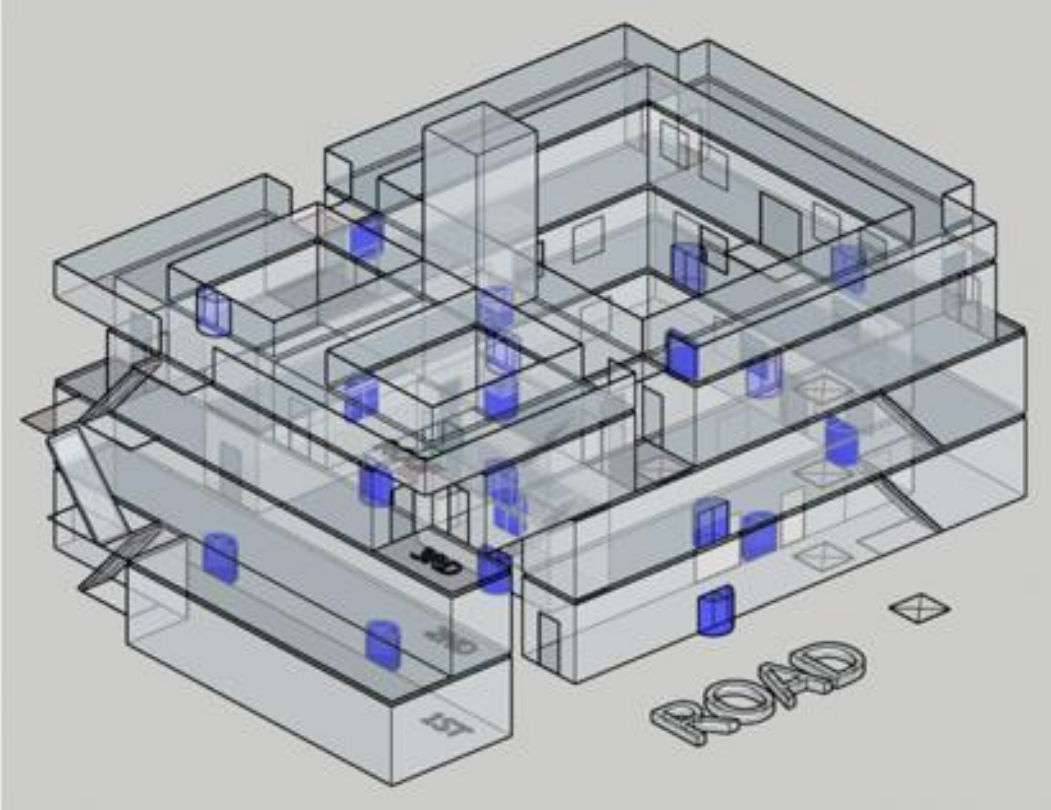  | | | | | | | | | | | |

| MUTC Hotel Trainer | | | | | | | | | | | | |
|---|---|---|---|---|---|---|---|---|---|---|---|---|
| **Dimensions (L x W x H) and volume** | 9.8 x 24.4 x 15.7 m = 3754.2 $m^3$<br>(32.2 x 80.1 x 51.5 ft = 132829.8 $ft^3$) | | **Fiducials** | | | | | | | | | |
| **Number of rooms** | 21 | **Metrics** | **A** | **B** | **C** | **D** | **E** | **F** | **G** | **H** | **I** | **J** |
| **Number of floors** | 5 | **Minimum traversal distance (m)** | Due to the variety of paths available to reach each fiducial half, difficulty metrics are not provided | | | | | | | | | |
| **Lighting** | 0-100K | **Minimum orientation changes** | | | | | | | | | | |
| **Walls** | Corrugated steel | **Difficulty rating** | | | | | | | | | | |
| **Floor** | Corrugated steel, wood | | | | | | | | | | | |
| **Type** | Indoor and “Outdoor” (Open roof) | | | | | | | | | | | |
| **Image** | 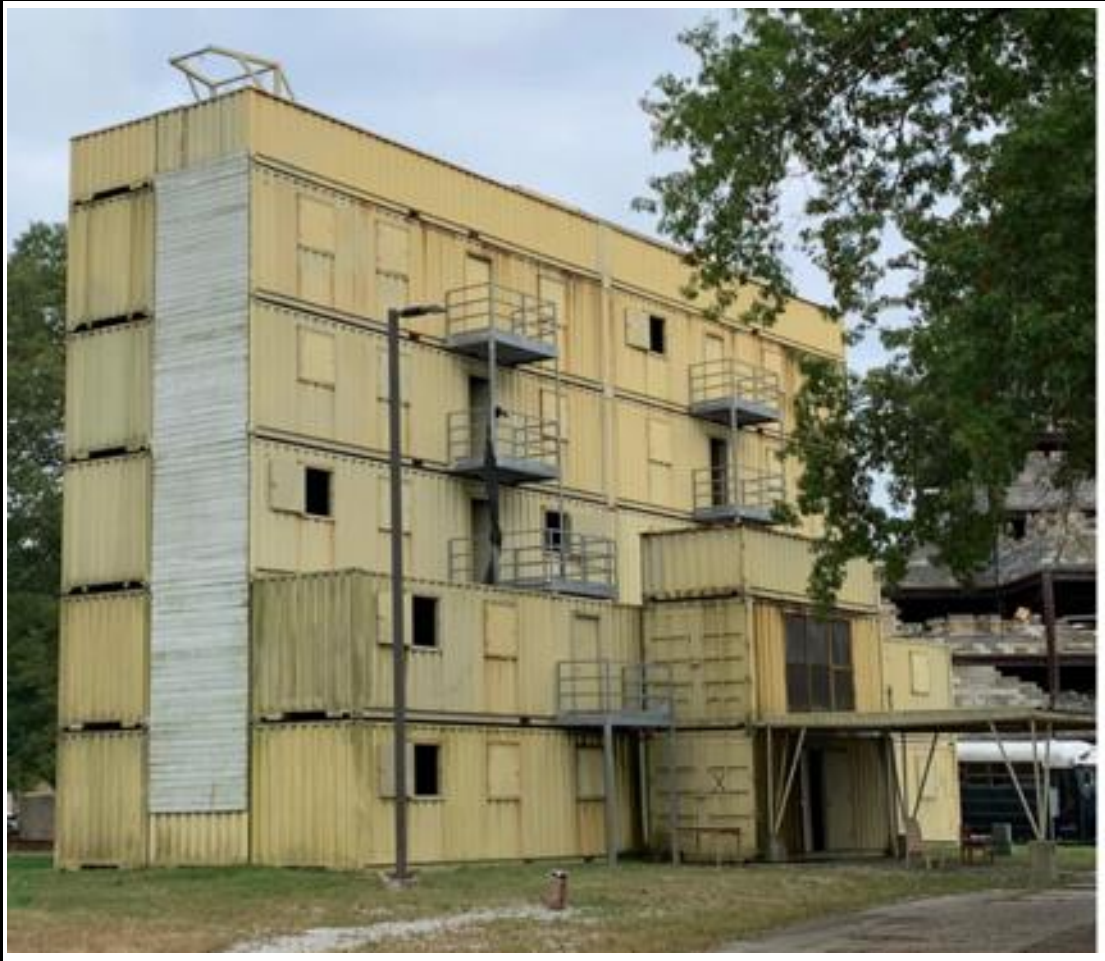 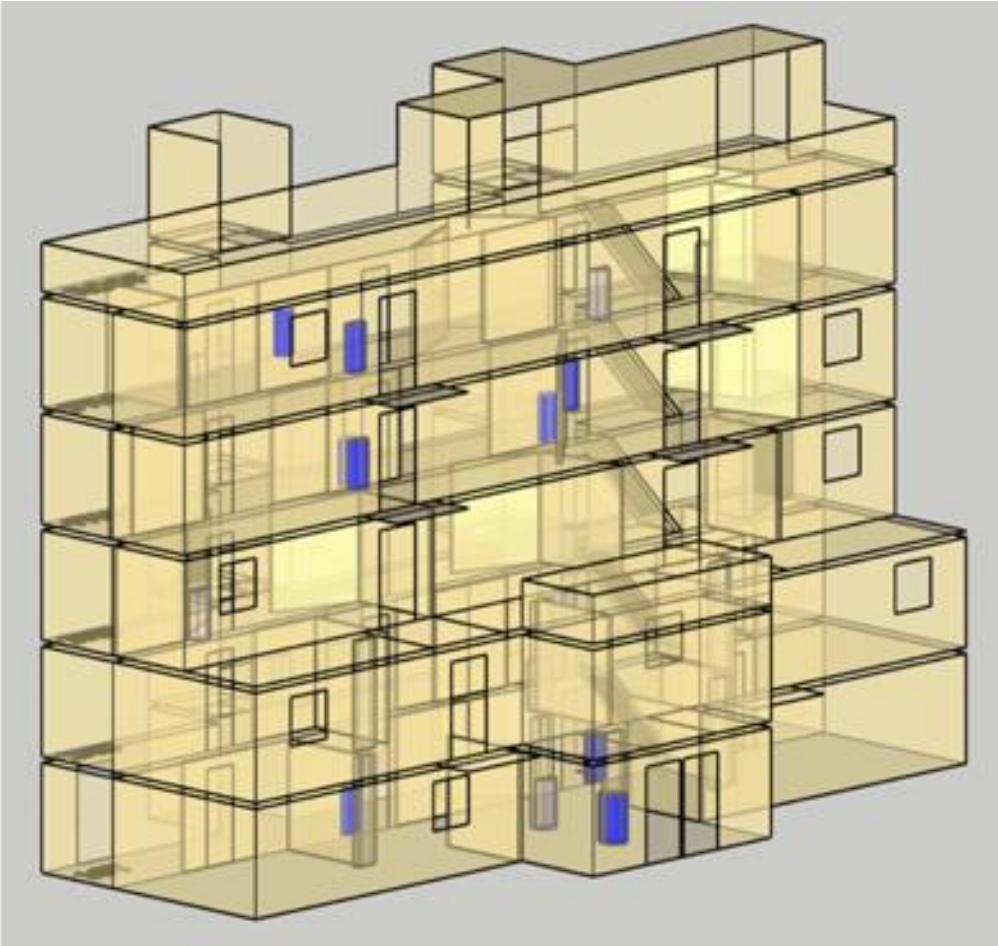 | | | | | | | | | | | |

## Performance data

Note: The results of all photogrammetric mapping was very poor, so only two examples are shown to illustrate this.

**MUTC Fire Trainer: One flight mapped with Flyability Inspector 3.0**

Point cloud map generated from one flight shown below with singulated and labeled fiducials.

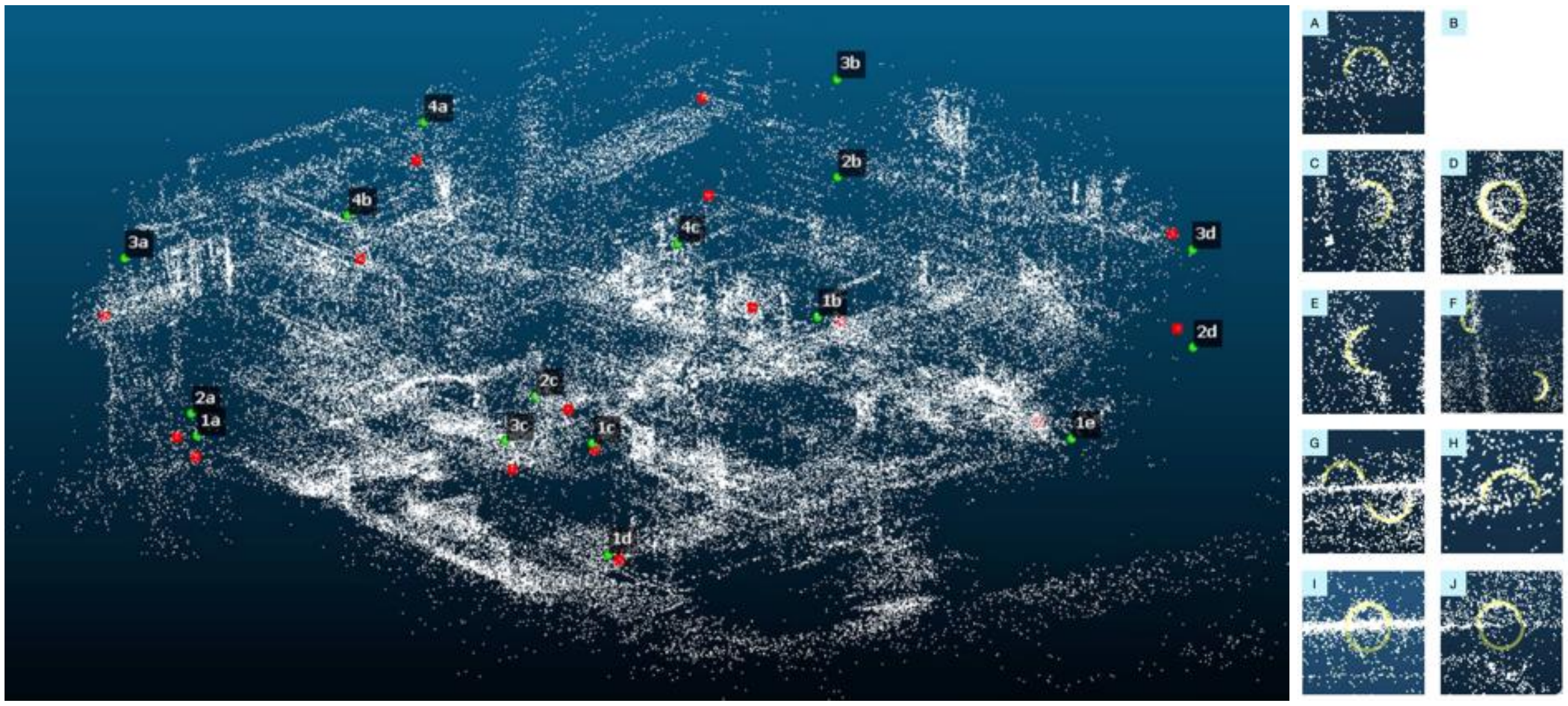

| sUAS | Flight | Metrics | Fiducials | | | | | | | | | | Map |
|---|---|---|---|---|---|---|---|---|---|---|---|---|---|
| | | | A | B | C | D | E | F | G | H | I | J | |
| Flyability Elios 2 GOV | 1 | Shape accuracy | I | | I | C | I | S | S | I | C | C | 30% |
| | | Coverage | // | X | // | ✓ | // | ✓ | ✓ | // | ✓ | ✓ | 100%* |
| | | Mapping time (min) | | | | | | | | | | | 9 |
| | | Processing time (min) | | | | | | | | | | | 9 |
| | | Global error (cm) | | | | | | | | | | | 80 (+/- 21)* |

*Note: Due to inaccuracies in the ground truth measurements of the fiducial positions, coverage and global error metrics were evaluated based on 16 corners in the environment (e.g., the vertices of walls, floors, and ceilings).

**MUTC Fire Trainer: One flight mapped with Pix4Dmapper**

Photogrammetric map generated from one flight using Pix4Dmapper shown below (isometric view) with singulated and labeled fiducials.

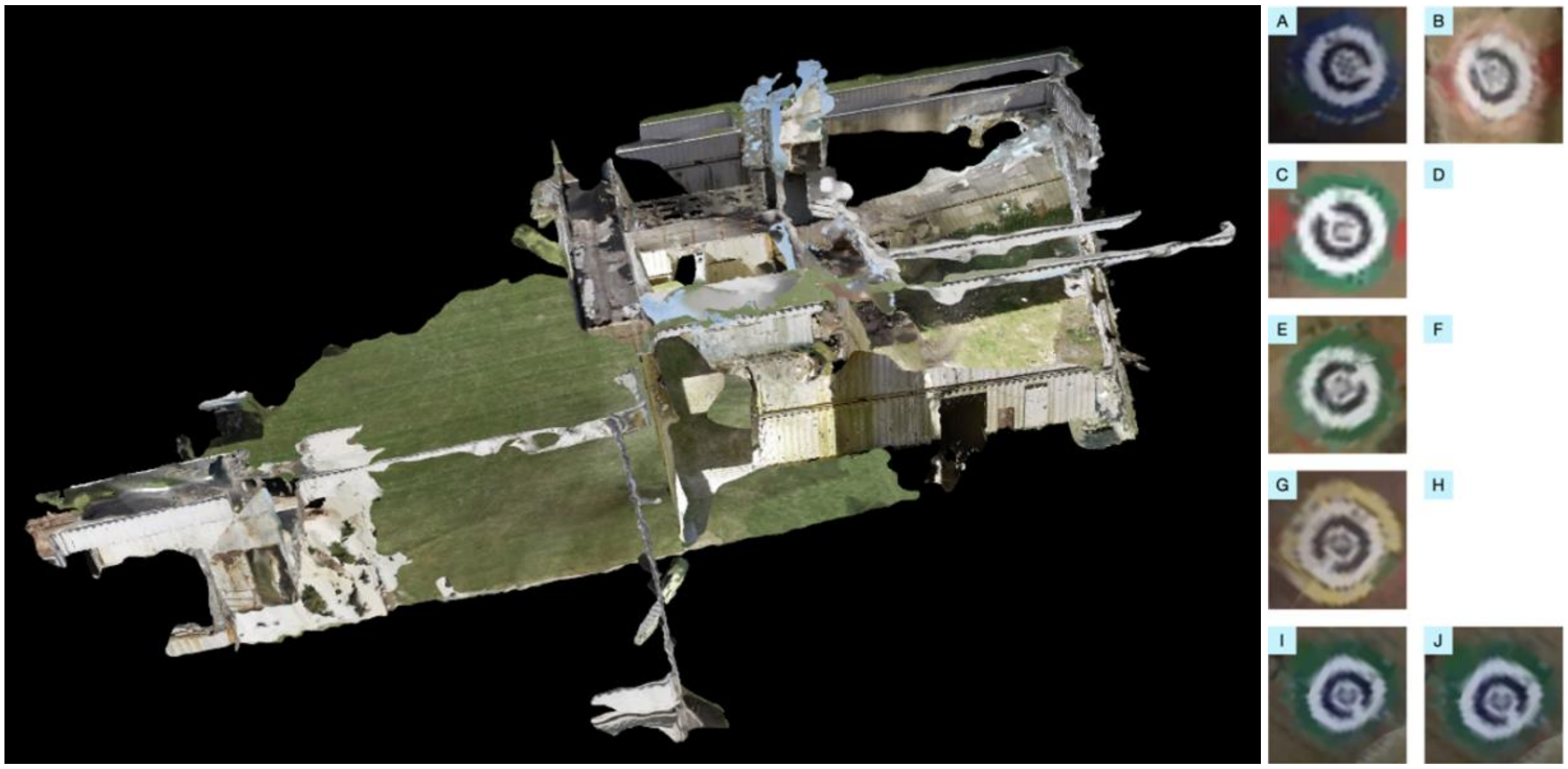

Only visual acuity metrics are provided due to the poor construction of the photogrammetric map.

| | | | Fiducials | | | | | | | | | | |
|---|---|---|---|---|---|---|---|---|---|---|---|---|---|
| **sUAS** | **Flight** | **Metrics** | **A** | **B** | **C** | **D** | **E** | **F** | **G** | **H** | **I** | **J** | **Map** |
| Flyability Elios 2 GOV | 1 | Visual acuity (mm) | 20 | 20 | 8 | | 8 | | 8 | | 8 | 8 | 11.4 (+/- 5.9) |
| | | Mapping time (min) | | | | | | | | | | | 9 |
| | | Processing time (min) | | | | | | | | | | | 480+ |

**MUTC Fire Trainer: Two flights mapped with Flyability Inspector 3.0**

Point cloud map generated from two flights merged automatically using CloudCompare shown below with singulated and labeled fiducials.

Flight 1:

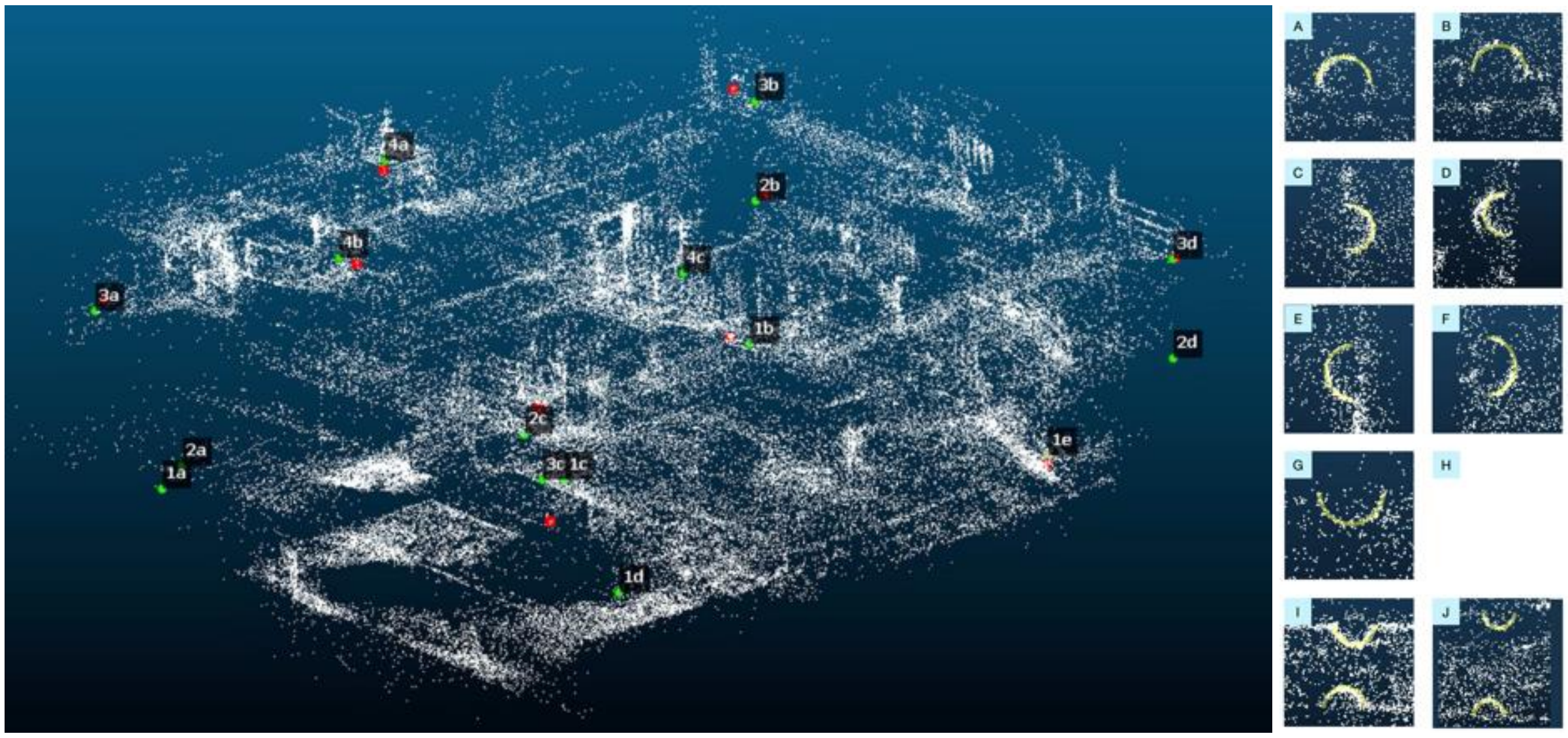

Flight 1+2:

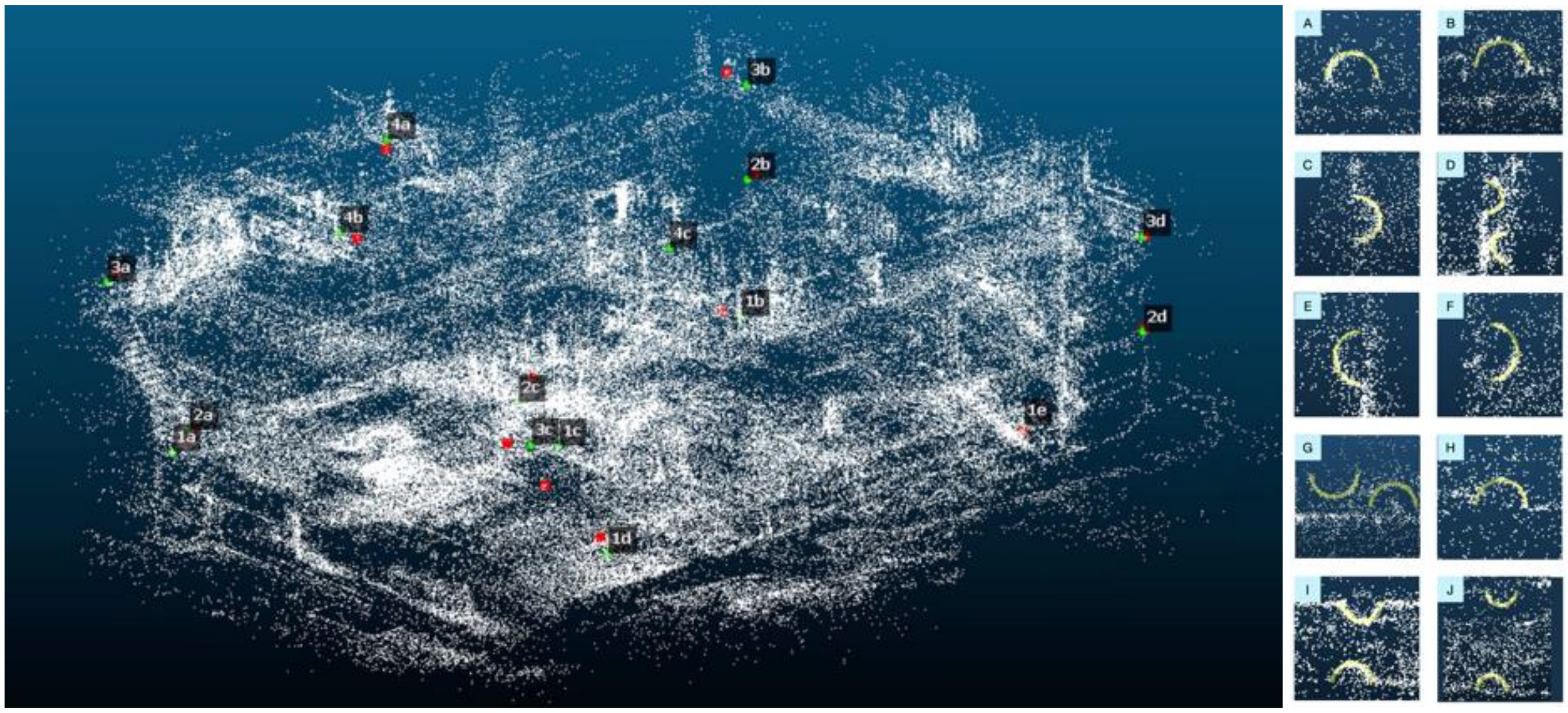

| sUAS | Flight | Metrics | Fiducials | | | | | | | | | | Map |
|---|---|---|---|---|---|---|---|---|---|---|---|---|---|
| | | | A | B | C | D | E | F | G | H | I | J | |
| Flyability Elios 2 GOV | 1 | Shape accuracy | I | I | I | I | I | I | I | | S | S | 0% |
| | | Coverage | // | // | // | // | // | // | // | X | ✓ | ✓ | 63%* |
| | | Mapping time (min) | | | | | | | | | | | 9 |
| | | Processing time (min) | | | | | | | | | | | 9 |
| | | Global error (cm) | | | | | | | | | | | 49 (+/- 10)* |
| Flyability Elios 2 GOV | 1+2 | Shape accuracy | I | I | I | S | I | I | S | I | S | S | 0% |
| | | Coverage | // | // | // | ✓ | // | // | ✓ | // | ✓ | ✓ | 94%* |
| | | Mapping time (min) | | | | | | | | | | | 18 |
| | | Processing time (min) | | | | | | | | | | | 18 |
| | | Global error (cm) | | | | | | | | | | | 42 (+/- 10)* |

*Note: Due to inaccuracies in the ground truth measurements of the fiducial positions, coverage and global error metrics were evaluated based on 16 corners in the environment (e.g., the vertices of walls, floors, and ceilings).

**MUTC Hotel Trainer: One flight mapped with Flyability Inspector 3.0**

Point cloud map generated from one flights shown below with singulated and labeled fiducials.

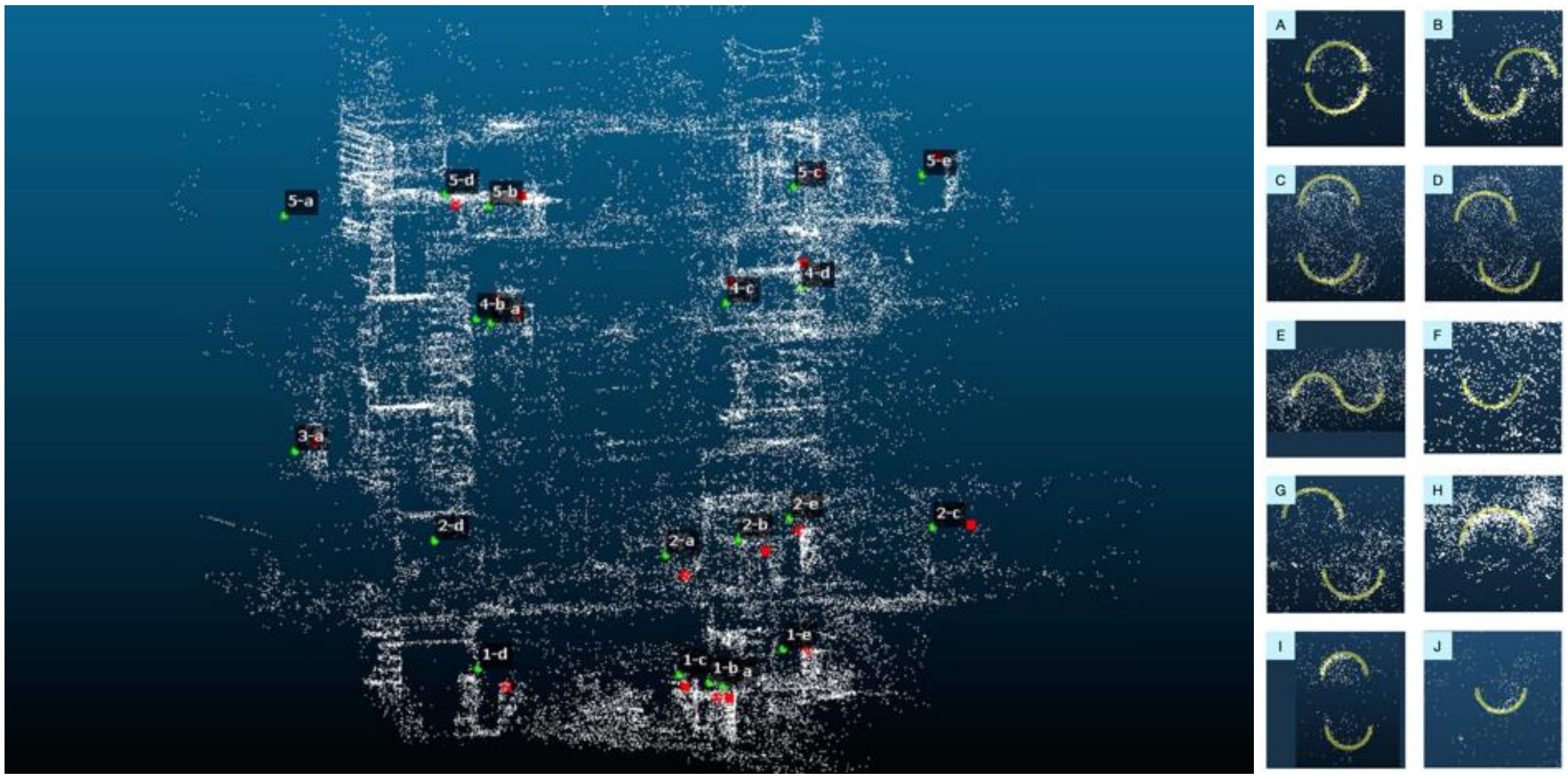

| sUAS | Flight | Metrics | Fiducials | | | | | | | | | | Map |
|---|---|---|---|---|---|---|---|---|---|---|---|---|---|
| | | | A | B | C | D | E | F | G | H | I | J | |
| Flyability Elios 2 GOV | 1 | Shape accuracy | C | S | S | S | S | I | S | I | S | I | 10% |
| | | Coverage | ✓ | ✓ | ✓ | ✓ | ✓ | // | ✓ | // | ✓ | // | 85% |
| | | Mapping time (min) | | | | | | | | | | | 9 |
| | | Processing time (min) | | | | | | | | | | | 9 |
| | | Global error (cm) | | | | | | | | | | | 38 (+/- 8) |

**MUTC Hotel Trainer: Two flights mapped with Flyability Inspector 3.0**

Point cloud map generated from two flights merged automatically using CloudCompare shown below with singulated and labeled fiducials.

Flight 1:

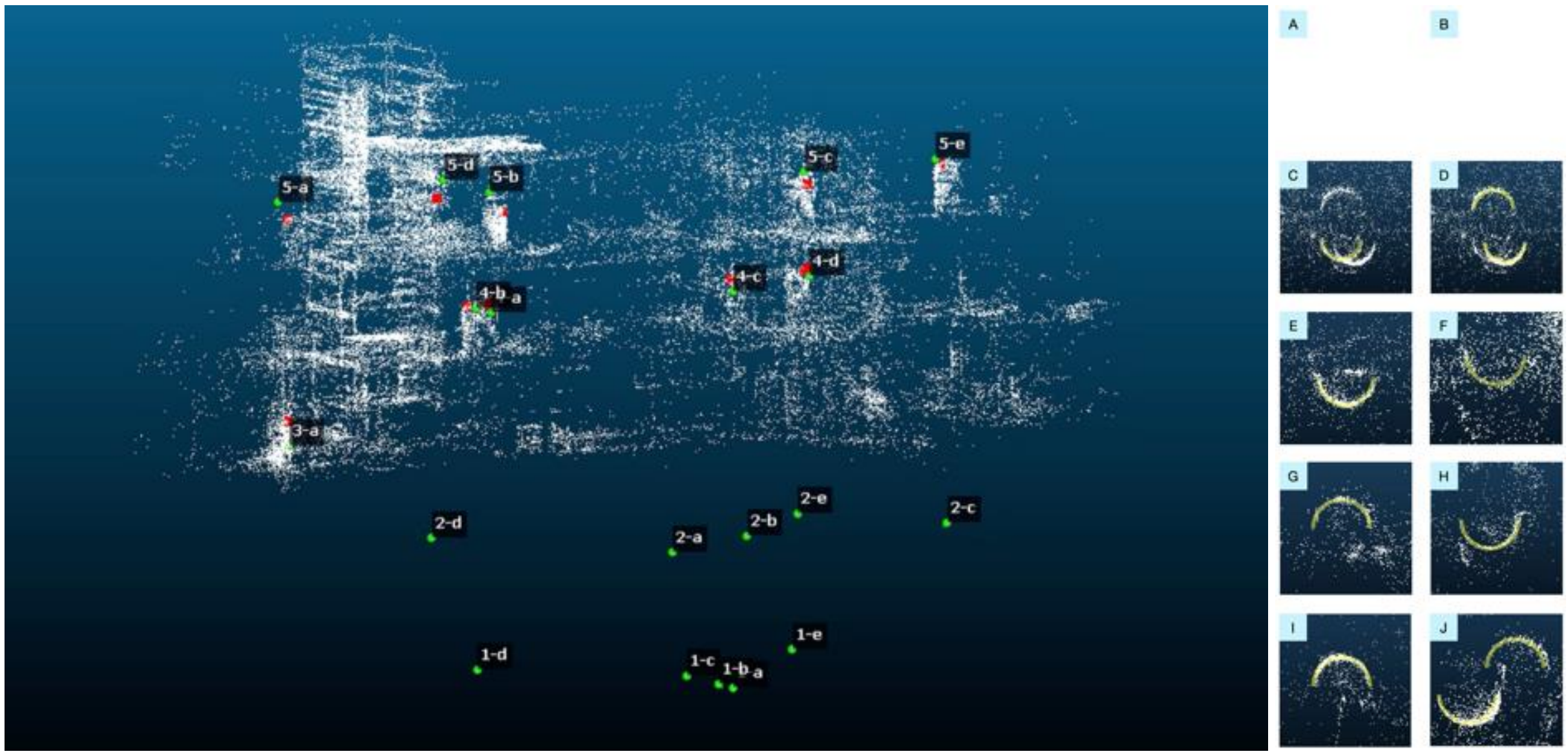

Flight 1+2:

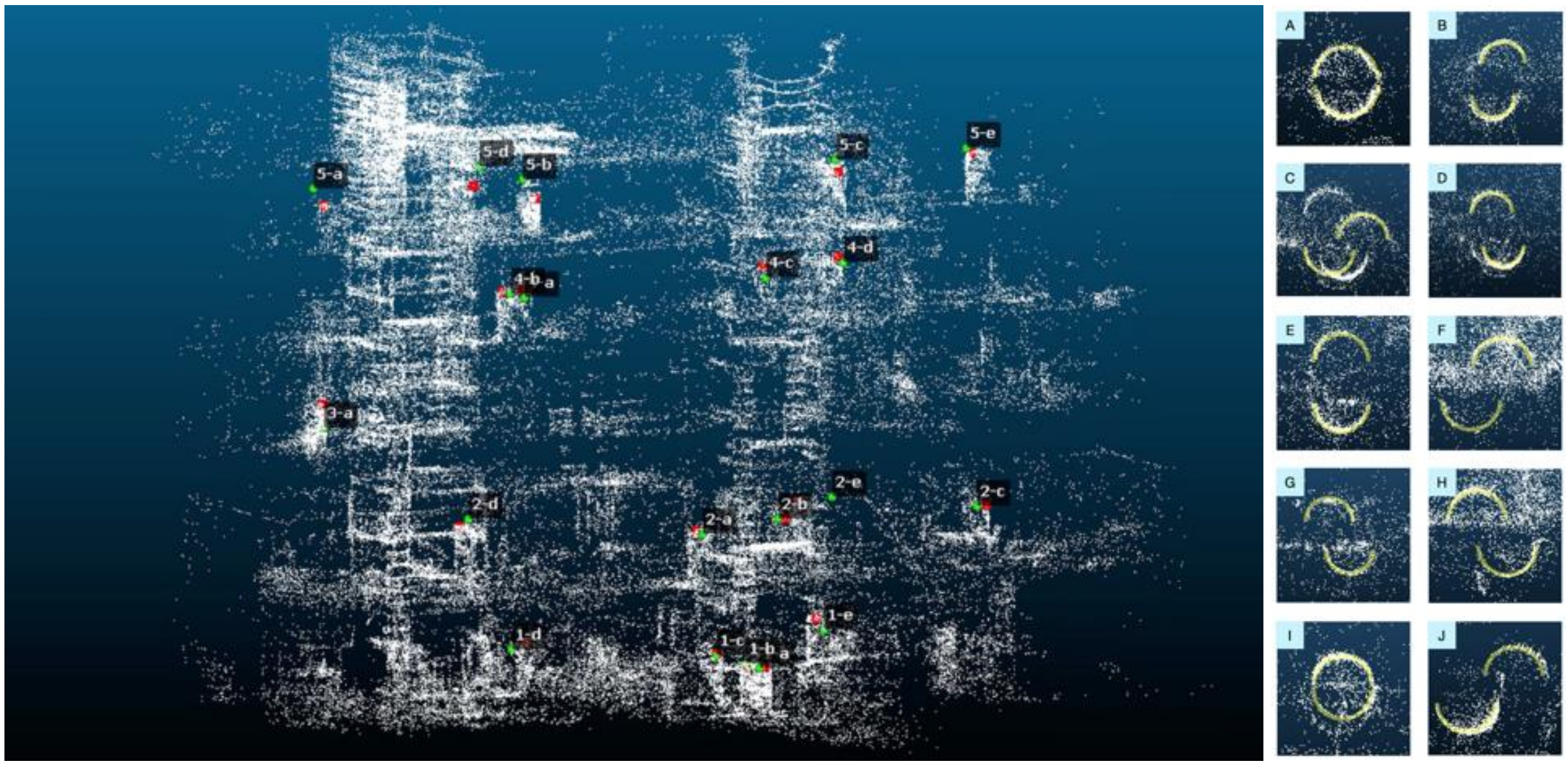

| sUAS | Flight | Metrics | Fiducials | | | | | | | | | | Map |
|---|---|---|---|---|---|---|---|---|---|---|---|---|---|
| | | | A | B | C | D | E | F | G | H | I | J | |
| Flyability Elios 2 GOV | 1 | Shape accuracy | | | I | S | I | I | I | I | I | S | 0% |
| | | Coverage | X | X | // | ✓ | // | // | // | // | // | ✓ | 50% |
| | | Mapping time (min) | | | | | | | | | | | 9 |
| | | Processing time (min) | | | | | | | | | | | 9 |
| | | Global error (cm) | | | | | | | | | | | 25 (+/- 8) |
| Flyability Elios 2 GOV | 1+2 | Shape accuracy | C | S | S | S | S | S | S | S | C | S | 20% |
| | | Coverage | ✓ | ✓ | ✓ | ✓ | ✓ | ✓ | ✓ | ✓ | ✓ | ✓ | 100% |
| | | Mapping time (min) | | | | | | | | | | | 18 |
| | | Processing time (min) | | | | | | | | | | | 18 |
| | | Global error (cm) | | | | | | | | | | | 28 (+/- 7) |

**MUTC Hotel Trainer: Two flight mapped with Pix4Dmapper**

Photogrammetric map generated from two flights merged manually using Pix4Dmapper shown below (isometric view) with singulated and labeled fiducials.

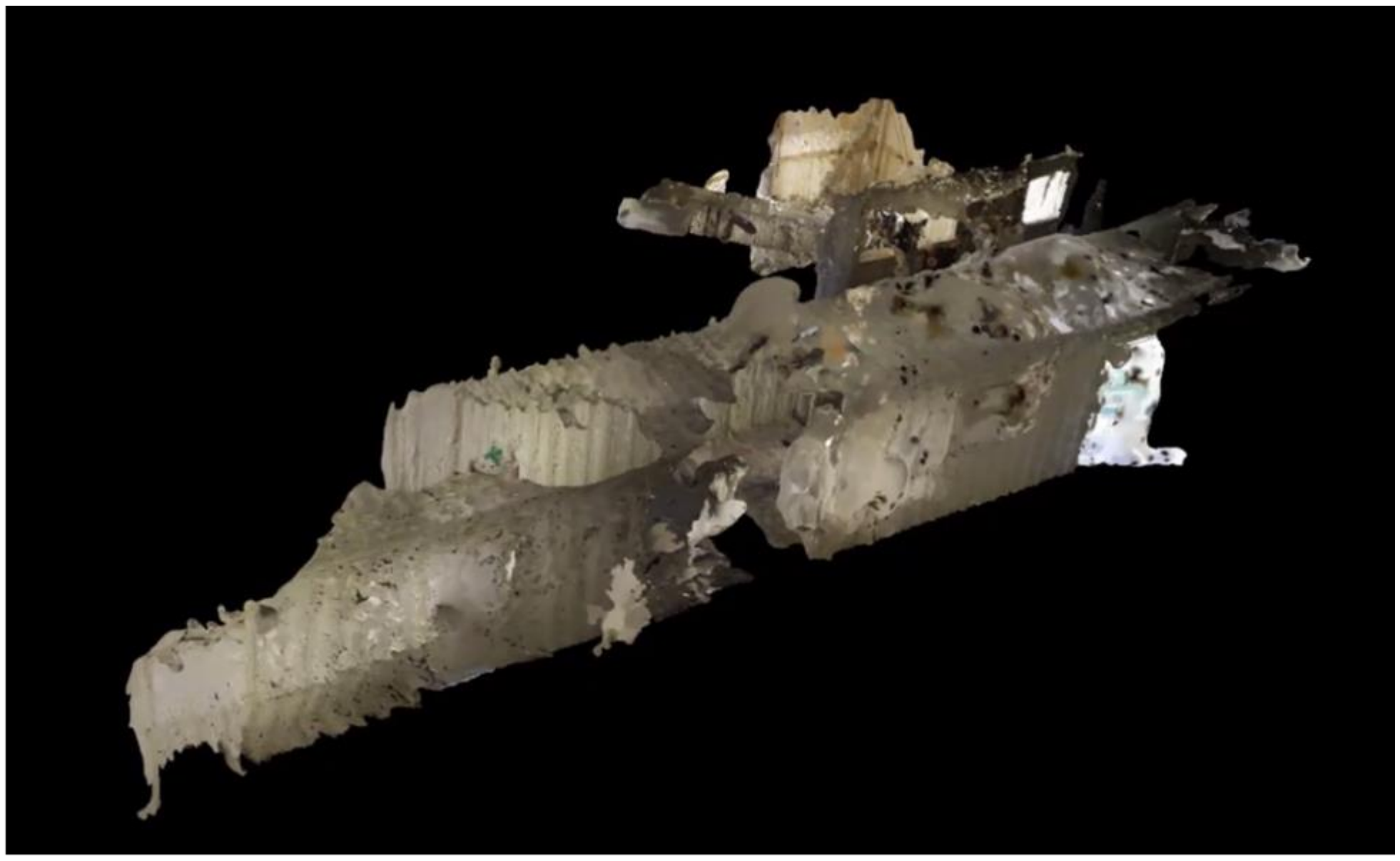

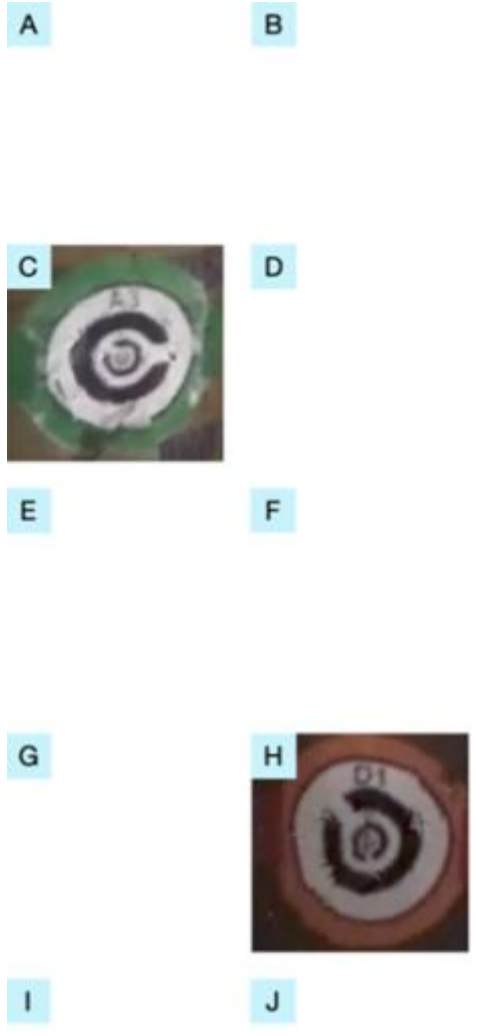

Only visual acuity metrics are provided due to the poor construction of the photogrammetric map.

| sUAS | Flight | Metrics | Fiducials | | | | | | | | | | Map |
|---|---|---|---|---|---|---|---|---|---|---|---|---|---|
| | | | A | B | C | D | E | F | G | H | I | J | |
| Flyability Elios 2 GOV | 1+2 | Visual acuity (mm) | | | 8 | | | | | 8 | | | 8 |
| | | Mapping time (min) | | | | | | | | | | | 18 |
| | | Processing time (min) | | | | | | | | | | | 480+ |

# Autonomy

## Non-Contextual Autonomy Ranking

Affiliated publications: [Hertel et al., 2022]

### Summary of Test Method

In this test, the user selects a set of mission-independent features (e.g., maximum range, number of smart behaviors) and measures those features for a set of sUAS in order to generate a series of non-contextual autonomy scores for comparison across sUAS. These data points can be gathered either by referencing vendor-provided specification sheets for each sUAS or measured and verified through experimentation using the test methods specified in this document (which is preferred for an ideal evaluation). The method used to derive the measure for each feature should be reported to delineate between assumed and demonstrated measures. The collected data is used to calculate several metrics including sUAS Autonomy Level ($N_{AL}$), sUAS Component Potential ($N_{CP}$), and ultimately the sUAS Combined NCAP Score.

The user can add or remove the mission-independent features they deem relevant. These are hardware-based features that their value can change from system to system, but does not change for a system when in different situations and missions. The following features are relevant examples for subterranean and constrained indoor operations, and can typically be derived from vendor-provided specification sheets and/or using the test methods specified in this document: flight time, charge time, stream resolution, field of view (FOV), maximum range, thermal camera resolution, weight, maximum flight speed, number of sensors, number of smart behaviors, and signal-to-noise ratio (SNR).

### Benchmarking Results

Evaluation and reasoning of autonomy levels $N_{AL}$ of each sUAS:

| sUAS | Perception | Modeling | Planning | Execution | $N_{AL}$ |
|---|---|---|---|---|---|
| Cleo Robotics Dronut X1P | 2 RGB camera, thermal camera, LiDAR | SLAM capabilities | Obstacle avoidance planning | Avoids obstacles, auto-land | 3 |
| Flyability Elios 2 GOV | RGB camera, IMU, thermal camera, 5 distance sensors | Models surroundings using distance sensors | None | None | 1 |
| Lumenier Nighthawk V3 | RGB camera, thermal camera | None | None | None | 0 |
| Parrot ANAFI USA GOV | 2 RGB cameras, IMU thermal camera, GPS | Uses GPS and IMU for positioning in maps, visual modeling of targets | Geofencing and autonomous navigation | Target tracking, return to home | 3 |
| Skydio X2D | 6 RGB cameras, thermal camera, GPS | Maps surrounding areas from camera imaging | Plans best path in environment | Obstacle avoidance, path execution | 3 |
| Teal Golden Eagle | RGB camera, thermal camera, GPS, IMU | None | None | None | 0 |
| Vantage Robotics Vesper | RGB camera, thermal camera, GPS | None | None | None | 0 |

Selected sUAS platform features used for $N_{CP}$ measurements:

| sUAS | Flight Time (min) | Charge Time (min) | Stream Res. | FOV | Max. Range (m) | Thermal Camera Res. | Weight (g) | Max. Flight Speed (m/s) | # of Sensors | # of Smart Behaviors |
|---|---|---|---|---|---|---|---|---|---|---|
| Cleo Robotics Dronut X1P | 15 | 50 | FHD | 100° | 2000 | N/A | 370 | 3 | 3 | 2 |
| Flyability Elios 2 GOV | 10 | 90 | FHD | 114° | 500 | 160x120 | 1450 | 6.5 | 10 | 7 |
| Lumenier Nighthawk V3 | 22 | N/A | FHD | N/A | 2000 | 620x512 | 1200 | 3 | 4 | 5 |
| Parrot ANAFI USA GOV | 32 | 120 | HD | 84° | 4000 | 320x256 | 500 | 14.7 | 10 | 5 |
| Skydio X2D | 23 | 120 | 4k | 200° | 3500 | 320x256 | 775 | 16 | 11 | 10 |
| Teal Golden Eagle | 30 | 45 | 4k | 90° | 3000 | 320x256 | 1044 | 22 | 7 | 3 |
| Vantage Robotics Vesper | 50 | N/A | FHD | 63° | 4000 | 320x240 | 697 | 20 | 6 | 2 |

$N_{CP}$ score of each sUAS using different combination methods and uniform weights (best in class = P rank 4 or higher):

| sUAS | $S_{max}$ | $S_{map}$ | $S_{zsc}$ | $S_{sum}$ | P | $N_{AL}$ |
|---|---|---|---|---|---|---|
| Cleo Robotics Dronut X1P | 0.18 (6) | 0.33 (5) | -0.98 (7) | 0.05 (7) | 2.48 (6) | 3 |
| Flyability Elios 2 GOV | 0.17 (7) | 0.30 (7) | 0.22 (2) | 0.05 (6) | 2.29 (7) | 1 |
| Lumenier Nighthawk V3 | 0.20 (5) | 0.30 (6) | -0.09 (5) | 0.06 (5) | 2.60 (5) | 0 |
| Parrot ANAFI USA GOV | 0.35 (4) | 0.50 (4) | 0.05 (4) | 0.09 (4) | 3.48 (4) | 3 |
| Skydio X2D | 0.53 (1) | 0.72 (1) | 0.93 (1) | 0.14 (1) | 4.63 (1) | 3 |
| Teal Golden Eagle | 0.39 (2) | 0.57 (2) | 0.05 (3) | 0.10 (3) | 3.81 (2) | 0 |
| Vantage Robotics Vesper | 0.38 (3) | 0.52 (3) | -0.27 (6) | 0.10 (2) | 3.56 (3) | 0 |
| **Best in class** | **Non-Contextual Autonomy Ranking** | | | | | |
| | Parrot ANAFI USA GOV<br>Skydio X2D<br>Teal Golden Eagle<br>Vantage Robotics Vesper | | | | | |

sUAS platform autonomy measure represented in the NCAP coordinate < $N_{AL}$ , $N_{CP}$ > with uniform weight values:

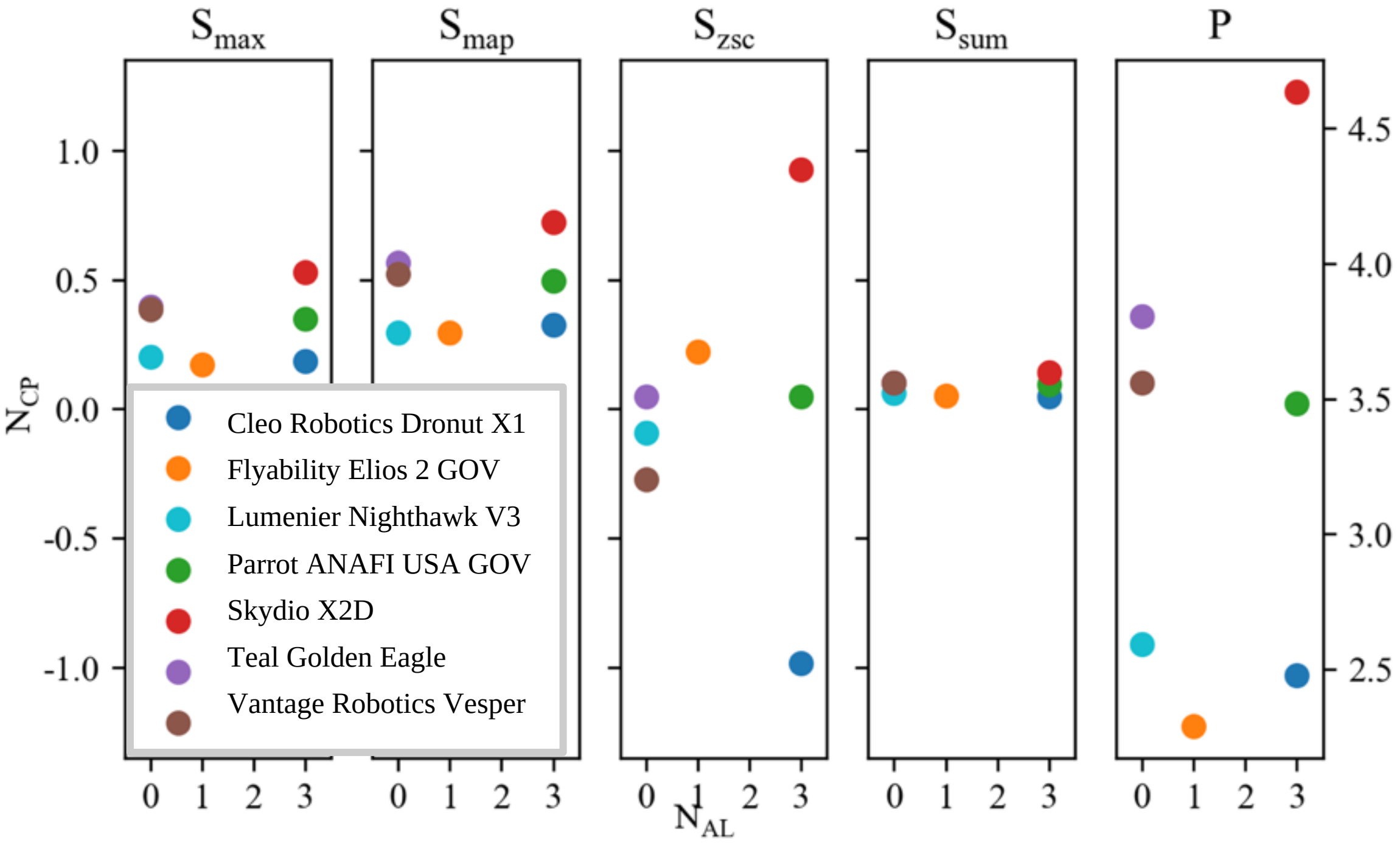

## Contextual Autonomy Ranking

Affiliated publications: [Donald et al., 2023]

### Summary of Test Method

In this test, the user selects a sub-task to be evaluated in a specific environment. Several examples have been described in this document including Runtime Endurance in Enclosed Spaces, Takeoff and Land/Perch, Navigation Through Apertures, Navigation Through Corridors, and the Room Clearing test. It should be noted that the proposed framework is not limited to these sets of tests and can be used for any sUAS mission. For each experiment, data should be collected according to three axes of Environmental Complexity (EC), Mission Complexity (MC), and Human Independence (HI). This is similar to the Autonomy Levels for Unmanned Systems (ALFUS) framework [Huang et al., 2007; Durst and Gray, 2014]. These three axes allow the user to categorize various factors of a mission that can affect a system's autonomy efficiently. The Environmental Complexity (EC) axis accounts for the differences in terrains and environments of a mission. In this axis, larger values correspond to more complex environments while smaller values correspond to simpler environments. The Mission Complexity (MC) axis accounts for the different levels of difficulty in movements, actions, or decisions required to complete a mission successfully. In this axis, larger values represent a mission requiring more complex decisions for completion while the smaller values represent simpler missions. The Human Independence (HI) axis accounts for the level of independence the sUAS offers to the operator for the successful completion of a mission. This axis also accounts for the types of actions which are able to be completed by the sUAS, and the complexities of those actions. Higher values on this axis indicate sUAS that can complete more complex movements while requiring more independence from the operator. The more difficult the tasks which the sUAS is able to perform results in a larger value along the HI axis. In addition, the larger the portion of the mission which is able to be completed by the sUAS without the operator, will also result in a larger value along the HI axis.

The user can add or remove the mission-specific features to each axis as they deem relevant. The following features are relevant examples for subterranean and constrained indoor operations: aperture/hallway cross-sectional area, ambient light level, verticality of the hallway, number of crashes, number of rollovers, completion percentage, static roll angle, static pitch angle, static vertical obstruction, static horizontal obstruction, coverage percentage, C's detected, duration, and obstructions. Note that metrics used in the evaluation of sub-tasks, as described in other sections of this document, can be used as features in this test for comparing different sUAS autonomy levels.

Scores for each sUAS, in each sub-task, alongside an example predictive score based upon a uniformly distributed weighting (best in class = predictive score of 0.85 or higher for systems with data from at least 3 tests):

| sUAS | Navigation: Through Corridors | Navigation: Through Apertures | Takeoff | Landing | Runtime Endurance: Indoor Movement | Room Clearing | Predictive Score |
|---|---|---|---|---|---|---|---|
| Cleo Robotics Dronut X1P | 0.90 | 1.0 | 0.71 | 0.87 | 0.76 | 0.73 | 0.82 |
| Flyability Elios 2 GOV | 1.0 | 1.0 | 1.0 | 1.0 | 0.5 | 0.76 | 0.85 |
| Lumenier Nighthawk V3 | 0.84 | 1.0 | 1.0 | 0.87 | - | - | 0.92 |
| Parrot ANAFI USA GOV | 0.83 | 0.83 | 1.0 | 1.0 | 0.5 | 0.79 | 0.80 |
| Skydio X2D | - | - | 0.75 | 0.97 | 0.65 | 0.75 | 0.77 |
| Teal Golden Eagle | - | - | 0.99 | 0.91 | - | - | 0.95 |
| Vantage Robotics Vesper | 0.80 | 1.0 | 0.82 | 0.89 | - | 0.85 | 0.87 |
| **Best in class** | **Contextual Autonomy Ranking** | | | | | | |
| | Flyability Elios 2 GOV<br>Lumenier Nighthawk V3<br>Vantage Robotics Vesper | | | | | | |

The following figure illustrates contextual autonomy score and ranking of the evaluated sUAS according to the above table.

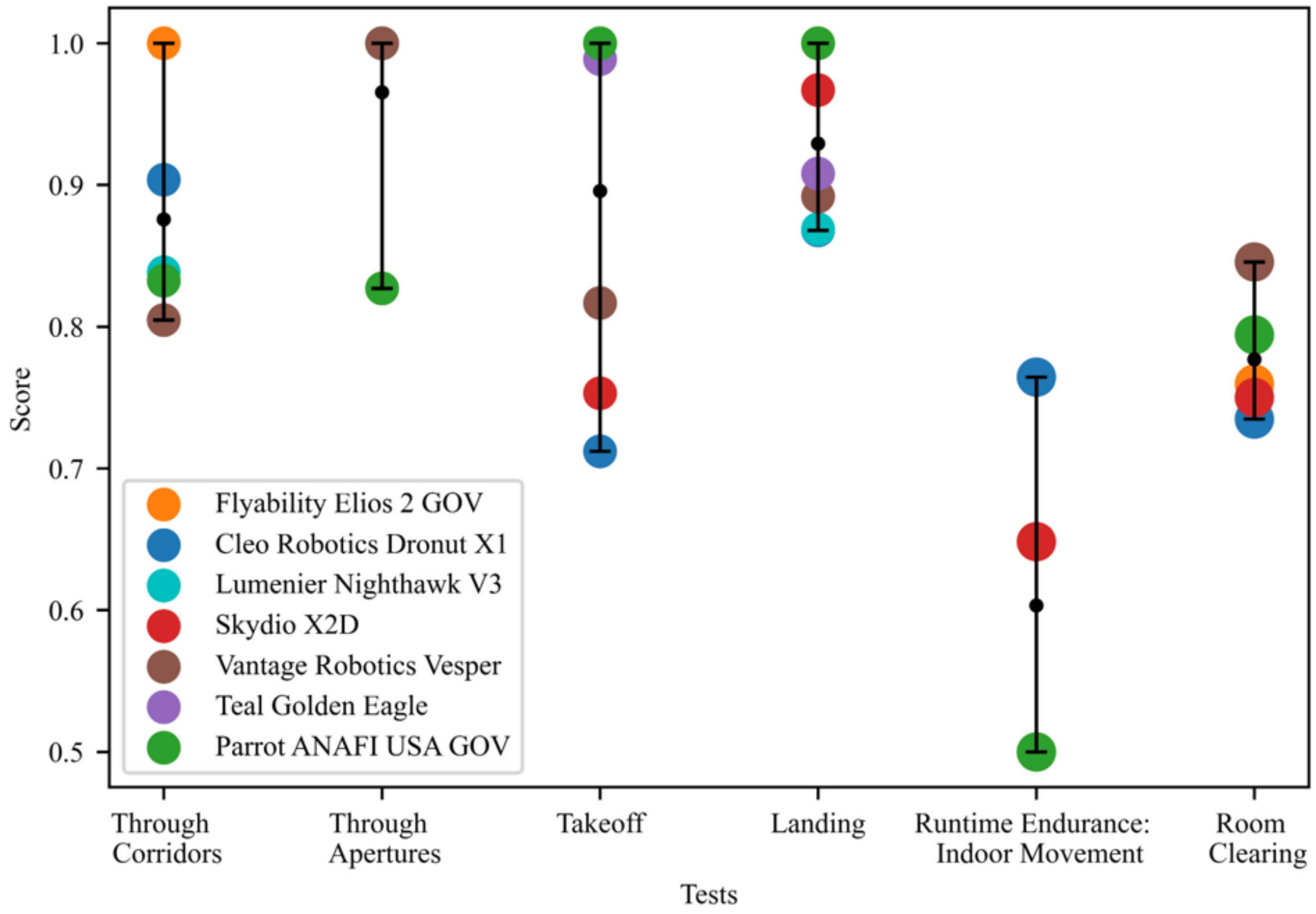

# Trust

## Characterizing Factors of Trust

### Summary of Test Method

This test method consists of recording videos of scripted sUAS operations and conducting surveys of participants to provide subjective feedback regarding their trust in the system. A set of relevant sUAS features that could impact trust are selected (e.g., performance, appearance, interaction type) along at least two conditions for each feature (e.g., appearance: with or without exposed propellers). A relevant task and environment description is also generated to match the use case. The videos are then scripted such that the task can be performed equivalently when sUAS are used that possess each set of relevant feature conditions. A series of videos are recorded, one for each set of feature conditions, showing both an external view of the operator commanding the sUAS and a camera feed from the onboard the sUAS. The videos should follow a similar structure in terms of timing, resolution, and camera angles such that they are similar enough to the others in the series aside from the feature condition that is active. The videos can be edited as needed in order to adjust their feature conditions (e.g., to make a sUAS appear to generate less noise than it actually does, replace the audio with that of a quieter sUAS or by adjusting the volume).

Participants read a description of a scenario involving sUAS performing an operationally relevant task (e.g., performing a mapping mission in the ruins of a subway line) in which there might be some obstacles, hazards, and other features of interest. In the scenario, the feature conditions that will be evaluated are described as options for how to perform the task with the sUAS. For example, if the impact of interaction type on trust is to be evaluated, then the scenario will describe that the sUAS can either be teleoperated remotely or while co-located with the sUAS. Participants then watch a video of the scenario showing the sUAS performing the task with one of the feature conditions active. The selection of which feature condition to show a participant is selected randomly from the set of generated videos, but the number of responses sought for each feature condition should be equivalent to balance the results. Participants can be recruited to perform this test in-person or online (e.g., Amazon Mechanical Turk, Prolific), noting that online questionnaires can yield higher response rates.

Two types of tests are specified that investigate the effects of different factors on human trust in a sUAS as an assistant in a cooperative task:

Feature Characterization: Each participant watches a single video/feature condition. After watching the video, participants respond to questionnaires to rate their trust in the sUAS shown in the video.

System Comparison: Each participant watches two videos (randomly ordered, but balanced across participants) with one or more feature conditions active, each video using a different sUAS platform to perform the task. After watching the videos, participants first answer a questionnaire to indicate their preference of which system they would prefer to have perform the task, provide a reason why, and then respond to questionnaires to rate their trust in one of the two sUAS (randomly selected, but balanced across participants) shown in the video.

## Benchmarking Results

Note: the Feature Characterization test was performed in order to determine what elements should be evaluated for trust, whereas the System Comparison test was performed as a means to compare platforms to one another. Thus, only data from the System Comparison test is shown in this report.

### System Comparison

Due to the delayed timing of receiving sUAS platforms for the DECISIVE program, trust evaluations were conducted to compare the Flyability Elios 2 GOV (a system received very early on) and a DJI Mavic 2 Pro (a system purchased prior to the program). The evaluation results explicitly reflect trust measures for these two sUAS, but the characteristics by which they are compared are shared by the other sUAS that are benchmarked throughout this report. As such, the results below are generalized to apply to these other systems.

Evaluations were performed to compare the following sUAS characteristics:

- Noise generated by the system: quiet vs. loud
- Protective hardware on the system: with propeller protection vs. without propeller protection
- Low light performance, i.e. when the system is in a dark environment: high visibility video transmission vs. low visibility video transmission

These characteristics are applied to each sUAS as follows, with characteristics matching the Flyability Elios 2 GOV in blue and those matching the DJI Mavic Pro 2 in orange:

| sUAS | Noise | Protective Hardware | Low Light Performance |
|---|---|---|---|
| Cleo Robotics Dronut X1P | Loud | Yes | Low |
| DJI Mavic Pro 2 | Quiet | No | Low |
| FLIR Black Hornet PRS | Quiet | No | Low |
| Flyability Elios 2 GOV | Loud | Yes | High |
| Lumenier Nighthawk V3 | Quiet | No | High |
| Parrot ANAFI USA GOV | Quiet | No | Low |
| Skydio X2D | Quiet | No | Low |
| Teal Golden Eagle | - | No | Low |
| Vantage Robotics Vesper | Loud | Yes | Low |

Videos were generated for the several unique combinations of characteristics as follows:

| sUAS | Protective Hardware | Lighting | Low Light Performance | Noise | Video |
|---|---|---|---|---|---|
| Flyability Elios 2 GOV | Yes | Light | High | High | A: https://youtu.be/9F0LaH4lOHo |
| | | Dark | High | High | B: https://youtu.be/7m9p3LO9or8 |
| | | | | Low | C: https://youtu.be/L7ZhB3ccda8 |
| DJI Mavic Pro 2 | No | Light | High | Low | D: https://youtu.be/b6H12nRdRNI |
| | No | Dark | Low | Low | E: https://youtu.be/eASF3gq1FR0 |

The following videos were used for each comparison in order to isolate the sUAS characteristic being evaluated:

- Noise: A vs. C (same sUAS, same lighting, same low light performance, different noise)
- Protective hardware: A vs. D (different sUAS and protective hardware, same lighting)
- Low light performance: B vs. E (different sUAS, same lighting, different low light performance)

Significant differences or a trend toward significant differences are presented when comparing the results of individual questionnaire items. Only one significant difference was found for the noise characteristic, so no conclusive evaluation data is presented. Multiple significant or trending towards significant differences were found for the protective hardware and performance characteristics; the results of those evaluations are shown below:

| Protective Hardware | | | | | |
|---|---|---|---|---|---|
| **Trust Measure** | **Items with significant or a trend toward significant differences** | **Flyability Elios 2 GOV Mean Score** | **DJI Mavic Pro 2 Mean Score** | **Mann-Whitney Test** | **More Trustworthy System** |
| **HCTM** | 1. I believe that there could be negative consequences when using the drone | 3.43 | 2.73 | U=375 p=0.09 | DJI Mavic Pro 2 |
| | 3. It is risky to interact with the drone | 3.53 | 2.93 | U=378 p=0.10 | DJI Mavic Pro 2 |
| | 4. I believe that the drone will act in my best interest | 4.93 | 4.23 | U=373 p=0.09 | Flyability Elios 2 GOV |
| | 6. I believe that the drone is interested in understanding my needs and preferences | 3.93 | 2.93 | U=345 p=0.05 | Flyability Elios 2 GOV |
| | 10. If I use the drone, I think I would be able to depend on it completely | 4.63 | 4.19 | U=292 p=0.08 | Flyability Elios 2 GOV |
| **CTPA** | 3. I am confident in the system. | 5.16 | 4.31 | U=329.5 p=0.07 | Flyability Elios 2 GOV |

| Low Light Performance | | | | | |
|---|---|---|---|---|---|
| **Trust Measure** | **Items with significant or a trend toward significant differences** | **Flyability Elios 2 GOV Mean Score** | **DJI Mavic Pro 2 Mean Score** | **Mann-Whitney Test** | **More Trustworthy System** |
| **HCTM** | 2. I feel I must be cautious when using the drone | 4.80 | 5.43 | U=360.5 p=0.08 | Flyability Elios 2 GOV |
| | 3. It is risky to interact with the drone | 3.50 | 4.22 | U=293 p=0.05 | Flyability Elios 2 GOV |
| | 6. I believe that the drone is interested in understanding my needs and preferences | 3.56 | 3.03 | U=371 p=0.11 | Flyability Elios 2 GOV |
| | 11. I can always rely on the drone for performing mapping task | 5.53 | 4.90 | U=373 p=0.12 | Flyability Elios 2 GOV |
| **CTPA** | 3. I am confident in the system. | 5.64 | 4.96 | U=329.5 p=0.07 | Flyability Elios 2 GOV |
| | 4. The system provides security. | 4.92 | 4.20 | U=333.5 p=0.08 | Flyability Elios 2 GOV |

Based on these results, it is suggested that the Flyability Elios 2 GOV is considered to be the more trustworthy system when compared to the DJI Mavic Pro 2. Generalizing these results to the other sUAS in order to determine a best in class evaluation across all platforms (i.e., those with protective hardware and those with higher performance in low light environments) is as follows:

| | **More Trustworthy System Per Possession of Protective Hardware** | **More Trustworthy System Per Its Performance in Low Light Environments** |
|---|---|---|
| **Best in class** | Cleo Robotics Dronut X1P<br>Flyability Elios 2 GOV<br>Vantage Robotics Vesper | Flyability Elios 2 GOV<br>Lumenier Nighthawk V3 |

# Situation Awareness

## Interface-Afforded Attention Allocation

Affiliated publications: [Choi et al., 2022]

### Summary of Test Method

This test method is used to evaluate the probability of attending to various situational elements (SEs) provided by the sUAS interface over a series of operationally relevant scenarios (ORS), using the SEEV model for analysis. In order to calculate SA by applying the SEEV model, each SE can be divided into the four SEEV parameters (Salience, Effort, Expectancy, and Value) to quantify the attention allocation proportion ($f_i$). The quantitative value is the weighted sum or multiplication of four parameters. Salience is weighted according to the color, size, and type of the SE. Effort is weighted according to the movement of the operator's eye to see the SE. Expectancy refers to the weight of the event frequency or changeability of SE. Value refers to the importance according to the mission of the task. SEs can be quantified by substituting numbers corresponding to information of SEs about Salience/Effort/Expectancy/Value into the SEEV model. Four Landolt Cs (Orange/Red/Green/Blue), two images (Radioactive/Oxygen), and four indicators (Altitude/Heading/Front Distance to Surface/System Battery) were identified as SEs from a series of operationally relevant scenarios (ORSs) were designed in order to mimic specific conditions in the environment, elements of interest, and mission tasks for sUAS operations in subterranean and constrained indoor spaces. Based on the result of the attention allocation proportion ($f_i$), we set out to compare $f_i$ among different platforms (depending on the platform, there are cases where there is no SE(s) among indicator SEs).

### Benchmarking Results

Graph of the attention allocation proportion ($f_i$) for each platform:

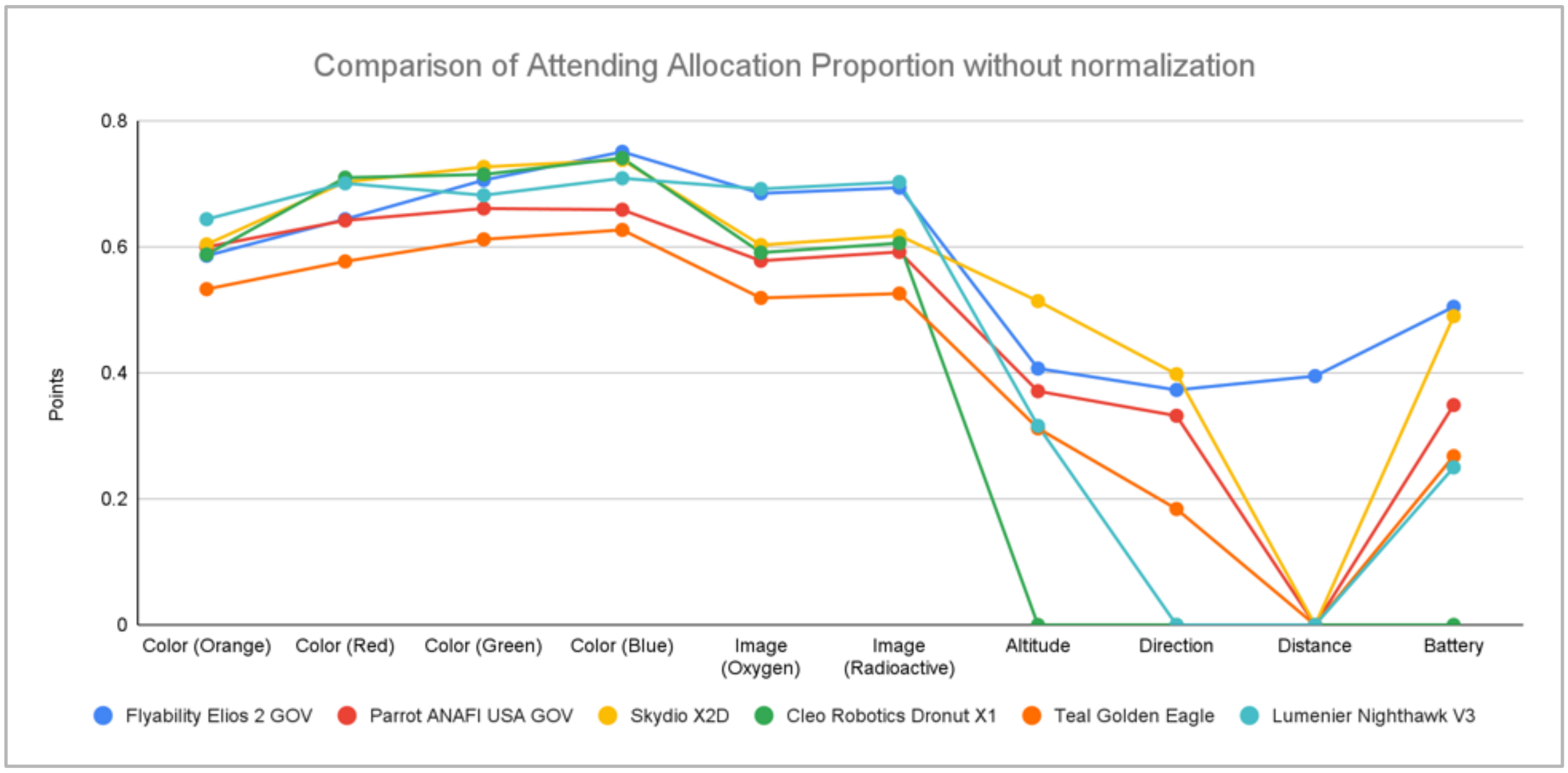

Data of the attention allocation for each platform (best in class = total score of 5.0 or higher):

| Name | SE | Cleo Robotics Dronut X1P | Flyability Elios 2 GOV | Lumenier Nighthawk V3 | Parrot ANAFI USA GOV | Skydio X2D | Teal Golden Eagle |
|---|---|---|---|---|---|---|---|
| **Color (Orange)** | **SE01** | 0.588 | 0.586 | 0.644 | 0.600 | 0.604 | 0.533 |
| **Color (Red)** | **SE02** | 0.710 | 0.644 | 0.701 | 0.642 | 0.703 | 0.577 |
| **Color (Green)** | **SE03** | 0.715 | 0.706 | 0.682 | 0.661 | 0.727 | 0.612 |
| **Color (Blue)** | **SE04** | 0.741 | 0.751 | 0.709 | 0.659 | 0.738 | 0.627 |
| **Image (Oxygen)** | **SE05** | 0.591 | 0.685 | 0.692 | 0.578 | 0.603 | 0.519 |
| **Image (Radioactive)** | **SE06** | 0.606 | 0.694 | 0.703 | 0.592 | 0.618 | 0.526 |
| **Altitude** | **SE07** | - | 0.407 | 0.316 | 0.371 | 0.514 | 0.312 |
| **Direction** | **SE08** | - | 0.373 | - | 0.332 | 0.398 | 0.184 |
| **Distance** | **SE09** | - | 0.395 | - | - | - | - |
| **Battery** | **SE10** | - | 0.505 | 0.250 | 0.349 | 0.490 | 0.268 |
| | **Total** | 3.951 | 5.746 | 4.697 | 4.784 | 5.395 | 4.158 |
| | **Best in class** | **Interface-Afforded Attention Allocation** | | | | | |
| | | Flyability Elios 2 GOV<br>Skydio X2D | | | | | |

## Situation Awareness (SA) Survey Comparison

Affiliated publications: [Choi et al., 2021]

### Summary of Test Method

According to the flight missions (Aviate/Navigate/Hazard detection), SEs are divided into required SEs and desire SEs based on the highest importance perceived by pilots. Therefore, the difference in abilities according to the missions can be emphasized by assigning weight required SEs and desired SEs. Through an experiment with a participant, each perception level (p(SE)), which is measured whether the participant is aware of the SE (i.e., undetected, detected, or comprehended), is obtained. The perception levels of SEs can be expressed as a metric of SA. The metric is the weighted sum of required SEs and desired SE. Existing models and assessment methods, including the MIDAS (Man-machine Integration Design and Analysis)-based SA model and Attention Allocation Model [Xu et al., 2013], are adopted and refined in order to make them more suitable for sUAS operations in subterranean and constrained indoor environments. The metric goes through the theoretical models and OSA is obtained.

Participants can be recruited to perform this test in-person or online (e.g., Amazon Mechanical Turk, Prolific), noting that online questionnaires can yield higher response rates. Participants watch an evaluation video of a sUAS flight with a specific focus to observe and comprehend the SEs of the operational environment. At certain points while watching the video, the video will pause and questions from the SAGAT questionnaire will pop up for the participant to answer. The participants' answers are evaluated for accuracy to determine a corresponding score, and SA values are calculated using quantitative assessment methods.

### Benchmarking Results

Boxplots of operator SA values from the MIDAS-based SA model (MDS) and the attention allocation SA model (AAM):

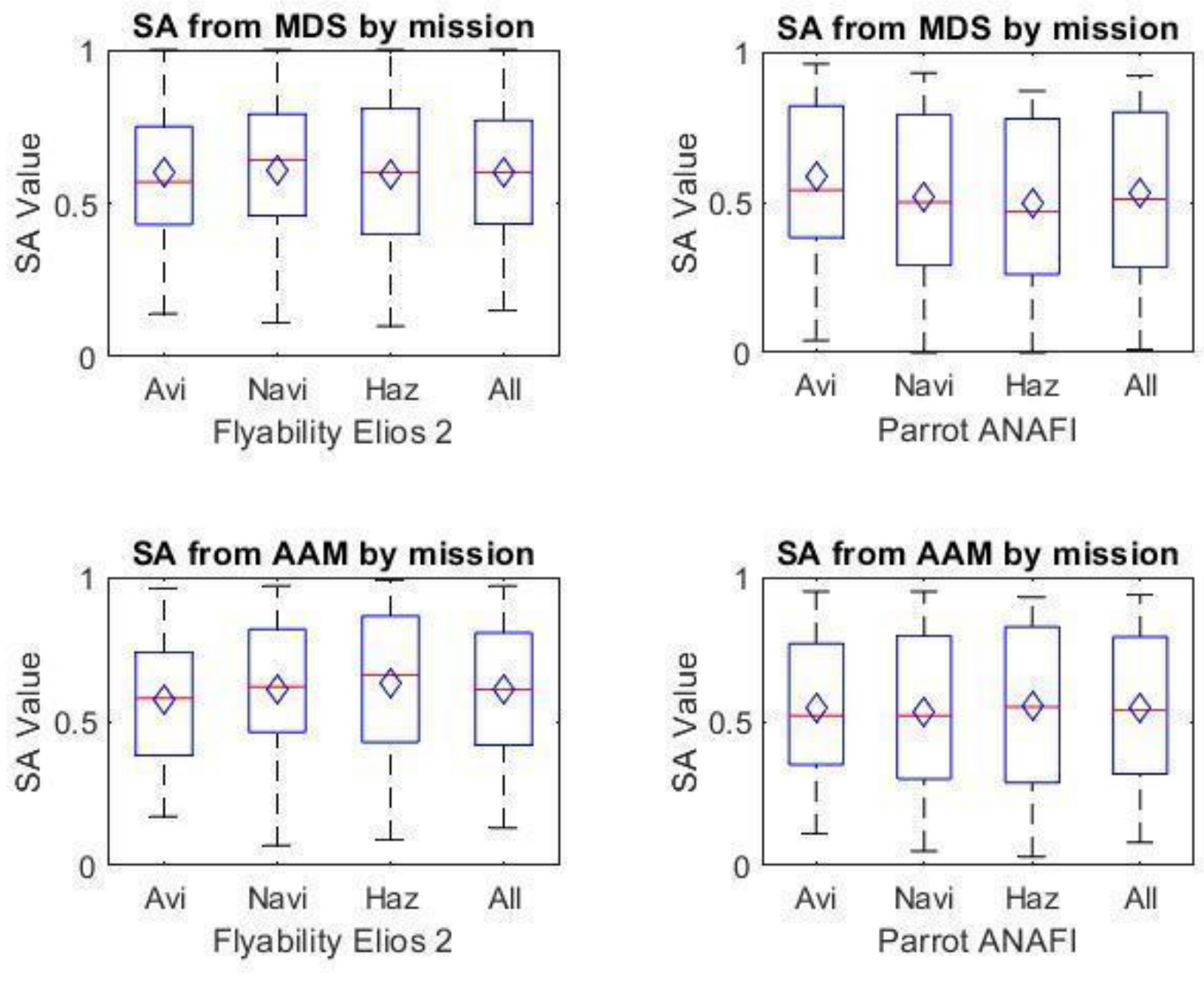

Note: ◊ denotes the mean of the result. Key: Avi = Aviate, Navi = Navigate, Haz = Hazard Detection, All = Overall

Data of Operator SA values:

| Mission | Flyability Elios 2 GOV | | Parrot ANAFI USA GOV | |
|---|---|---|---|---|
| | MDS | AAM | MDS | AAM |
| | $\mu, \sigma$ | $\mu, \sigma$ | $\mu, \sigma$ | $\mu, \sigma$ |
| **Aviate** | 0.22, 0.60 | 0.22, 0.59 | 0.27, 0.59 | 0.26, 0.55 |
| **Navigate** | 0.23, 0.61 | 0.24, 0.61 | 0.29, 0.51 | 0.28, 0.54 |
| **Hazard** | 0.26, 0.59 | 0.25, 0.63 | 0.28, 0.49 | 0.28, 0.55 |
| **Overall** | 0.23, 0.60 | 0.23, 0.61 | 0.27, 0.53 | 0.27, 0.54 |

The OSA calculated for participants of the experiment are tabulated in terms of mean and standard deviation, denoted by $\mu$ and $\sigma$, respectively. The results are grouped based on the mission and OSA values for each sUAS are calculated using both MIDAS-based SA model and Attention Allocation Model.

# References

[Choi et al., 2021] Choi, Minseop, John Houle, Nathan Letteri, and Thanuka L. Wickramarathne. "On The Use of Small-Scale Unmanned Autonomous Systems for Decision-Support in Subterranean Environments: The Case of Operator Situational Awareness Assessment." In 2021 IEEE International Symposium on Technologies for Homeland Security (HST), pp. 1-6. IEEE, 2021.

[Choi et al., 2022] Choi, Minseop, John Houle, and Thanuka L. Wickramarathne. "On the Development of Quantitative Operator Situational Awareness Assessment Methods for Small-Scale Unmanned Aircraft Systems." In 2022 25th International Conference on Information Fusion (FUSION), pp. 1-8. IEEE, 2022.

[Durst and Gray, 2014] Durst, Phillip J., and Wendell Gray. Levels of autonomy and autonomous system performance assessment for intelligent unmanned systems. ENGINEER RESEARCH AND DEVELOPMENT CENTER VICKSBURG MS GEOTECHNICAL AND STRUCTURES LAB, 2014.

[Donald et al., 2023] Donald, Ryan, Peter Gavriel, Adam Norton, and S. Reza Ahmadzadeh. Contextual Autonomy Evaluation of Unmanned Aerial Vehicles in Subterranean Environments. In Proceedings of the 9th International Conference on Automation, Robotics and Applications (ICARA) 2023, February 2023.

[Hertel et al., 2022] Hertel, Brendan, Ryan Donald, Christian Dumas, and S. Reza Ahmadzadeh. "Methods for Combining and Representing Non-Contextual Autonomy Scores for Unmanned Aerial Systems." In 2022 8th International Conference on Automation, Robotics and Applications (ICARA), pp. 135-139. IEEE, 2022.

[Huang et al., 2007] Huang, Hui-Min, Kerry Pavek, Mark Ragon, Jeffry Jones, Elena Messina, and James Albus. "Characterizing unmanned system autonomy: Contextual autonomous capability and level of autonomy analyses." In *Unmanned Systems Technology IX*, vol. 6561, pp. 509-517. SPIE, 2007.

[Meriaux and Jerath, 2022] Meriaux, Edwin and Kshitij Jerath. “Evaluation of Navigation and Trajectory-following Capabilities of Small Unmanned Aerial Systems.” In 2022 IEEE International Symposium on Technologies for Homeland Security (HST), pp. 1-8. IEEE, 2022.

[Norton et al., 2021] Norton, Adam, Peter Gavriel, Brendan Donoghue, and Holly Yanco. "Test Methods to Evaluate Mapping Capabilities of Small Unmanned Aerial Systems in Constrained Indoor and Subterranean Environments." In 2021 IEEE International Symposium on Technologies for Homeland Security (HST), pp. 1-8. IEEE, 2021.

[Norton et al., 2022] Norton, Adam, Reza Ahmadzadeh, Kshitij Jerath, Paul Robinette, Jay Weitzen, Thanuka Wickramarathne, Holly Yanco, Minseop Choi, Ryan Donald, Brendan Donoghue, Christian Dumas, Peter Gavriel, Alden Giedraitis, Brendan Hertel, Jack Houle, Nathan Letteri, Edwin Meriaux, Zahra Rezaei, Rakshith Singh, Gregg Willcox, and Naye Yoni. “DECISIVE Test Methods Handbook: Test Methods for Evaluating sUAS in Subterranean and Constrained Indoor Environments, Version 1.1.” arXiv preprint arXiv:2211.01801, November 2022.

[Xu et al., 2013] Xu, Wu, Wanyan Xiaoru, and Zhuang Damin. “Attention allocation modeling under multi-factor condition.” In Journal of Beijing University of Aeronautics and Astronautics, 2013.